%% file: paper.tex
\PassOptionsToPackage{dvipsnames}{xcolor}

\documentclass[sigconf,screen,balance=true]{acmart}

\acmSubmissionID{604}

\copyrightyear{2024}
\acmYear{2024}
\setcopyright{rightsretained}
\acmDOI{10.1145/3641519.3657445}
\usepackage[capitalize,nameinlink]{cleveref}
\usepackage{xspace}
\usepackage[percent]{overpic}
\usepackage{pict2e} 
\usepackage{colortbl} 
\usepackage[caption=false]{subfig}
\usepackage{rotating}
\usepackage{tikz}
\usepackage{pifont}
\usepackage{tabularx}
\usetikzlibrary{positioning}
\usepackage{enumitem}
\setlist[itemize]{align=parleft,left=0.5em..1.5em}

\usepackage{graphicx} 

\acmConference[SIGGRAPH Conference Papers '24]{Special Interest Group on Computer Graphics and Interactive Techniques Conference Conference Papers '24}{July 27-August 1, 2024}{Denver, CO, USA}
\acmBooktitle{Special Interest Group on Computer Graphics and Interactive Techniques Conference Conference Papers '24 (SIGGRAPH Conference Papers '24), July 27-August 1, 2024, Denver, CO, USA}
\acmISBN{979-8-4007-0525-0/24/07}

\begin{document}

\input{macros}


\title[\rgbtooxx: Image decomposition and synthesis using material- and lighting-aware diffusion models]{\rgbtooxx: Image decomposition and synthesis using\\material- and lighting-aware diffusion models}

\author{Zheng Zeng}
\email{zhengzeng@ucsb.edu}
\orcid{0000-0001-9025-9427}
\affiliation{
    \institution{Adobe Research}
    \institution{University of California, Santa Barbara}
    \country{USA}
}

\author{Valentin Deschaintre}
\email{deschain@adobe.com}
\orcid{0000-0002-6219-3747}
\affiliation{
    \institution{Adobe Research}
    \country{United Kingdom}
}

\author{Iliyan Georgiev}
\email{igeorgiev@adobe.com}
\orcid{0000-0002-9655-2138}
\affiliation{
    \institution{Adobe Research}
    \country{United Kingdom}
}

\author{Yannick Hold-Geoffroy}
\email{holdgeof@adobe.com}
\orcid{0000-0002-1060-6941}
\affiliation{
    \institution{Adobe Research}
    \country{Canada}
}

\author{Yiwei Hu}
\email{yiwhu@adobe.com}
\orcid{0000-0002-3674-295X}
\affiliation{
    \institution{Adobe Research}
    \country{USA}
}

\author{Fujun Luan}
\email{fluan@adobe.com}
\orcid{0000-0001-5926-6266}
\affiliation{
    \institution{Adobe Research}
    \country{USA}
}

\author{Ling-Qi Yan}
\email{lingqi@cs.ucsb.edu}
\orcid{0000-0002-9379-094X}
\affiliation{
    \institution{University of California, Santa Barbara}
    \country{USA}
}

\author{Miloš Hašan}
\email{mihasan@adobe.com}
\orcid{0000-0003-3808-6092}
\affiliation{
    \institution{Adobe Research}
    \country{USA}
}

\renewcommand{\shortauthors}{Zeng et al.}


\begin{abstract}
    The three areas of realistic forward rendering, per-pixel inverse rendering, and generative image synthesis may seem like separate and unrelated sub-fields of graphics and vision. However, recent work has demonstrated improved estimation of per-pixel intrinsic channels (albedo, roughness, metallicity) based on a diffusion architecture; we call this the \rgbtoxx problem. We further show that the reverse problem of synthesizing realistic images given intrinsic channels, \xxtorgb, can also be addressed in a diffusion framework. Focusing on the image domain of interior scenes, we introduce an improved diffusion model for \rgbtoxx, which also estimates lighting, as well as the first diffusion \xxtorgb model capable of synthesizing realistic images from (full or partial) intrinsic channels. Our \xxtorgb model explores a middle ground between traditional rendering and generative models: We can specify only certain appearance properties that should be followed, and give freedom to the model to hallucinate a plausible version of the rest. This flexibility allows using a mix of heterogeneous training datasets that differ in the available channels. We use multiple existing datasets and extend them with our own synthetic and real data, resulting in a model capable of extracting scene properties better than previous work and of generating highly realistic images of interior scenes.
\end{abstract}




\begin{teaserfigure}
    \centering
    \vspace{-2mm}
    \includegraphics{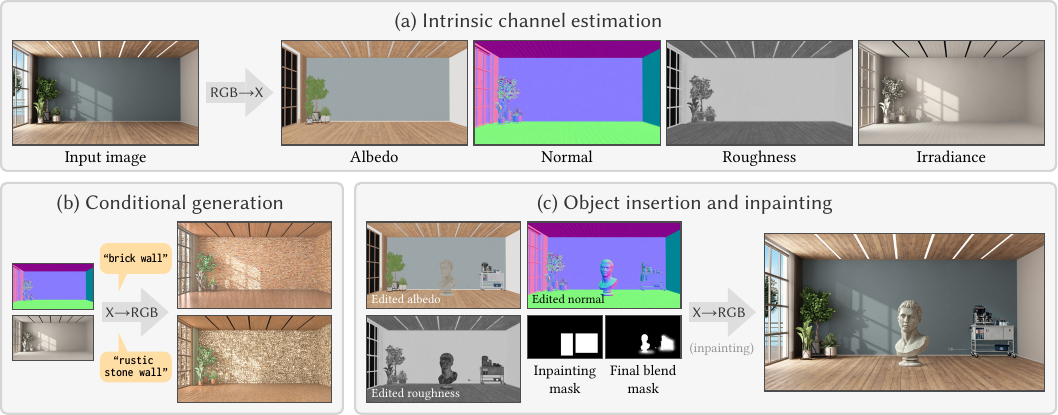}
    \caption{
        We present models for image decomposition into intrinsic channels (\rgbtoxx) and image synthesis from such channels (\xxtorgb) in a unified conditional diffusion framework. (a) Our \rgbtoxx model produces clean, plausible estimates of the intrinsic channels \xx. (b) New realistic images can be produced using our \xxtorgb model. Here we use a subset of the estimated channels, plus a text prompt. (c) We insert synthetic objects into the estimated channels and use an in-painting version of our \xxtorgb model, with appropriate masks, to synthesize a final composite image with matching lighting and shadows.
    }
    \label{fig:teaser}
    \vspace{1mm}
\end{teaserfigure}

\maketitle

\section{Introduction}

Estimating geometric, shading, and lighting information from images has been long studied by the computer vision community, since classical work on intrinsic image decomposition. This problem is inherently difficult due to its under-constrained nature, including the ambiguity between illumination and materials \cite{Grosse2009}. More recent work has focused on the related problem of per-pixel inverse rendering \cite{Li2020iris,IrisFormer}. This has produced physical material and lighting estimations, specifically diffuse albedos, specular roughness and metallicity, as well as various spatially varying representations of lighting. We refer to all of these information buffers as \emph{intrinsic channels} and denote them using the symbol \xx, and the problem of estimating them as \rgbtoxx.

On the other hand, computer graphics, and especially the sub-field of physically based rendering, has long focused on the reverse task of turning detailed scene descriptions (comprising geometry, lighting, and materials) into realistic images. State-of-the-art rendering methods employ Monte Carlo light-transport simulation~\cite{PBRT}, commonly followed by a neural denoiser that encapsulates priors about plausible noise-free images. We refer to the problem of synthesizing an image from a given description as \xxtorgb.

A recent approach to producing highly realistic images, very different from traditional rendering, is based on generative models for image synthesis, especially based on large diffusion models \cite{StableDiffusion,Dalle2}. These models operate by iteratively denoising an image, pushing the neural-denoiser approach to the limit by starting from pure noise.

These three areas may seem unrelated, but we believe they should be studied in a unified way. We explore the connections between diffusion models, rendering, and intrinsic channel estimation, focusing on both material/light estimation and image synthesis conditioned on material/lighting, all in the same diffusion framework.

Recent work has demonstrated improved estimation of intrinsic channels based on a diffusion architecture. \citet{Kocsis2023} observe that further progress in this domain is likely to use generative modeling, due to the under-constrained and ambiguous nature of the problem. We follow this direction further. In addition to a new model for \rgbtoxx which improves upon that of \citet{Kocsis2023}, we also introduce a first \xxtorgb diffusion model which synthesizes realistic images from (full or partial) intrinsic channels. Much like \rgbtoxx, the \xxtorgb problem requires a strong (ideally generative) prior to guide synthesis towards a plausible image, even with incomplete or overly simple intrinsic-channel information \xx.

Typical generative models are simple to use, but hard to precisely control. On the other hand, traditional rendering is precise but requires full scene specification, which is limiting. Our \xxtorgb model explores a middle ground where we specify only certain appearance properties that should be followed, and give freedom to the model to hallucinate a plausible version of the rest.

Our intrinsic channels \xx contain per-pixel albedo, normal vector, roughness, as well as \emph{lighting} information which we represent as per-pixel irradiance on the scene surfaces. Furthermore, our \xxtorgb model is trained using channel dropout, which enables it to synthesize images using any subset of channels as input. This in turn makes it possible to use a mix of heterogeneous training datasets that differ in the available channels. We use multiple existing datasets and add our own synthetic and real data---a key advantage allowing us to expand training data beyond that of previous models. This paper makes the following contributions:
\begin{itemize}
    \item
          An \rgbtoxx model improving upon previous work \cite{Kocsis2023} by using more training data from multiple heterogeneous datasets and adding support for lighting estimation;
    \item
          An \xxtorgb model capable of synthesizing realistic images from given intrinsic channels \xx, supporting partial information and optional text prompts. We combine existing datasets and add a new, high-quality interior scene dataset to achieve high realism.
\end{itemize}
In summary, we propose a unified diffusion-based framework that enables realistic image analysis (intrinsic channel estimation describing geometric, material, and lighting information) and synthesis (realistic rendering given the intrinsic channels), demonstrated in the domain of realistic indoor scene images; see \Cref{fig:teaser}.

Our work is the first step towards unified frameworks for both image decomposition and synthesis. We believe it can bring benefits to a wide range of downstream editing tasks, including material editing, relighting, and realistic rendering from simple/under-specified scene definitions.

\section{Related Work}
\label{sec:relatedwork}

\paragraph{Generative models for images}

Over the last decade, deep-learning-based image generation has rapidly progressed, notably with the advent of generative adversarial networks (GANs) \cite{goodfellow2014generative} and the subsequent body of research that improves both quality and stability \cite{karras2020analyzing, pan2019recent, gui2021review}. However, the adversarial-based approach of GANs is prone to mode collapse, making them challenging to train. More recently, diffusion models have been shown to scale to training sets of hundreds of millions of images and produce extremely high-quality images~\cite{Dalle2, StableDiffusion}. However, such models are costly to train, prompting research to fine-tune pre-trained models for various domains or conditioning~\cite{Alchemist, ControlNet, hu2021lora}, rather than training from scratch. We leverage the recent progress in this area to design our network architectures on top of Stable Diffusion v2.1~\cite{StableDiffusion}, adding conditioning and dropout as a means for flexible input at test time.

\paragraph{Intrinsic decomposition}

The problem of intrinsic image decomposition was defined almost five decades ago by \citet{barrow1978recovering} as a way to approximate an image $I$ as a combination of diffuse reflectance (albedo), diffuse shading (irradiance), and
optionally a specular term. Priors are necessary to estimate multiple values per pixel. Early priors include the retinex theory \cite{land1971lightness} which states that shading tends to have slower variation than reflectance. Pre-2009 methods are summarized by \citet{Grosse2009}, while more recent methods are summarized by \citet{garces2022survey}. We compare our albedo estimates to the most recent method of~\citet{OrdinalShading}.

Several recent works extend the traditional intrinsic decomposition to estimate more values per pixel, including specular roughness and/or metallicity, and lighting representations. Their training datasets focus on interior scenes. \citet{Li2020iris} are the first to use a large synthetic dataset of paired RGB renderings and decompositions to train a convolutional architecture for intrinsic channel estimation. The synthetic dataset used to train this method was later improved and released as \emph{OpenRooms} \cite{OpenRooms}. A further improvement was achieved by a switch from convolutional to vision transformer architectures \cite{IrisFormer}. More recently, \citet{InteriorVerse} introduce a new, more realistic synthetic interior dataset, and trained a convolutional architecture outperforming the method of \citet{Li2020iris}, mostly due to the more realistic dataset.

A more recent alternative is to extract intrinsic images from pre-trained models such as StyleGAN~\cite{karras2019style} or pre-trained diffusion models~\cite{du2023generative, lee2023exploiting, bhattad2024stylegan}. In this spirit, intrinsic image diffusion \cite{Kocsis2023} proposes to fine-tune a general-purpose diffusion model to the per-pixel inverse rendering problem, going beyond previous methods by leveraging priors learned for image generation instead of predicting an average of the plausible solutions at each pixel. Their model is trained on \datasetIV~\cite{InteriorVerse}, a synthetic dataset of interior renderings. We further extend this work by training a similar \rgbtoxx model with a different architecture on more data sources and additional intrinsic buffers. We further couple it with a new \xxtorgb model synthesizing realistic images from these buffers, effectively closing the loop back to RGB.

\paragraph{Normal estimation}

Estimating per-pixel normal is related to intrinsic decomposition as it estimates 3D information for each pixel which is highly relevant to shading. However, this problem is typically studied in isolation from intrinsic images and has recently received limited attention compared to depth estimation. To demonstrate the competitiveness of our method, we consider an internal method, PVT-normal, based on Pyramid Vision Transformer \cite{wang2022pvt} and trained on datasets similar to MiDaS \cite{Ranftl2022, birkl2023midas} to estimate normals. In our tests, PVT-normal outperforms the currently available state-of-the-art normal estimation methods. This model is not specific to interior scenes and is trained on a diverse dataset.

\paragraph{Neural image synthesis from decompositions}

Several previous works have explored problems similar to our \xxtorgb problem. Deep Shading \cite{DeepShading} solves the problem of learning screen-space shading effects (e.g., ambient occlusion, image-based lighting, subsurface scattering) using a CNN-based architecture learned on synthetic data, resulting in fast rendering, competitive or better than hand-tuned screen-space shaders. Deep Illumination \cite{DeepIllumination} is an approach based on a conditional GAN learned per scene, efficiently predicting global illumination given screen-space intrinsic buffers, while direct illumination is computed analytically. \citet{InteriorVerse} introduce a screen-space ray-tracing approach to synthesize images from intrinsic channels. In contrast, our approach jointly considers image decomposition and synthesis, does not require any ray tracing, and its models are general across the interior scene domain.

\paragraph{Relighting}

Single-image scene relighting methods have been proposed using both explicit \cite{Outcast, yu2020self, pandey2021total} and implicit \cite{rudnev2022nerf, wang2023fegr} representations. These works are limited to simple lighting: a single directional light source or low-order spherical harmonics. Closer to our work, \citet{Li2022relighting} build a per-pixel inverse rendering method to relight interior scenes from a single image. Furthermore, they introduce a hybrid neural and classical rendering system that synthesizes relit images given intrinsic channels and lighting information, similar to our \xxtorgb. While we believe our framework can be part of a toolbox for relighting, we do not specifically focus on solving the relighting problem, which poses challenges beyond our scope.

\begin{table}[t]
    \centering
    \newcommand{\available}{\textcolor{Green}{\ding{51}}}
    \newcommand{\unreliable}{\textcolor{BurntOrange}{\ding{51}}}
    \newcommand{\unavailable}{\textcolor{Red}{\ding{55}}}
    \caption{
        We combine four heterogeneous datasets (ours in bold), each providing a subset of the channels we need for training. For each dataset we mark channels as available (\available), unavailable (\unavailable), or available but not fully reliable (\unreliable). We also include representative images from the datasets.
        \datasetSI is an RGB-only dataset for which we estimated the intrinsic channels using our \rgbtoxx model.
    }
    \scalebox{0.88}{%
        \setlength{\tabcolsep}{2pt}%
        \begin{tabularx}{1.13\linewidth}{lcccccccc}
            \toprule
            \textbf{Dataset}    & \textbf{Size} & \textbf{Albedo} & \textbf{Normal} & \textbf{Roughness} & \textbf{Metallic.} & \textbf{Irrad.} \\
            \midrule
            \datasetIV\;        & 50,097        & \available      & \available      & \unreliable        & \unreliable        & \unavailable    \\
            \datasetHS          & 73,819        & \unreliable     & \available      & \unavailable       & \unavailable       & \available      \\
            \textbf{\datasetEM} & 17,000        & \available      & \available      & \available         & \available         & \unavailable    \\
            \textbf{\datasetSI} & 50,000        & \available      & \available      & \unreliable        & \unreliable        & \available      \\
            \bottomrule
        \end{tabularx}
    }
    \\[2mm]
    \setlength{\tabcolsep}{2pt}%
    \small
    \begin{tabular}{cccc}
        \includegraphics[width=0.23\linewidth,height =0.23\linewidth]{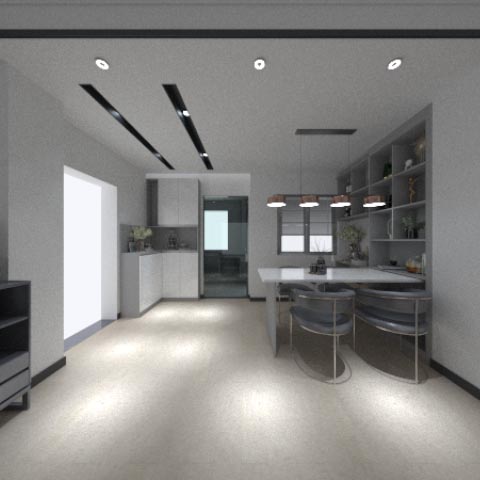} &
        \includegraphics[width=0.23\linewidth,height =0.23\linewidth]{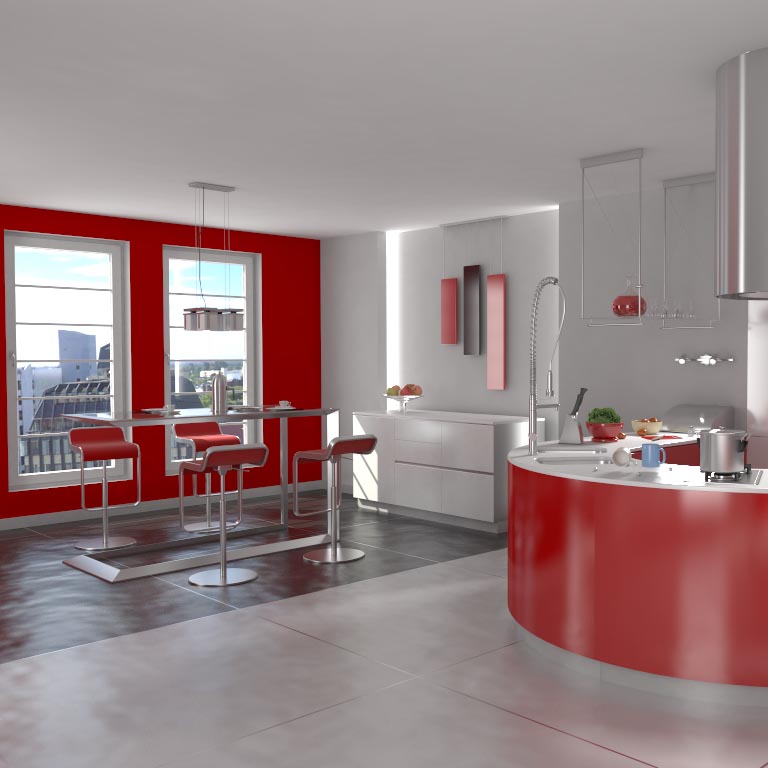}      &
        \includegraphics[width=0.23\linewidth,height =0.23\linewidth]{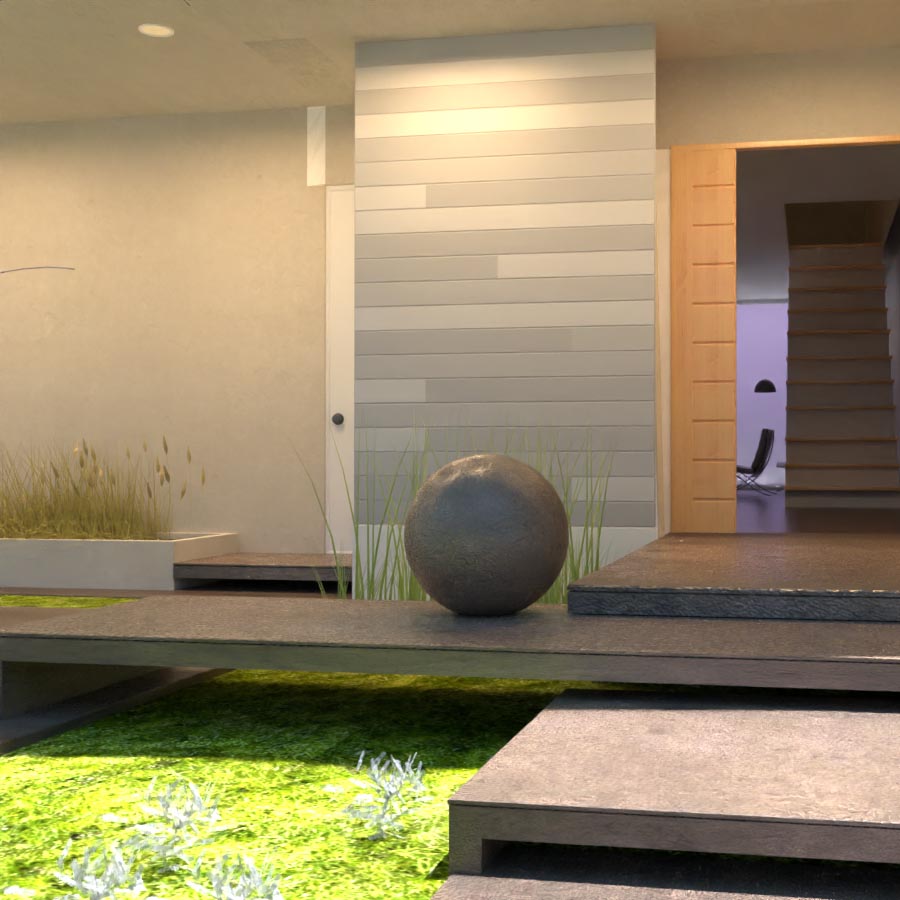}    &
        \includegraphics[width=0.23\linewidth,height =0.23\linewidth]{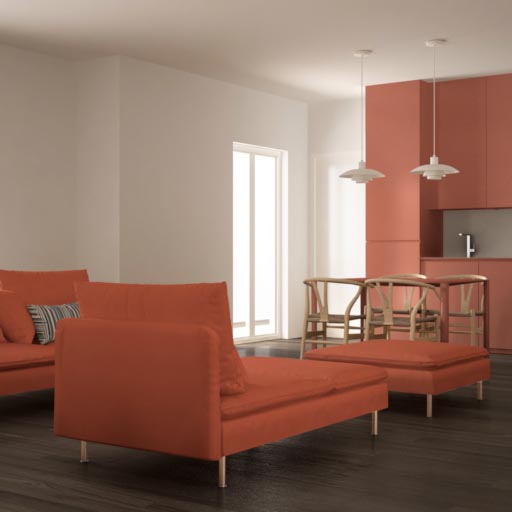}                                                                     \\[-0.5mm]
        \datasetIV                                                                                            & \datasetHS & \textbf{\datasetEM} & \textbf{\datasetSI}
    \end{tabular}
    \label{tab:datasets}
\end{table}

\section{Intrinsic channels and datasets}

In this section, we discuss the intrinsic channels \xx used in our models, and the datasets with paired RGB images and intrinsic channels that we used or prepared ourselves.

\subsection{Intrinsic channels}
\label{sec:Channels}

In our \rgbtoxx and \xxtorgb models, we use the following channels:
\begin{itemize}
    \item
          Normal vector $\normal \in \mathbb{R}^{H \times W \times 3}$ specifying geometric information in camera space;
    \item
          Albedo $\albedo \in \mathbb{R}^{H \times W \times 3}$, also commonly referred to as base color, which specifies the diffuse albedo for dielectric opaque surfaces and specular albedo for metallic surfaces;
    \item
          Roughness $\roughness \in \mathbb{R}^{H \times W}$, typically understood as the square root of the parameter $\alpha$ in GGX or Beckmann microfacet distributions \cite{Walter2007}. High roughness means more matte materials while low roughness means shinier;
    \item
          Metallicity $\metallic \in \mathbb{R}^{H \times W}$, typically defined as a linear blend weight interpolating between treating the surface as dielectric and metallic; and
    \item
          Diffuse irradiance $\irra \in \mathbb{R}^{H \times W \times 3}$, which serves as a lighting representation. It represents the amount of light reaching a surface point integrated over the upper cosine-weighted hemisphere.
\end{itemize}
We also contemplated adding a per-pixel depth channel, but eventually found it unnecessary, as depth can be estimated from normals, and normals typically contain more information about high-frequency local variations.

Unlike a material system in a traditional rendering framework, the above properties are fairly imprecise. For example, they cannot represent glass. Instead, we treat glass as having zero roughness and metallicity. This usually does not pose problems: the model infers from context that an object is a window or a glass cabinet, and plausibly inpaints objects or illumination behind the glass.

All intrinsic channels in our datasets have the same resolution as the corresponding RGB images, and are estimated at full resolution by \rgbtoxx. However, it is sometimes beneficial to condition \xxtorgb on downsampled channels, as discussed in \Cref{sec:x2rgb}.

\subsection{Datasets}
\label{sec:Datasets}

To train our models, we ideally desire a large, high-quality image dataset, containing paired information for all of the channels we require: normal \normal, albedo \albedo, roughness \roughness, metallicity \metallic, diffuse irradiance \irra, the corresponding RGB image \image (ideally a real photograph or at least a very realistic render), and a text caption describing the image. However, no existing dataset satisfies these requirements, and we instead piece together datasets with partial information and construct new datasets to fill the gaps. \Cref{tab:datasets} summarizes the size and channel availability of the datasets we use.

\datasetIV~\cite{InteriorVerse} is a synthetic indoor scene dataset, containing over 50,000 rendered images with \normal, \albedo, \roughness, and \metallic channels in addition the rendered images \image. There are a few issues with this dataset. First, the rendered images contain noise; this does not pose problems for \rgbtoxx estimation, but the \xxtorgb synthesis model learns to reproduce the undesirable noise. We resolve this by applying an off-the-shelf denoiser (NVIDIA OptiX denoiser~\cite{optix}). Furthermore, we found that roughness and metallicity values are often dubious, and decided not use them for this dataset. The dataset also has a synthetic style, which the \xxtorgb model would learn to imitate if trained exclusively on it. The small variety of objects and materials causes some biases, e.g., green albedo has strong correlation with plants, so a green-albedo wall would be synthesized with a leafy texture if trained solely on \datasetIV.

\datasetHS~\cite{hypersim} is another synthetic photorealistic dataset comprising over 70,000 rendered images, with \normal, \albedo, and most importantly \irra data available. This dataset does not include other material information like roughness and metallicity, and sometimes bakes specular shading into the albedo. Fortunately, this is not common enough to preclude us from using the albedo data. While \datasetHS expands the scene appearance variety over \datasetIV, it is still not sufficient for highly realistic synthesis.

We complete these with two of our own datasets. The first is \datasetEM, a synthetic dataset generated similarly to \datasetIV by rendering synthetic scenes created by artists, randomly placing cameras along pre-recorded camera paths, and rendering 17,000 images of 85 indoor scenes. The main benefit of \datasetEM is that it provides us with roughness \roughness and metallicity \metallic, for which this dataset is currently the only reliable source.

To further enhance the training data and help our \xxtorgb model synthesize realistic images, we use 50,000 high-quality commercial interior scene images. These images come from photographs or high-quality renderings, with no additional channels  available. We therefore estimate normals, albedo, roughness, metallicity, and diffuse irradiance using our \rgbtoxx model. The combination of images and estimated channels form our \datasetSI dataset.

To better preserve the existing text-understanding abilities of the base diffusion model during fine-tuning for \xxtorgb, we precompute image captions for all images in all of the above datasets, using the BLIP-2 model \cite{Blip2}.

\begin{figure*}[t]
    \centering
    \includegraphics{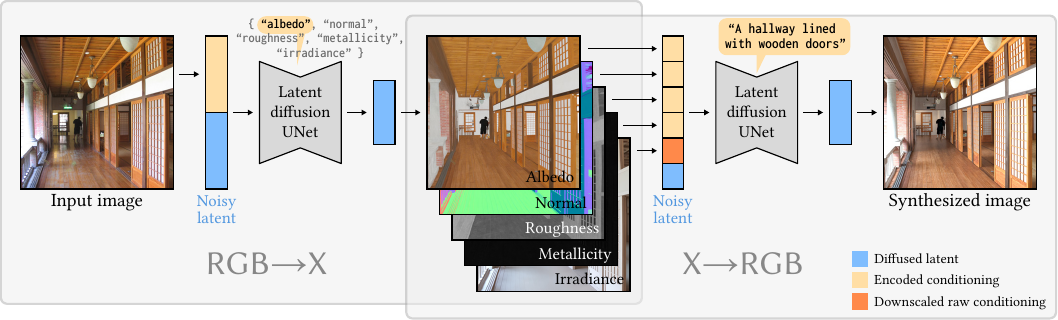}
    \caption{
        High-level overview of our two diffusion models. \textbf{Left:} The \rgbtoxx model takes the input image, encoded into latent space by the pre-trained encoder, concatenated with the diffusion latent. We repurpose the text prompt to a switch choosing the desired output channel; this allows training with datasets containing any subset of the supported channels. \textbf{Right:} The \xxtorgb model concatenates the input intrinsic channels, again encoded by the pre-trained encoder, with the diffusion latent. One exception is the irradiance (lighting) channel, which is downsampled to latent resolution rather than passed through the encoder. This model can accept usual text prompts. All input conditions to \xxtorgb are optional.
    }
    \label{fig:rgb2x}
\end{figure*}

\section{The \rgbtoxx model}
\label{sec:rgb2x}

In this section, we describe our \rgbtoxx model to estimate the intrinsic channels \xx from an input RGB image \image. The output contains all channels discussed in \Cref{sec:Channels}. Similarly to \citet{Kocsis2023}, we fine-tune a pre-trained text-to-image latent diffusion model, Stable Diffusion 2.1 \cite{StableDiffusion}. \Cref{fig:rgb2x} shows a high-level overview of our model.

Like Stable Diffusion, our \rgbtoxx model operates on a latent space with a pre-trained encoder $\mathcal{E}$ and decoder $\mathcal{D}$. During inference, it takes $\mathcal{E}(\image)$ as input condition and iteratively denoises a Gaussian-noise latent image $\mathbf{z}_T^{\xx}$ to produce the target latent image $\mathbf{z}_0^{\xx}$ encoding the intrinsic channels \xx. During training, we optimize the following loss function $L_\theta$ with v-prediction~\cite{salimans2022progressive} as we find v-prediction to give better results than noise $\epsilon$ prediction:
\begin{equation}
    \mathbf{v}_t^{\rgbtoxx} = \sqrt{\bar{\alpha}_t} \epsilon-\sqrt{1-\bar{\alpha}_t} \mathbf{z}_0^{\xx}  \; ,
\end{equation}
\begin{equation}
    L_\theta=\left\|\mathbf{v}_t^{\rgbtoxx} -\hat{\mathbf{v}}_\theta^{\rgbtoxx}\left(t, \mathbf{z}_t^{\xx} \mathcal{E}\left(\image\right), \tau\left(\mathbf{p}^{\xx}\right)\right)\right\|_2^2  \; .
\end{equation}
Here $t$ is the noise amount (``time step'') drawn uniformly during training, $\epsilon \sim \mathcal{N}(0,1)$, $\bar{\alpha}_t$ is a scalar function of $t$, $\hat{\mathbf{v}}_\theta^{\rgbtoxx}$ is our \rgbtoxx diffusion model with parameters $\theta$, $\mathbf{z}_0^{\xx}$ is the target latent, $\mathbf{z}_t^{\xx}$ is the noisy latent after adding noise $\epsilon$ at time step $t$ to $\mathbf{z}_0^{\xx}$. Further, $\mathbf{p}^{\xx}$ is the text prompt computed for $\image$, and $\tau$ is the CLIP text encoder~\cite{radford2021learning} that encodes the prompt into a text-embedding vector. This CLIP embedding is used as a specialized context for cross-attention layers of the model.

\paragraph{Image and intrinsic channel encoding}

As our model operates in latent space, we encode the input image as $\mathcal{E}\left(\image\right)$ and concatenate it to the noisy latent $\mathbf{z}_t^{\xx}$ as input to our  $\hat{\mathbf{v}}_\theta^{\rgbtoxx}$ model. We use the frozen encoder $\mathcal{E}$ from the original Stable Diffusion model, which we also found to work well for all our intrinsic images encoding.

\paragraph{Handling multiple output channels}

The output of the original Stable Diffusion model is a 4-channel latent image which can be decoded into a single \rgb image. As we aim to produce additional output channels (albedo \albedo, normal \normal, roughness \roughness, metallicity \metallic, and lighting \irra) we may expect that a larger latent vector may help better encode the information as done in previous work~\cite{Kocsis2023}. However, we find that extending the number of latent channels of the original model leads to lower-quality results. Indeed, adding more latent channels to the operating latent space of a diffusion model forces us to re-train both input and output convolutional layers from scratch. In a way, the model is suddenly ``shocked'' into a new domain, making the training more challenging.

We train our model with various datasets to increase variety, as described in \Cref{sec:Datasets}, but this comes with the additional issue of heterogeneous intrinsic channels, which is challenging for our approach that stacks all intrinsic channels into a larger latent. A straightforward approach would be to only include the loss for available maps in each training iteration. We however found this approach to perform poorly.

Our solution is to produce a single intrinsic channel at a time and repurpose the input text prompt (which does not serve any other purpose in our \rgbtoxx task) as a ``switch'' to control the diffusion-model output. Previous work~\cite{brooks2022instructpix2pix,Alchemist} shows that it is possible to use specially designed prompts as instructions to control a single diffusion model to perform different tasks explicitly. Inspired by this design, we use five fixed prompts acting as switches. More specifically, one unit of data is collated as $\{\mathbf{g}, \mathbf{p}^{\xx}, \image\}$ where $\mathbf{g} \in \{\normal, \albedo, \roughness, \metallic, \irra\}$ and $\mathbf{p}^{\xx} \in \{$ \texttt{``normal''}, \texttt{``albedo''},\texttt{``roughness''}, \texttt{``metallicity''},  \texttt{``irradiance''}$\}$ is set accordingly. We find that this approach performs similarly to fine-tuning separate models for each output modality in $\{\normal, \albedo, \roughness, \metallic, \irra\}$ while fine-tuning and storing only a single network's weights.

\section{The \xxtorgb model}
\label{sec:x2rgb}

We now describe our \xxtorgb model, performing realistic \rgb image synthesis from intrinsic channels \xx, illustrated in \Cref{fig:rgb2x}. Much like for \rgbtoxx, we fine-tune a diffusion model starting from Stable Diffusion 2.1 with several different considerations.

In the \xxtorgb case, we define the target latent variable as $\mathbf{z}_0^{RGB} = \mathcal{E}\left(\image\right)$, directly encoding the image \image. We provide the input condition \xx through concatenation of the encoded input intrinsic channels, adjusted to take various dataset properties in consideration as described below. When using all intrinsic images, the input latent vector is defined as
\begin{equation}
    \mathbf{z}_t^{\xx} = (\mathcal{E}({\normal}), \mathcal{E}({\albedo}), \mathcal{E}({\roughness}), \mathcal{E}({\metallic}), \mathcal{E}({\irra})).
\end{equation}
We train our \xxtorgb model by minimizing the loss function $L_\theta'$:
\begin{equation}
    \mathbf{v}_t^{\xxtorgb} = \sqrt{\bar{\alpha}_t} \epsilon-\sqrt{1-\bar{\alpha}_t} \mathbf{z}_0^{RGB}  \; ,
\end{equation}
\begin{equation}
    L_\theta'=\left\|\mathbf{v}_t^{\xxtorgb} -\hat{\mathbf{v}}_\theta^{\xxtorgb}\left(t, \mathbf{z}_t^{RGB}, \mathbf{z}_t^{\xx}, \tau\left(\mathbf{p}\right)\right)\right\|_2^2 \; .
\end{equation}
Here, $\hat{\mathbf{v}}_\theta^{\xxtorgb}$ is our \xxtorgb diffusion model with parameters $\theta$. We concatenate the noisy latent $\mathbf{z}_t^{RGB}$ and the conditioning latent $\mathbf{z}_t^{\xx}$ together before feeding it into $\hat{\mathbf{v}}_\theta^{\xxtorgb}$. The CLIP text embedding $\tau\left(\mathbf{p}\right)$ is used as the context for cross-attention layers of the model. For \xxtorgb the text embedding is used as an additional control as is usual in Diffusion Models.

While our \rgbtoxx model required a solution to output multiple modalities, the \xxtorgb model only requires changing the input layers to handle the additional conditional latent channels. Indeed, as in the original Stable Diffusion, the output remains a single \rgb image. During training, only the newly added weights of the input convolutional layer need to be trained from scratch to handle the additional conditions, which does not ``shock'' the model out of its normal denoising ability for $\mathbf{z}_t^{RGB}$.

\paragraph{Handling heterogeneous data}

Still, the problem of different intrinsic data channels missing from different datasets remains. To resolve this issue, we follow the observations made in previous works~\cite{ho2022classifier,huang2023composer} which propose to jointly train a conditional and unconditional diffusion model through condition channel dropout to improve sample quality and enable image generation with any subset of conditions. We therefore use an intrinsic channel drop-out strategy; with it, our conditioning latent $\mathbf{z}_t^{\xx}$ can be rewritten as:
\begin{equation}
    \mathcal{P}(x) \in \{\mathcal{E}(x), 0\}
\end{equation}
\begin{equation}
    \mathbf{z}_t^{\xx}=(\mathcal{P}(\normal) \mathbin,  \mathcal{P}(\albedo) \mathbin,  \mathcal{P}(\roughness) \mathbin,  \mathcal{P}(\metallic)\mathbin,  \mathcal{P}(\irra)).
\end{equation}
This approach lets us handle heterogeneous datasets during training, and choose which inputs to provide at inference; for example, providing no albedo or no lighting will result in the model generating plausible images, using its prior to compensate for the missing information (\Cref{fig:x2rgb-text}).

\paragraph{Low-resolution lighting}

Our \rgbtoxx model succeeds in estimating highly detailed lighting in the form of a diffuse irradiance image \irra, closely following high-resolution geometry and normals. While this could be beneficial for some applications, using these detailed lighting buffers for \xxtorgb presents an issue if we want to actually \emph{edit} the detailed normals and control the lighting using a coarser interpretation of \irra.
In other words, we would like to provide the lighting as a ``hint'' to the \xxtorgb model, rather than a precise per-pixel control. Instead of encoding the full resolution lighting \irra into the latent space as for other conditions, we simply downsample it into the same resolution as the latent. By doing so, we provide the \xxtorgb model with a coarser hint of lighting without pixel detail, while still achieving adherence to the overall lighting condition. This is important, e.g., when editing the normals in \Cref{fig:mat-replacement}.

\paragraph{Fine-tuning for inpainting.} To enable local editing applications (shown in \Cref{fig:teaser,fig:mat-replacement}) we fine-tune our \xxtorgb model to support inpainting by simply adding a masked image and mask channels to the model input. We downsample the mask to the latent-space resolution and concatenate it to the conditioning latent $\mathbf{z}_t^{\xx}$.

\section{Results}
\label{sec:results}

\paragraph{Note about picking results from a generative model}

Applying generative models to the \rgbtoxx and \xxtorgb problems means that the output is not unique but sampled from a distribution. While we could evaluate a number of samples and take their mean \cite{Kocsis2023}, we do not recommend this approach, as it can blur details that have been reasonably estimated within each sample. Instead, we pick a single sample to display in the paper, and provide more samples in the supplementary materials. Albedo, lighting, and normal samples are typically usable, while more attention is required for roughness and metallicity due to the lack of reliable training data and the inherent ambiguity of these properties.

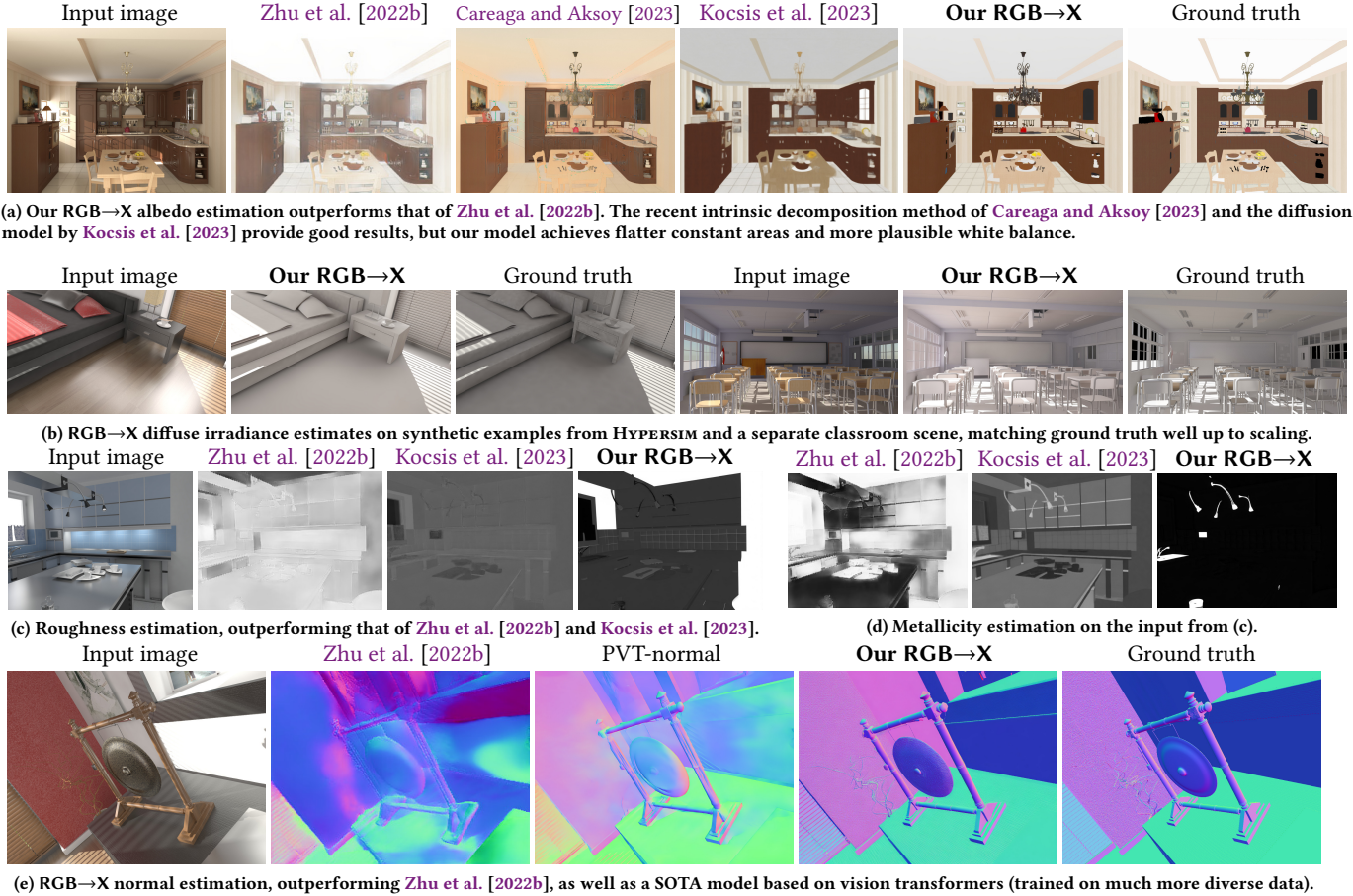
\begin{figure*}[t]
    \centering
    \input{figures/rgb2x_synthetic/rgb2x_synthetic}
    \caption{
        \textbf{Synthetic data comparison} of our \rgbtoxx model against previous methods \cite{InteriorVerse,OrdinalShading} and a known ground truth. All input images and ground truths are from \datasetHS, except for the classroom scene (c).
    }
    \label{fig:rgb2x_synthetic}
\end{figure*}

\begin{figure*}[t]
    \centering
    \input{figures/rgb2x_real/rgb2x_real}
    \caption{
        \textbf{Real-data comparison} of our \rgbtoxx model to previous methods.
    }
    \label{fig:rgb2x_real}
\end{figure*}

\subsection{\rgbtoxx on synthetic and real inputs}

\Cref{fig:rgb2x_synthetic,fig:rgb2x_real} show our results on intrinsic channel estimation for synthetic and real examples. None of the synthetic input examples were part of the training data. Please see the supplementary materials for many more results.

\paragraph{Albedo}

We compare albedo estimation to previous work in \Cref{fig:rgb2x_synthetic}(a) for synthetic and \Cref{fig:rgb2x_real}(a, b) for real inputs. Generally, we find that our model is best at removing reflections, highlights, shadows, and color cast from the inputs, while providing the flattest estimates for albedo regions that should indeed be constant. The method of \citet{InteriorVerse} performs worse on both synthetic and real inputs, hinting at the limitations of non-generative models, nor have designs incorporating special knowledge about the albedo estimation problem. The recent intrinsic decomposition method of \citet{OrdinalShading} provides good results, but our model achieves flatter constant areas and a more plausible white balance. Although they also show impressive results, the same is true for the diffusion model by \citet{Kocsis2023}. For example, on the bedroom photo in \Cref{fig:rgb2x_real}(a, top row), our model is the only one correctly predicting that all bed-linen pixels should have identical white albedo. The challenging real image in \Cref{fig:rgb2x_real}(b) also results in a very clean albedo estimate that outperforms other methods, though our model removes some wear from the wooden floor, possibly due to training on synthetic materials without wear.

\paragraph{Diffuse irradiance (lighting)}

In \Cref{fig:rgb2x_synthetic}(b), we see that our model produces diffuse irradiance estimates that closely match ground truth on synthetic data, even on inputs with intricate shadow patterns, and with very little to no leaking of material properties into the estimation. The color in the irradiance is also plausibly shifted away from pure white under colored lighting. Our estimates are also realistic and plausible on real inputs, such as in \Cref{fig:rgb2x_real}(b). \citet{OrdinalShading} do not provide irradiance directly, so we divide the original image by their predicted albedo and use the resulting approximate irradiance as a baseline.

\paragraph{Metallicity and roughness}

As shown in \Cref{fig:rgb2x_synthetic}(c,d) and \Cref{fig:rgb2x_real}(c,d), our \rgbtoxx model generates much more plausible roughness and metallicity for a given input image than the previous publicly available state of the art \cite{InteriorVerse, Kocsis2023}. These material properties are challenging to recover accurately, for two reasons. First, the amount of reliable training data for them is the lowest. Second, they only impact surface reflection significantly if lit by appropriate high-frequency illumination; otherwise the model has to revert to prior knowledge, estimating what the object could be and whether such objects tend to be rough or metallic. These issues translate to a higher sampling variance of our model, and a lower yield of ``good'' samples. We show this variability in our estimations in the supplemental materials.

\paragraph{Normals}

In synthetic tests (\Cref{fig:rgb2x_synthetic}(e)) as well as real ones (\Cref{fig:rgb2x_real}(e)), we show that our model estimates normals plausibly, including high-frequency geometry, while correctly predicting flat normals for flat surfaces even if they have texture or high-frequency lighting. Our results outperform those of \citet{InteriorVerse} and slightly improve on the state-of-the-art PVT-normal. While we observe that our model normal estimation generalizes reasonably well (see more examples in supplementary materials), we do not claim general improvement in this space, as PVT-normal is specifically designed to work well across general images. We provide this comparison for the sake of completeness.

\paragraph{Quantitative comparisons}

For albedo, normal, roughness and metallicity estimation, we compare to the corresponding previous methods in \Cref{tab:quantitative}. We find that our \rgbtoxx has the best PSNR and LPIPS values on all channels, with the exception of irradiance for which we do not have existing methods to compare to.

\subsection{\xxtorgb model results}

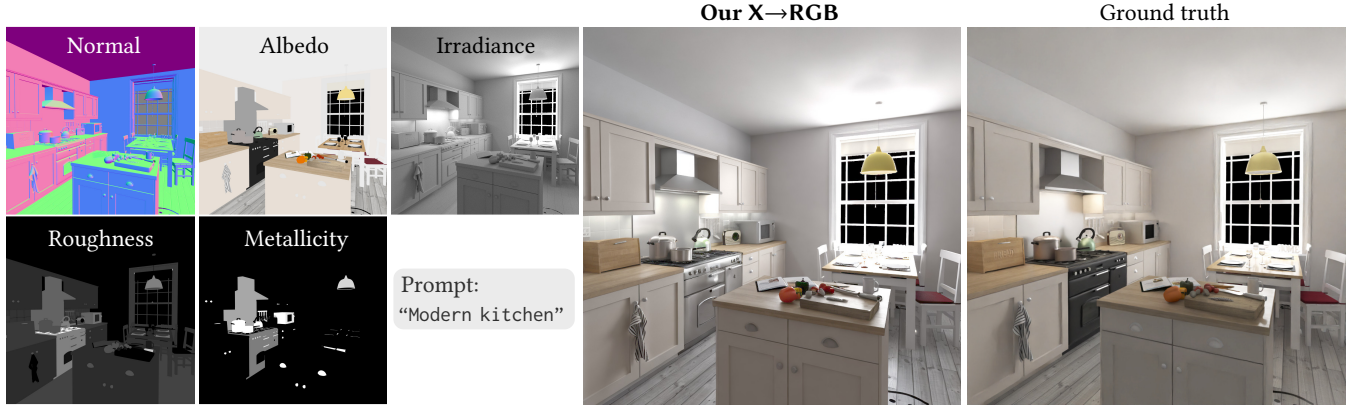
\begin{figure*}[t]
    \centering
    \input{figures/x2rgb/x2rgb1}
    \caption{
        Our \xxtorgb result on the synthetic kitchen scene~\cite{kitchen} which is not part of our training data. We rendered all intrinsic channels, shown on the left, and fed them into the model, along with a text prompt. The result matches the path-traced reference well. There are some differences, e.g., \xxtorgb makes the stove brighter than the requested albedo, likely because dark metallic materials are rare in the training data.
    }
    \label{fig:x2rgb-pathtrace}
    \vspace{.5cm}
\end{figure*}

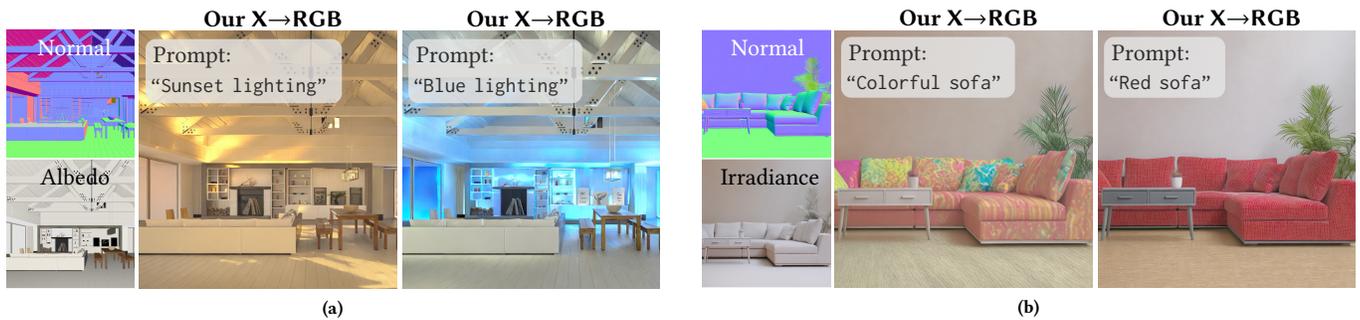
\begin{figure*}[t]
    \centering
    \input{figures/x2rgb/x2rgb2}
    \caption{
        \xxtorgb synthesis given normal and albedo channels only, demonstrating lighting and color-control use of text prompts. (a)~Starting from normal and albedo only, we show that the lighting can be controlled by text prompts to some extent. (b)~Starting from normal and albedo only, we similarly show the color of objects can be controlled by text prompts to some extent.
    }
    \label{fig:x2rgb-text}
    \vspace{.5cm}
\end{figure*}

\begin{figure*}[t]
    \centering
    \input{figures/mat-replacement/mat-replacement}
    \caption{
        \rgbtoxx and \xxtorgb models used in combination for material replacement. We show three edit examples. \textbf{Top left:} We change both normal and albedo, resulting in a fuzzier, bumpier red couch. \textbf{Top right:} We edit the right wall albedo of the Cornell box to blue and show that the colour bleeding in the rightmost box is  correctly updated. \textbf{Bottom:} We first change the normal to introduce wood planks geometry instead of the original carpet, and then also add wood albedo to edit the floor appearance.
    }
    \label{fig:mat-replacement}
\end{figure*}
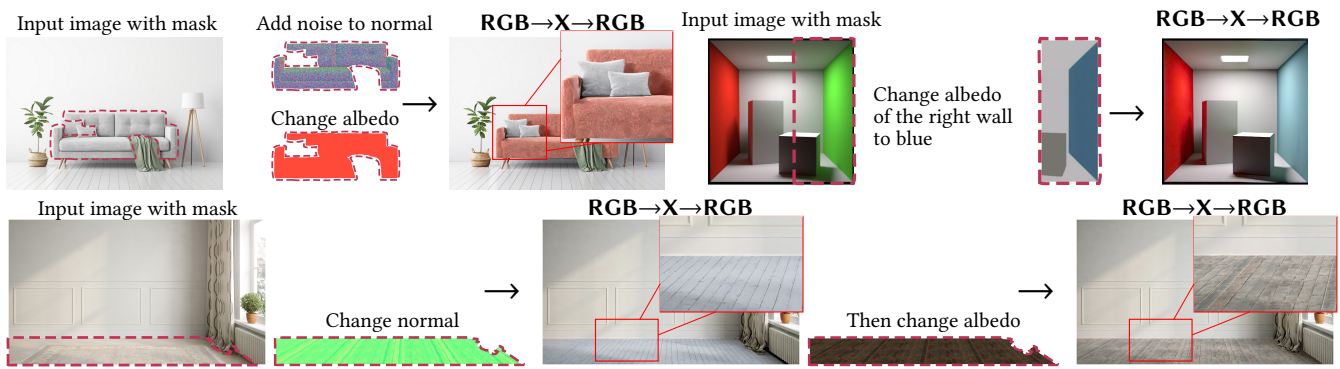

\paragraph{Comparison to path tracing reference}

In \Cref{fig:x2rgb-pathtrace}, we validate that our \xxtorgb model produces results closely matching traditional Monte Carlo path tracing, as long as the input channels \xx are not far from the training distribution of synthetic interiors. Here, we use a common synthetic kitchen scene, not part of our training data. We use all intrinsic channels (shown on the left) and feed them into our model, along with a text prompt. The result matches the path-traced reference well in terms of material appearance and global illumination. Differences can also be noted: for example, the stove has a dark metallic material in the input channels, which is rare in the training data. Our model generates a brighter aluminum material, matching the metallicity instead of the albedo channel.

\paragraph{Subsets of input channels and text prompts.}

\Cref{fig:x2rgb-text} demonstrates the ability of our \xxtorgb model to generate plausible images by specifying only a subset of the appearance properties as input. Furthermore, text prompts can be used for additional control. Here, we control the lighting (a) or object colors (b).  Generally, text control works well only when there are only few objects (e.g., one sofa and a few pillows). It is hard to control the color of a specific object by text, but this issue is a common challenge for all diffusion models.

\begin{table}[t]
    \centering
    \caption{
        Numerical evaluation of \rgbtoxx against existing methods. Albedo, normal, and irradiance evaluations are conducted on the \datasetHS test set. Roughness and metallicity are evaluated on the \datasetEM test set.
    }
    \scalebox{0.82}{%
        \setlength{\tabcolsep}{2pt}%
        \begin{tabularx}{1.21\linewidth}{ccccccc}
            \toprule
            \multicolumn{1}{c}{\textbf{Method}}        & \multicolumn{2}{c}{\textbf{Albedo}}    & \multicolumn{2}{c}{\textbf{Method}}            & \multicolumn{2}{c}{\textbf{Normal}}                                                                                                                                                          \\
            \cmidrule(lr){2-3}
            \cmidrule(lr){6-7}
            \multicolumn{1}{c}{}                       & \textbf{PSNR}$\uparrow$                & \multicolumn{1}{c}{\textbf{LPIPS}$\downarrow$} & \multicolumn{2}{c}{}                      & \textbf{PSNR}$\uparrow$                        & \multicolumn{1}{c}{\textbf{LPIPS}$\downarrow$}                                                  \\
            \cmidrule[0.4pt](lr){1-3}
            \cmidrule[0.4pt](lr){4-7}
            \multicolumn{1}{c}{\textbf{Our \rgbtoxx}}  & \textbf{17.4}                          & \multicolumn{1}{c}{\textbf{0.18}}              & \multicolumn{2}{c}{\textbf{Our \rgbtoxx}} & \textbf{19.8}                                  & \multicolumn{1}{c}{\textbf{0.18}}                                                               \\
            \multicolumn{1}{c}{\citet{InteriorVerse}}  & 11.7                                   & \multicolumn{1}{c}{0.54}                       & \multicolumn{2}{c}{\citet{InteriorVerse}} & 16.5                                           & \multicolumn{1}{c}{0.45}                                                                        \\
            \multicolumn{1}{c}{\citet{OrdinalShading}} & 13.5                                   & \multicolumn{1}{c}{0.34}                       & \multicolumn{2}{c}{PVT-normal}            & 18.8                                           & \multicolumn{1}{c}{0.30}                                                                        \\
            \multicolumn{1}{c}{\citet{Kocsis2023}}     & 12.1                                   & \multicolumn{1}{c}{0.41}                       & \multicolumn{2}{c}{}                      &                                                & \multicolumn{1}{c}{}                                                                            \\
            \midrule
            \multicolumn{1}{c}{\textbf{Method}}        & \multicolumn{2}{c}{\textbf{Roughness}} & \multicolumn{2}{c}{\textbf{Metallicity}}       & \multicolumn{2}{c}{\textbf{Irradiance}}                                                                                                                                                      \\
            \cmidrule(lr){2-3}
            \cmidrule(lr){4-5}
            \cmidrule(lr){6-7}
            \multicolumn{1}{c}{}                       & \textbf{PSNR}$\uparrow$                & \multicolumn{1}{c}{\textbf{LPIPS}$\downarrow$} & \textbf{PSNR}$\uparrow$                   & \multicolumn{1}{c}{\textbf{LPIPS}$\downarrow$} & \textbf{PSNR}$\uparrow$                        & \multicolumn{1}{c}{\textbf{LPIPS}$\downarrow$} \\
            \cmidrule[0.4pt](lr){1-7}
            \multicolumn{1}{c}{\textbf{Our \rgbtoxx}}  & \textbf{11.2}                          & \multicolumn{1}{c}{\textbf{0.52}}              & \textbf{12.1}                             & \multicolumn{1}{c}{\textbf{0.44}}              & \textbf{14.1}                                  & \multicolumn{1}{c}{\textbf{0.22}}              \\
            \multicolumn{1}{c}{\citet{InteriorVerse}}  & 4.4                                    & \multicolumn{1}{c}{0.77}                       & 2.22                                      & \multicolumn{1}{c}{0.82}                       & N/A                                            & N/A                                            \\
            \multicolumn{1}{c}{\citet{Kocsis2023}}     & 10.3                                   & \multicolumn{1}{c}{0.57}                       & 8.63                                      & \multicolumn{1}{c}{0.75}                       & N/A                                            & N/A                                            \\
            \bottomrule
        \end{tabularx}
    }%
    \label{tab:quantitative}%
    \vspace{-3mm}
\end{table}

\subsection{Applications}

\paragraph{Material replacement}

In the top left example of \Cref{fig:mat-replacement}, we edit the normal and albedo of the sofa (estimated by \rgbtoxx), and re-synthesize the image with our inpainting \xxtorgb model, resulting in a fuzzier, bumpier red couch. On the top right, we apply intrinsic estimation to the classic Cornell box image and edit the right wall albedo to blue. We observe that the color bleeding in the rightmost box is correctly updated. The inpainting mask here includes a larger region, allowing for the color-bleeding correction. In the bottom example, we  change the normal and albedo of the original room to edit the floor appearance to a wood floor.

\paragraph{Object insertion}

In \Cref{fig:teaser}(c), we use our framework to insert new synthetic objects into an RGB image. We render the intrinsic channels of the new objects and composite them into the estimated channels. We use our inpainting \xxtorgb model with rectangular masks to produce a composite with correct lighting and shadows, which we finally blend with the original image using a tighter mask. The statue and coffee cart integrate well into the scene.

\section{Conclusion}
\label{sec:conclusion}

In this paper, we explored a unified diffusion framework for intrinsic channel estimation from images (termed \rgbtoxx) and synthesizing realistic images from such channels (\xxtorgb). Our intrinsic information \xx contains albedo, normals, roughness, metallicity, and lighting (irradiance). Our \rgbtoxx model matches or exceeds the quality of previous methods, which are specialized to subsets of our intrinsic channels. Our \xxtorgb model is capable of synthesizing realistic final images, even if we specify only certain appearance properties that should be followed, and give freedom to the model to generate the rest. We show combining both models enables applications such as material editing and object insertion. We believe our work is the first step towards unified diffusion frameworks capable of both image decomposition and rendering, which can bring benefits for a wide range of downstream editing tasks.

\paragraph{Acknowledgements} Ling-Qi Yan is supported by funds from Adobe, Intel, Linctex, Meta and XVerse.


\bibliographystyle{ACM-Reference-Format}
\bibliography{references}


\appendix

\section{Note about traditional rendering for \xxtorgb}

Note that our \xxtorgb problem cannot be easily solved using traditional rendering. Intrinsic channels (even if all are present) do not contain sufficient information to render a realistic image using traditional techniques, which require full 3D geometry (not just normals) and explicit light/material definitions, including for parts of the scene that are not directly seen by the camera. Screen-space ray tracing / occlusion methods yield only rough approximations, nowhere near the capabilities of our \xxtorgb model.

Furthermore, when the given intrinsic channels are imperfect or partial, traditional rendering is completely out of the question, but our generative model can still produce reasonable results, possibly controlled with appropriate text prompts.

\section{Implementation}

\paragraph{Training details}

We finetune pre-trained Stable Diffusion 2.1 for both \rgbtoxx and \xxtorgb models. Both models are trained on the \datasetIV, \datasetHS, and \datasetEM datasets. We train the \xxtorgb model additionally on the \datasetSI dataset, which is constructed from RGB images using our \rgbtoxx model. Both models are trained with a batch size of $256$ and using the AdamW optimizer~\cite{loshchilov2017decoupled} with a learning rate of $1\mathrm{e}{-5}$. We use a random crop of $512 \times 512$ for training and avoid using a random horizontal flip since it disrupts the camera-space normals. The training of each model takes around 100 hours and the fine-tuning to enable inpainting takes around 20 hours on 8 A100 GPUs.

\paragraph{Inference details}

We use the DDIM sampler~\cite{song2020denoising} with 50 steps for all of our results. We follow the suggestions proposed by ~\citet{lin2024common} to avoid over-exposure of the generated images.  Despite training at $512 \times 512$ crops, we can run test images at larger resolutions (e.g. 1080p). For comparison, the method of ~\citet{InteriorVerse} runs on a resolution of $320 \times 240$ (the default resolution noted in their code).

\paragraph{Classifier-free guidance}

Classifier-free guidance (CFG) is commonly used in diffusion models to improve text prompt alignment. For \xxtorgb, we use classifier-free guidance (CFG) similar to  InstructPix2Pix~\cite{brooks2022instructpix2pix}; for \rgbtoxx, we do not use CFG, since we found that it impairs the quality of the \rgbtoxx model and does not provide any benefits, since we do not use text prompts for this model in the usual sense.

\section{Discussion and Limitations}
\label{sec:discussion}

As our model relies on synthetic dataset for training, different challenges arise. In particular we find the various datasets to present their respective flaws. For example \datasetHS sometimes bakes shading into its albedo, and \datasetIV provides metallic and roughness parameters that are not reliable. Further, the available renderings tend to be noisy and the data often presents aliasing artifacts. While we try to take this into account with our heterogeneous training approach, higher quality, consistent datasets would be beneficial for improved quality. As scene materials datasets tend to be focused on interior scenes, they encode significant bias such as green color being most often for plants, or the fact that e.g. wooden curtains are not common, potentially limiting the editing freedom. Finally, the dropout rate and the probability of picking data from different sources during training can affect the resulting models, guiding them towards preferring some kinds of inputs over others.

In most of these cases, larger, more diverse data would be beneficial in reducing these issues.

In recent generative models a trade-off exists between diversity and adherence to input condition, which can be controlled with Classifier-Free Guidance (CFG). CFG however does not work in our context, and it would be interesting to define such a control mechanism.

As our networks are trained on 512 $\times$ 512 resolutions, we can process larger images, but find that quality degrades beyond a 2K resolution; perhaps a coarse-to-fine approach could be used to handle even larger images.

\begin{figure*}[t]
    \centering
    \input{supp/failure-cases/failure-cases}
    \caption{Failure cases of our \xxtorgb model on the \textsc{MIT Indoor Scene Recognition} real photo dataset~\cite{quattoni2009recognizing}.}
    \label{fig:failure-cases}
\end{figure*}

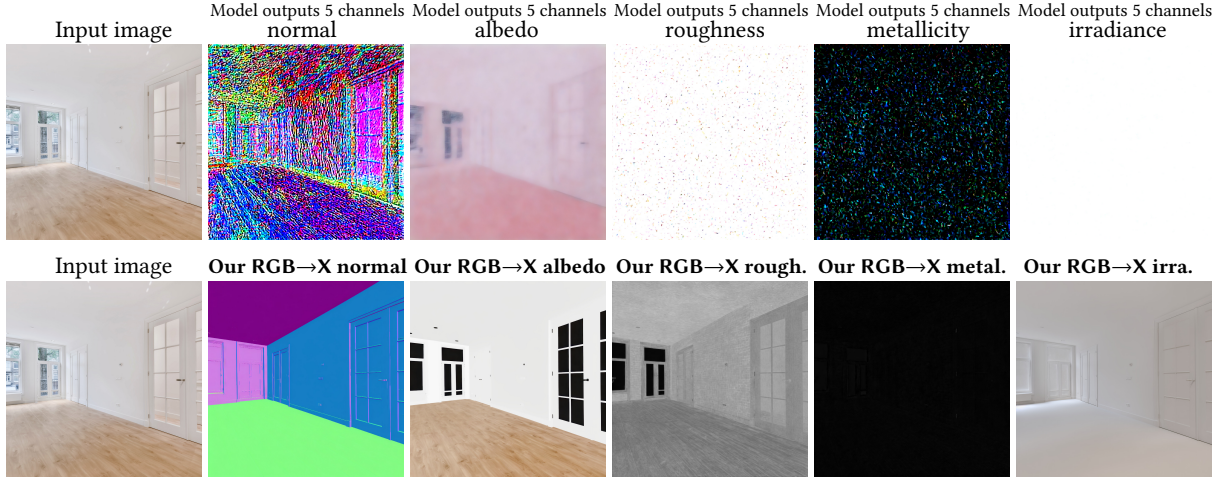
\begin{figure*}[t]
    \centering
    \input{supp/ablation-rgb2x/ablation-rgb2x-5}
    \caption{
        The model in the top row outputs a larger latent vector for 5 channels (normal, albedo, roughness, metallicity, and irradiance) at once. However, we find this model is hard to train and performs poorly even after 100 epochs of training.
    }
    \label{fig:ablation-rgb2x-5}
\end{figure*}

\begin{figure*}[t]
    \centering
    \input{supp/ablation-rgb2x/ablation-rgb2x-4}
    \caption{
        The model in the top row outputs a larger latent vector for 4 channels (albedo, roughness, metallicity, and irradiance) at once; compared to the model outputs 5 channels in Figure~\ref{fig:ablation-rgb2x-5} (top), this model starts to generate reasonable results in albedo after 100 epochs of training. Still, it performs poorly on the other channels.
    }
    \label{fig:ablation-rgb2x-4}
\end{figure*}
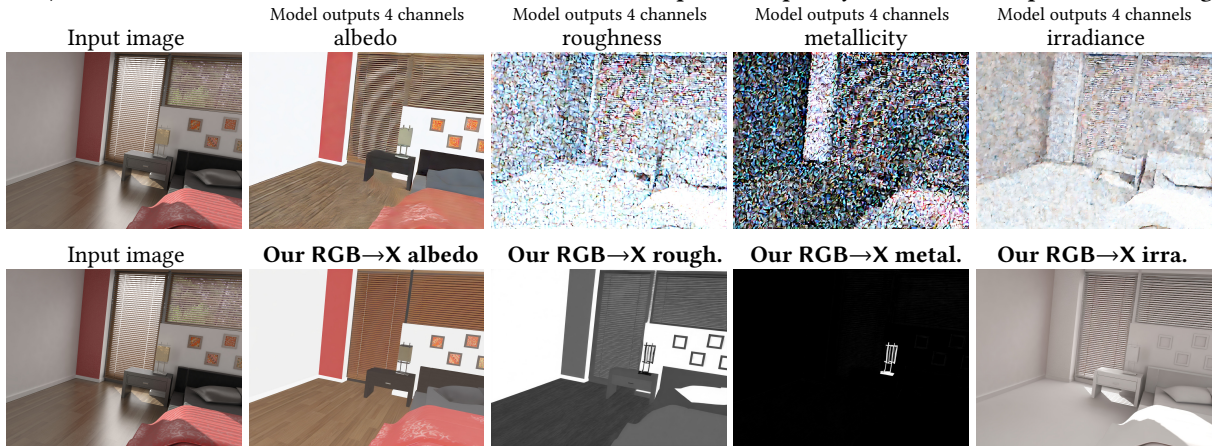

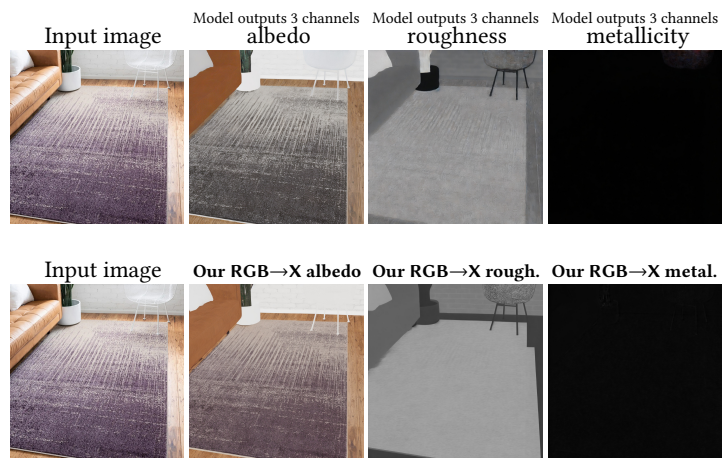
\begin{figure*}[t]
    \centering
    \input{supp/ablation-rgb2x/ablation-rgb2x-3}
    \caption{
        The model in the top row outputs a larger latent vector for 3 channels (albedo, roughness, and metallicity). Compared to models in Figure~\ref{fig:ablation-rgb2x-5} and Figure~\ref{fig:ablation-rgb2x-4}, this model can produce flat and clean albedo channel and decent roughness and metallicity channel. However, compared to our model trained with the same number of epochs, this model produce a color shift in albedo channel, and distortions in roughness channel.
    }
    \label{fig:ablation-rgb2x-3}
\end{figure*}

\end{document}

%% file: macros.tex
\newcommand{\xx}{\textsf{X}\xspace}
\newcommand{\rgb}{\textsf{RGB}\xspace}
\newcommand{\rgbtoxx}{\textsf{RGB$\rightarrow$X}\xspace}
\newcommand{\xxtorgb}{\textsf{X$\rightarrow$RGB}\xspace}
\newcommand{\rgbtooxx}{\textsf{RGB$\leftrightarrow$X}\xspace}
\newcommand{\rgbtoxxtorgb}{\textsf{RGB$\rightarrow$X$\rightarrow$RGB}\xspace}

\newcommand{\datasetIV}{\textsc{InteriorVerse}\xspace}
\newcommand{\datasetHS}{\textsc{Hypersim}\xspace}
\newcommand{\datasetEM}{\textsc{Evermotion}\xspace}
\newcommand{\datasetSI}{\textsc{ImageDecomp}\xspace}

\newcommand{\albedo}{\ensuremath{\mathbf{a}}\xspace}
\newcommand{\normal}{\ensuremath{\mathbf{n}}\xspace}
\newcommand{\roughness}{\ensuremath{\mathbf{r}}\xspace}
\newcommand{\metallic}{\ensuremath{\mathbf{m}}\xspace}
\newcommand{\irra}{\ensuremath{\mathbf{E}}\xspace}
\newcommand{\image}{\ensuremath{\mathbf{I}}\xspace}

%% file: figures/rgb2x_synthetic/rgb2x_synthetic.tex
\begin{minipage}{\linewidth}
    \begin{minipage}{1\linewidth}
        \centering
        \subfloat[Our \rgbtoxx albedo estimation outperforms that of \citet{InteriorVerse}. The recent intrinsic decomposition method of \citet{OrdinalShading} and the diffusion model by \citet{Kocsis2023} provide good results, but our model achieves flatter constant areas and more plausible white balance.]{
            \begin{overpic}[width=0.165\linewidth]{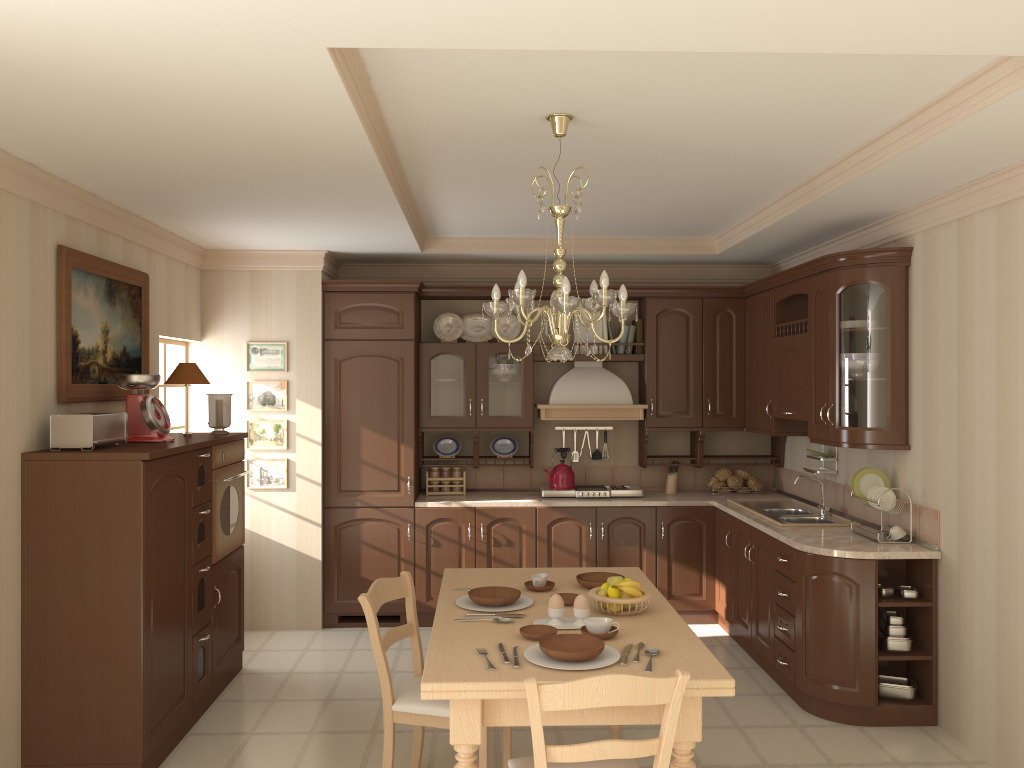}
                \put(25,79){Input image}
            \end{overpic}
            \hfill
            \begin{overpic}[width=0.165\linewidth]{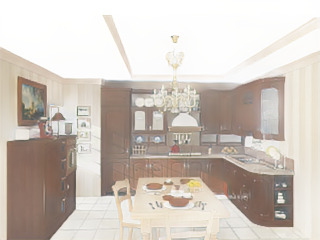}
                \put(13,79){\citet{InteriorVerse}}
            \end{overpic}
            \hfill
            \begin{overpic}[width=0.165\linewidth]{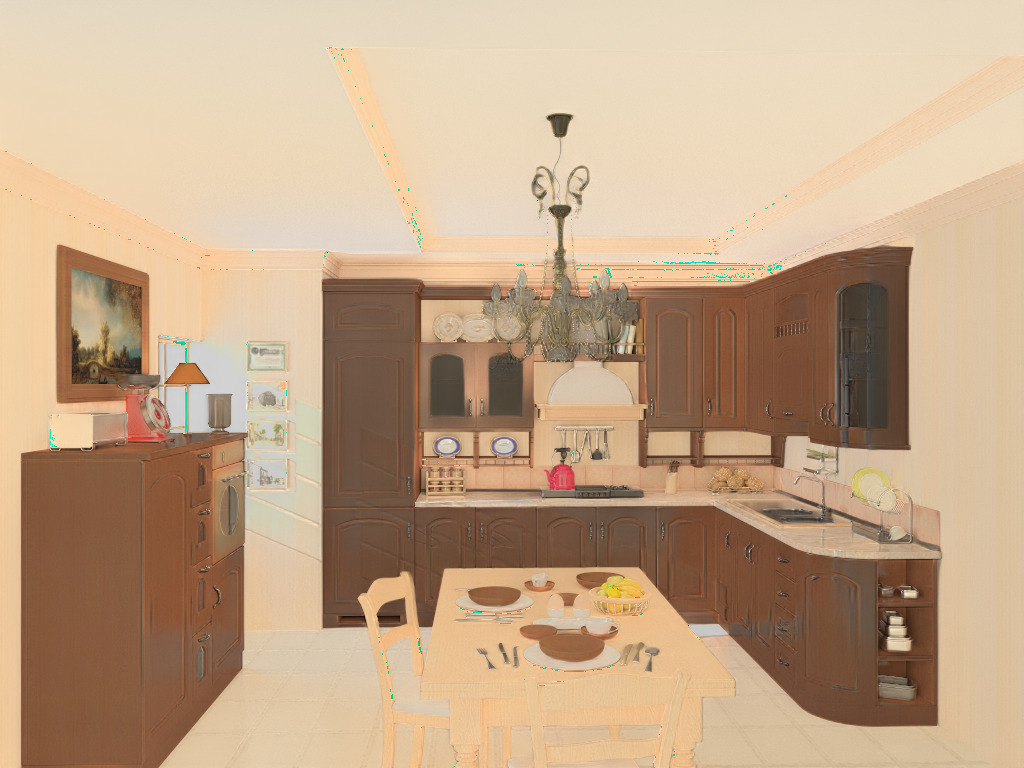}
                \put(0,79){\small\citet{OrdinalShading}}
            \end{overpic}
            \hfill
            \begin{overpic}[width=0.165\linewidth]{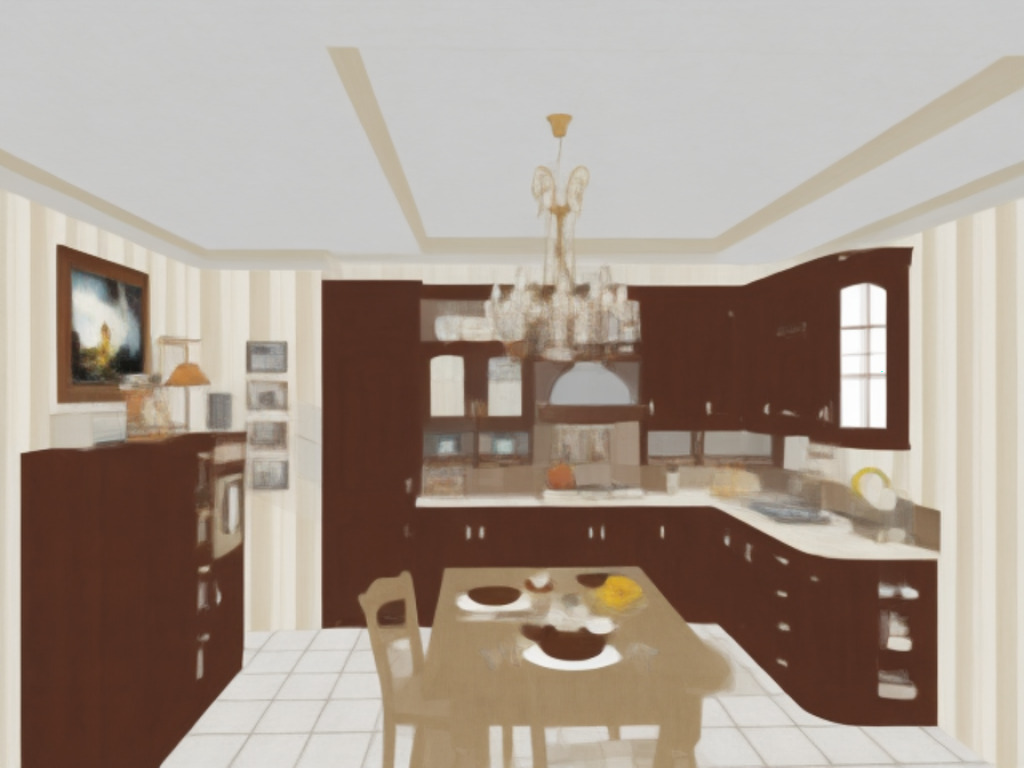}
                \put(9,79){\citet{Kocsis2023}}
            \end{overpic}
            \hfill
            \begin{overpic}[width=0.165\linewidth]{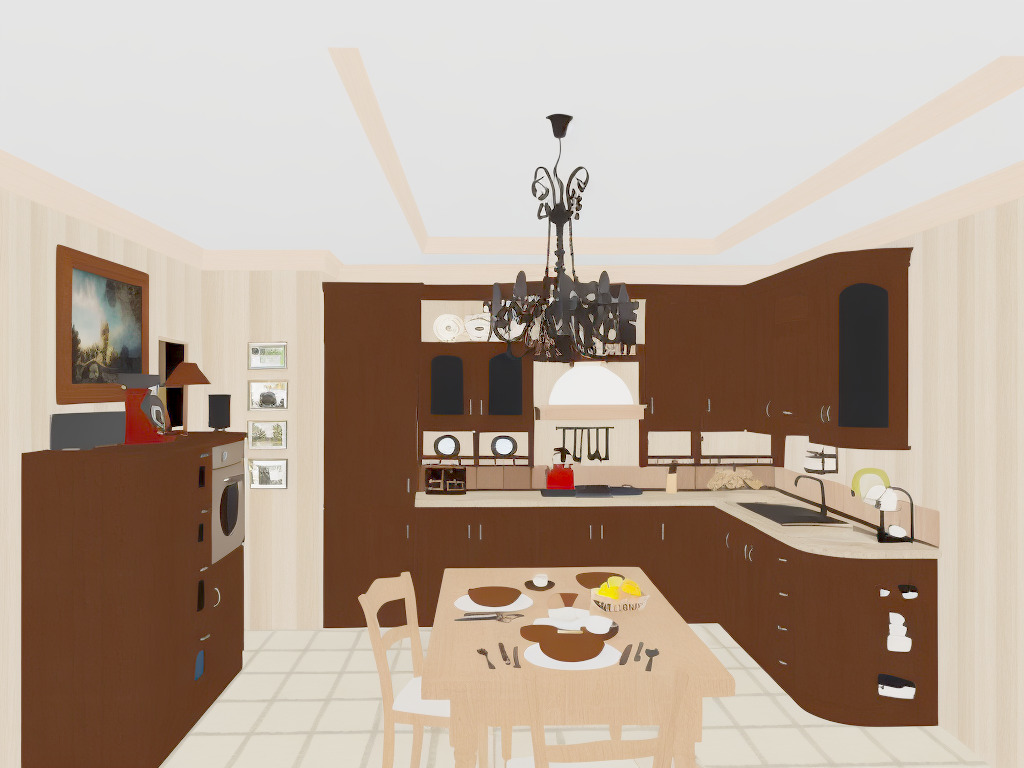}
                \put(19,79){\textbf{Our \rgbtoxx}}
            \end{overpic}
            \hfill
            \begin{overpic}[width=0.165\linewidth]{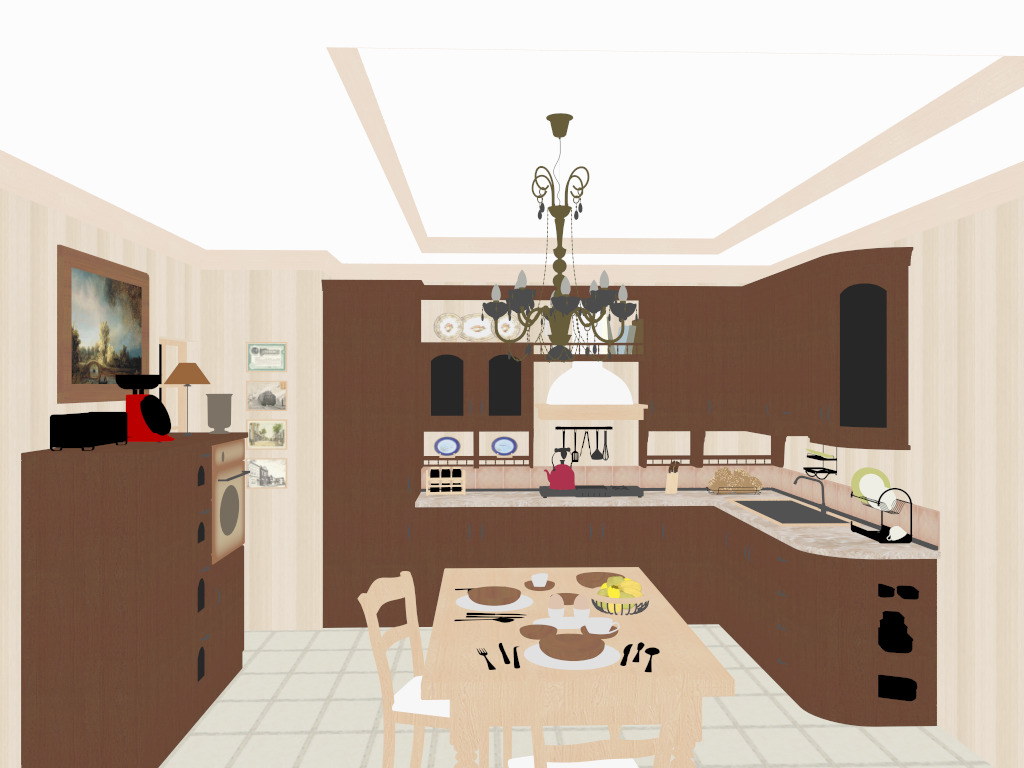}
                \put(20,79){Ground truth}
            \end{overpic}
        }
    \end{minipage}\par\smallskip
    \begin{minipage}{1\linewidth}
        \begin{minipage}{\linewidth}
        \end{minipage}\par\smallskip
        \centering
        \subfloat[\rgbtoxx diffuse irradiance estimates on synthetic examples from \datasetHS and a separate classroom scene, matching ground truth well up to scaling.]{
            \begin{overpic}[width=0.165\linewidth]{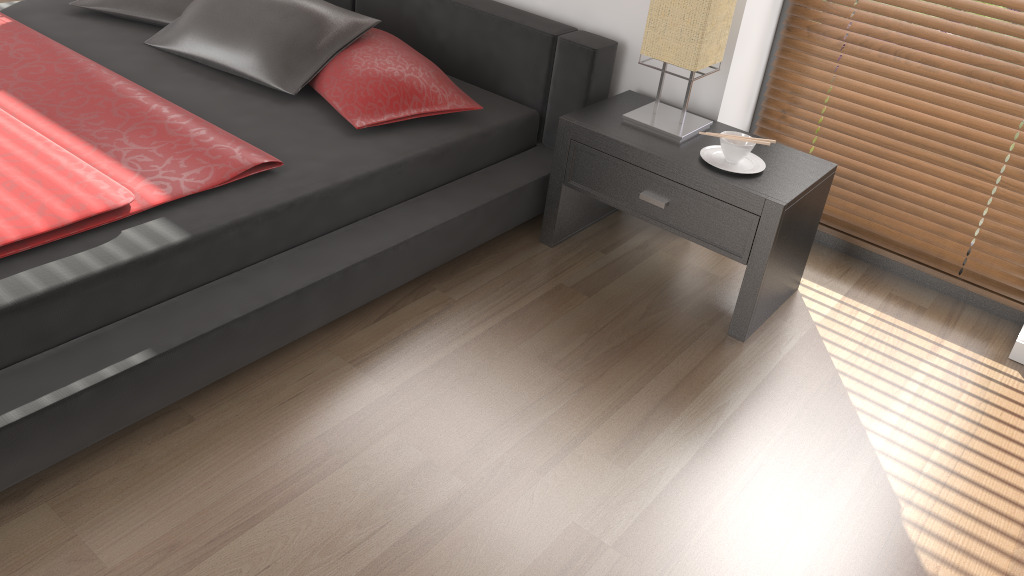}
                \put(25,60){Input image}
            \end{overpic}
            \hfill
            \begin{overpic}[width=0.165\linewidth]{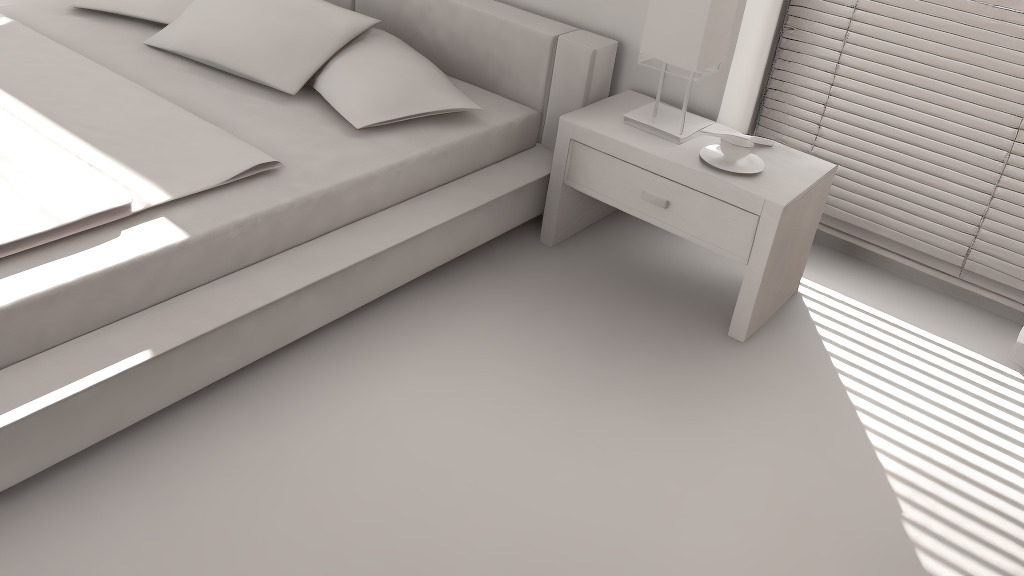}
                \put(17,60){\textbf{Our \rgbtoxx}}
            \end{overpic}
            \hfill
            \begin{overpic}[width=0.165\linewidth]{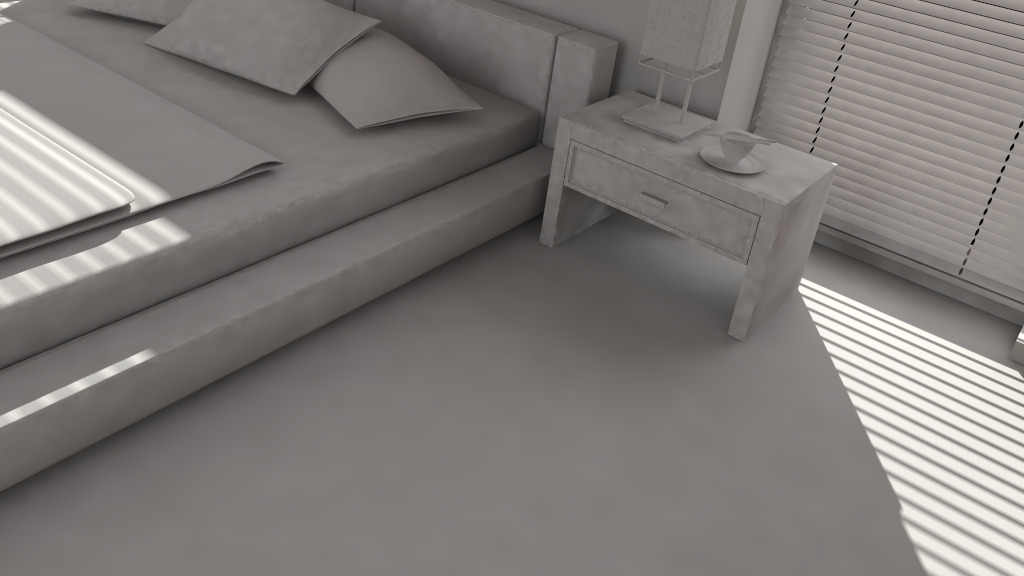}
                \put(22,60){Ground truth}
            \end{overpic}
            \hfill
            \begin{overpic}[width=0.165\linewidth]{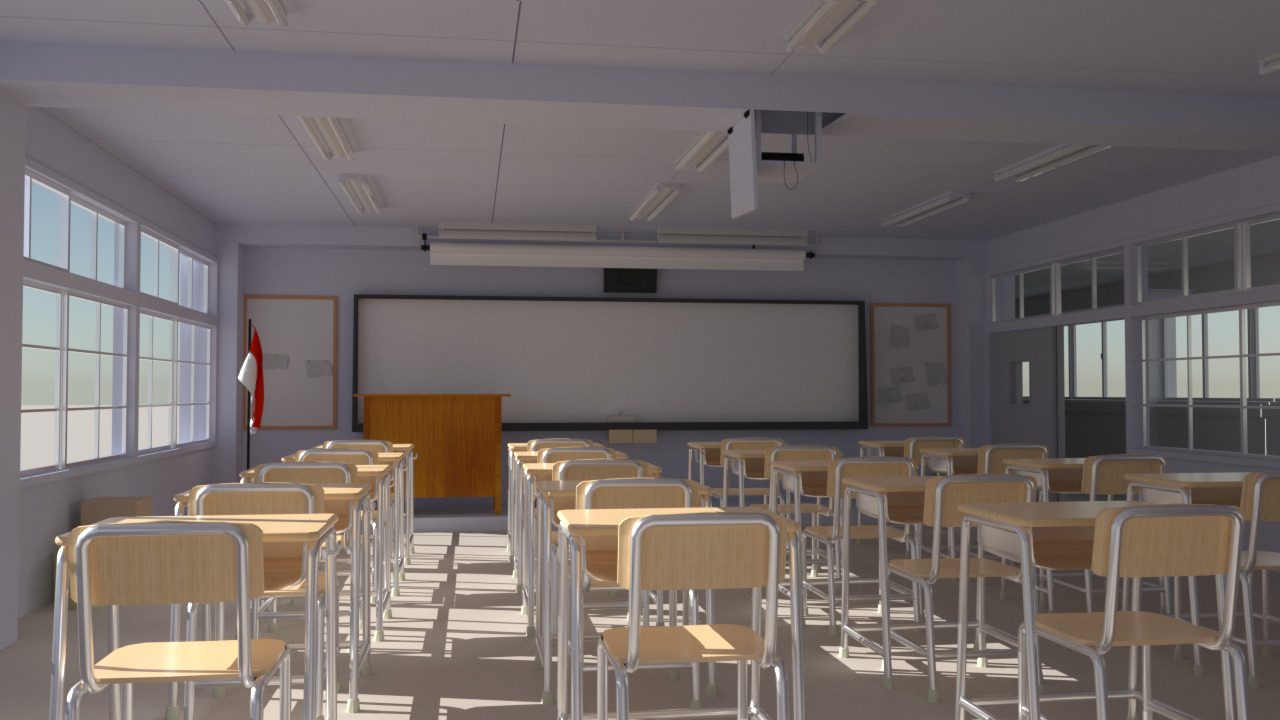}
                \put(25,60){Input image}
            \end{overpic}
            \hfill
            \begin{overpic}[width=0.165\linewidth]{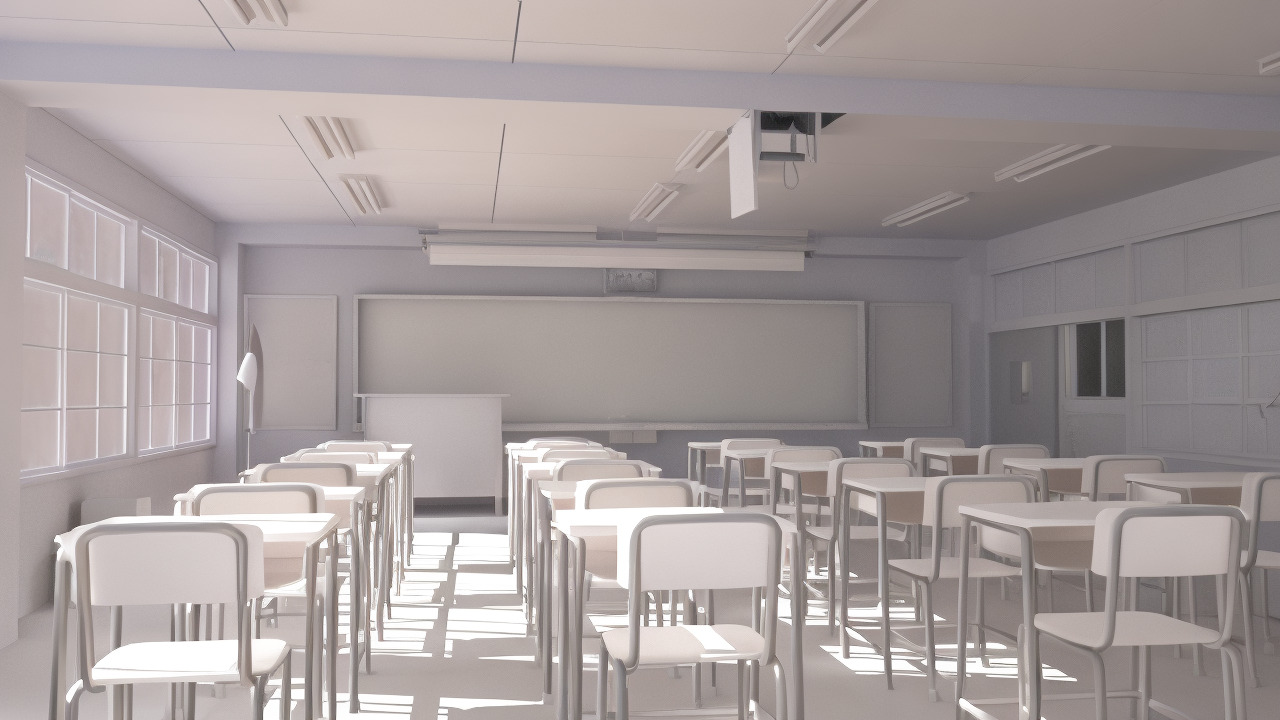}
                \put(17,60){\textbf{Our \rgbtoxx}}
            \end{overpic}
            \hfill
            \begin{overpic}[width=0.165\linewidth]{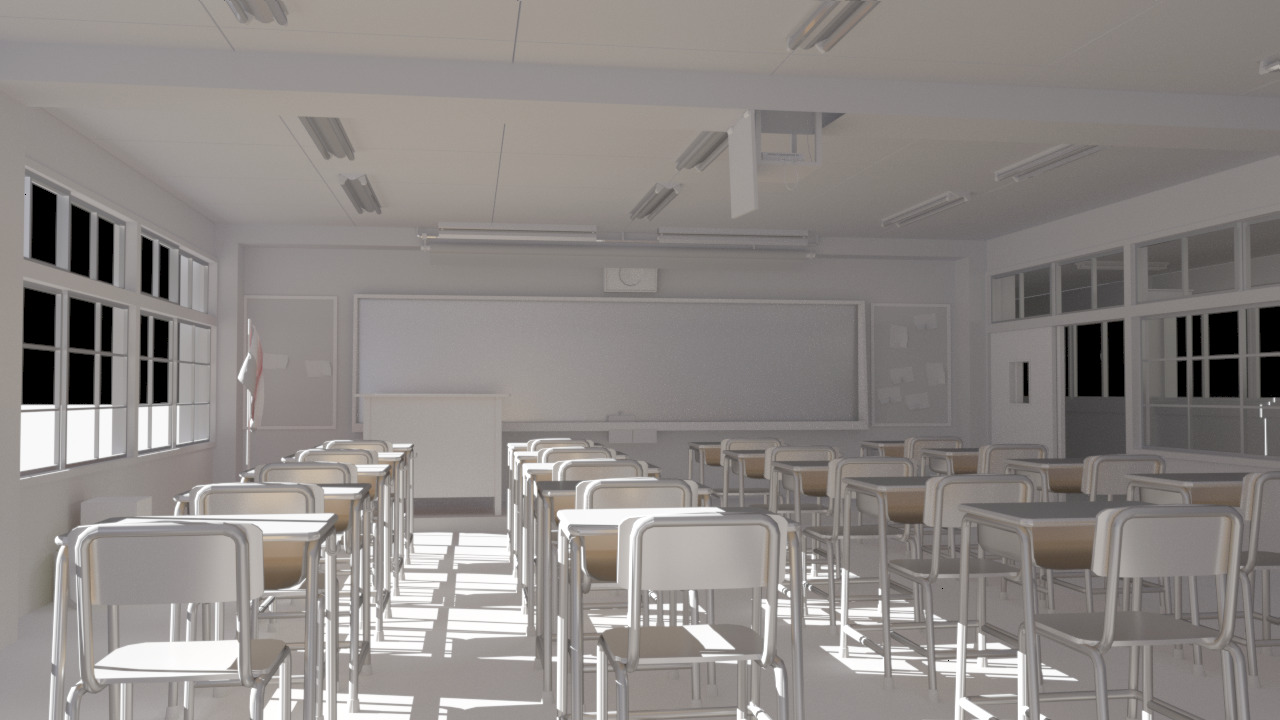}
                \put(22,60){Ground truth}
            \end{overpic}
        }
    \end{minipage}\par\smallskip
    \begin{minipage}{1\linewidth}
        \begin{minipage}{\linewidth}
        \end{minipage}\par\medskip
        \begin{minipage}{0.58\linewidth}
            \centering
            \subfloat[Roughness estimation, outperforming that of \citet{InteriorVerse} and \citet{Kocsis2023}.]{
                \begin{overpic}[width=0.24\linewidth]{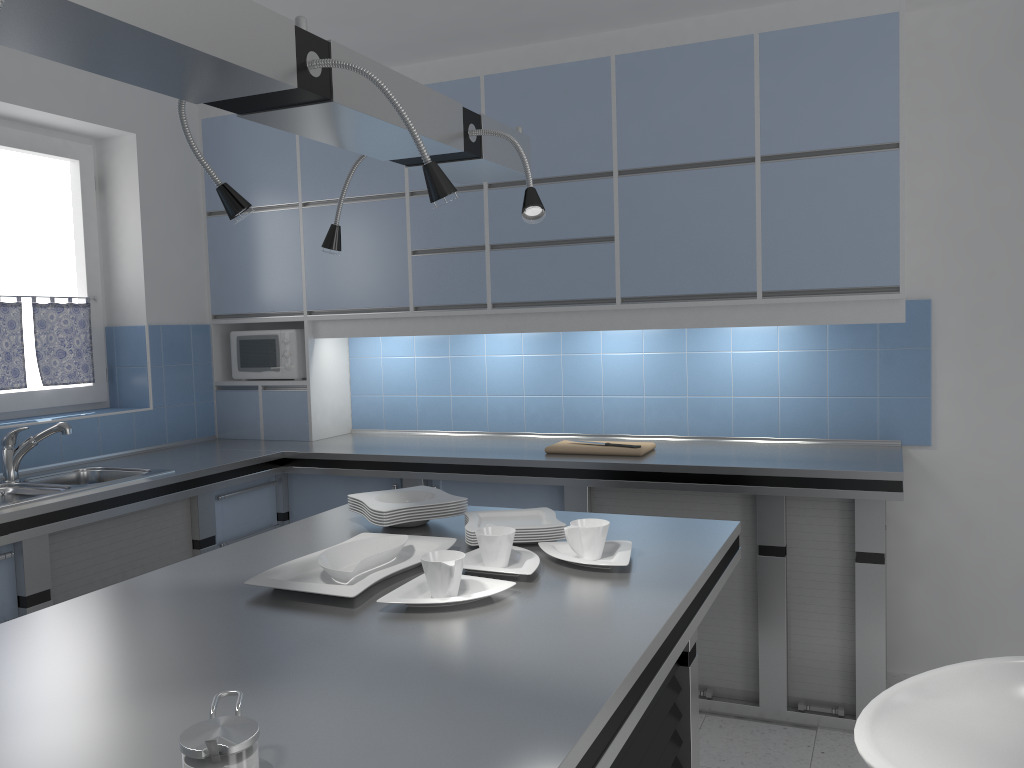}
                    \put(22,79){Input image}
                \end{overpic}
                \hfill
                \begin{overpic}[width=0.24\linewidth]{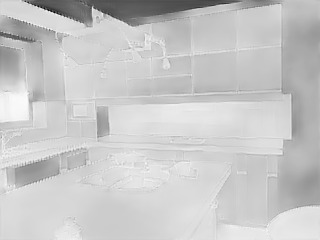}
                    \put(5,79){\citet{InteriorVerse}}
                \end{overpic}
                \hfill
                \begin{overpic}[width=0.24\linewidth]{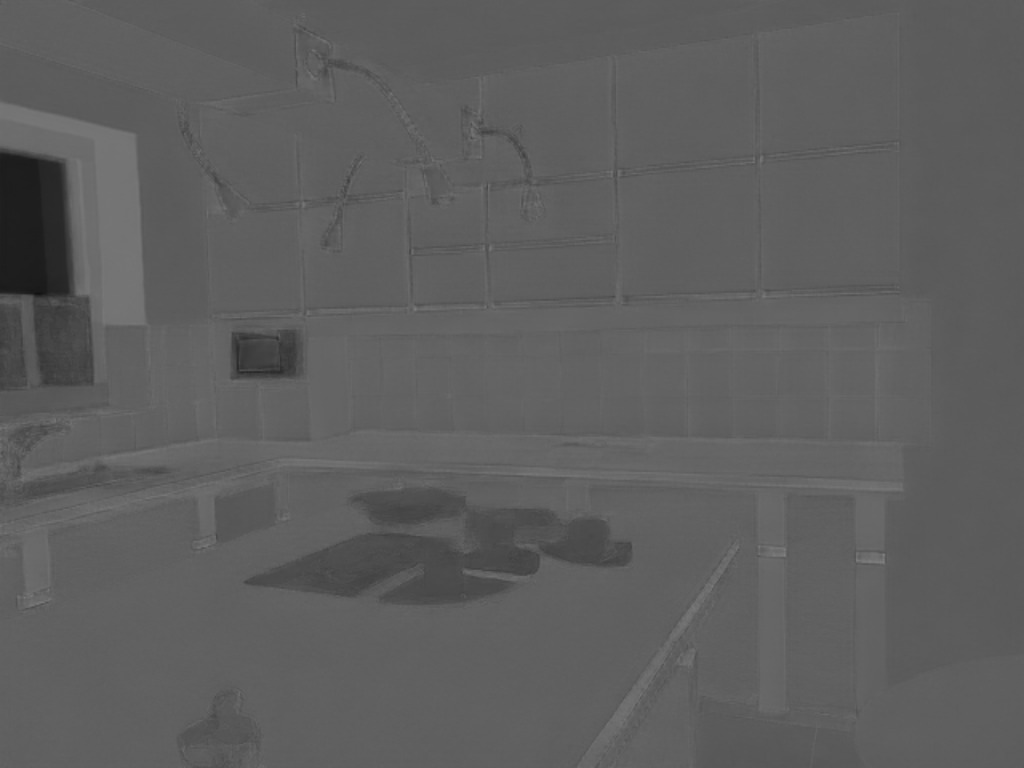}
                    \put(4,79){\citet{Kocsis2023}}
                \end{overpic}
                \hfill
                \begin{overpic}[width=0.24\linewidth]{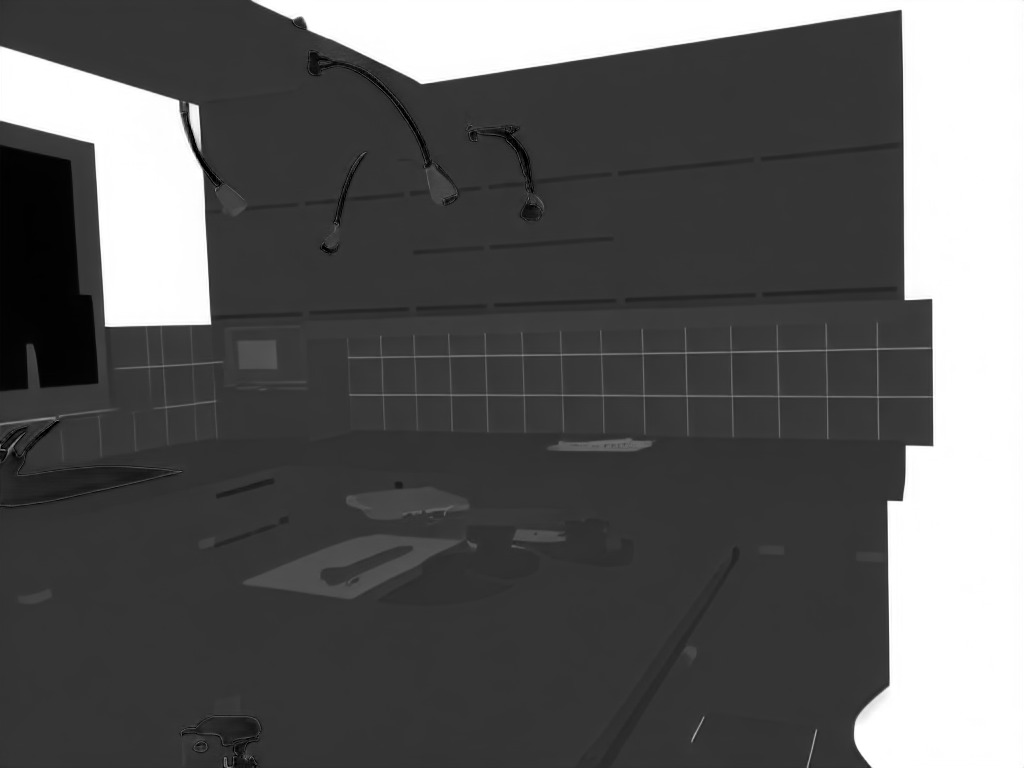}
                    \put(10,79){\textbf{Our \rgbtoxx}}
                \end{overpic}
            }
        \end{minipage}
        \hfill
        \begin{minipage}{0.41\linewidth}
            \centering
            \subfloat[Metallicity estimation on the input from (c).]{
                \begin{overpic}[width=0.33\linewidth]{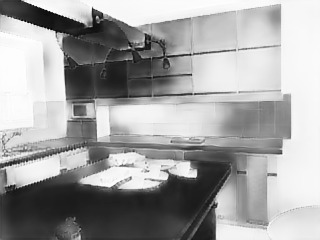}
                    \put(4,79){\citet{InteriorVerse}}
                \end{overpic}
                \hfill
                \begin{overpic}[width=0.33\linewidth]{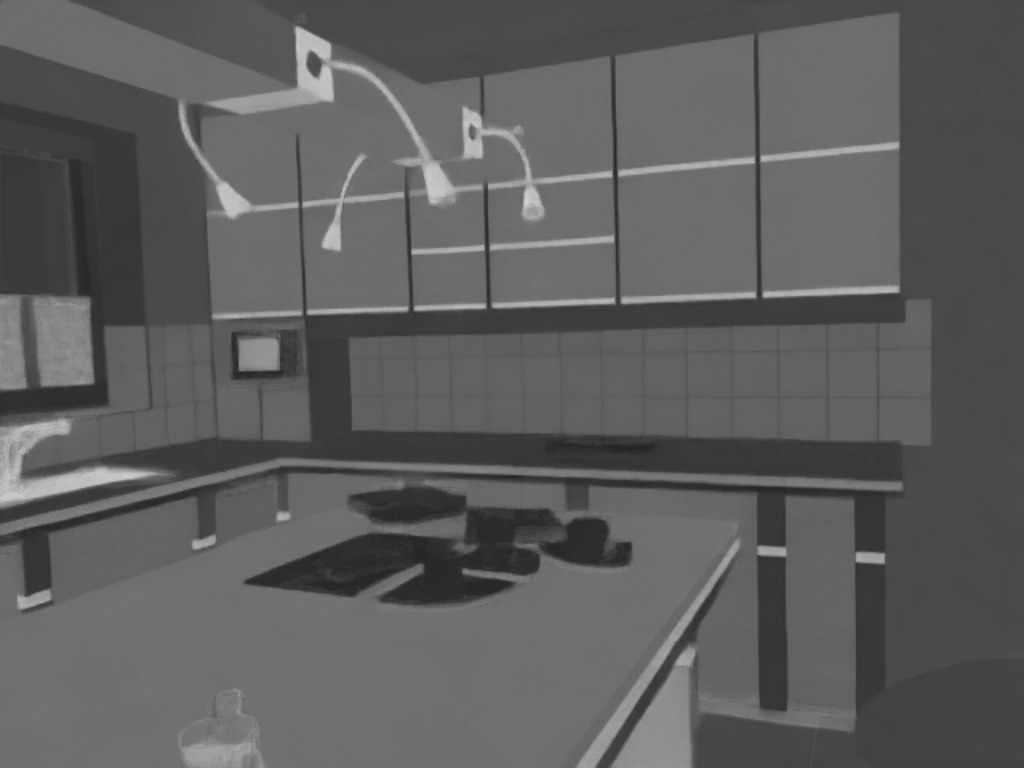}
                    \put(3,79){\citet{Kocsis2023}}
                \end{overpic}
                \hfill
                \begin{overpic}[width=0.33\linewidth]{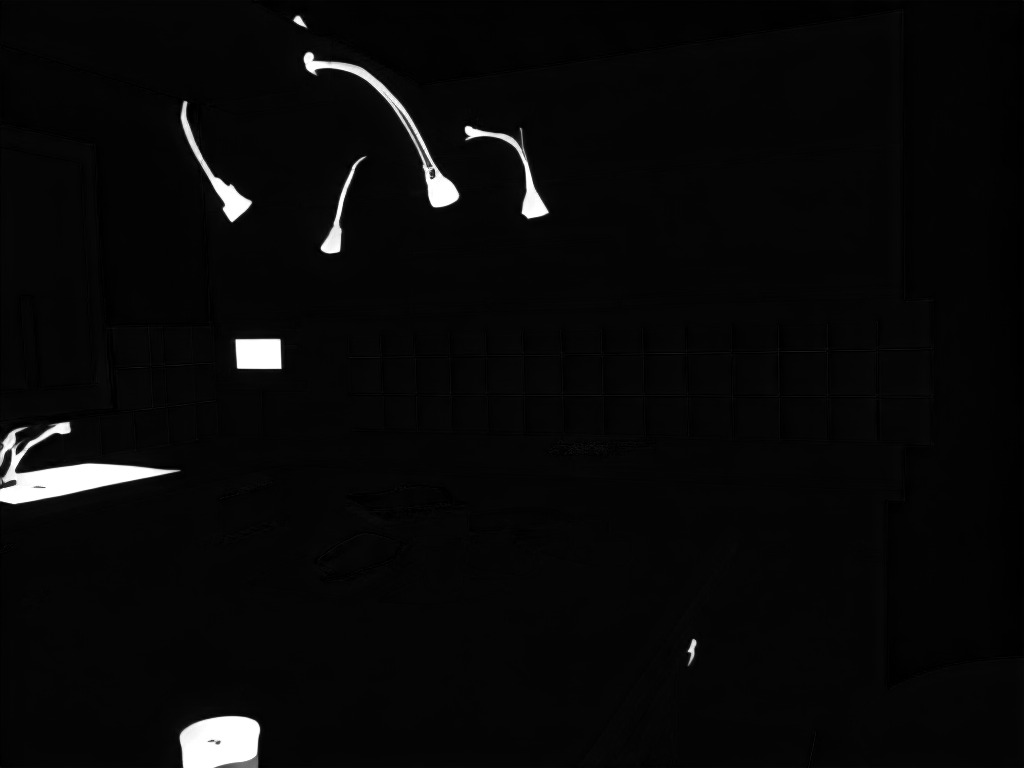}
                    \put(10,79){\textbf{Our \rgbtoxx}}
                \end{overpic}
            }
        \end{minipage}
    \end{minipage}\par\bigskip\medskip
    \begin{minipage}{1\linewidth}
        \centering
        \subfloat[\rgbtoxx normal estimation, outperforming \citet{InteriorVerse}, as well as a SOTA model based on vision transformers (trained on much more diverse data).]{
            \begin{overpic}[width=0.195\linewidth]{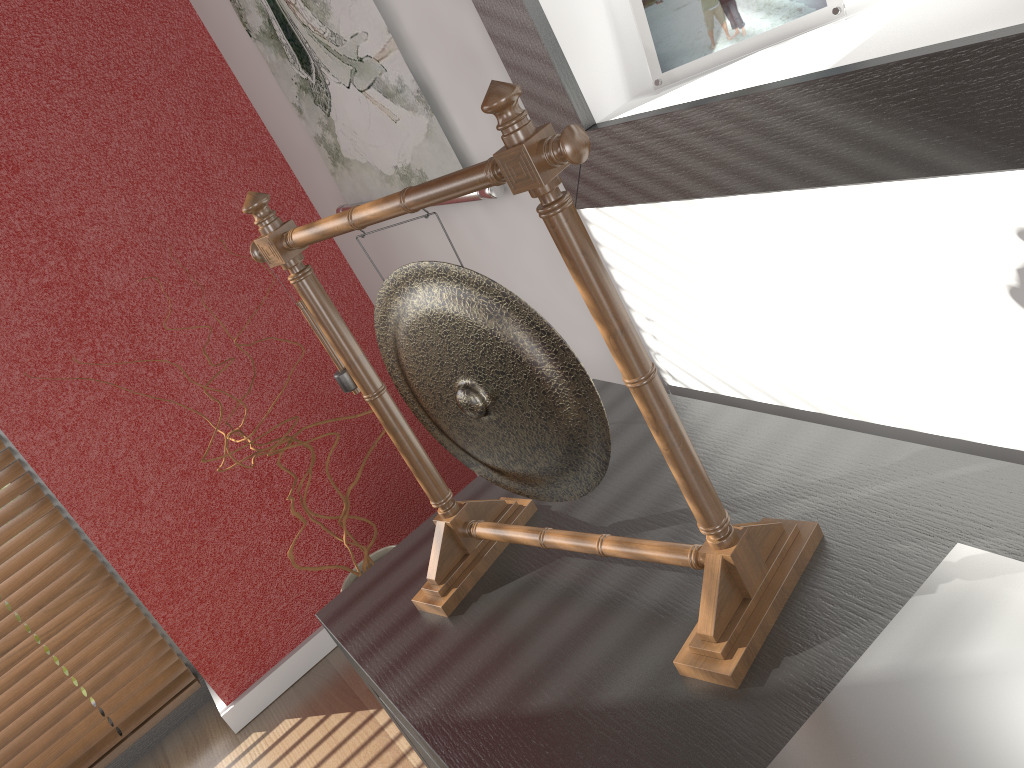}
                \put(29,79){Input image}
            \end{overpic}
            \hfill
            \begin{overpic}[width=0.195\linewidth]{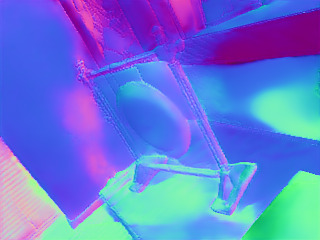}
                \put(20,79){\citet{InteriorVerse}}
            \end{overpic}
            \hfill
            \begin{overpic}[width=0.195\linewidth]{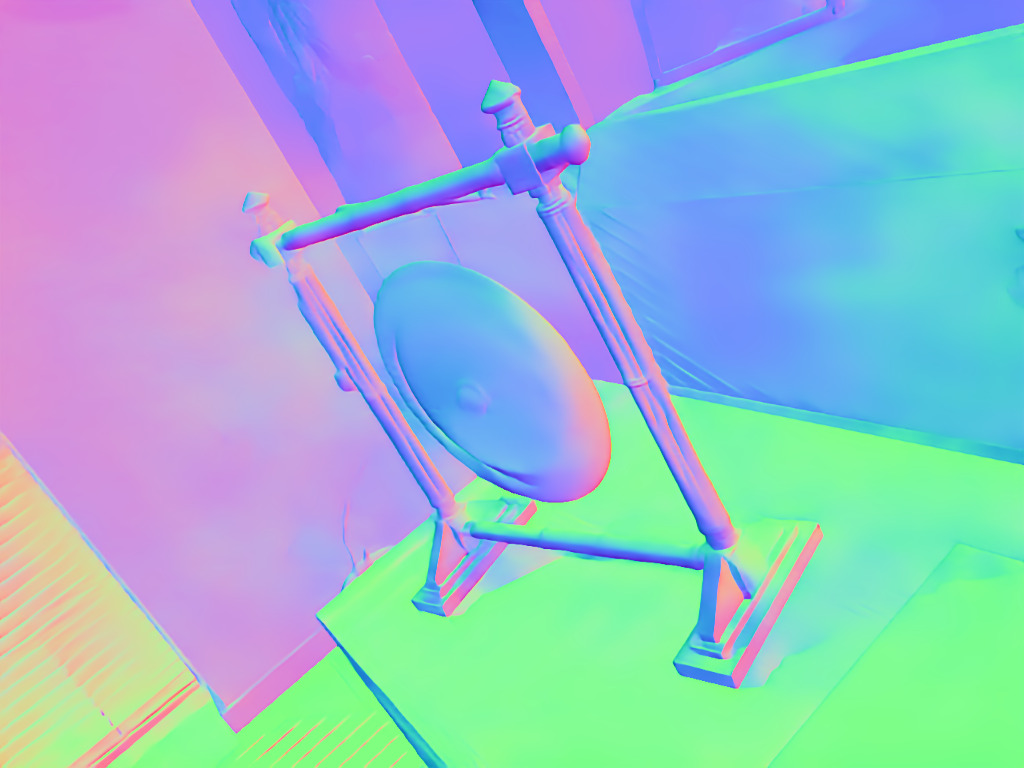}
                \put(26,79){PVT-normal}
            \end{overpic}
            \hfill
            \begin{overpic}[width=0.195\linewidth]{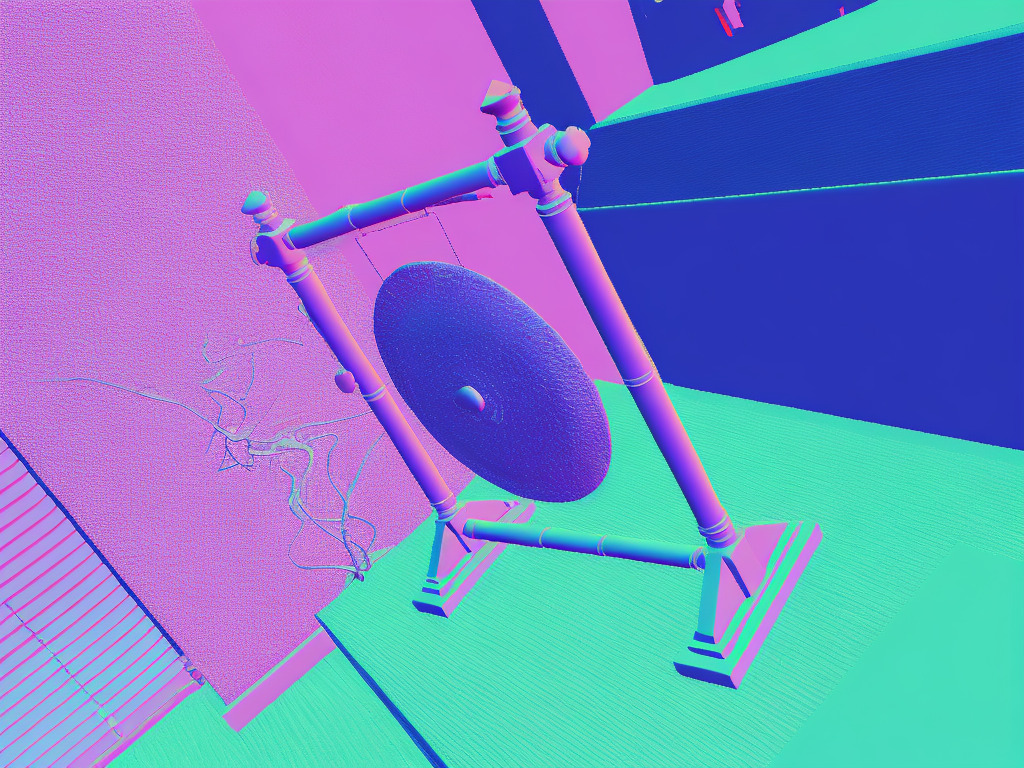}
                \put(22,79){\textbf{Our \rgbtoxx}}
            \end{overpic}
            \hfill
            \begin{overpic}[width=0.195\linewidth]{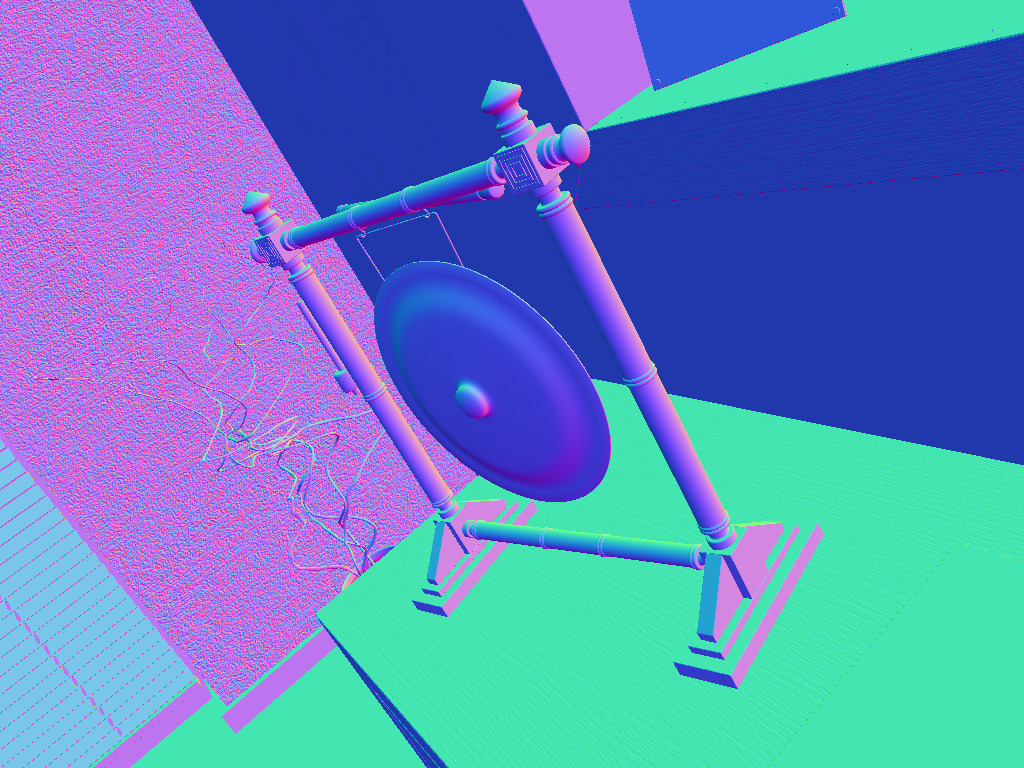}
                \put(25,79){Ground truth}
            \end{overpic}
        }
    \end{minipage}\par
\end{minipage}

%% file: figures/rgb2x_real/rgb2x_real.tex
\begin{minipage}{1\linewidth}
    \begin{minipage}{1\linewidth}
        \centering
        \subfloat[Albedo estimation on an image from the \textsc{IIW} real-photo dataset~\cite{bell2014intrinsic}. Our \rgbtoxx model clearly outperforms that of \citet{InteriorVerse}, while those of \citet{Kocsis2023} and \citet{OrdinalShading} provide reasonable estimates. Nevertheless, our results show more plausible white balance and flatness, e.g., correctly predicting that all bed-linen pixels should have identical white albedo.]{
            \begin{overpic}[width=0.195\linewidth]{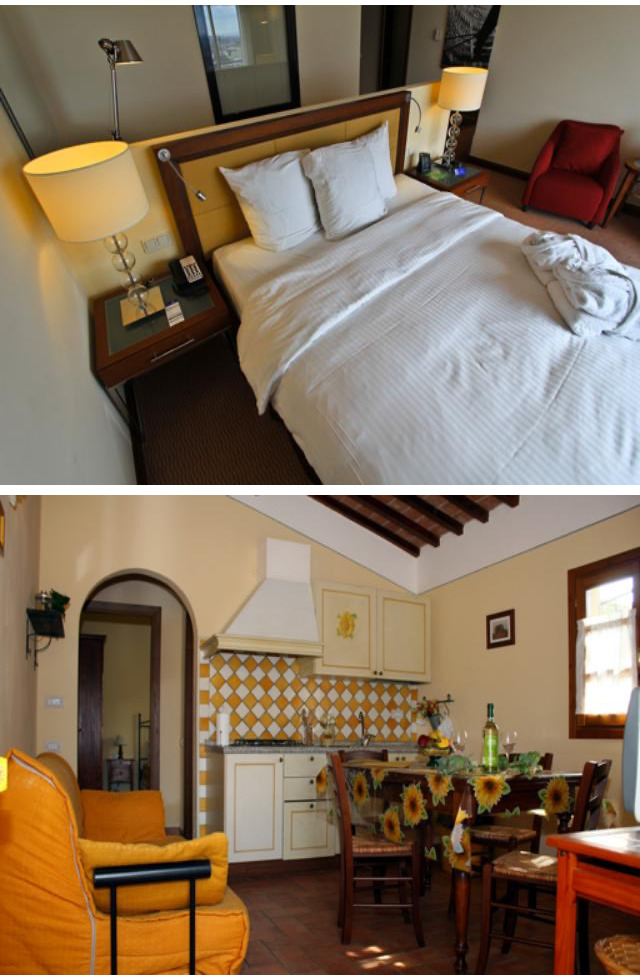}
                \put(19,102){Input image}
            \end{overpic}
            \hfill
            \begin{overpic}[width=0.195\linewidth]{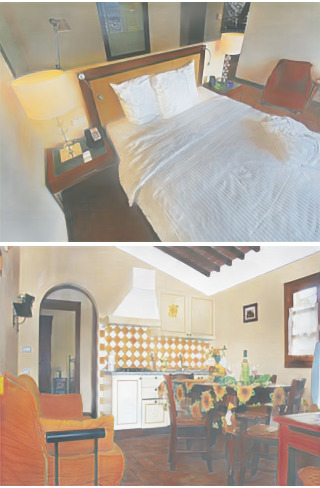}
                \put(13,102){\citet{InteriorVerse}}
            \end{overpic}
            \hfill
            \begin{overpic}[width=0.195\linewidth,]{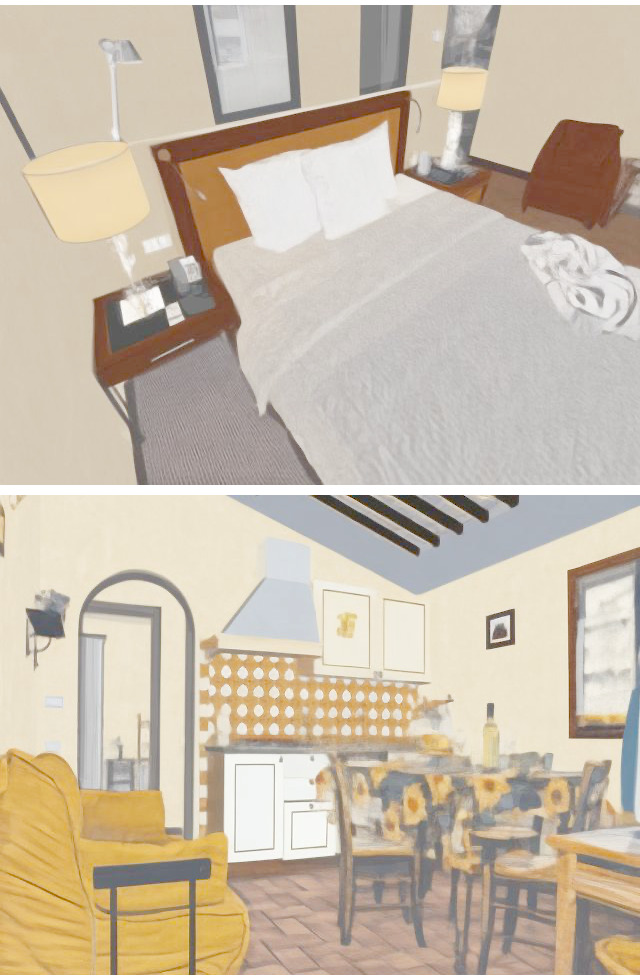}
                \put(11,102){\citet{Kocsis2023}}
            \end{overpic}
            \hfill
            \begin{overpic}[width=0.195\linewidth]{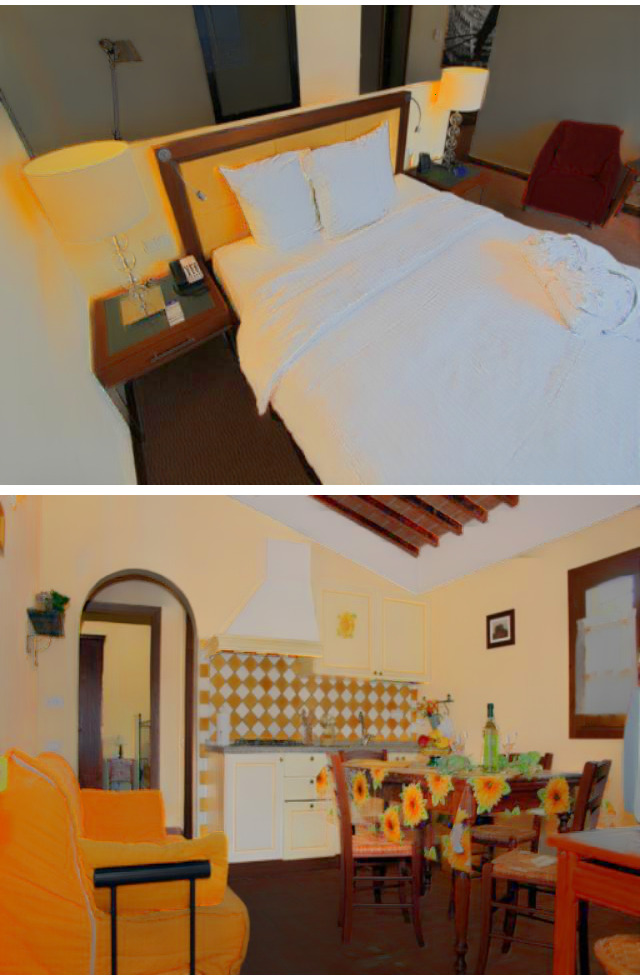}
                \put(1,102){\citet{OrdinalShading}}
            \end{overpic}
            \hfill
            \begin{overpic}[width=0.195\linewidth]{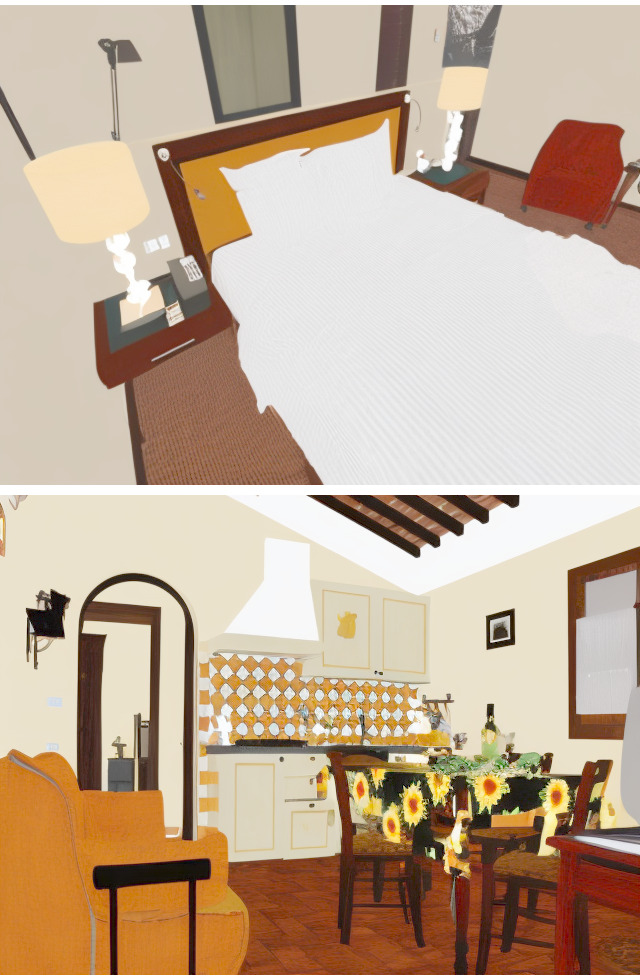}
                \put(12,102){\textbf{Our \rgbtoxx}}
            \end{overpic}\par\smallskip
        }
    \end{minipage}\par\smallskip
    \begin{minipage}{1\linewidth}
        \begin{minipage}{\linewidth}
        \end{minipage}\par\medskip
        \centering
        \subfloat[Albedo and irradiance estimation on an image from the \textsc{MIT Indoor Scene Recognition} dataset. Our \rgbtoxx result is the cleanest and has the flattest albedo regions. The result of \citet{OrdinalShading} however better preserves the imperfections in the floor albedo.]{
            \begin{overpic}[width=0.165\linewidth]{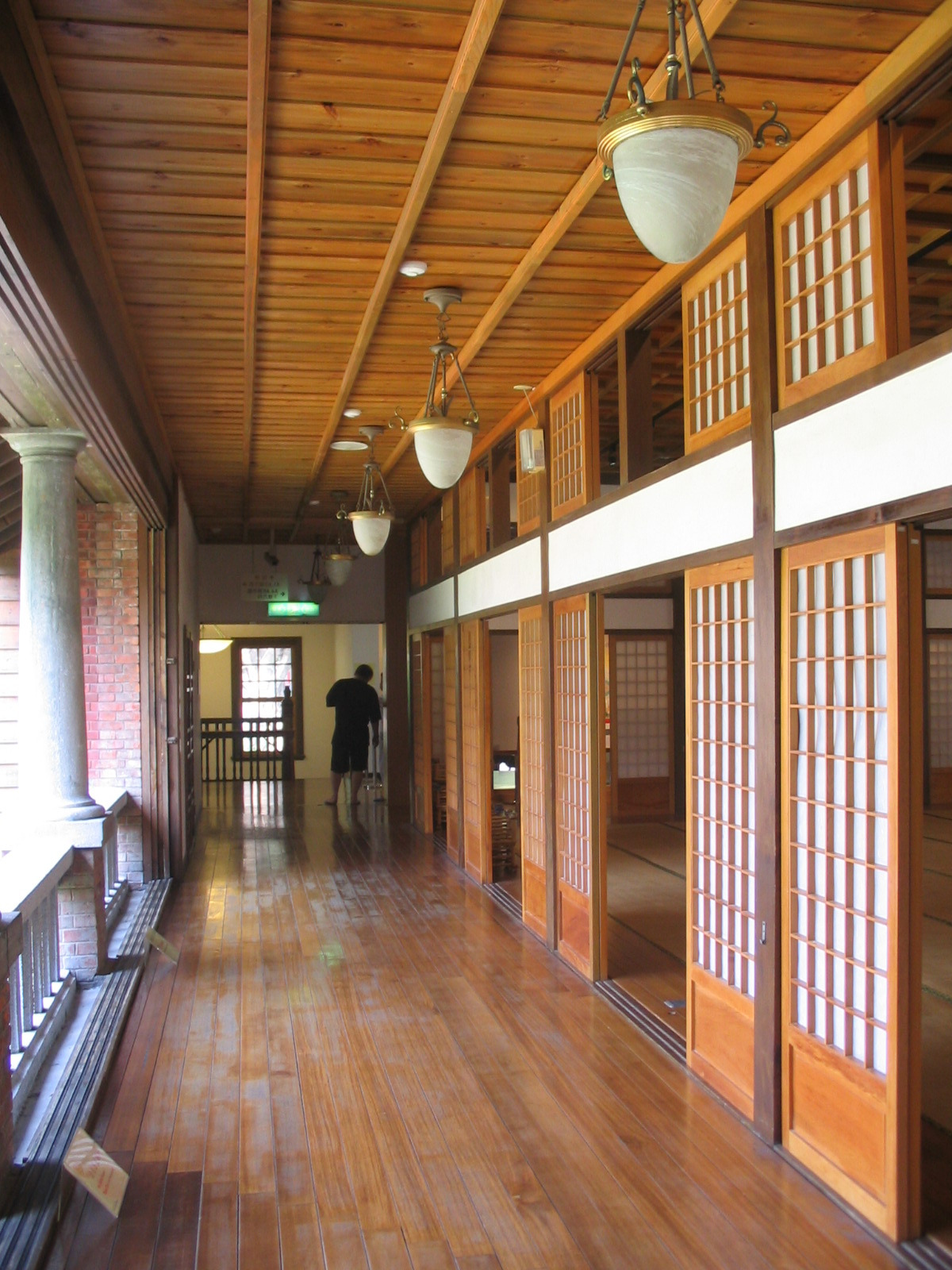}
                \put(18,103){Input image}
            \end{overpic}
            \hfill
            \begin{overpic}[width=0.165\linewidth]{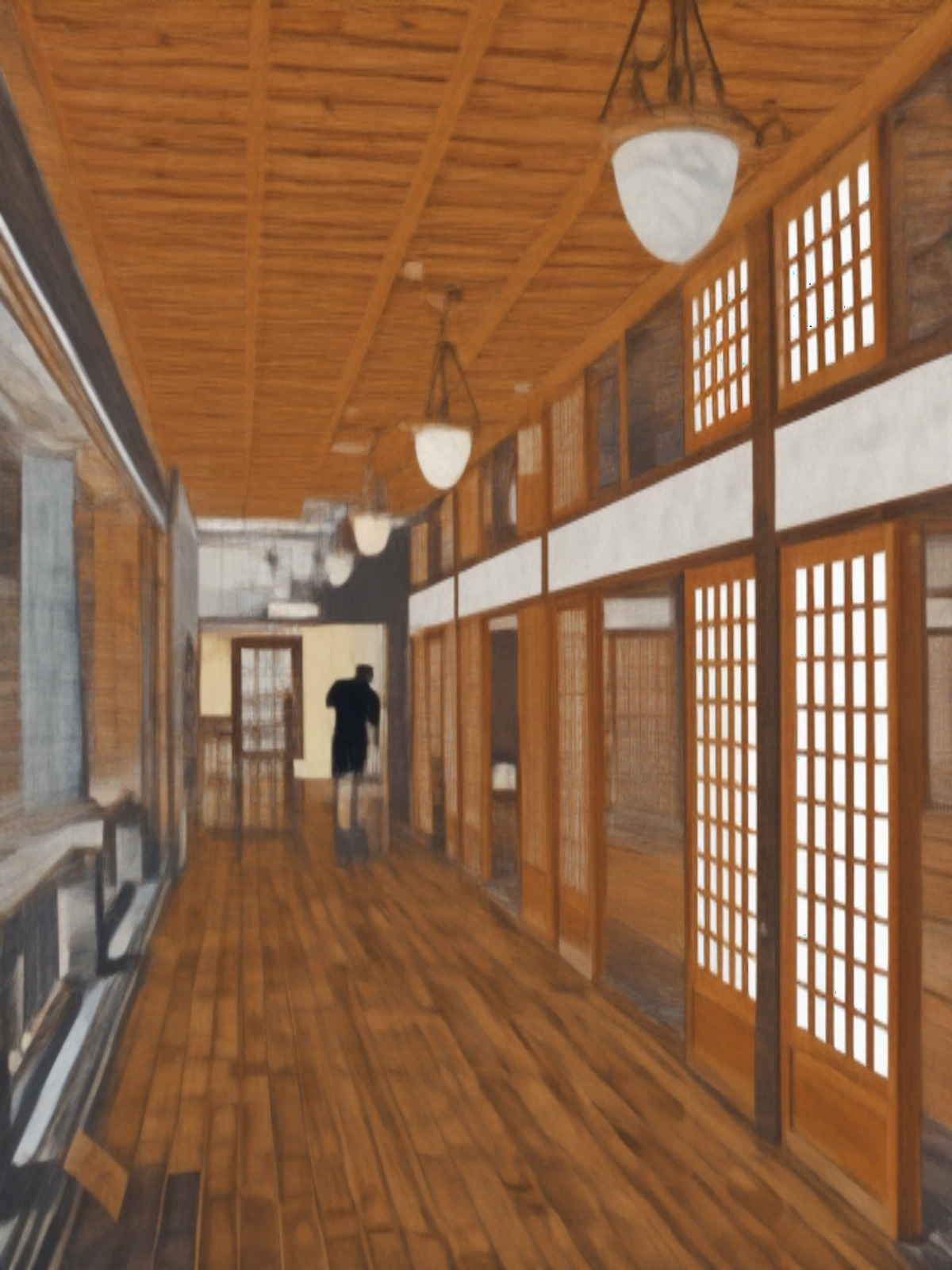}
                \put(6,103){\citet{Kocsis2023}}
            \end{overpic}
            \hfill
            \begin{overpic}[width=0.165\linewidth,]{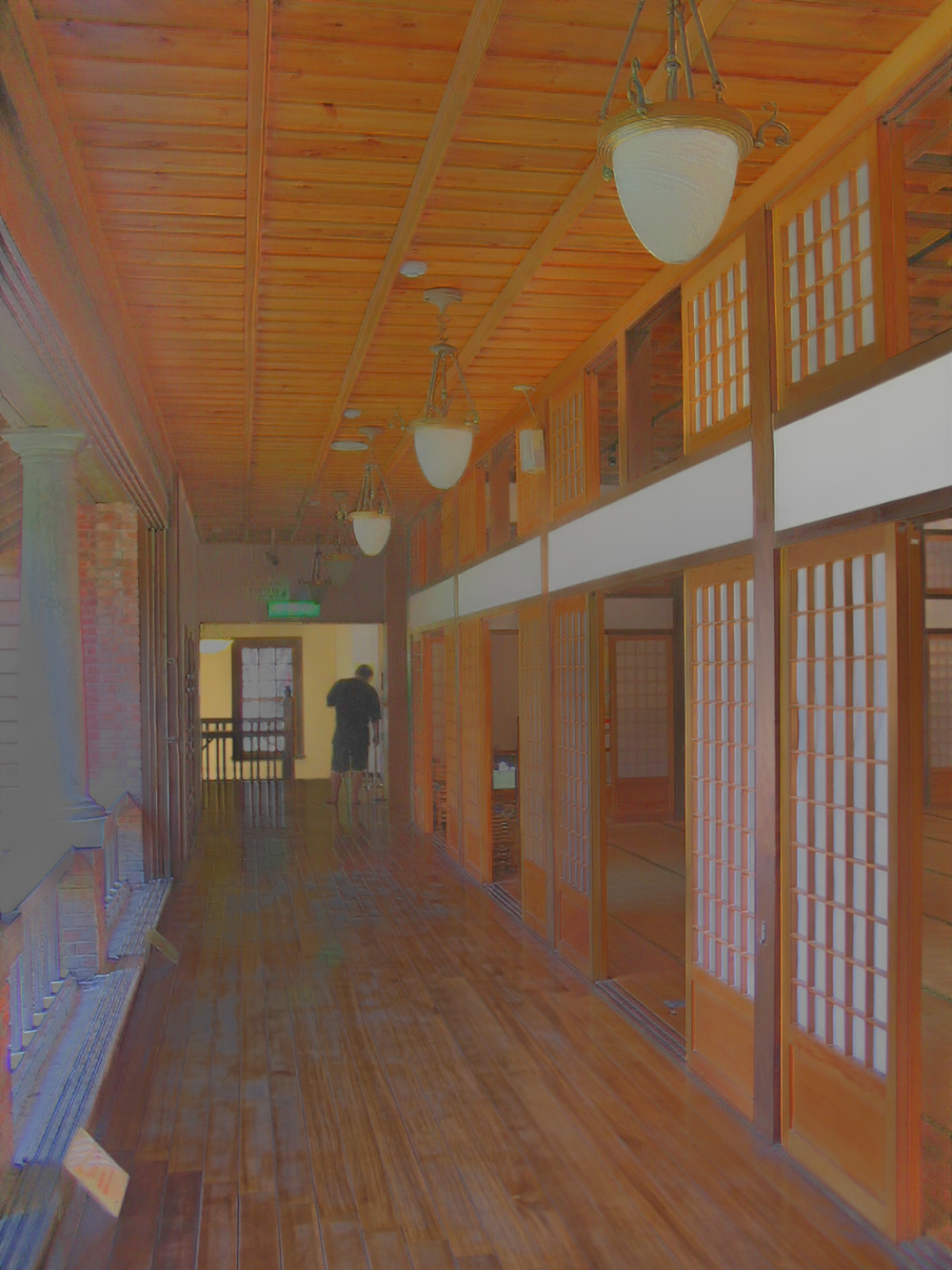}
                \put(4,103){\footnotesize\citet{OrdinalShading}}
            \end{overpic}
            \hfill
            \begin{overpic}[width=0.165\linewidth,]{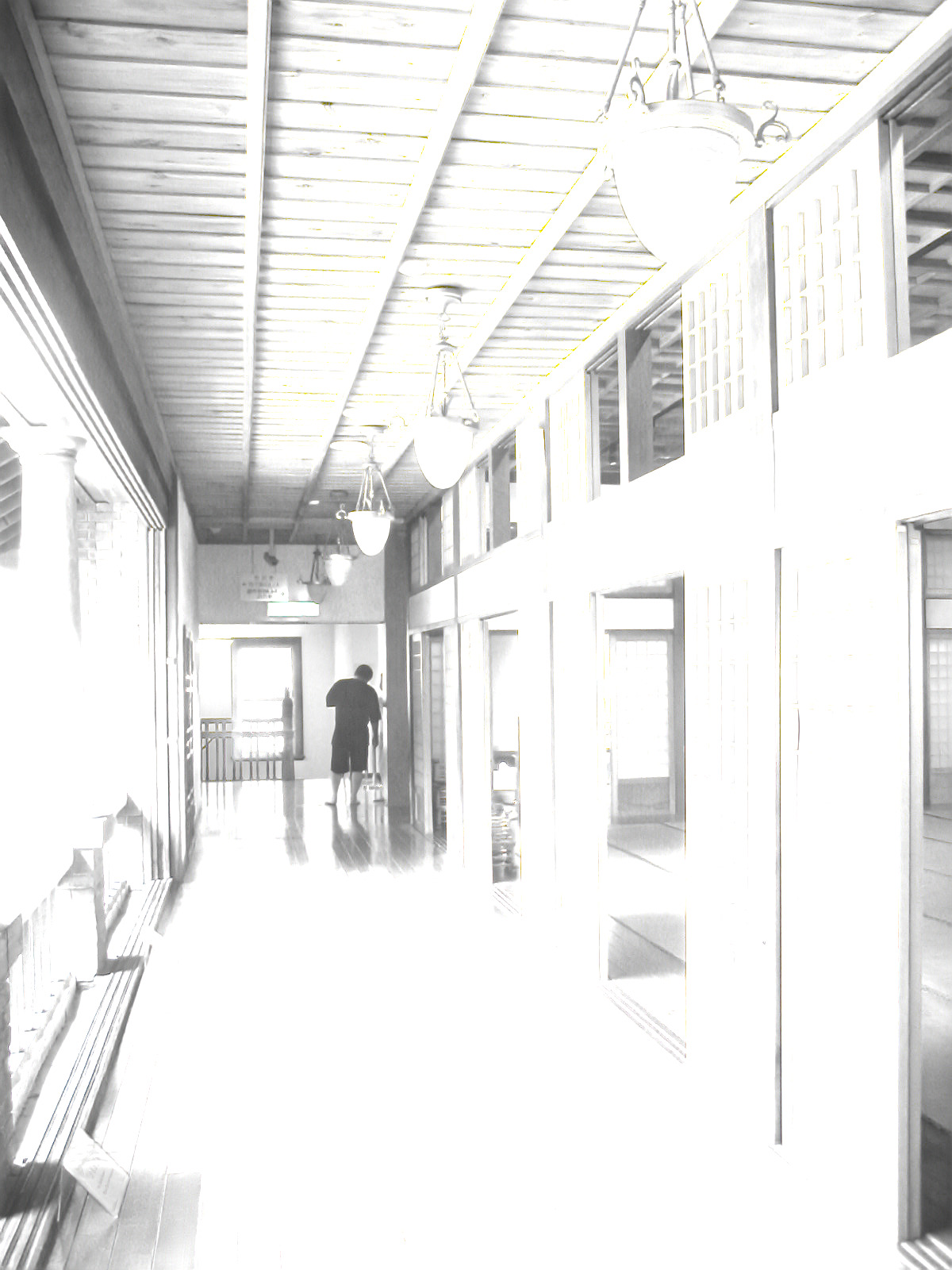}
                \put(7,110){\footnotesize Input image divided by}
                \put(4,103){\footnotesize\citet{OrdinalShading}}
            \end{overpic}
            \hfill
            \begin{overpic}[width=0.165\linewidth]{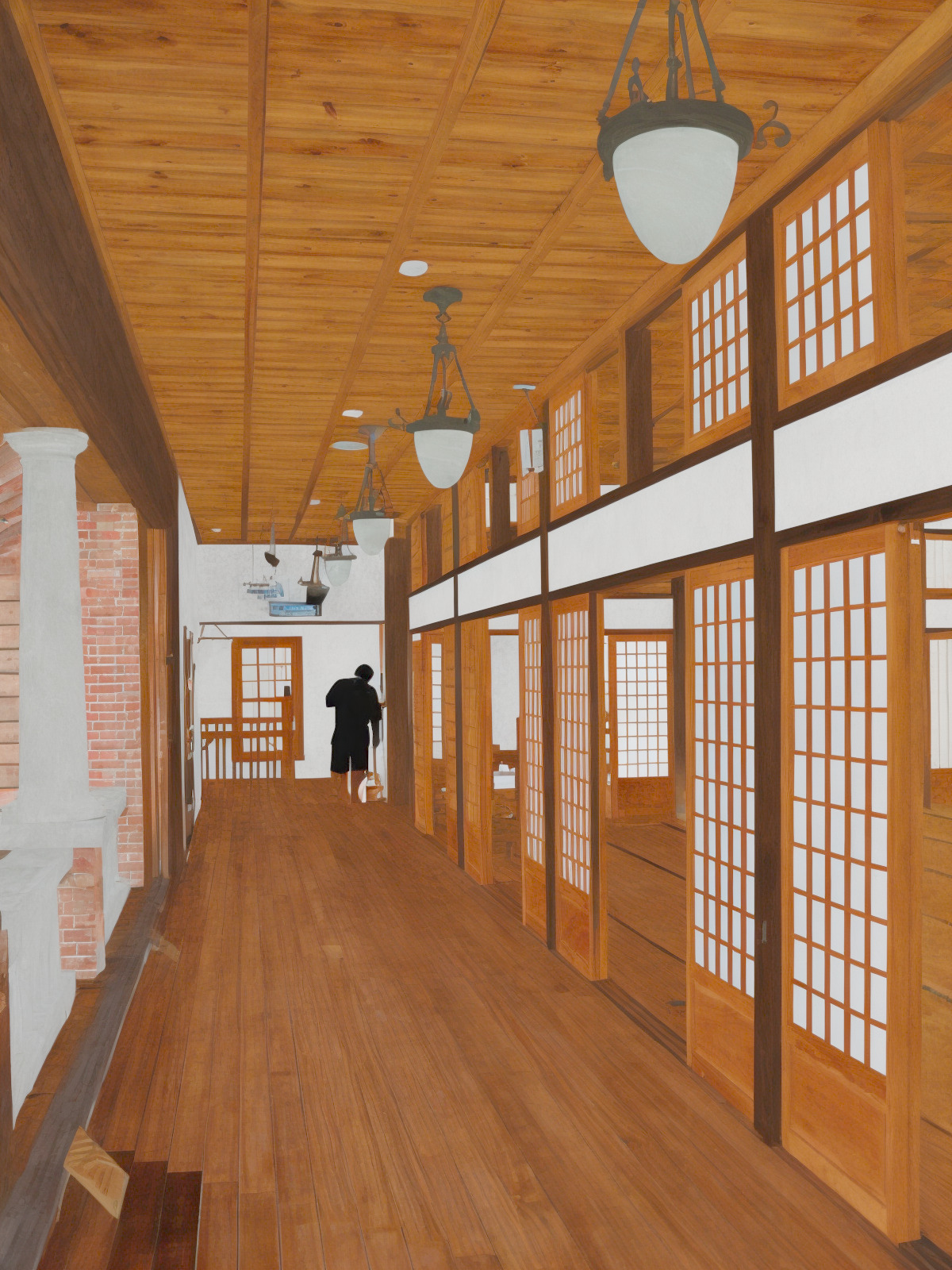}
                \put(10,103){\footnotesize \textbf{Our \rgbtoxx albedo}}
            \end{overpic}
            \hfill
            \begin{overpic}[width=0.165\linewidth]{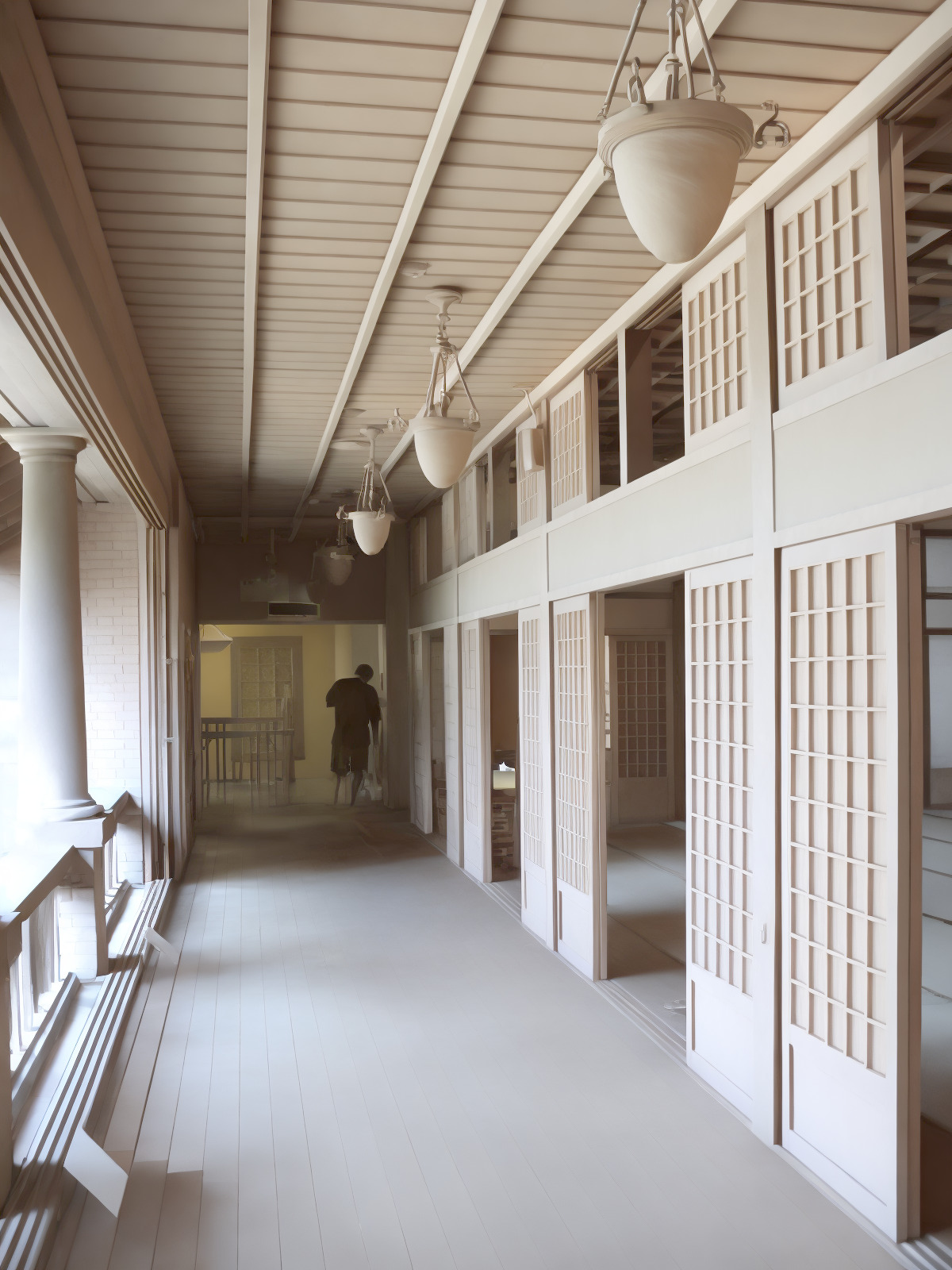}
                \put(5,103){\footnotesize \textbf{Our \rgbtoxx irradiance}}
            \end{overpic}
        }
    \end{minipage}\par\medskip
    \begin{minipage}{1\linewidth}
        \begin{minipage}{\linewidth}
        \end{minipage}\par\bigskip
        \begin{minipage}{0.57\linewidth}
            \centering
            \subfloat[Our \rgbtoxx roughness estimation on the \textsc{IIW} dataset outperforms previous methods.]{
                \begin{overpic}[width=0.245\linewidth]{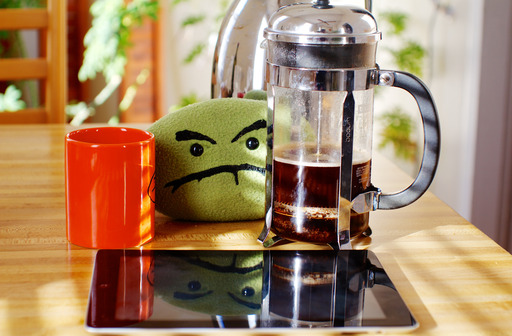}
                    \put(20,69){Input image}
                \end{overpic}
                \hfill
                \begin{overpic}[width=0.245\linewidth]{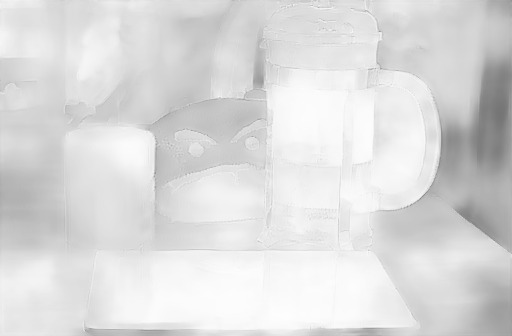}
                    \put(5,69){\citet{InteriorVerse}}
                \end{overpic}
                \hfill
                \begin{overpic}[width=0.245\linewidth]{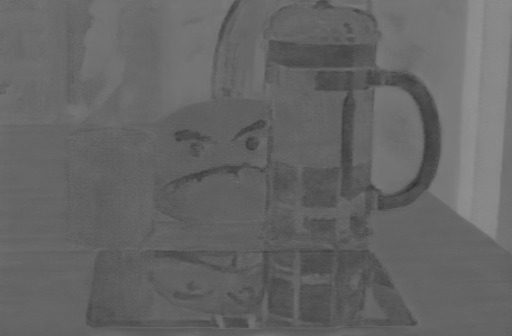}
                    \put(3,69){\citet{Kocsis2023}}
                \end{overpic}
                \hfill
                \begin{overpic}[width=0.245\linewidth]{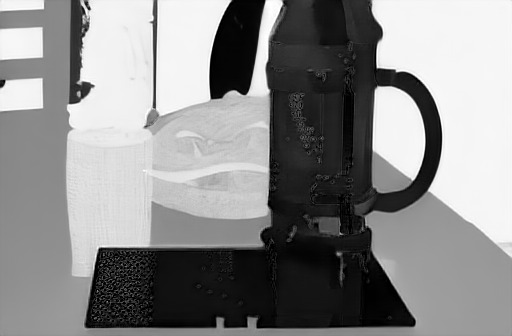}
                    \put(10,69){\textbf{Our \rgbtoxx}}
                \end{overpic}
            }
        \end{minipage}
        \hfill
        \begin{minipage}{0.42\linewidth}
            \centering
            \subfloat[Metallicity estimates on the same image as (c).]{
                \begin{overpic}[width=0.33\linewidth]{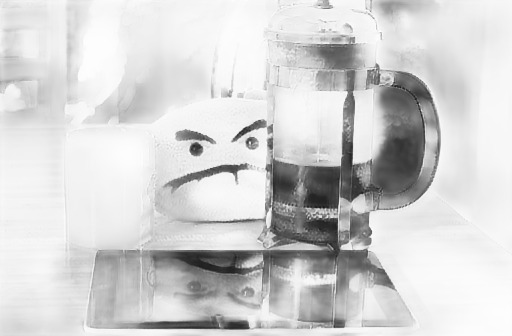}
                    \put(5,69){\citet{InteriorVerse}}
                \end{overpic}
                \hfill
                \begin{overpic}[width=0.33\linewidth]{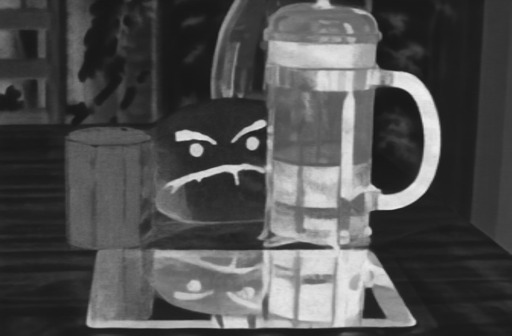}
                    \put(3,69){\citet{Kocsis2023}}
                \end{overpic}
                \hfill
                \begin{overpic}[width=0.33\linewidth]{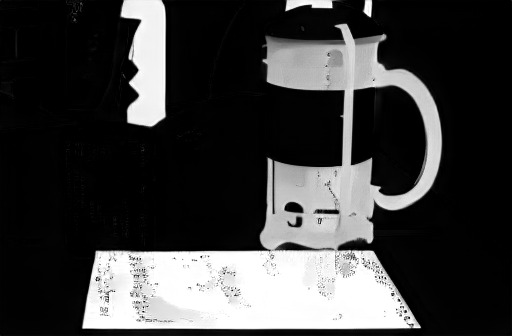}
                    \put(10,69){\textbf{Our \rgbtoxx}}
                \end{overpic}
            }
        \end{minipage}
    \end{minipage}\par\smallskip
    \begin{minipage}{1\linewidth}
        \begin{minipage}{\linewidth}
        \end{minipage}\par\medskip
        \centering
        \subfloat[Normal estimation on an image from the \textsc{MIT Indoor Scene Recognition} real-photo dataset~\cite{quattoni2009recognizing}. Our estimate is similar to that of PVT-normal, with slightly better results on flat areas. Note that our model is specialized to indoor scene training data, unlike PVT-normal.]{
            \begin{overpic}[width=0.245\linewidth]{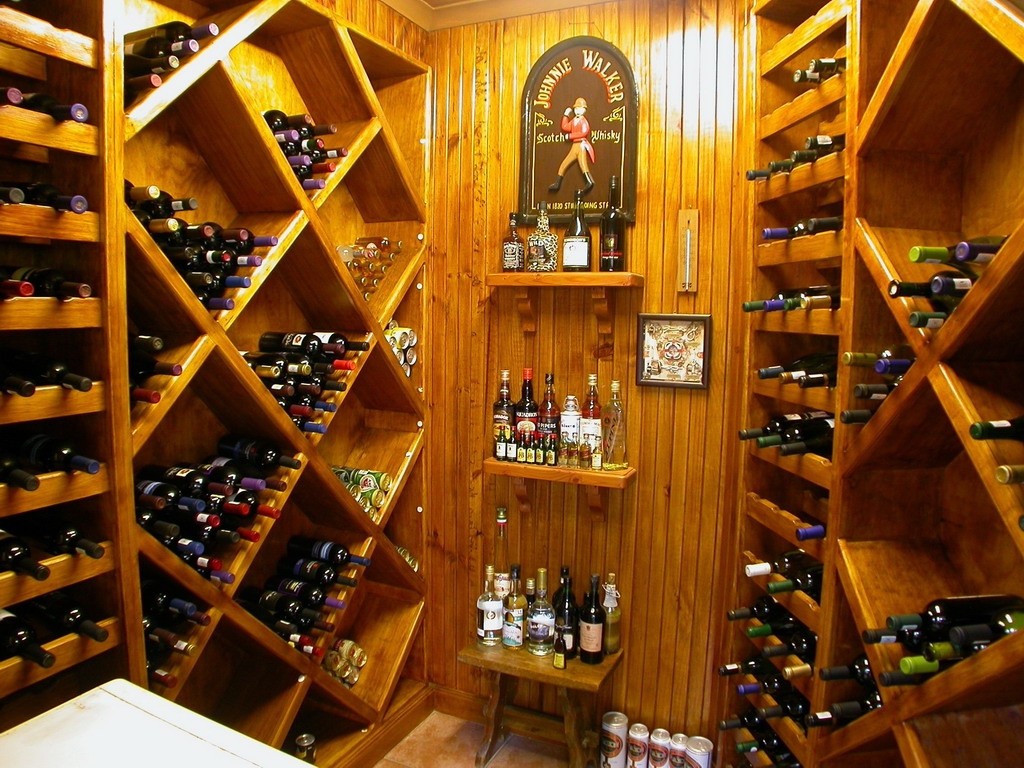}
                \put(32,79){Input image}
            \end{overpic}
            \hfill
            \begin{overpic}[width=0.245\linewidth]{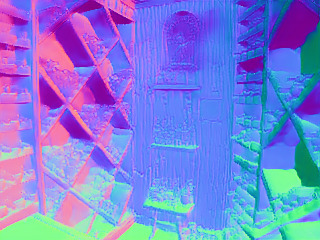}
                \put(24,79){\citet{InteriorVerse}}
            \end{overpic}
            \hfill
            \begin{overpic}[width=0.245\linewidth]{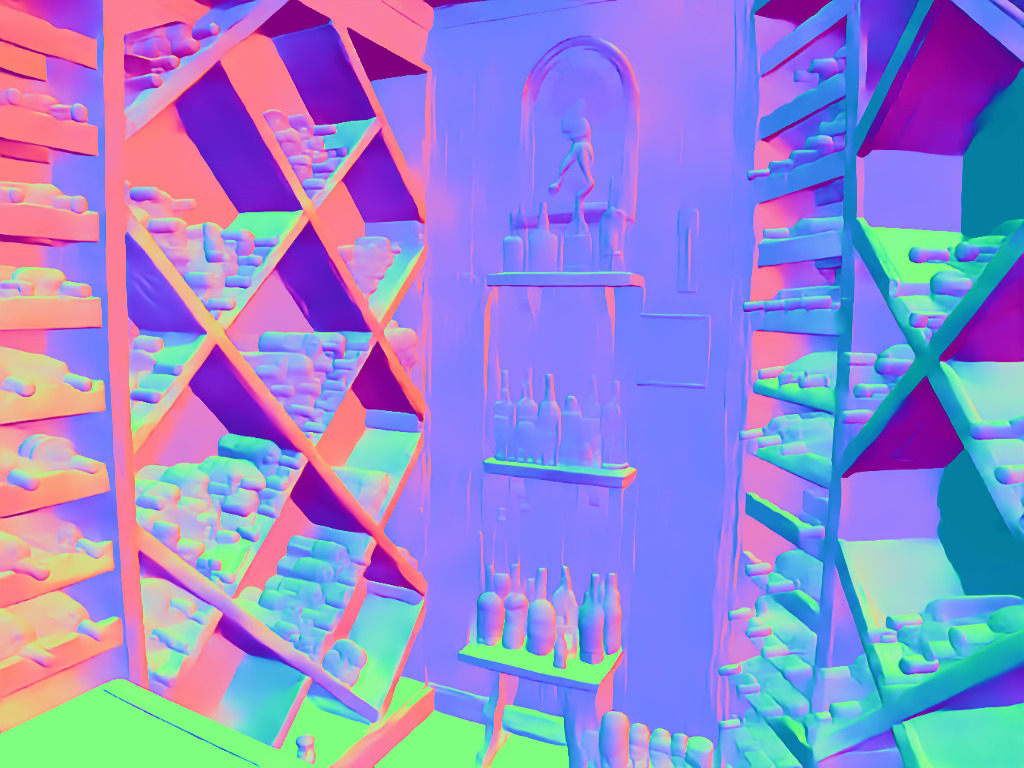}
                \put(32,79){PVT-normal}
            \end{overpic}
            \hfill
            \begin{overpic}[width=0.245\linewidth]{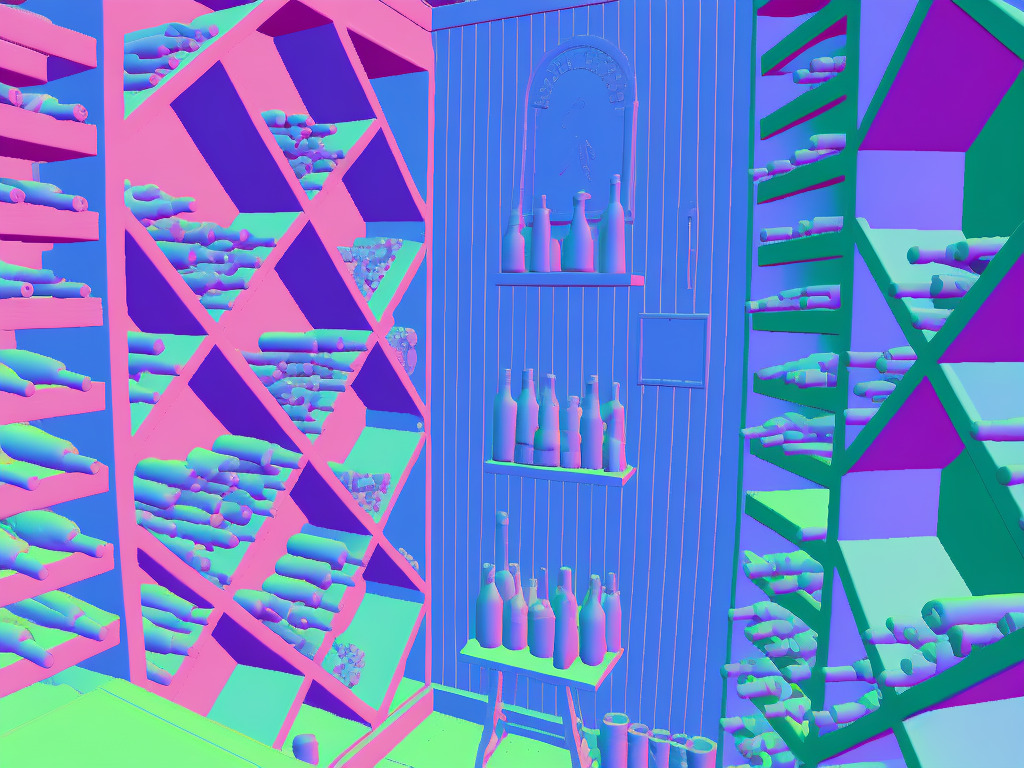}
                \put(26,79){\textbf{Our \rgbtoxx}}
            \end{overpic}
        }
    \end{minipage}\par
\end{minipage}

%% file: figures/x2rgb/x2rgb1.tex
\begin{minipage}{1\linewidth}
    \centering
    \begin{overpic}[width=0.14\linewidth]{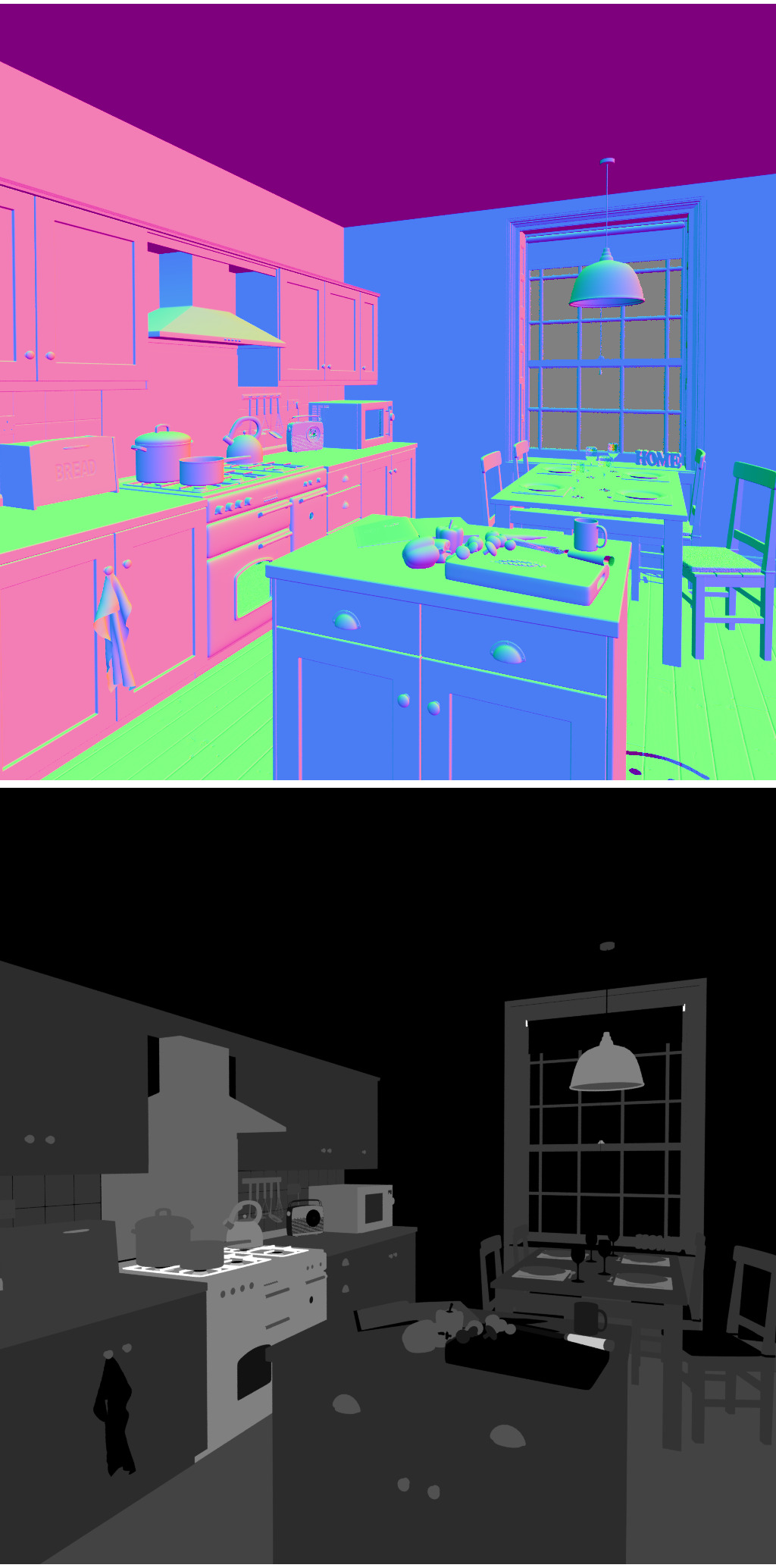}
        \put(16,93){\textcolor{white}{Normal}}
        \put(11,42){\textcolor{white}{Roughness}}
    \end{overpic}
    \hfill
    \begin{overpic}[width=0.14\linewidth]{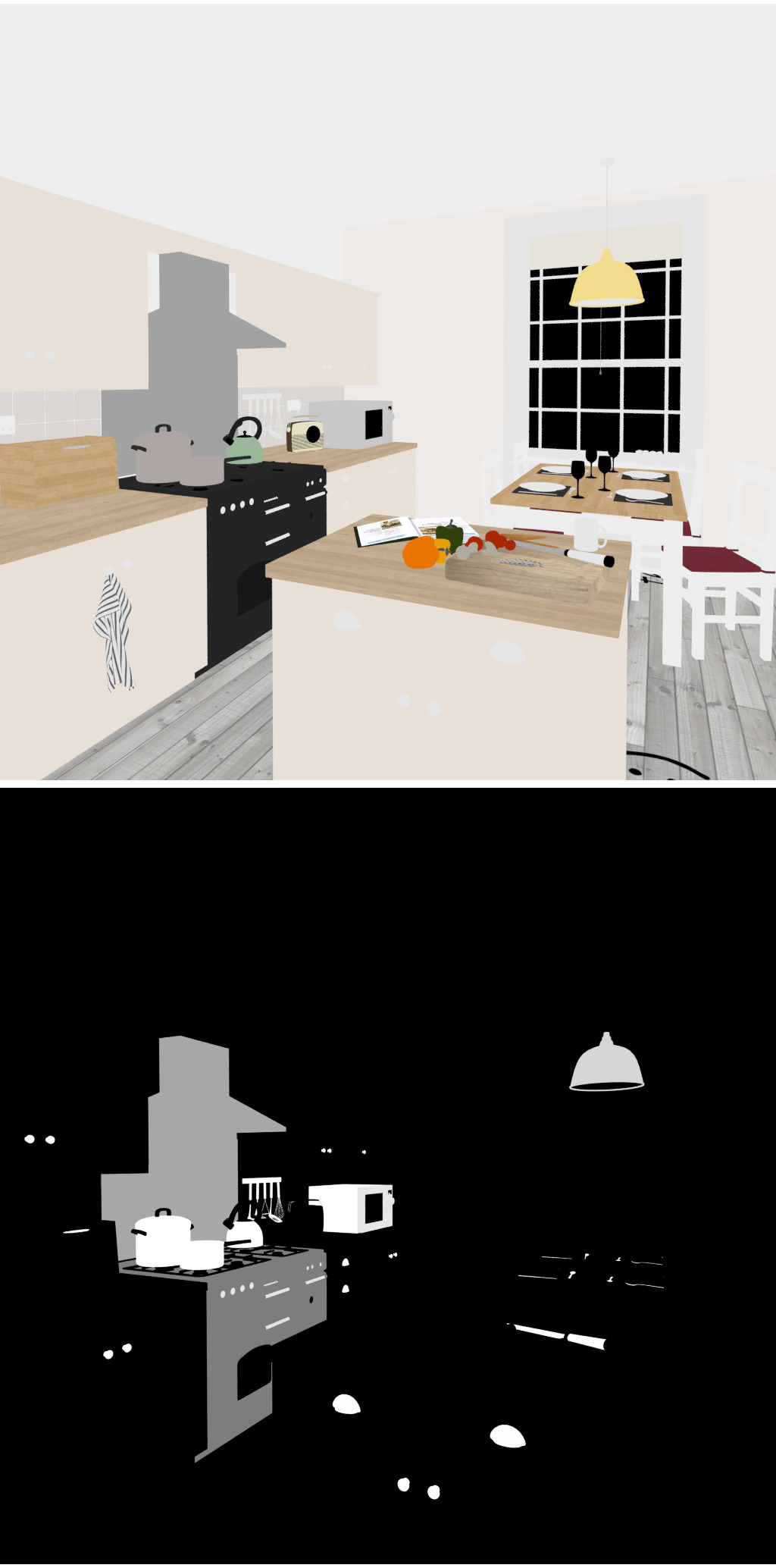}
        \put(16,93){Albedo}
        \put(12,42){\textcolor{white}{Metallicity}}
    \end{overpic}
    \hfill
    \begin{overpic}[width=0.14\linewidth]{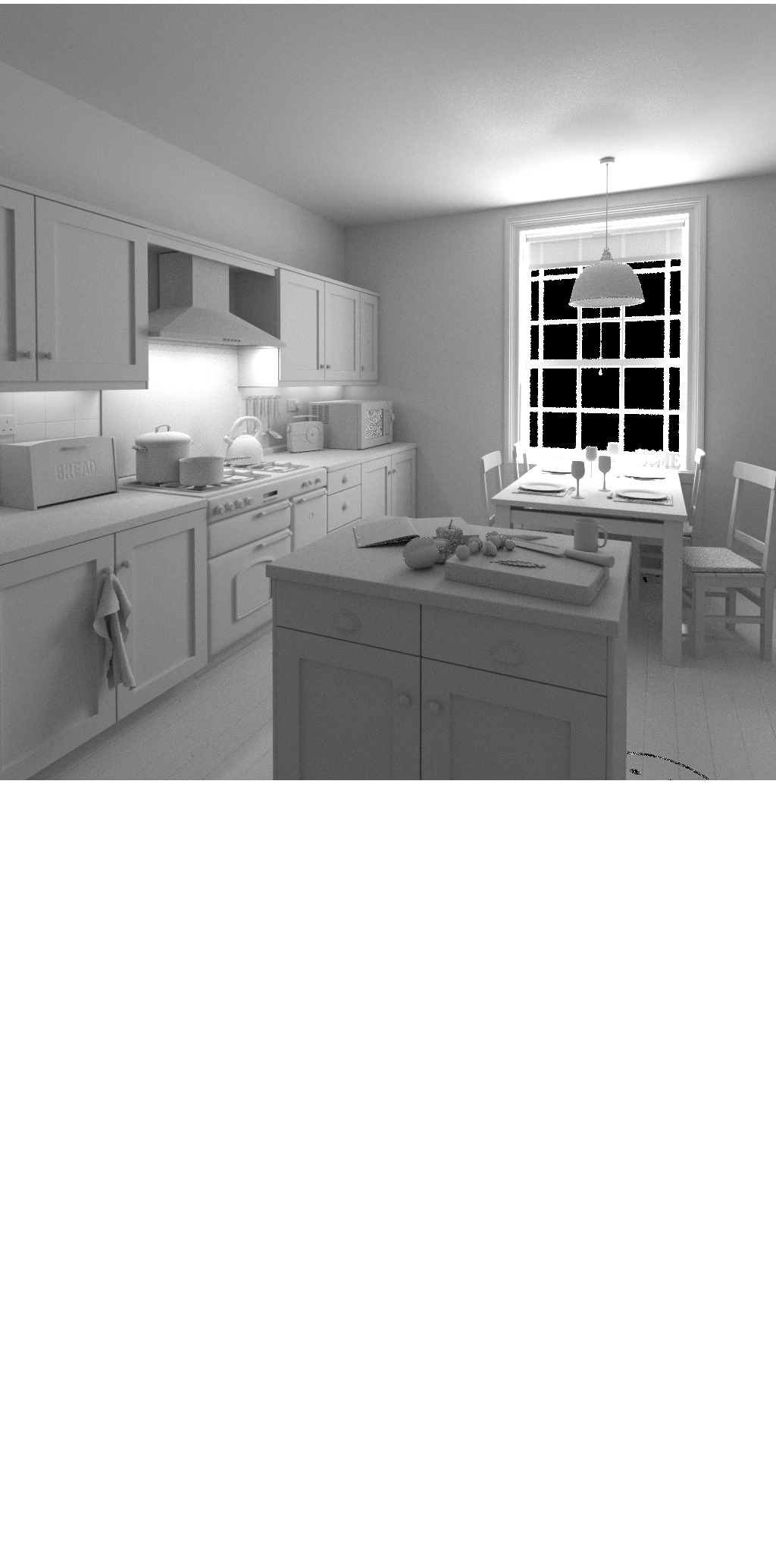}
        \put(12,93){Irradiance}
        \put(-1,20){
            \begin{tikzpicture}
                \node[fill=black!10,fill opacity=0.8,rounded
                    corners=1ex, text width=2.2cm,align=left] {Prompt: \\ \texttt{\small``Modern kitchen''}};
            \end{tikzpicture}
        }
    \end{overpic}
    \hfill
    \begin{overpic}[width=0.282\linewidth]{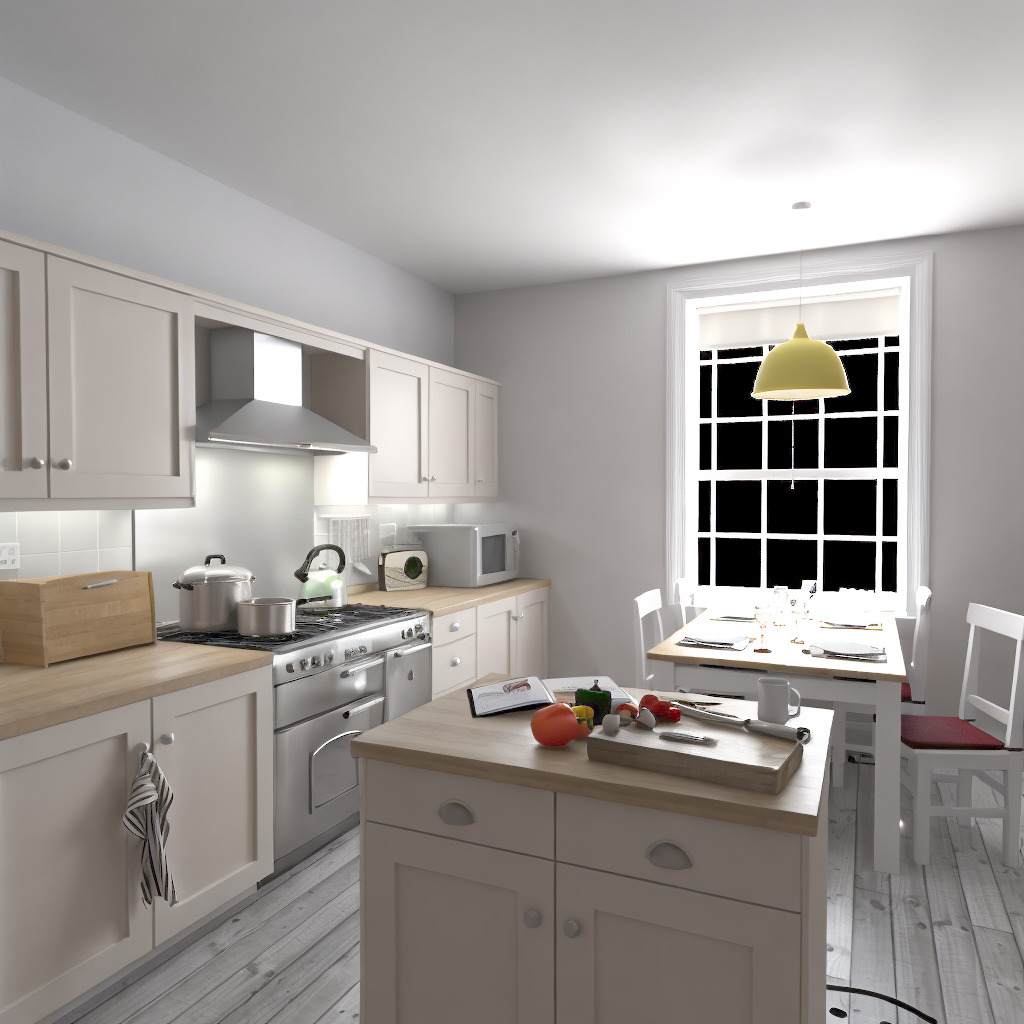}
        \put(31,102){\textbf{Our \xxtorgb}}
    \end{overpic}
    \hfill
    \begin{overpic}[width=0.282\linewidth]{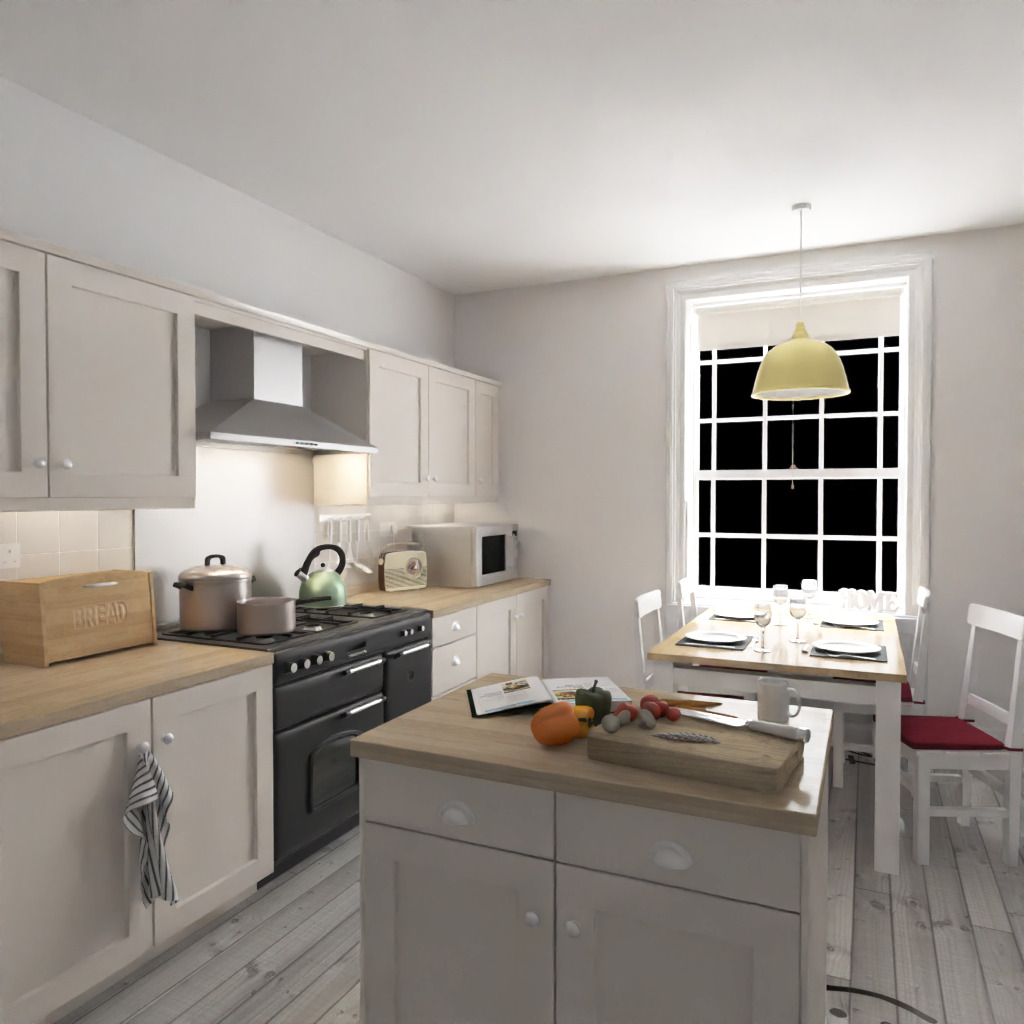}
        \put(35,102){Ground truth}
    \end{overpic}
\end{minipage}

%% file: figures/x2rgb/x2rgb2.tex
\begin{minipage}{1\linewidth}
    \begin{minipage}{0.48\linewidth}
        \subfloat[]{
            \begin{overpic}[width=0.199\linewidth]{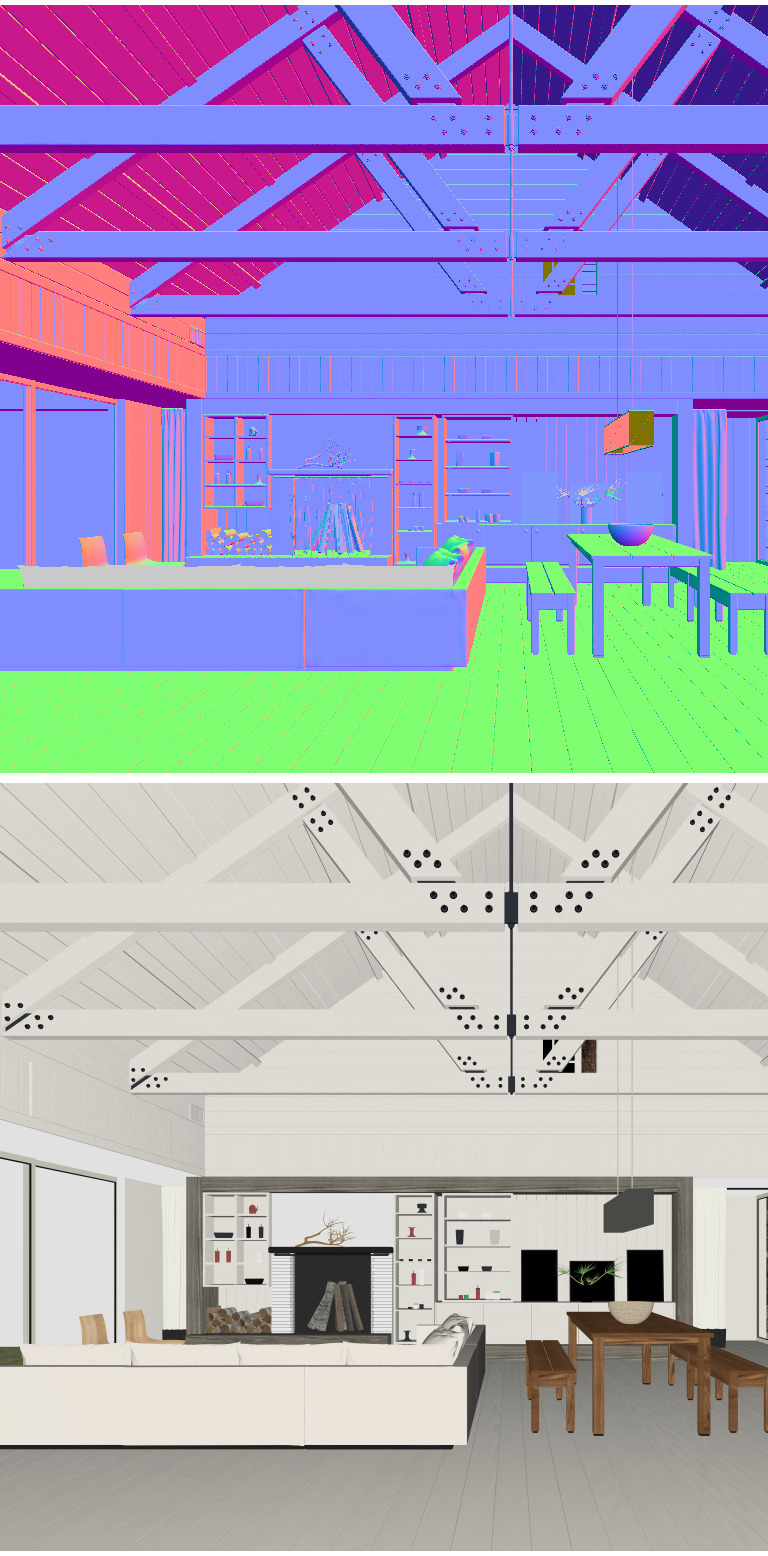}
                \put(12,90){\textcolor{white}{Normal}}
                \put(13,40){Albedo}
            \end{overpic}
            \hfill
            \begin{overpic}[width=0.4\linewidth]{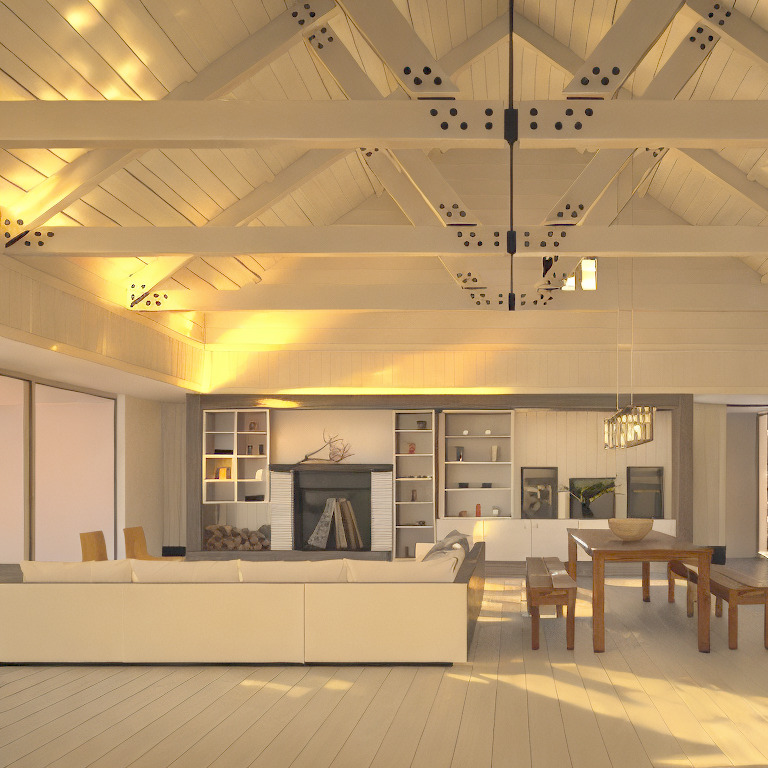}
                \put(25,102){\textbf{Our \xxtorgb}}
                \put(0,72){
                    \begin{tikzpicture}
                        \node[fill=black!10,fill opacity=0.8,rounded
                            corners=1ex, text width=2.4cm,align=left] {Prompt: \\ \texttt{\small``Sunset lighting''}};
                    \end{tikzpicture}
                }
            \end{overpic}
            \hfill
            \begin{overpic}[width=0.4\linewidth]{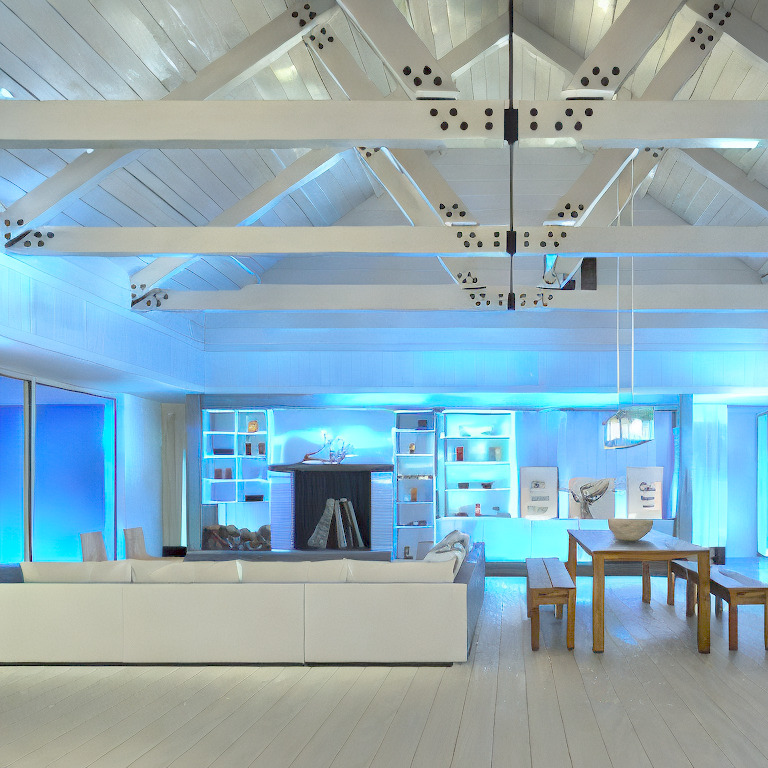}
                \put(25,102){\textbf{Our \xxtorgb}}
                \put(0,72){
                    \begin{tikzpicture}
                        \node[fill=black!10,fill opacity=0.8,rounded
                            corners=1ex, text width=2.1cm,align=left] {Prompt: \\ \texttt{\small``Blue lighting''}};
                    \end{tikzpicture}
                }
            \end{overpic}
        }
    \end{minipage}
    \hfill
    \begin{minipage}{0.48\linewidth}
        \subfloat[]{
            \begin{overpic}[width=0.199\linewidth]{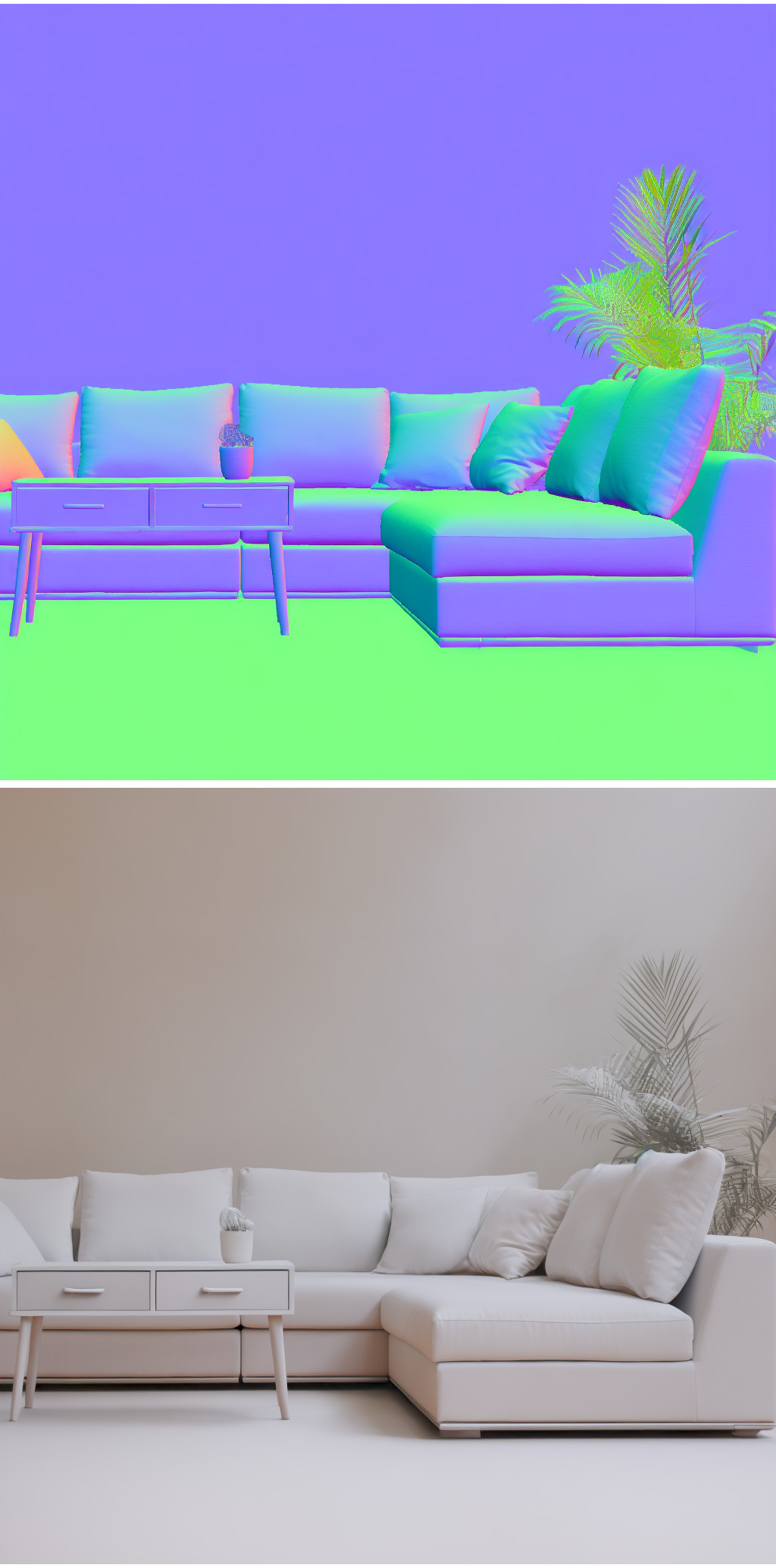}
                \put(11,90){\textcolor{white}{Normal}}
                \put(7,40){Irradiance}
            \end{overpic}
            \hfill
            \begin{overpic}[width=0.4\linewidth]{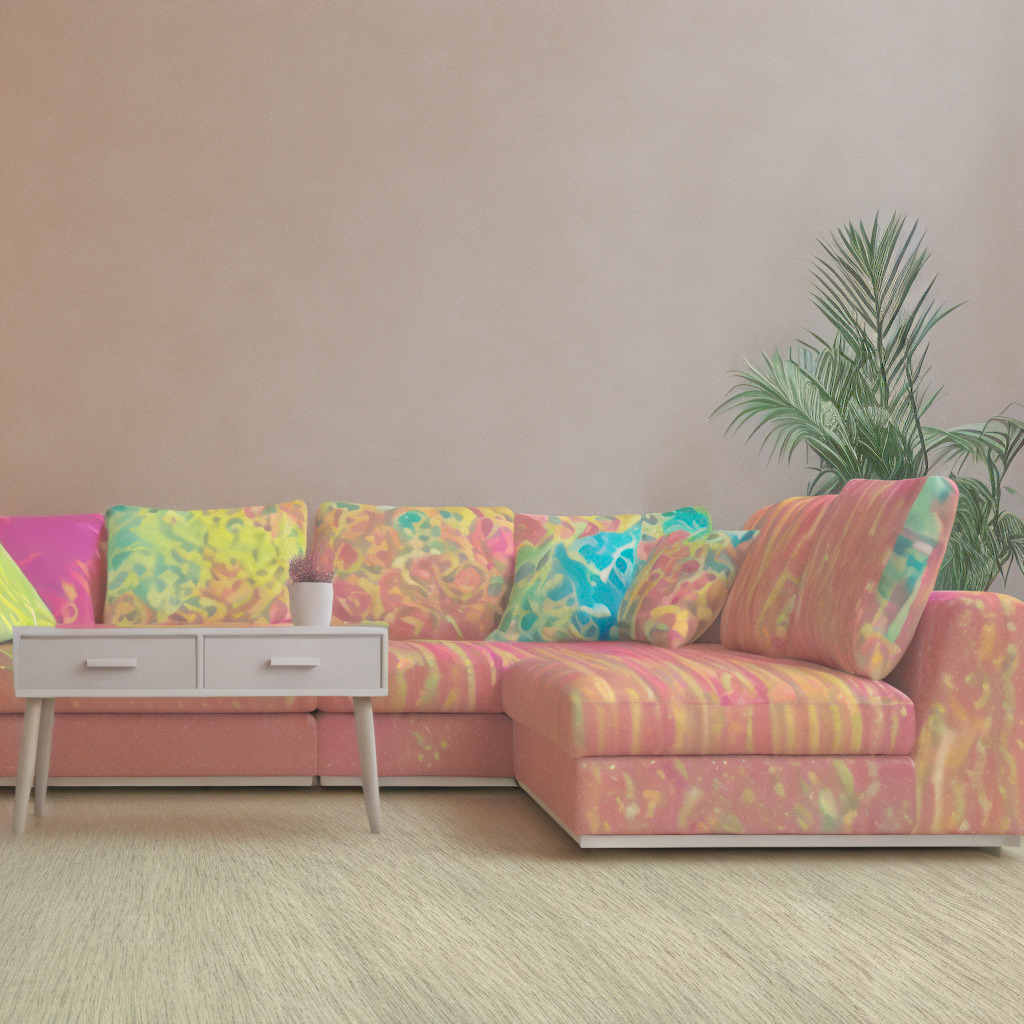}
                \put(25,102){\textbf{Our \xxtorgb}}
                \put(0,74){
                    \begin{tikzpicture}
                        \node[fill=black!10,fill opacity=0.8,rounded
                            corners=1ex, text width=2.1cm,align=left] {Prompt: \\ \texttt{\small``Colorful sofa''}};
                    \end{tikzpicture}
                }
            \end{overpic}
            \hfill
            \begin{overpic}[width=0.4\linewidth]{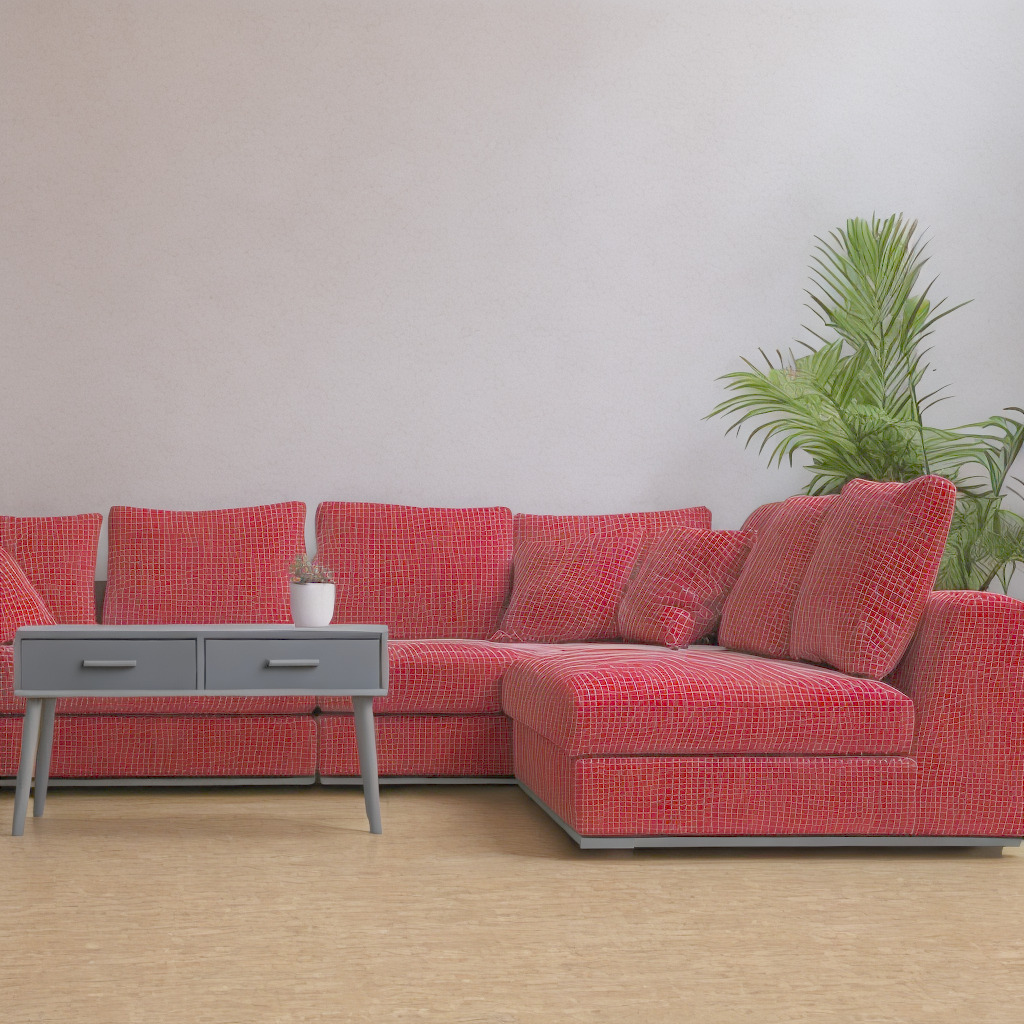}
                \put(25,102){\textbf{Our \xxtorgb}}
                \put(0,74){
                    \begin{tikzpicture}
                        \node[fill=black!10,fill opacity=0.8,rounded
                            corners=1ex, text width=1.4cm,align=left] {Prompt: \\ \texttt{\small``Red sofa''}};
                    \end{tikzpicture}
                }
            \end{overpic}
        }
    \end{minipage}
\end{minipage}

%% file: figures/mat-replacement/mat-replacement.tex
\begin{minipage}{1\linewidth}
    \begin{minipage}{1\linewidth}
        \begin{minipage}{\linewidth}
        \end{minipage}\par\smallskip
        \begin{overpic}[width=0.161\linewidth]{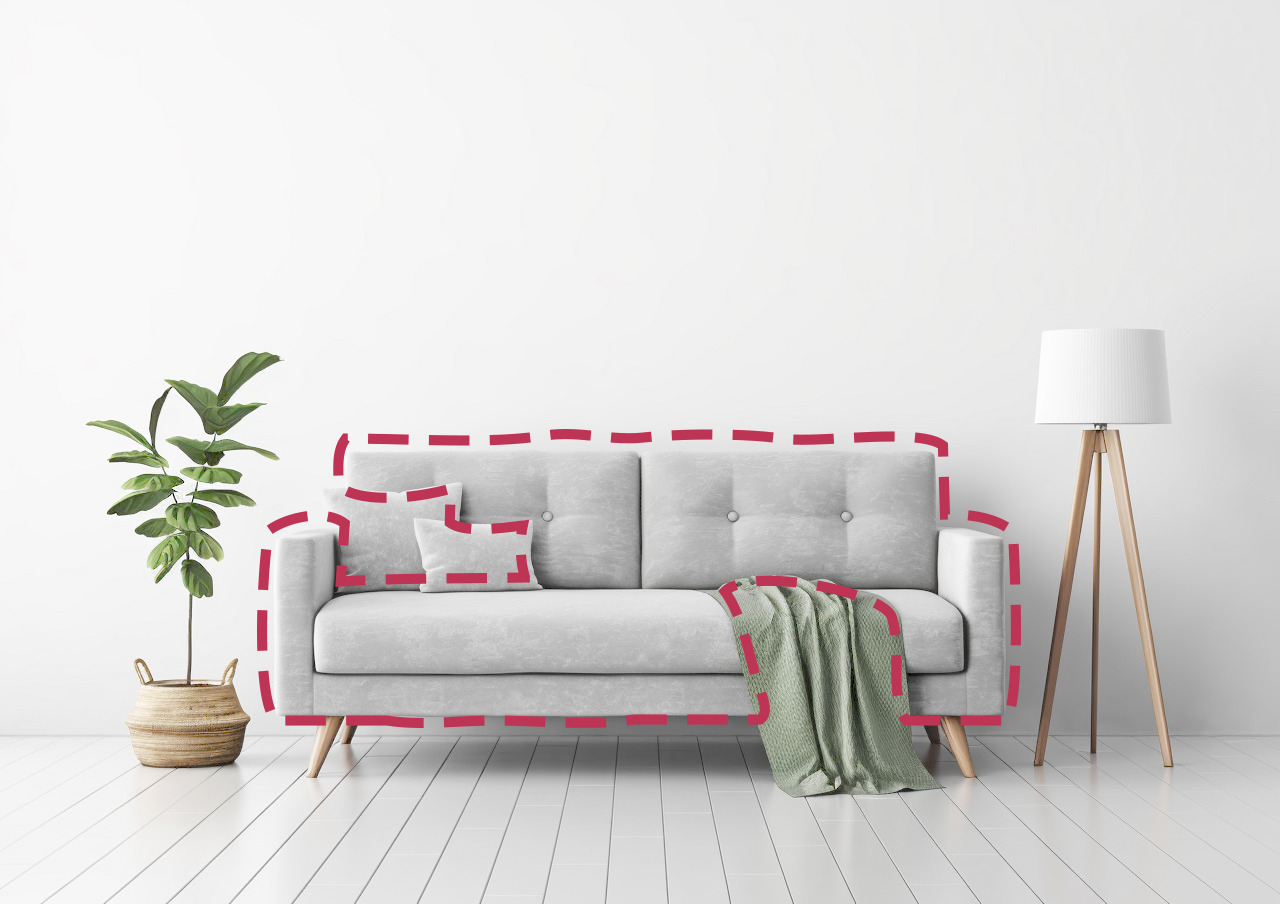}
            \put(5,74){\small Input image with mask}
        \end{overpic}
        \begin{overpic}[width=0.161\linewidth]{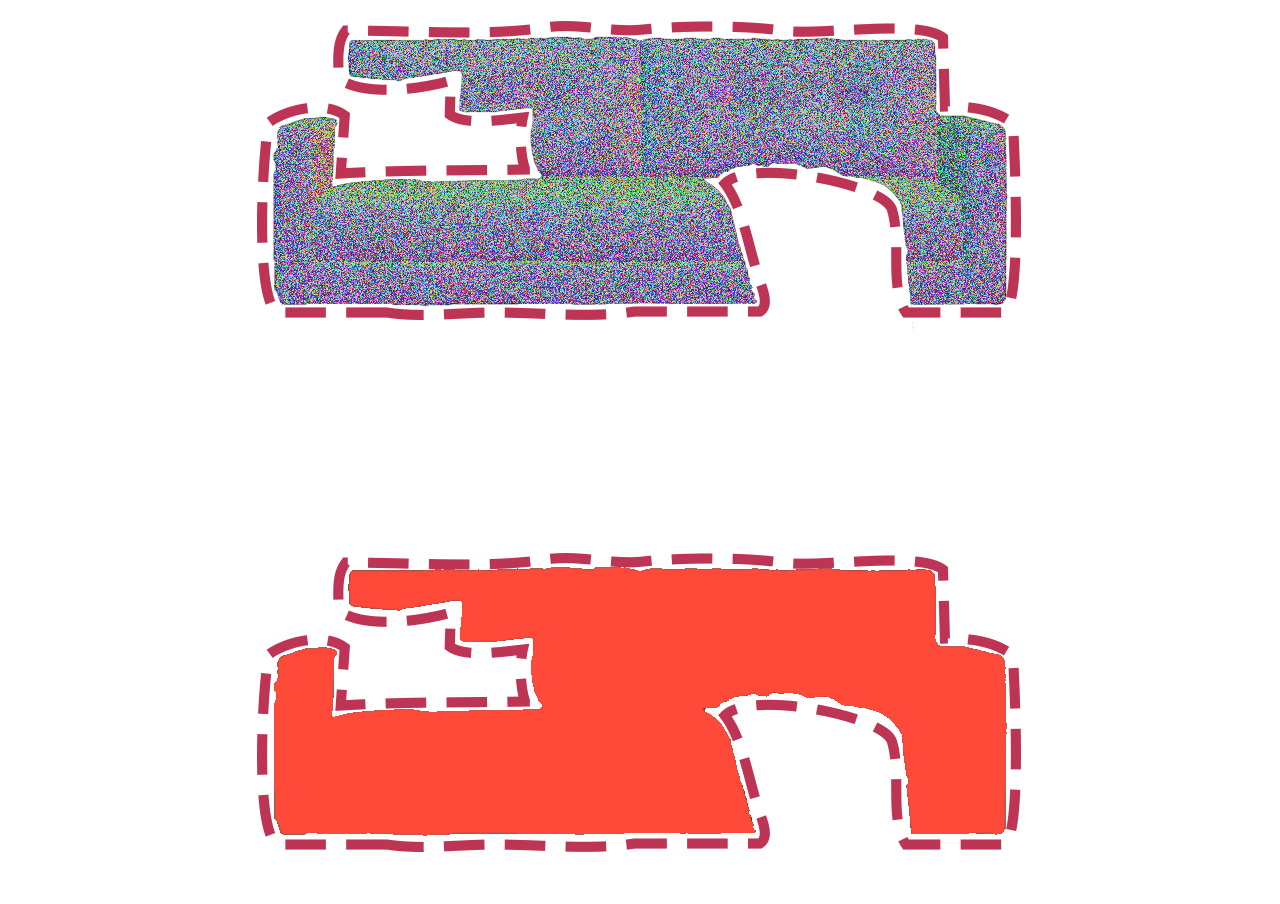}
            \put(13,74){\small Add noise to normal}
            \put(20,30){\small Change albedo}
            \put(80,35){\huge$\rightarrow$}
        \end{overpic}
        \begin{overpic}[width=0.161\linewidth]{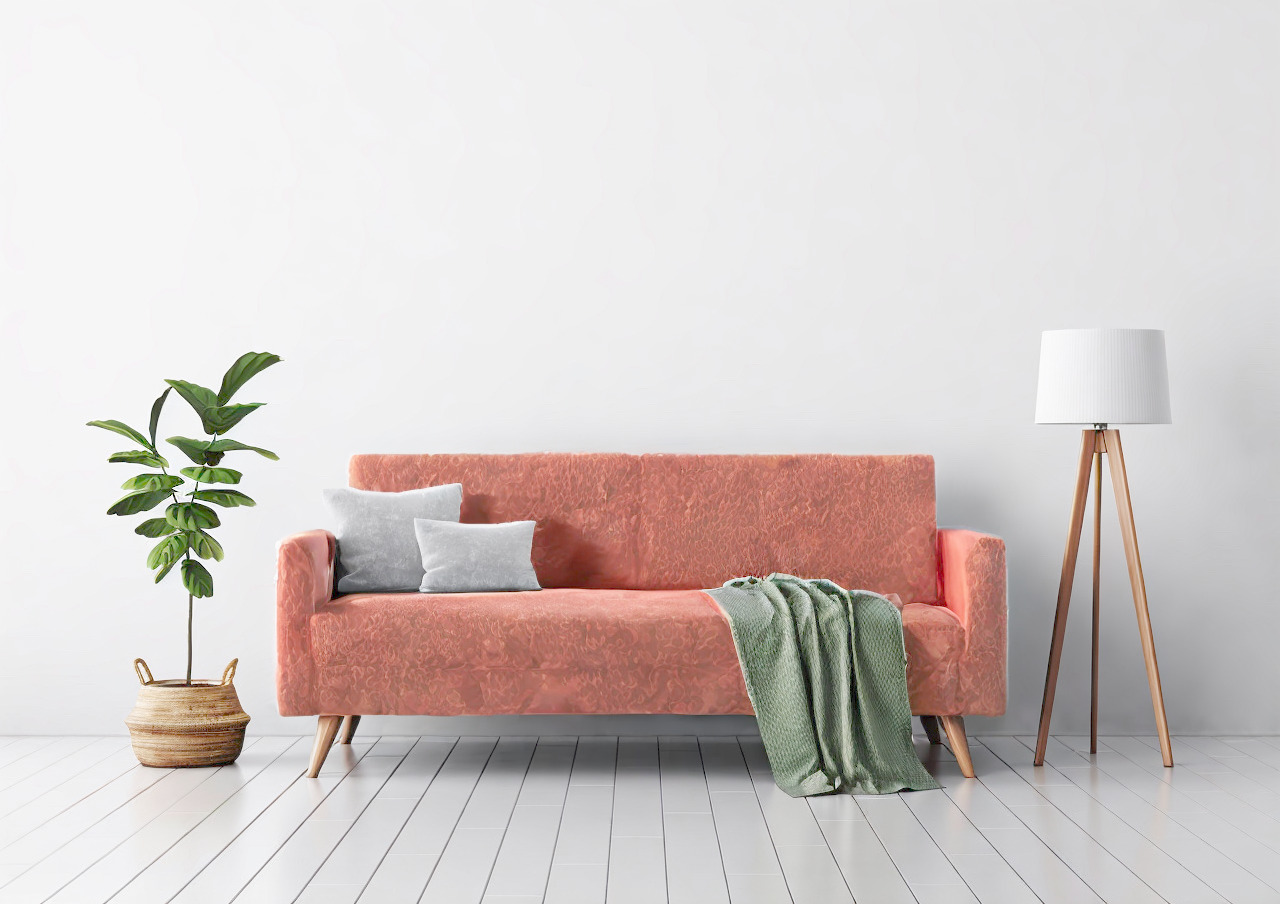}
            \put(14,74){\textbf{\rgbtoxxtorgb}}
            \put(52,22){\color{red}%
                \frame{\includegraphics[scale=.125]{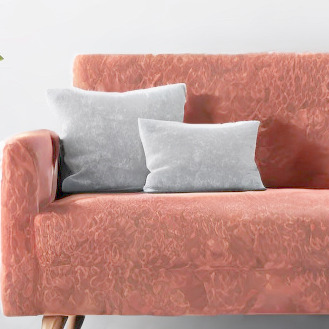}}}
            \put(20,14){\linethickness{0.15mm}\color{red}%
                \polygon(0,0)(25,0)(25,25)(0,25)}
            \put(35,39){\linethickness{0.1mm}\color{red}\line(1,1.3){16.75}}%
            \put(45,18){\linethickness{0.1mm}\color{red}\line(1,.15){25.5}}%
        \end{overpic}
        \begin{overpic}[width=0.165\linewidth]{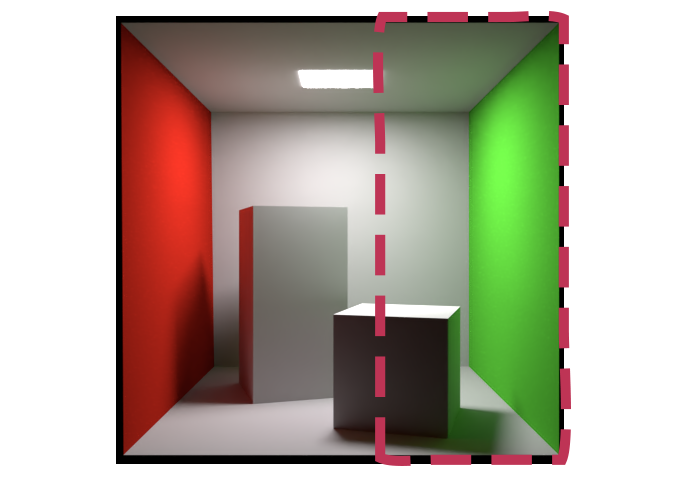}
            \put(5,74){\small Input image with mask}
        \end{overpic}
        \begin{overpic}[width=0.165\linewidth]{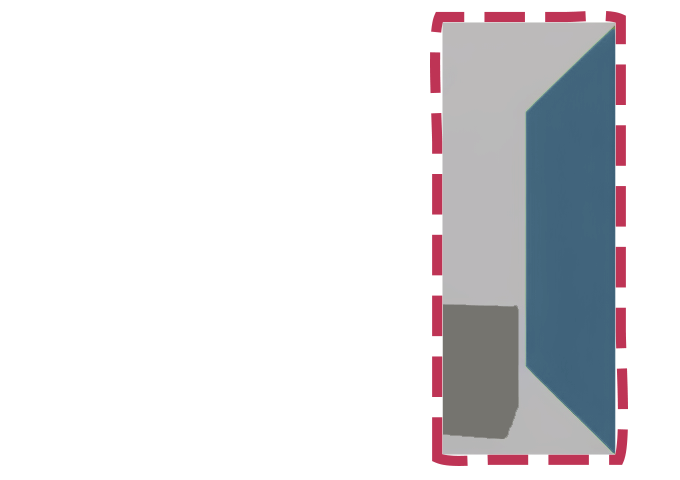}
            \put(-11,40){\small Change albedo}
            \put(-11,30){\small of the right wall}
            \put(-11,20){\small to blue}
            \put(95,30){\huge$\rightarrow$}
        \end{overpic}
        \begin{overpic}[width=0.165\linewidth]{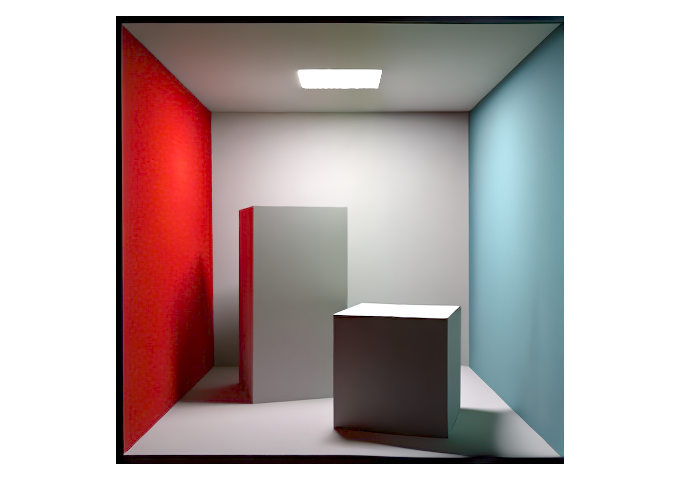}
            \put(14,74){\textbf{\rgbtoxxtorgb}}
        \end{overpic}\par\bigskip\smallskip
        \begin{overpic}[width=0.195\linewidth]{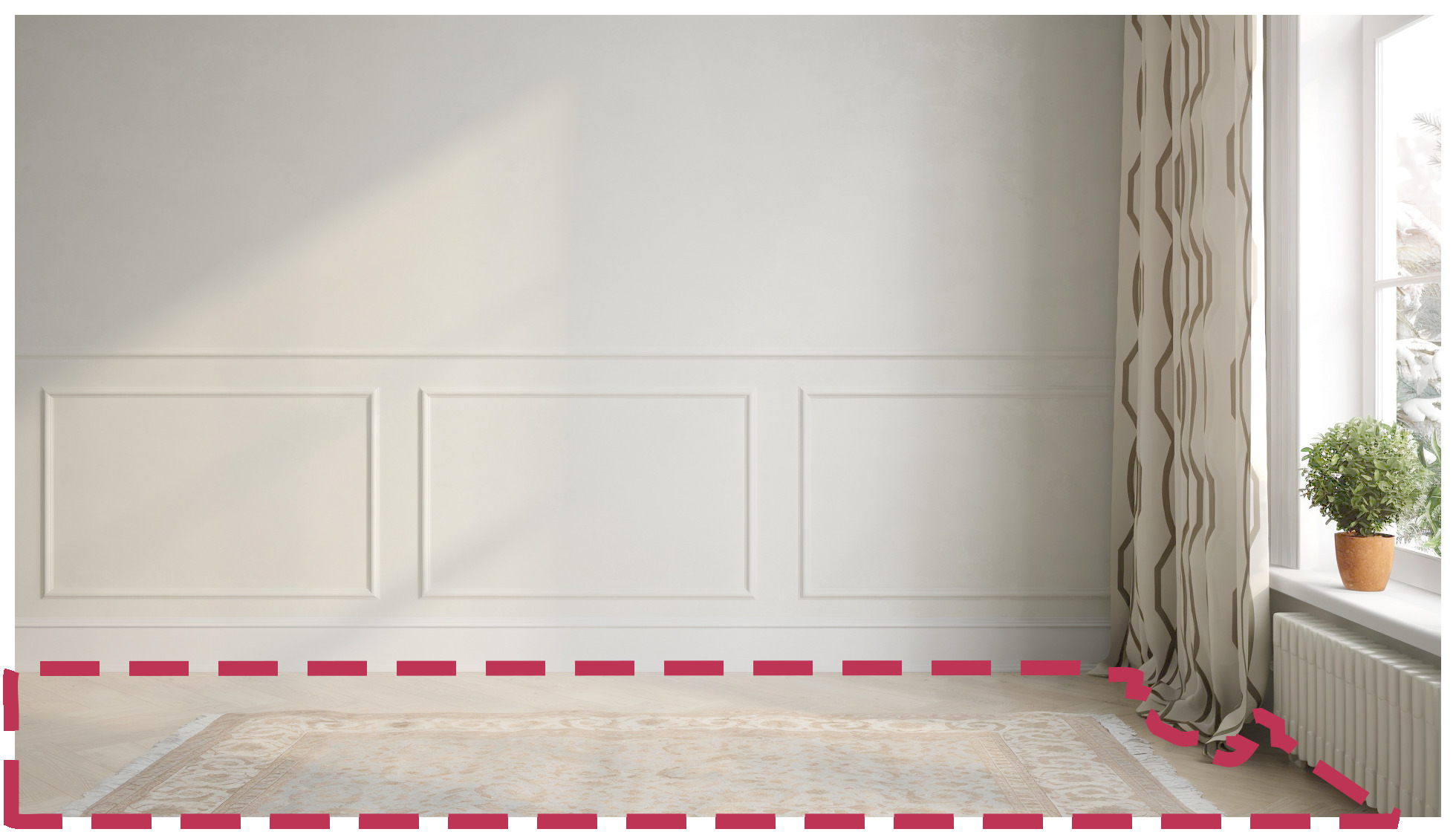}
            \put(12,58){\small Input image with mask}
        \end{overpic}
        \begin{overpic}[width=0.195\linewidth]{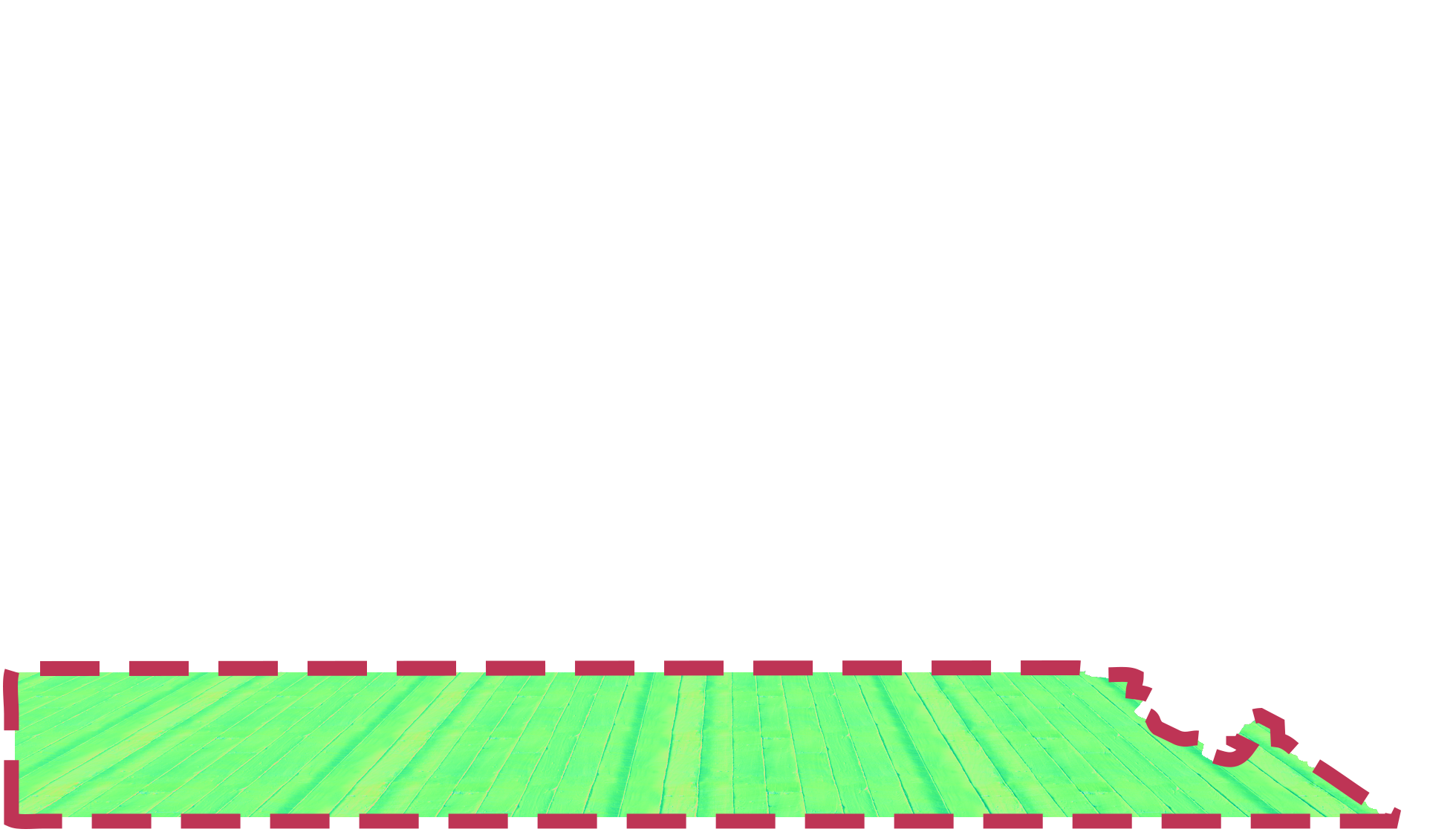}
            \put(20,15){\small Change normal}
            \put(80,25){\huge$\rightarrow$}
        \end{overpic}
        \begin{overpic}[width=0.195\linewidth]{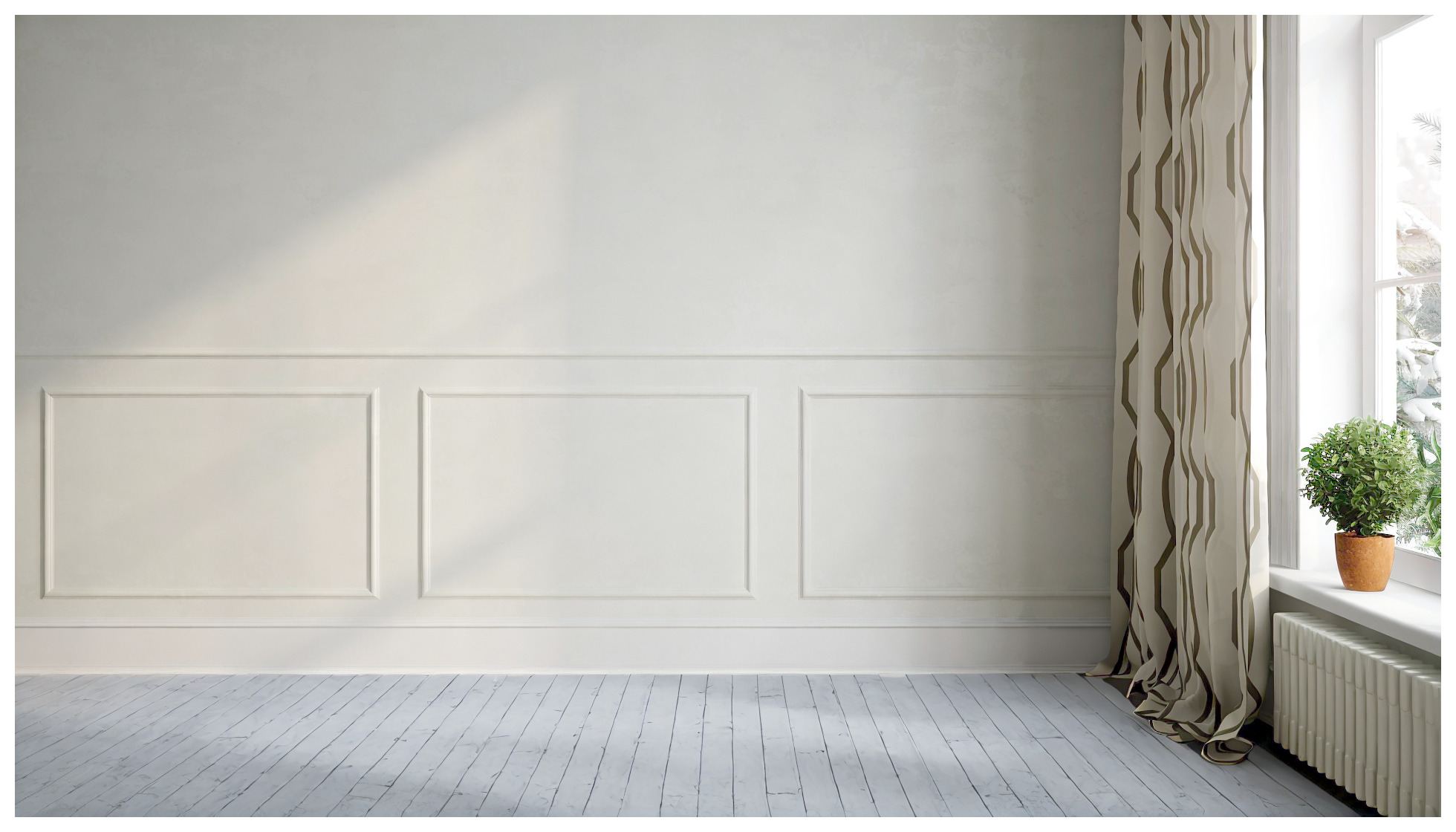}
            \put(18,58){\textbf{\rgbtoxxtorgb}}
            \put(46,21){\color{red}%
                \frame{\includegraphics[scale=.45]{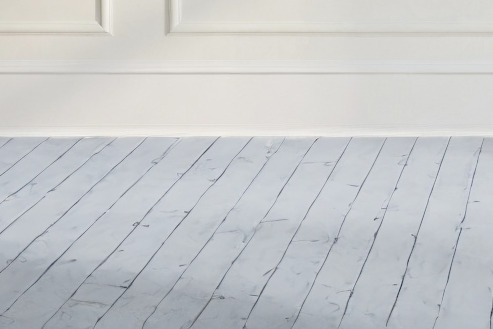}}}
            \put(21,2){\linethickness{0.15mm}\color{red}%
                \polygon(0,0)(24,0)(24,16)(0,16)}
            \put(38, 18){\linethickness{0.1mm}\color{red}\line(1,1.75){7.75}}%
            \put(45,12){\linethickness{0.1mm}\color{red}\line(1,.4){22}}%

        \end{overpic}
        \begin{overpic}[width=0.195\linewidth]{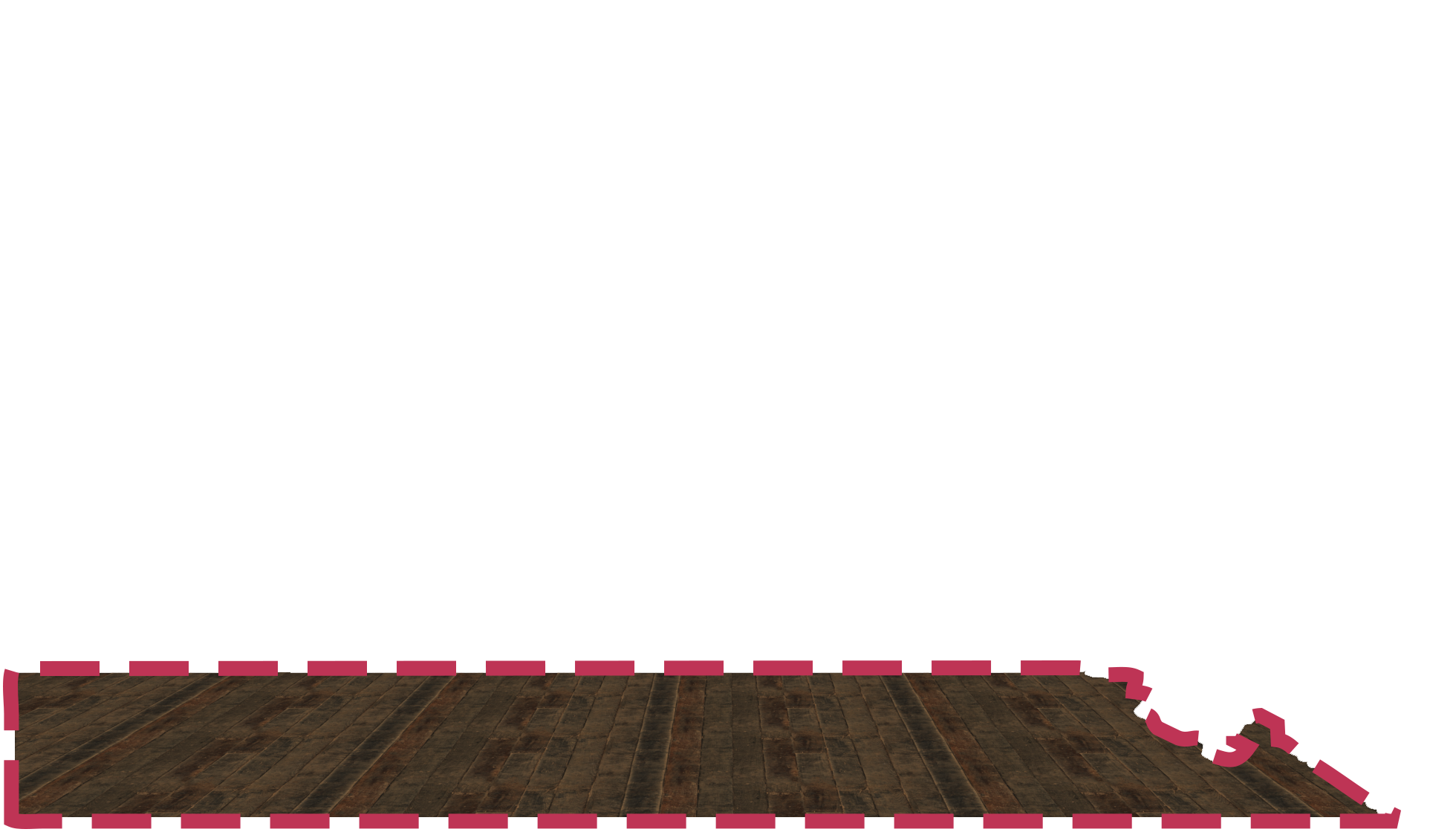}
            \put(15,15){\small Then change albedo}
            \put(80,25){\huge$\rightarrow$}
        \end{overpic}
        \begin{overpic}[width=0.195\linewidth]{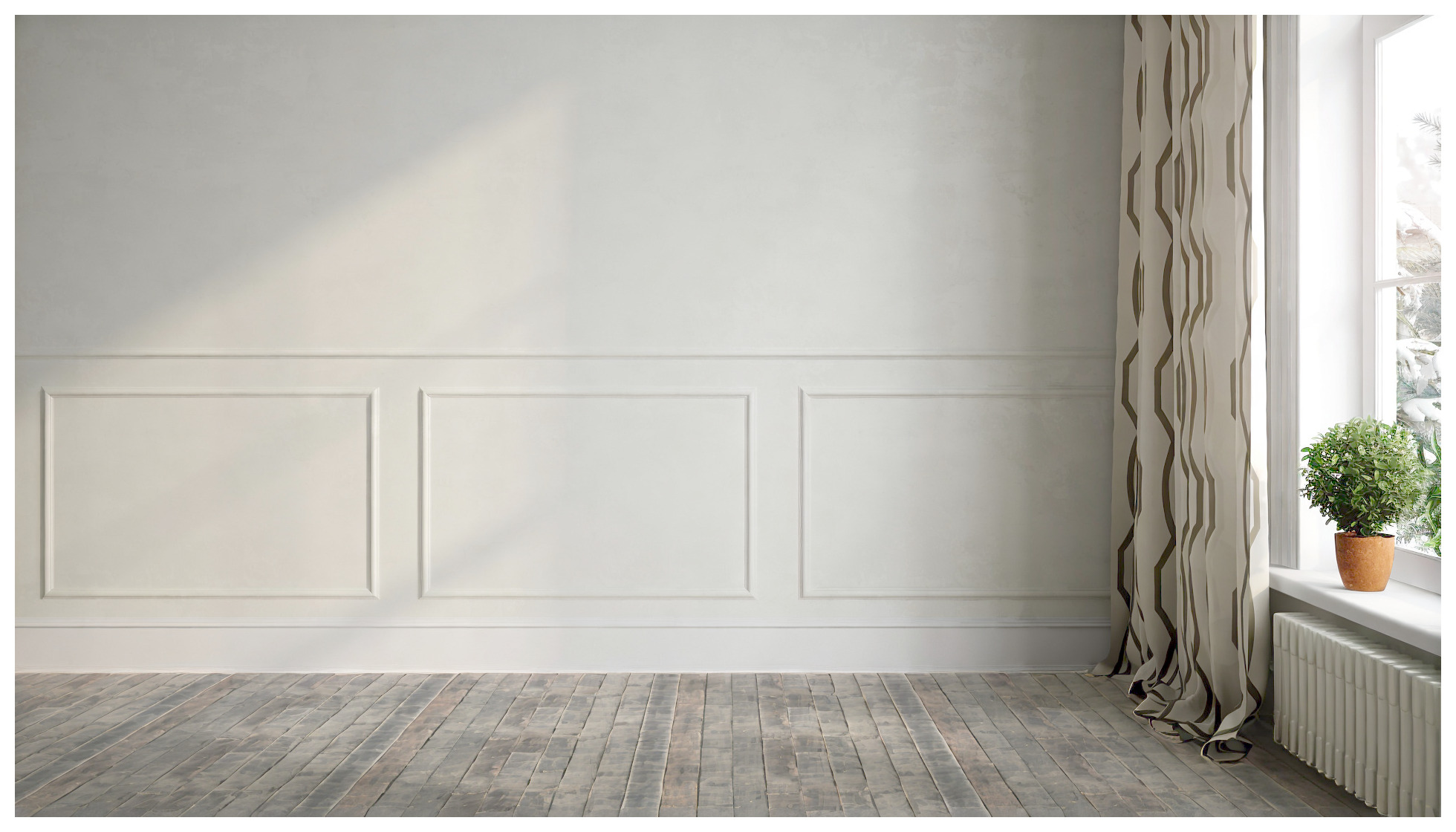}
            \put(18,58){\textbf{\rgbtoxxtorgb}}
            \put(46,21){\color{red}%
                \frame{\includegraphics[scale=.45]{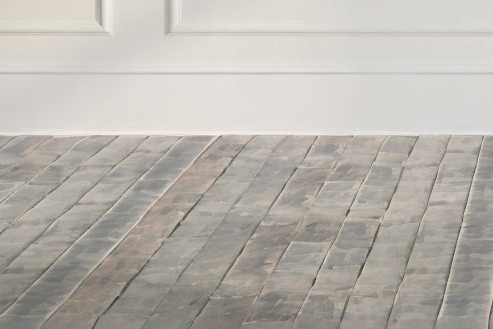}}}
            \put(21,2){\linethickness{0.15mm}\color{red}%
                \polygon(0,0)(24,0)(24,16)(0,16)}
            \put(38, 18){\linethickness{0.1mm}\color{red}\line(1,1.75){7.75}}%
            \put(45,12){\linethickness{0.1mm}\color{red}\line(1,.4){22}}%

        \end{overpic}
    \end{minipage}\par
\end{minipage}

%% file: supp/failure-cases/failure-cases.tex
\begin{minipage}{1\linewidth}
    \begin{minipage}{1\linewidth}
        \begin{minipage}{\linewidth}
        \end{minipage}\par\medskip
        \centering
        \subfloat[A real photo with a resolution of 2048 $\times$ 1536 and large object scales.]{
            \begin{overpic}[width=0.166\linewidth]{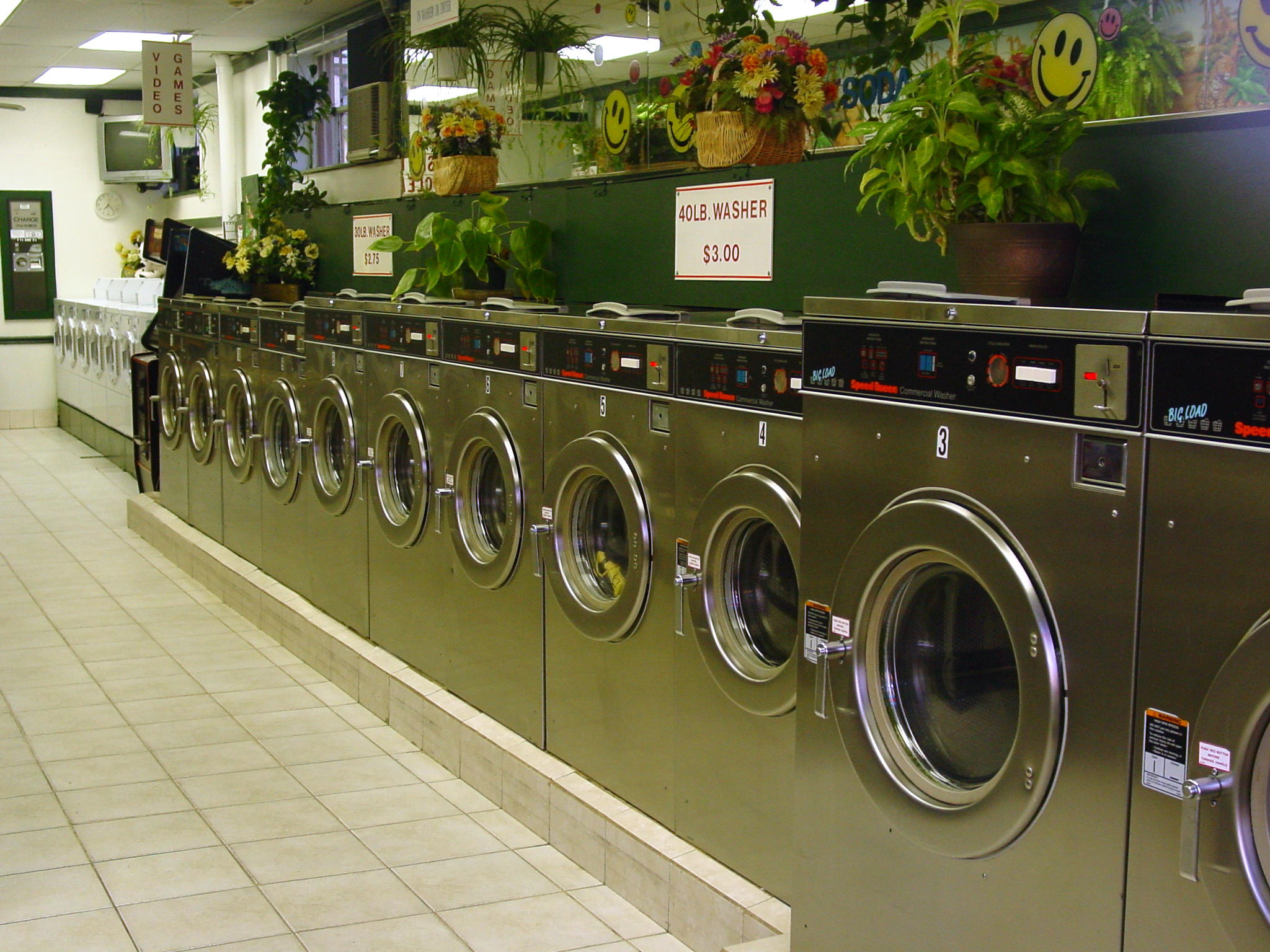}
                \put(25,79){Input image}
            \end{overpic}
            \hfill
            \begin{overpic}[width=0.166\linewidth]{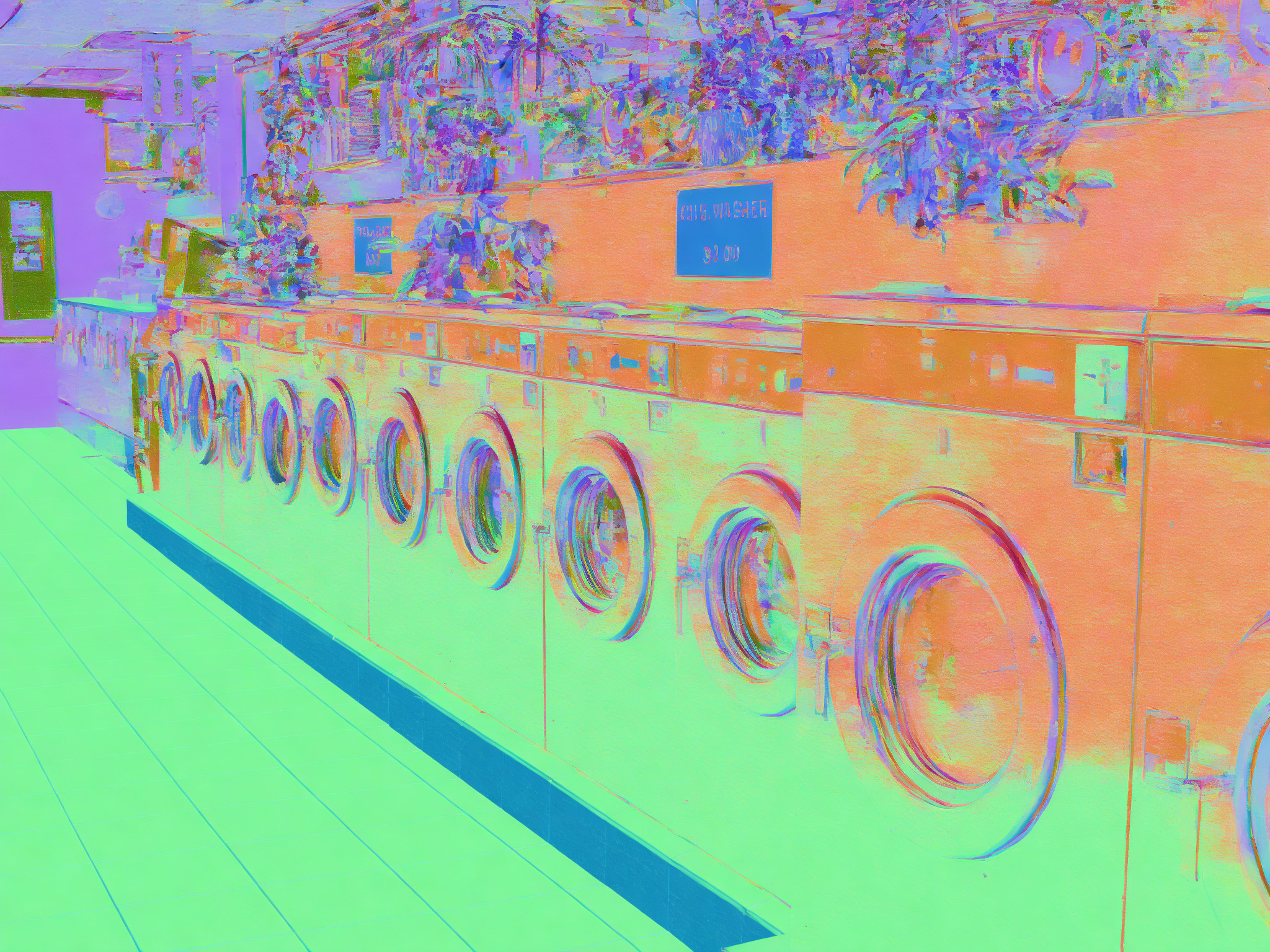}
                \put(0,79){\textbf{Our \rgbtoxx normal} }
            \end{overpic}
            \hfill
            \begin{overpic}[width=0.166\linewidth]{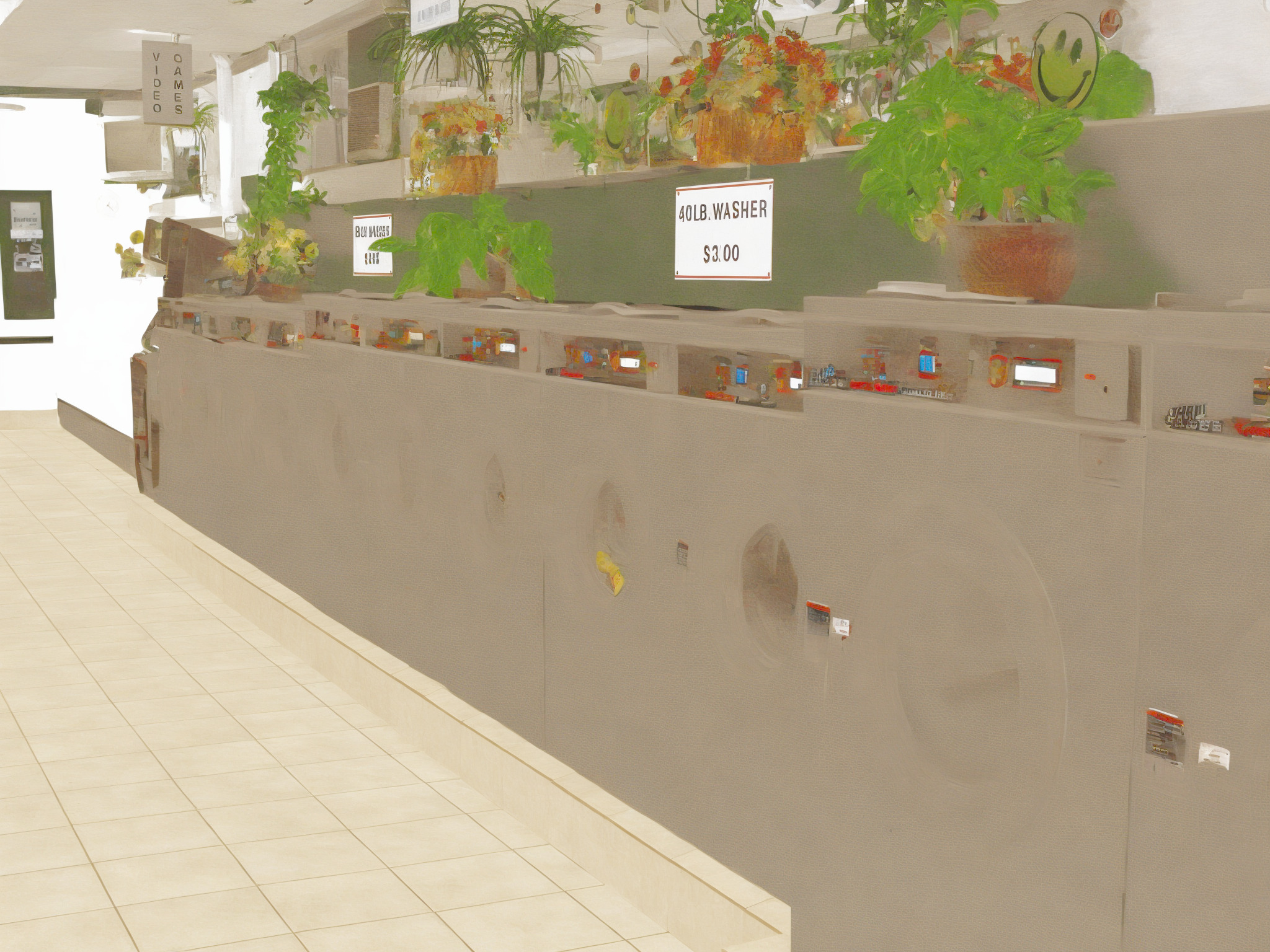}
                \put(2,79){\textbf{Our \rgbtoxx albedo}}
            \end{overpic}
            \hfill
            \begin{overpic}[width=0.166\linewidth]{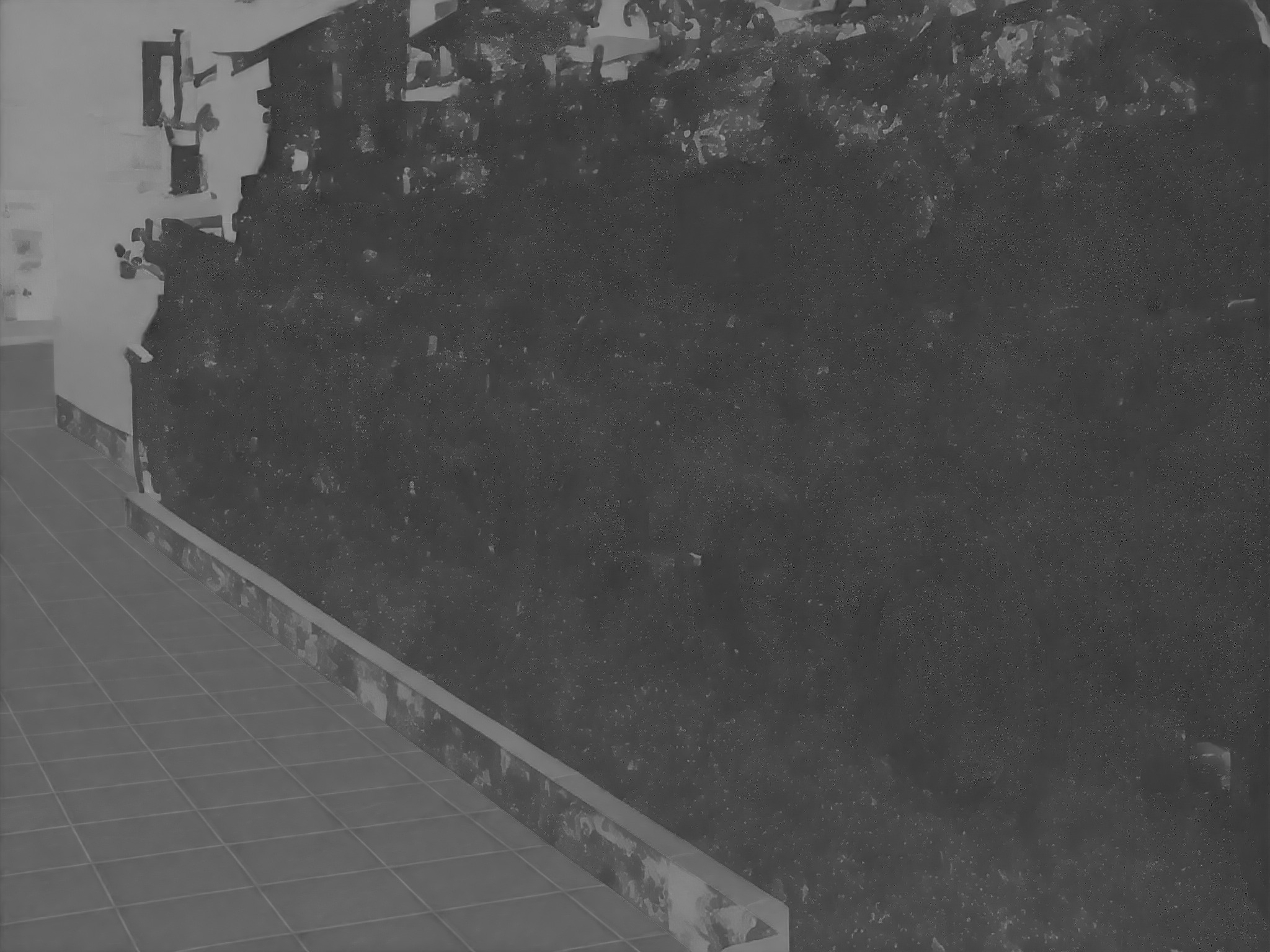}
                \put(2,79){\textbf{Our \rgbtoxx rough.}}
            \end{overpic}
            \hfill
            \begin{overpic}[width=0.166\linewidth]{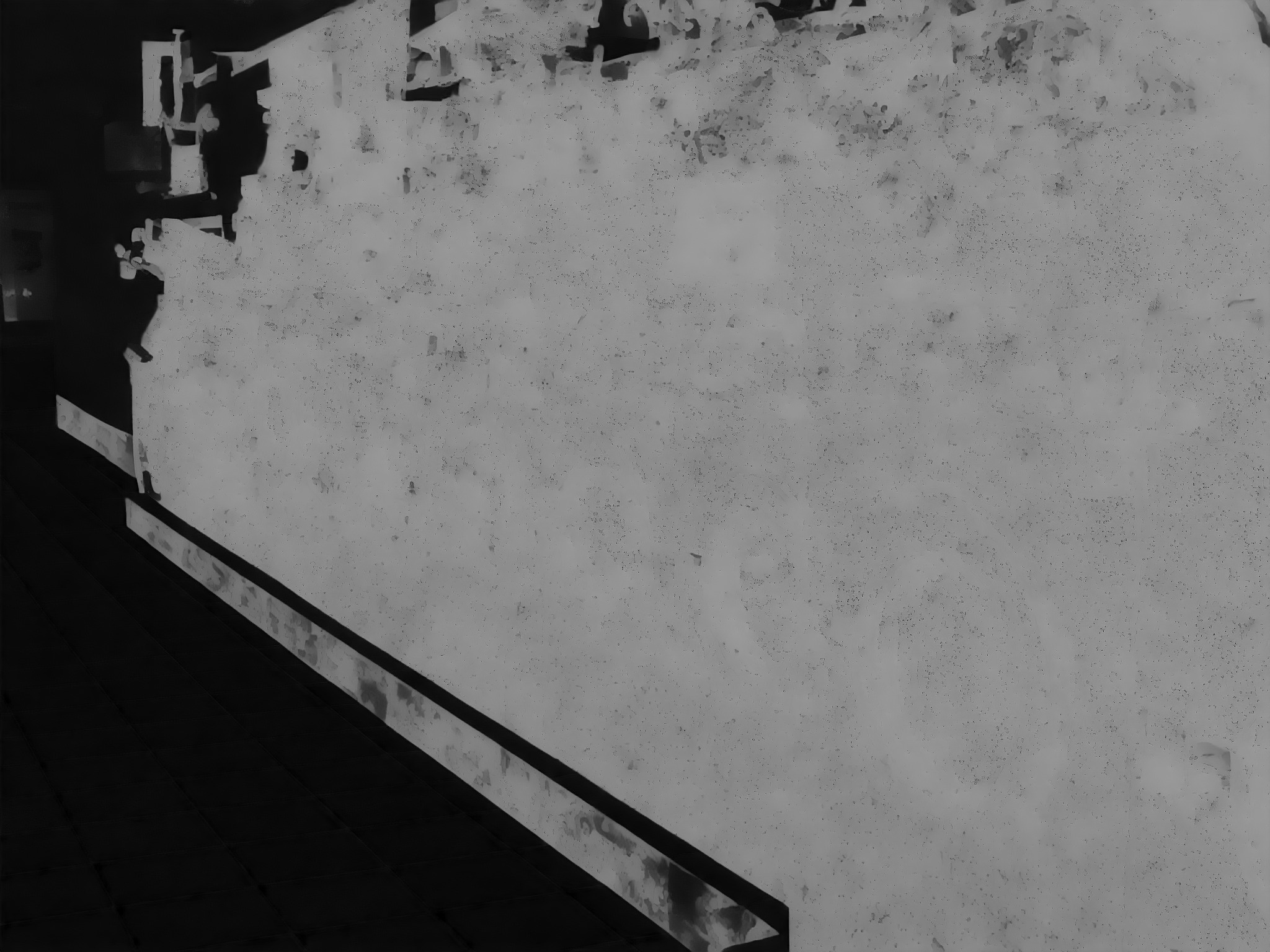}
                \put(2,79){\textbf{Our \rgbtoxx metal.}}
            \end{overpic}
            \hfill
            \begin{overpic}[width=0.166\linewidth]{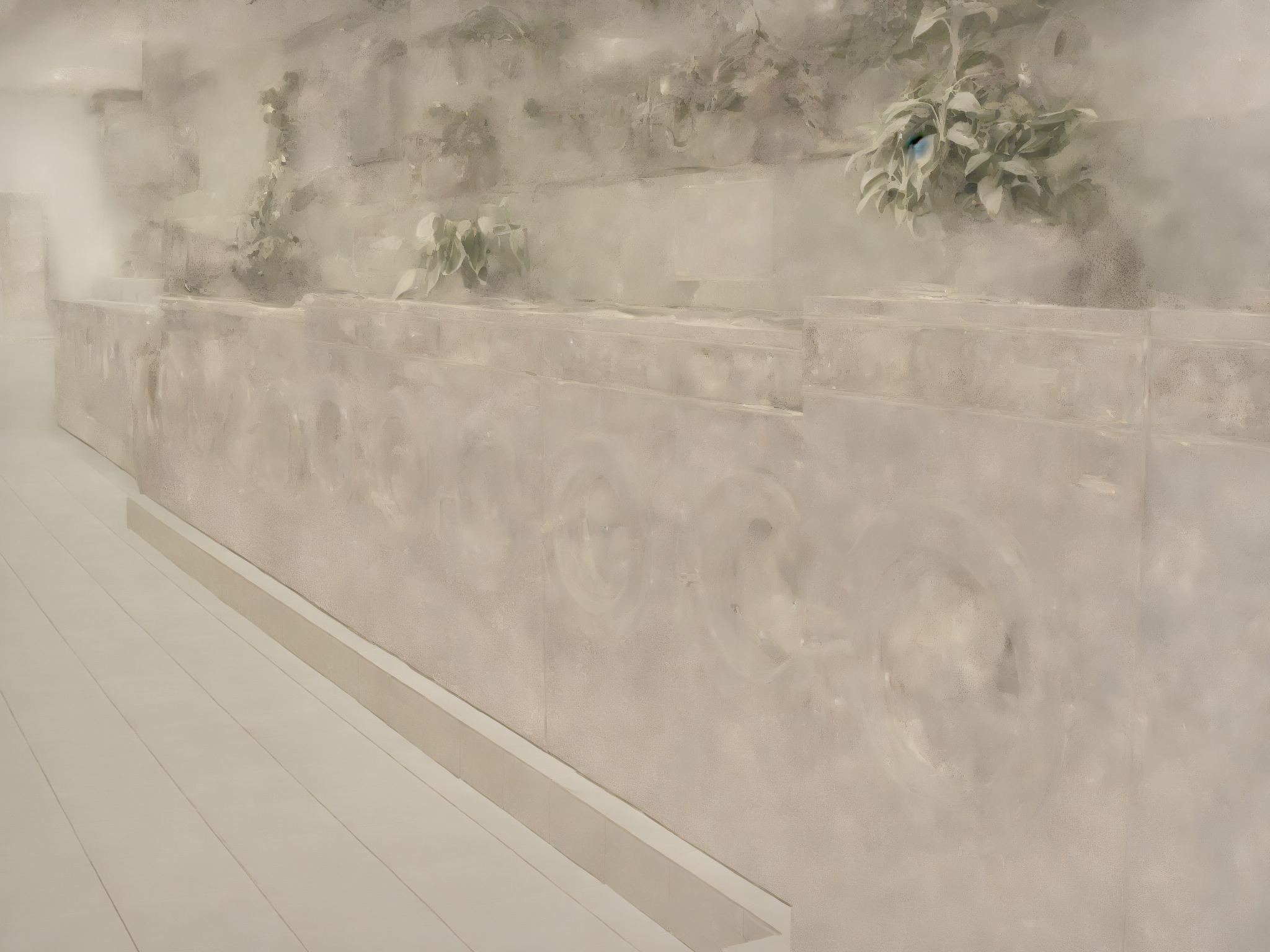}
                \put(4,79){\textbf{Our \rgbtoxx irra.}}
            \end{overpic}
        }
    \end{minipage}\par
    \begin{minipage}{1\linewidth}
        \begin{minipage}{\linewidth}
        \end{minipage}\par\medskip
        \centering
        \subfloat[A real photo full of humans that is out of our data distribution.]{
            \begin{overpic}[width=0.166\linewidth]{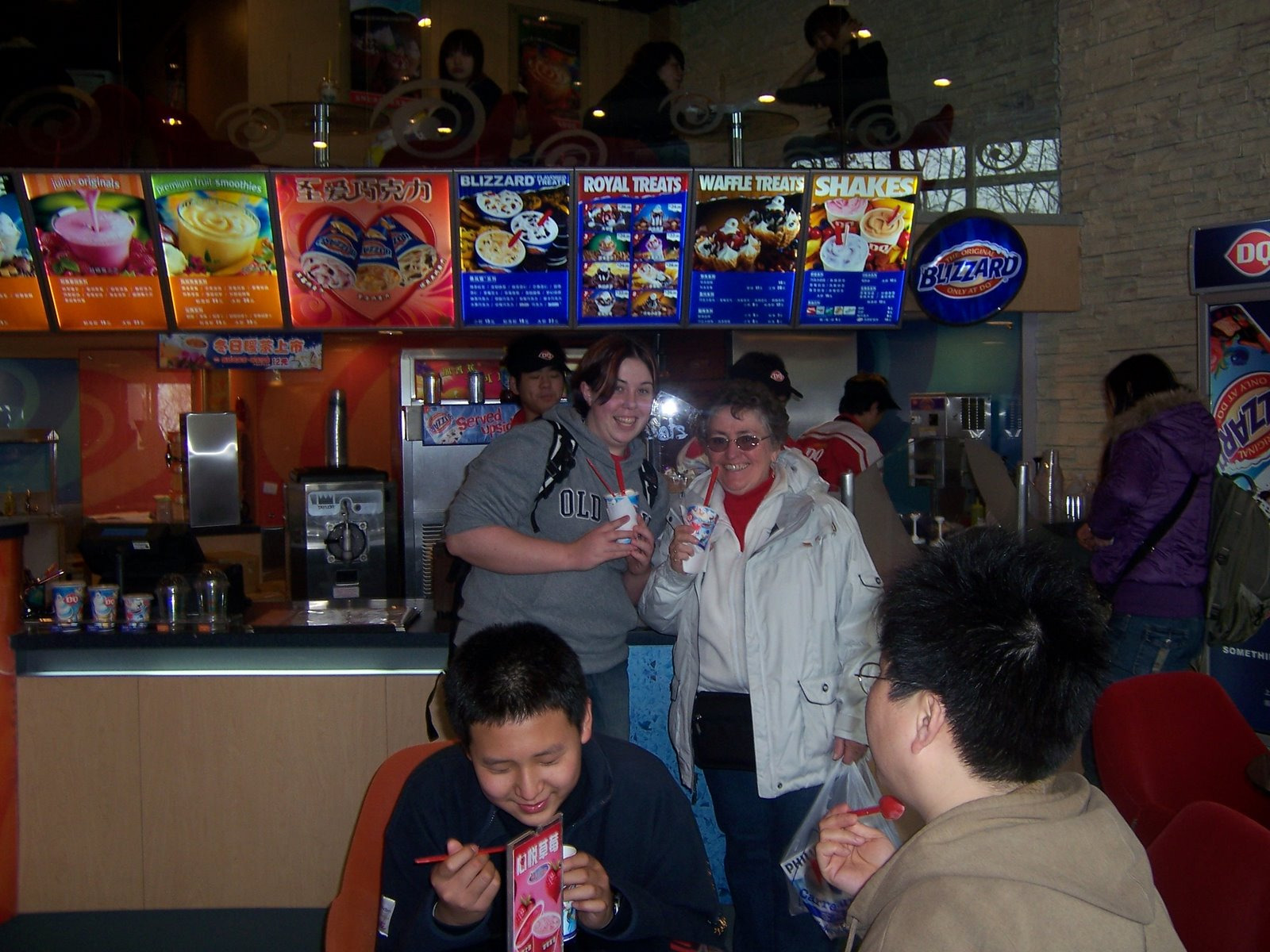}
                \put(25,79){Input image}
            \end{overpic}
            \hfill
            \begin{overpic}[width=0.166\linewidth]{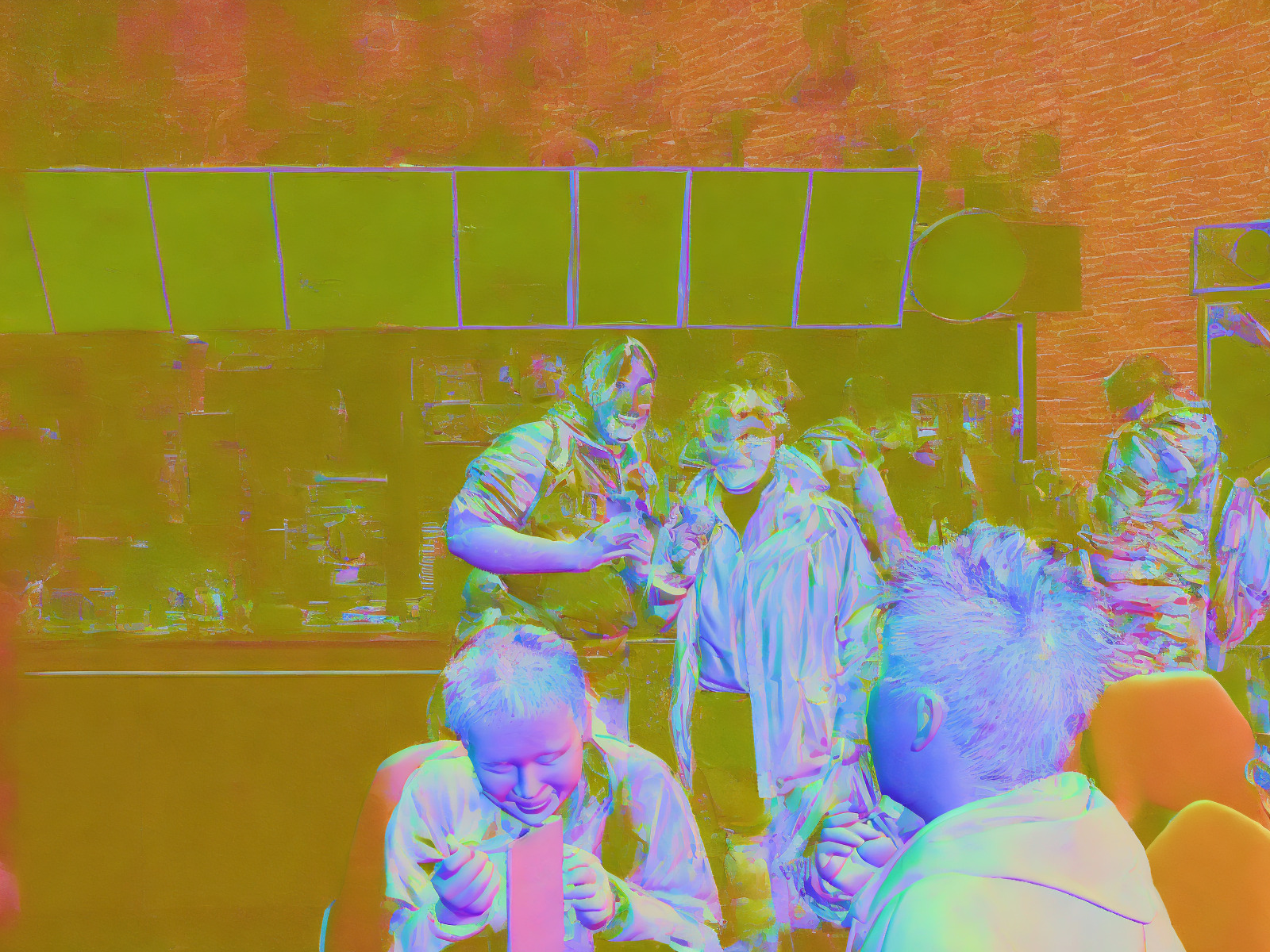}
                \put(0,79){\textbf{Our \rgbtoxx normal} }
            \end{overpic}
            \hfill
            \begin{overpic}[width=0.166\linewidth]{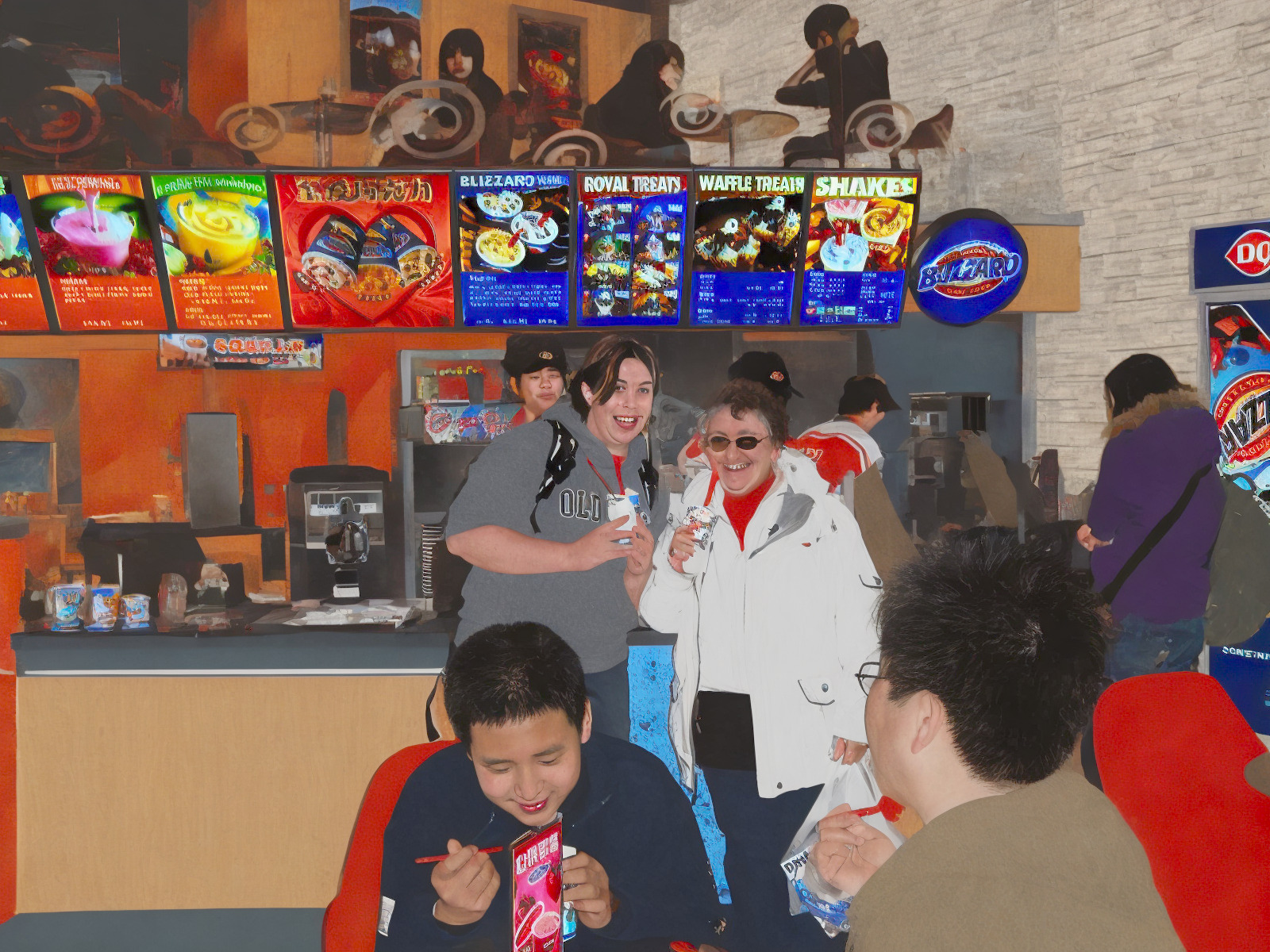}
                \put(2,79){\textbf{Our \rgbtoxx albedo}}
            \end{overpic}
            \hfill
            \begin{overpic}[width=0.166\linewidth]{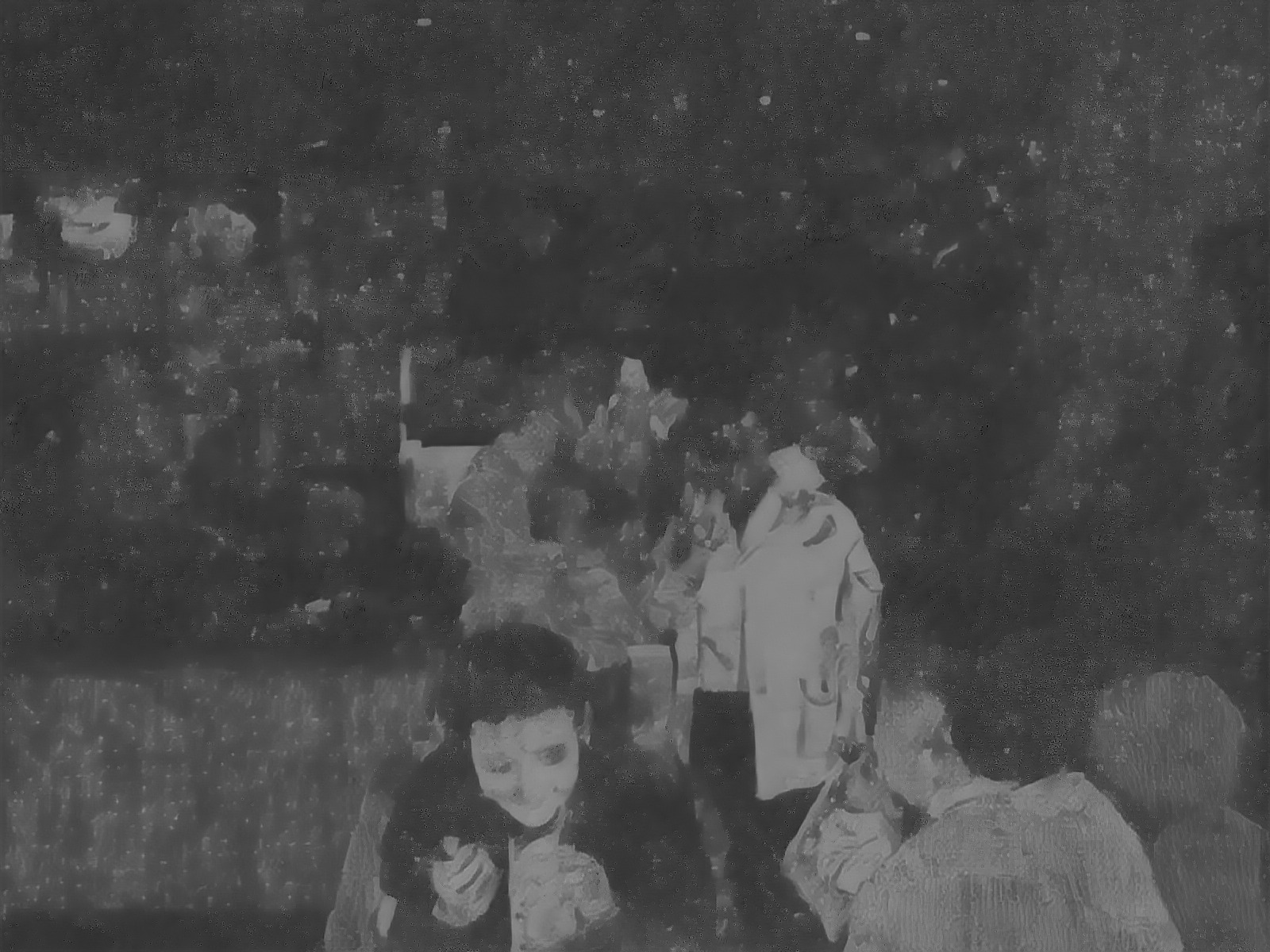}
                \put(2,79){\textbf{Our \rgbtoxx rough.}}
            \end{overpic}
            \hfill
            \begin{overpic}[width=0.166\linewidth]{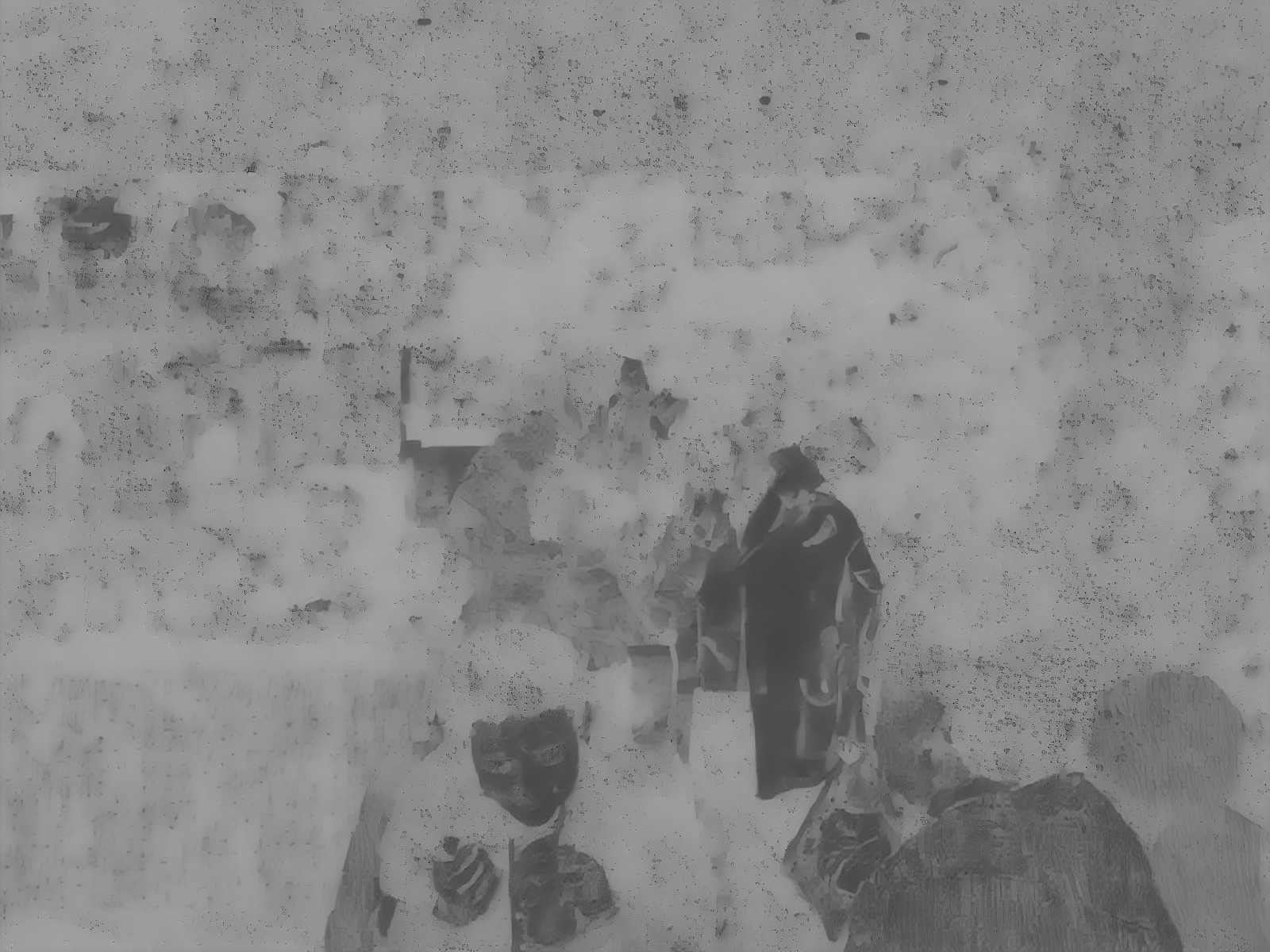}
                \put(2,79){\textbf{Our \rgbtoxx metal.}}
            \end{overpic}
            \hfill
            \begin{overpic}[width=0.166\linewidth]{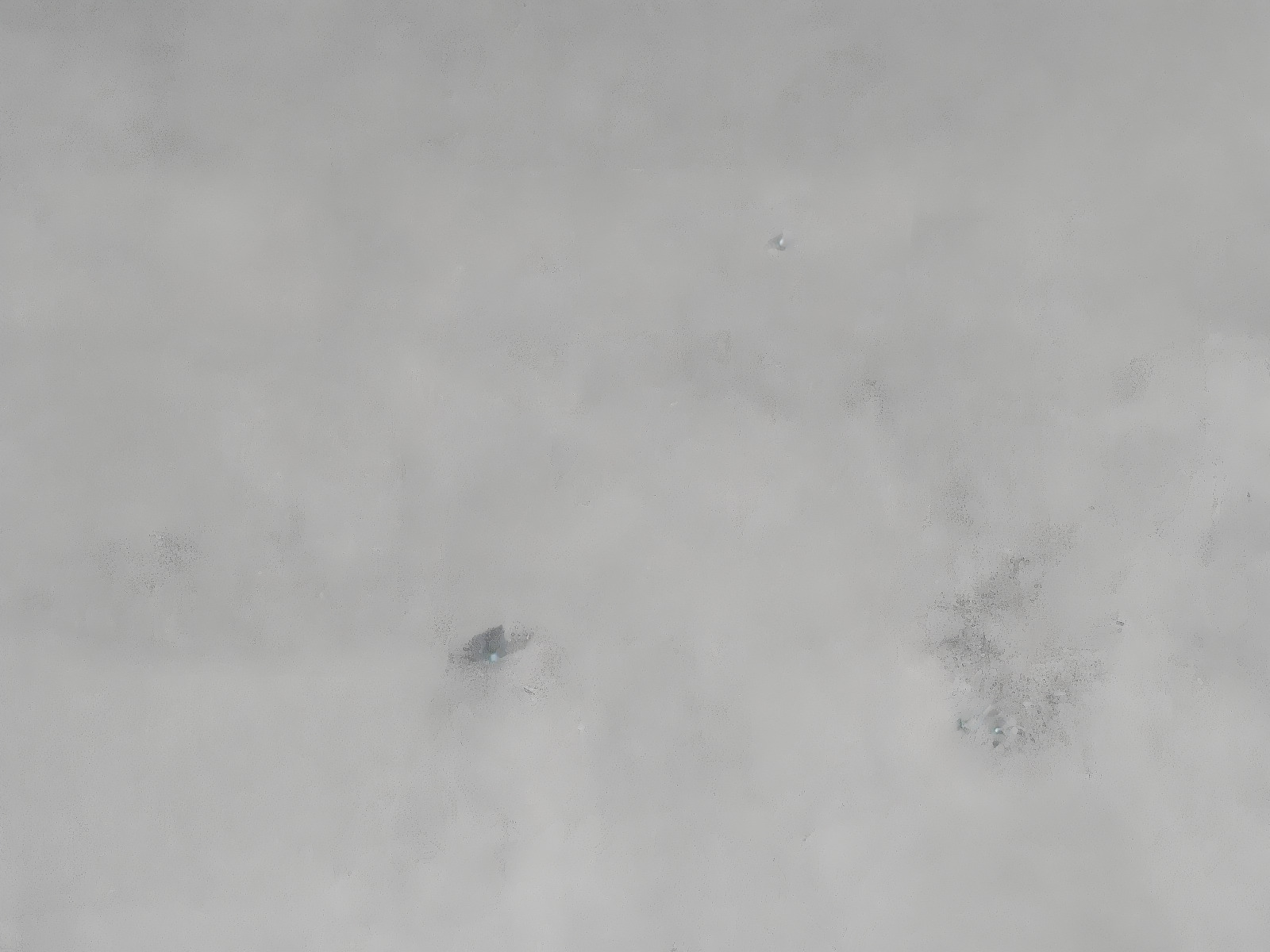}
                \put(4,79){\textbf{Our \rgbtoxx irra.}}
            \end{overpic}
        }
    \end{minipage}\par
    \begin{minipage}{1\linewidth}
        \begin{minipage}{\linewidth}
        \end{minipage}\par\medskip
        \centering
        \subfloat[A real photo with a resolution of 2048 $\times$ 1536, most of which is covered by mirrors and strong highlights.]{
            \begin{overpic}[width=0.166\linewidth]{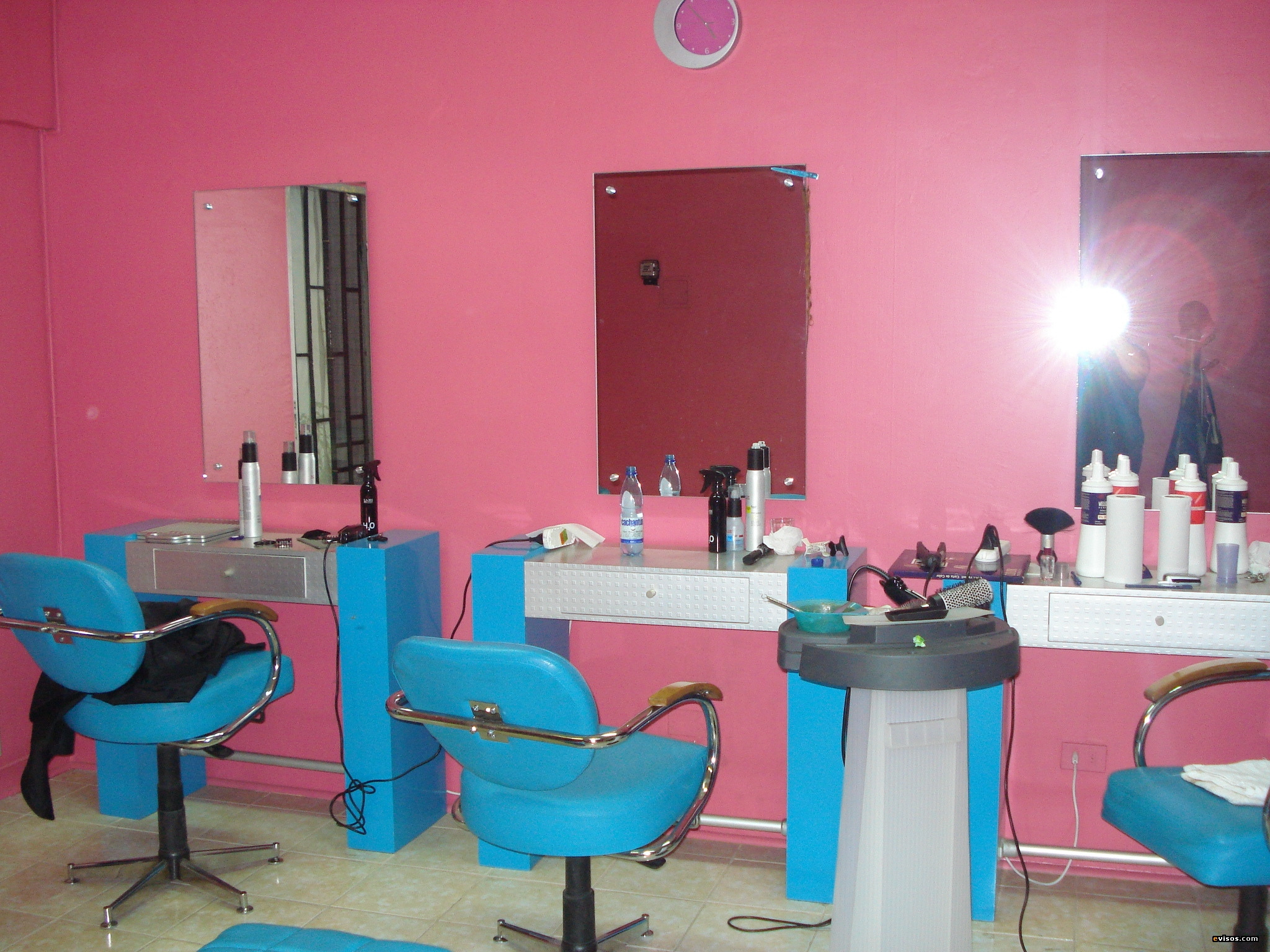}
                \put(25,79){Input image}
            \end{overpic}
            \hfill
            \begin{overpic}[width=0.166\linewidth]{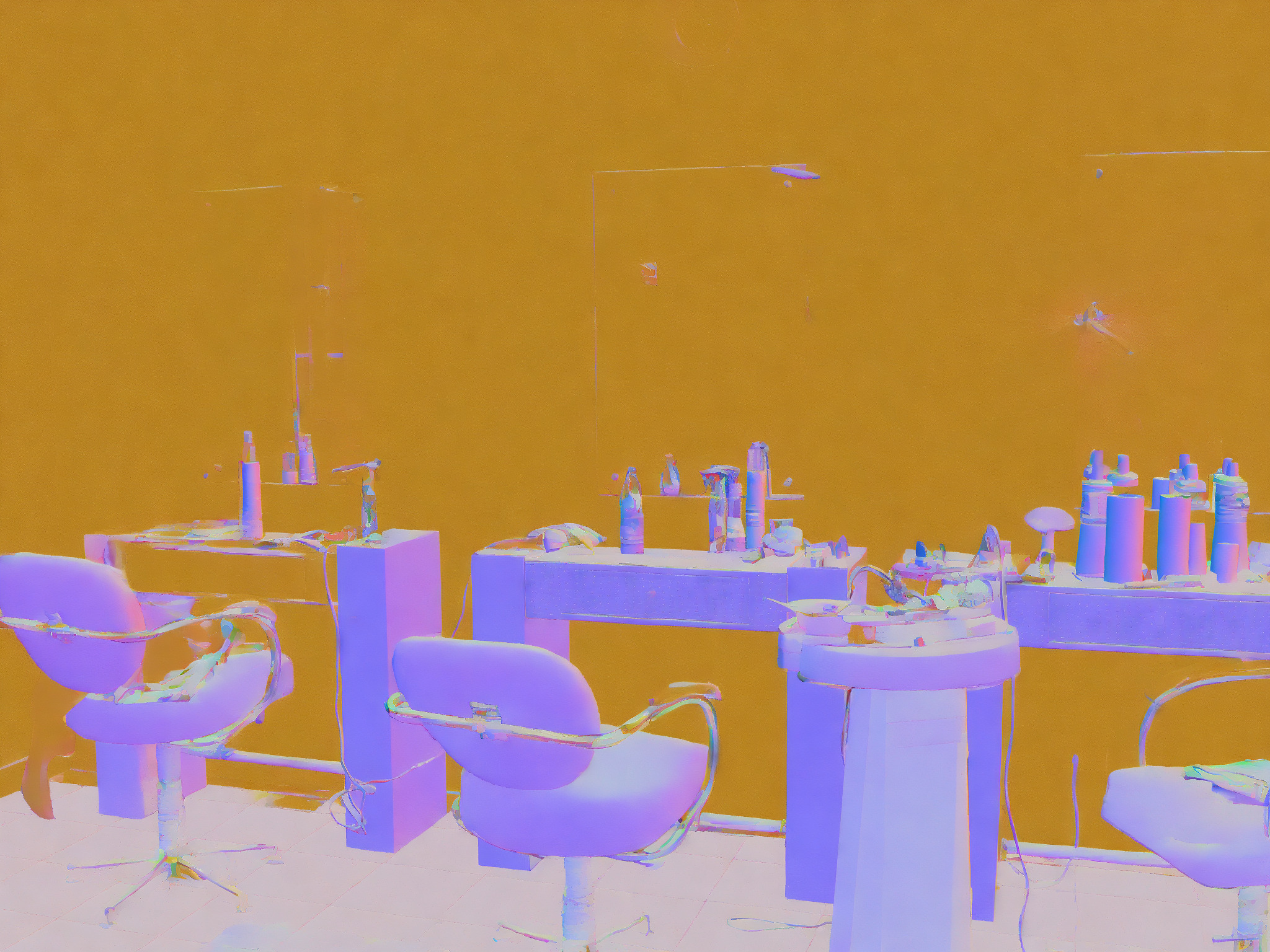}
                \put(0,79){\textbf{Our \rgbtoxx normal} }
            \end{overpic}
            \hfill
            \begin{overpic}[width=0.166\linewidth]{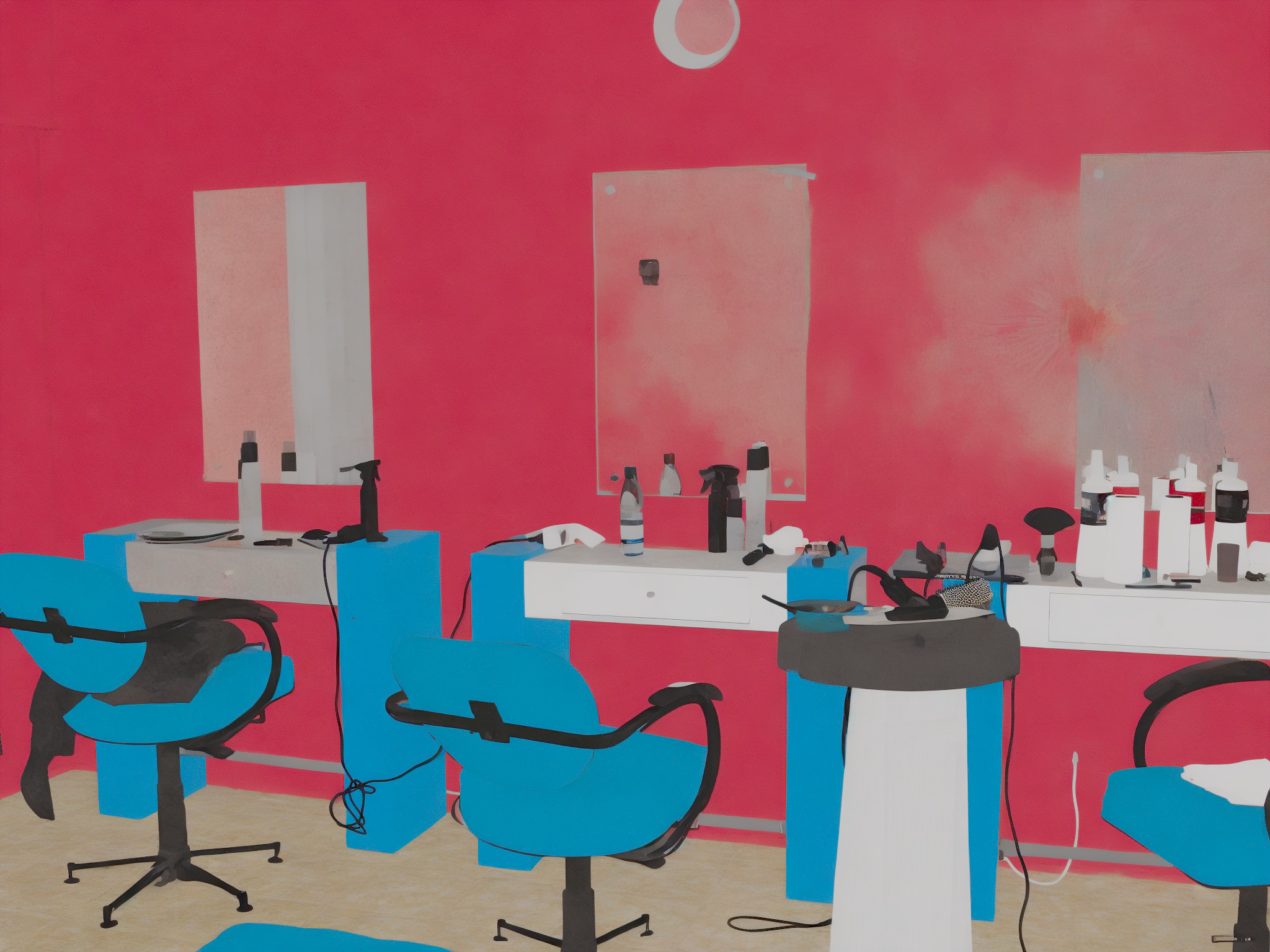}
                \put(2,79){\textbf{Our \rgbtoxx albedo}}
            \end{overpic}
            \hfill
            \begin{overpic}[width=0.166\linewidth]{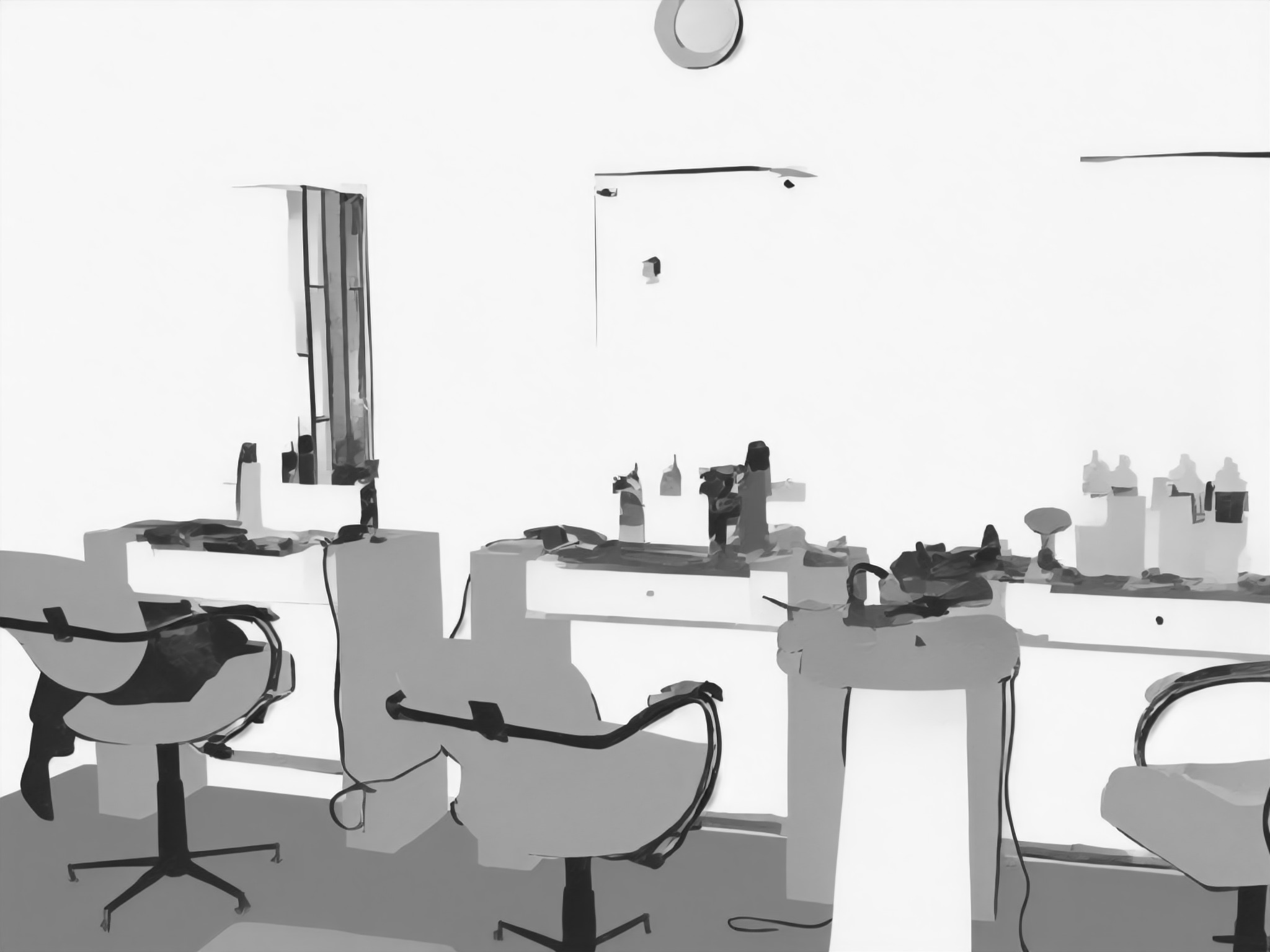}
                \put(2,79){\textbf{Our \rgbtoxx rough.}}
            \end{overpic}
            \hfill
            \begin{overpic}[width=0.166\linewidth]{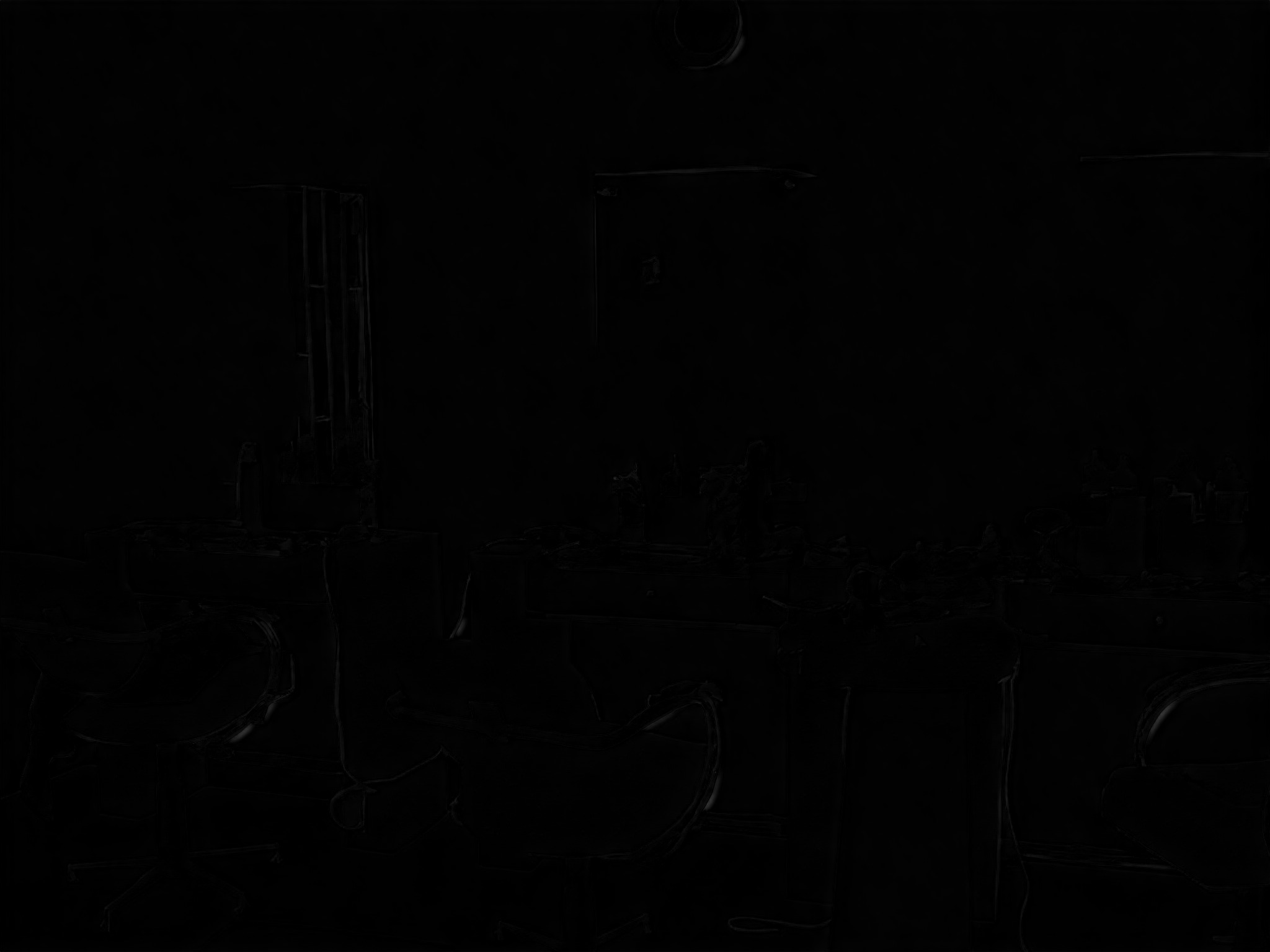}
                \put(2,79){\textbf{Our \rgbtoxx metal.}}
            \end{overpic}
            \hfill
            \begin{overpic}[width=0.166\linewidth]{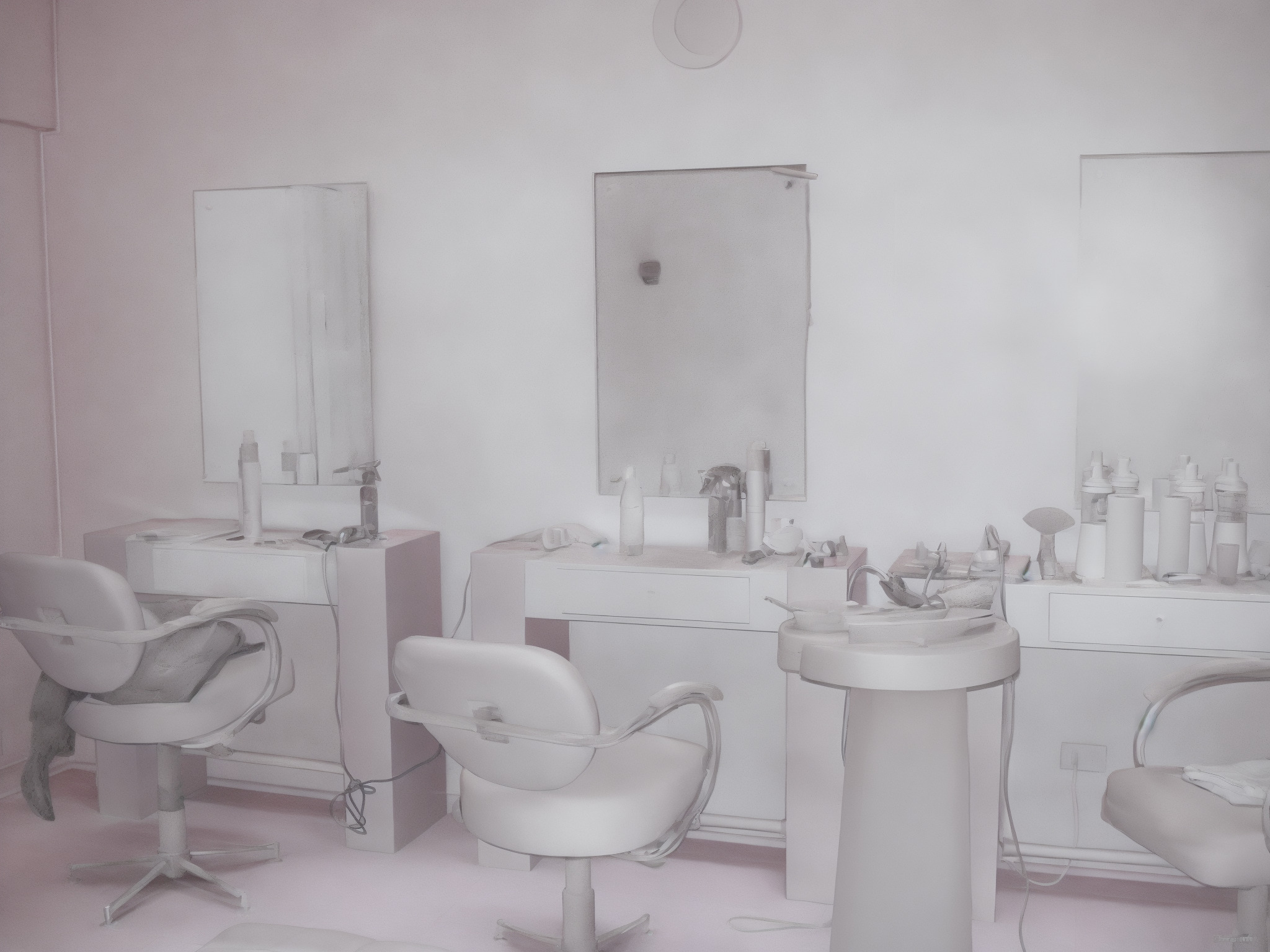}
                \put(4,79){\textbf{Our \rgbtoxx irra.}}
            \end{overpic}
        }
    \end{minipage}\par
    \begin{minipage}{1\linewidth}
        \begin{minipage}{\linewidth}
        \end{minipage}\par\medskip
        \centering
        \subfloat[A real photo although indoor, is high-resolution (2048 $\times$ 1536), extremely distorted, and grainy.]{
            \begin{overpic}[width=0.166\linewidth]{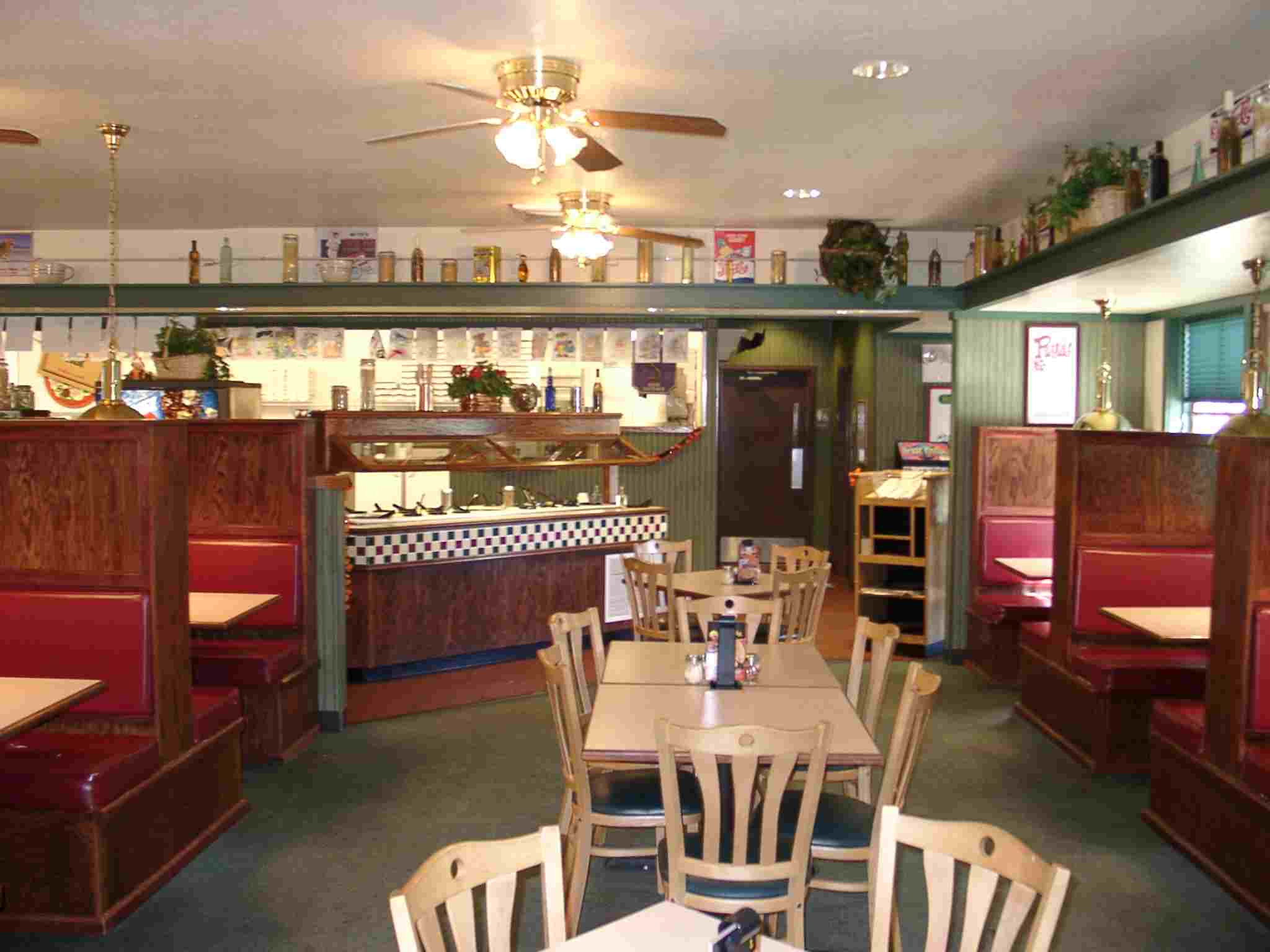}
                \put(25,79){Input image}
            \end{overpic}
            \hfill
            \begin{overpic}[width=0.166\linewidth]{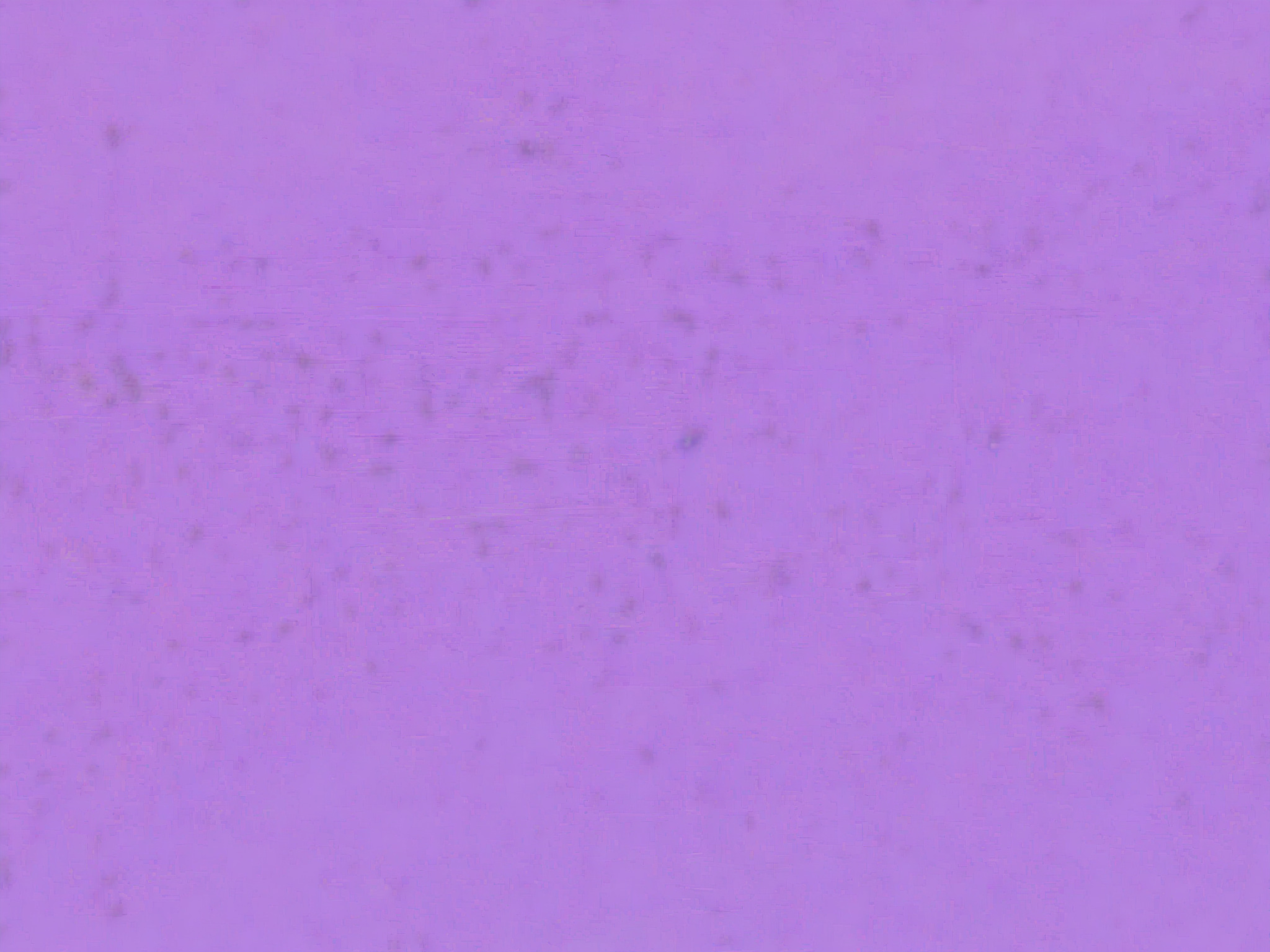}
                \put(0,79){\textbf{Our \rgbtoxx normal} }
            \end{overpic}
            \hfill
            \begin{overpic}[width=0.166\linewidth]{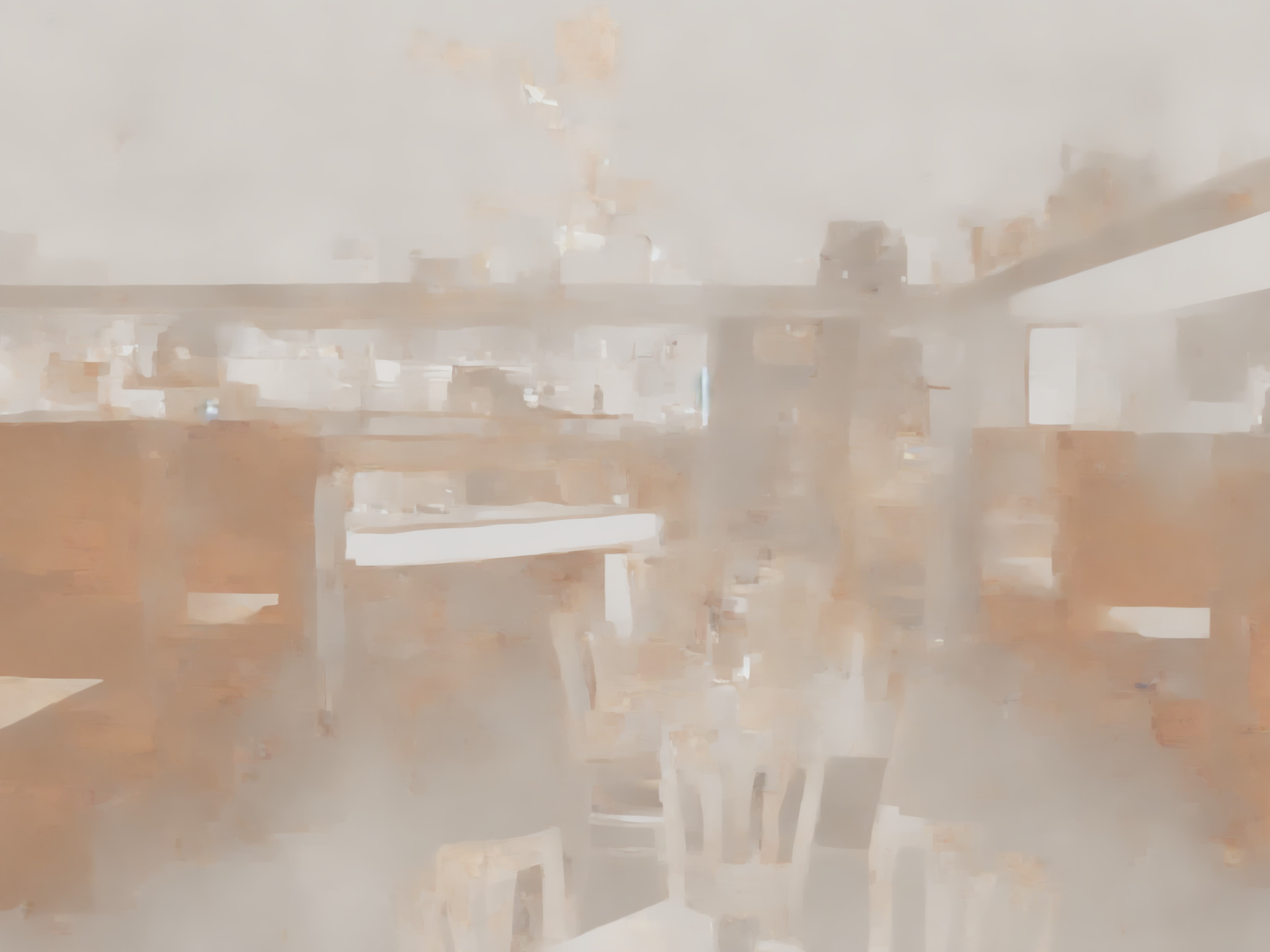}
                \put(2,79){\textbf{Our \rgbtoxx albedo}}
            \end{overpic}
            \hfill
            \begin{overpic}[width=0.166\linewidth]{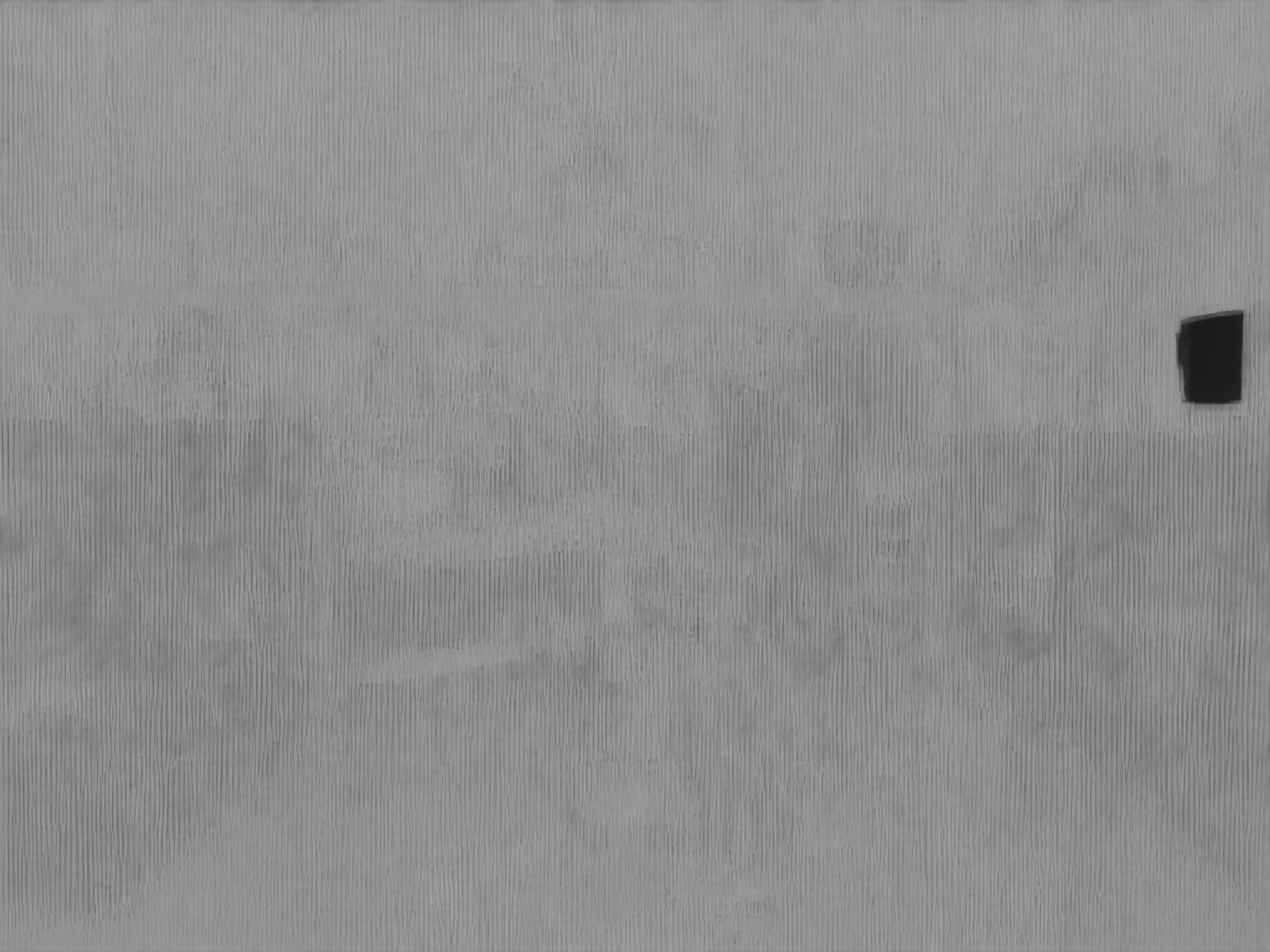}
                \put(2,79){\textbf{Our \rgbtoxx rough.}}
            \end{overpic}
            \hfill
            \begin{overpic}[width=0.166\linewidth]{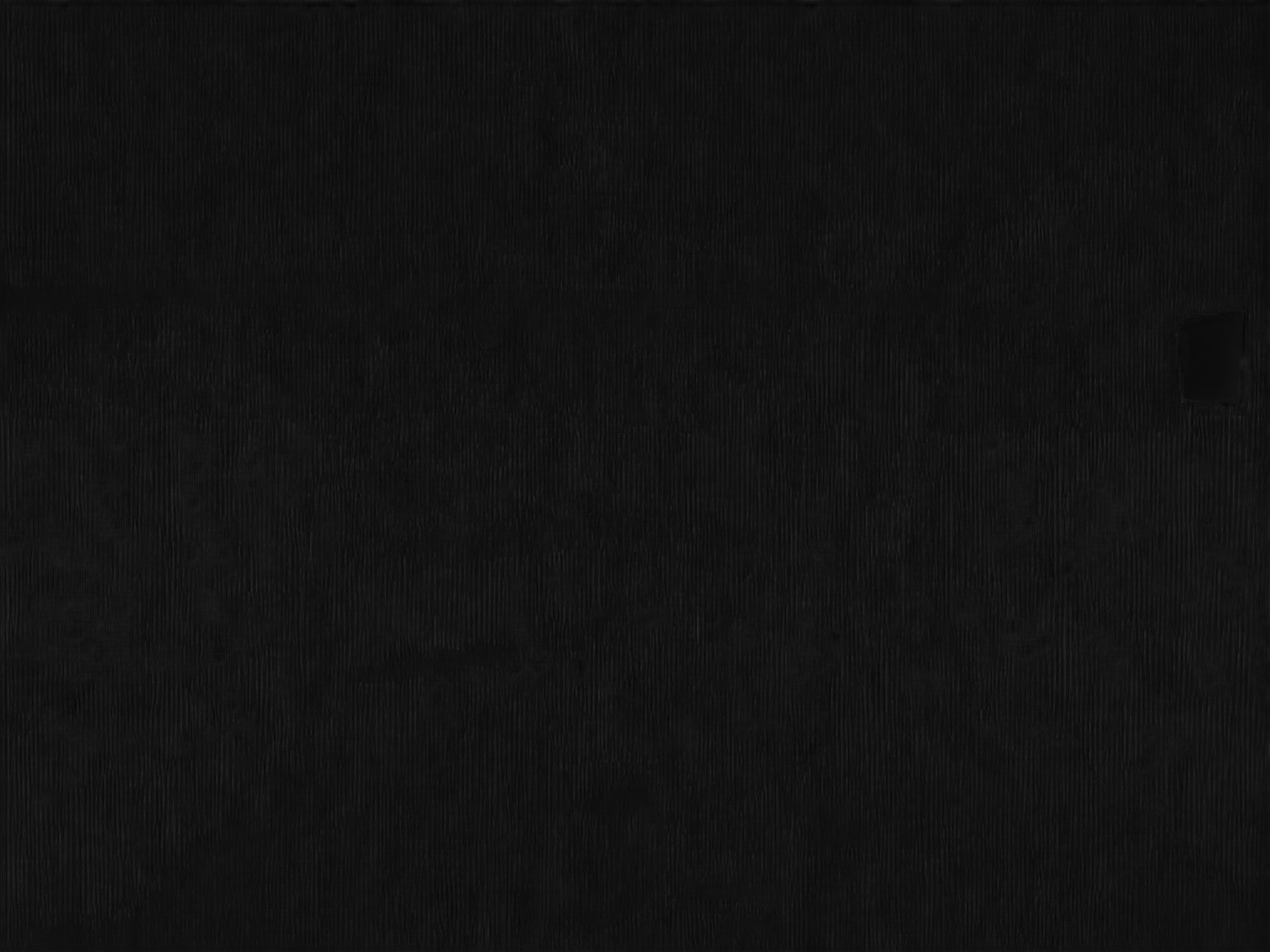}
                \put(2,79){\textbf{Our \rgbtoxx metal.}}
            \end{overpic}
            \hfill
            \begin{overpic}[width=0.166\linewidth]{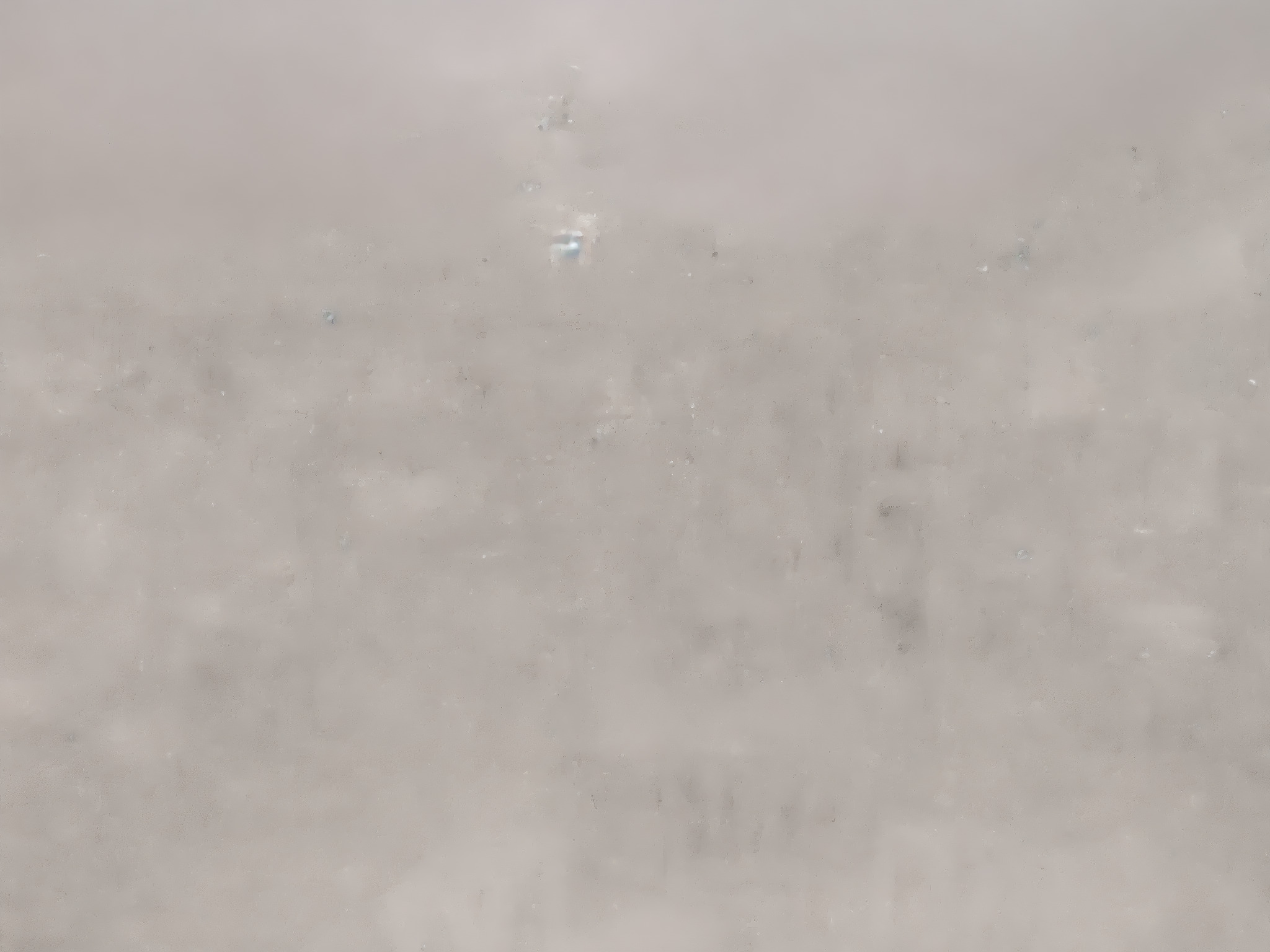}
                \put(4,79){\textbf{Our \rgbtoxx irra.}}
            \end{overpic}
        }
    \end{minipage}\par\smallskip
    \begin{minipage}{1\linewidth}
        \begin{minipage}{\linewidth}
        \end{minipage}\par\medskip
        \centering
        \subfloat[An outdoor real photo, which is out of the our data distribution; this one also contains weird blocking artifacts.]{
            \begin{overpic}[width=0.166\linewidth]{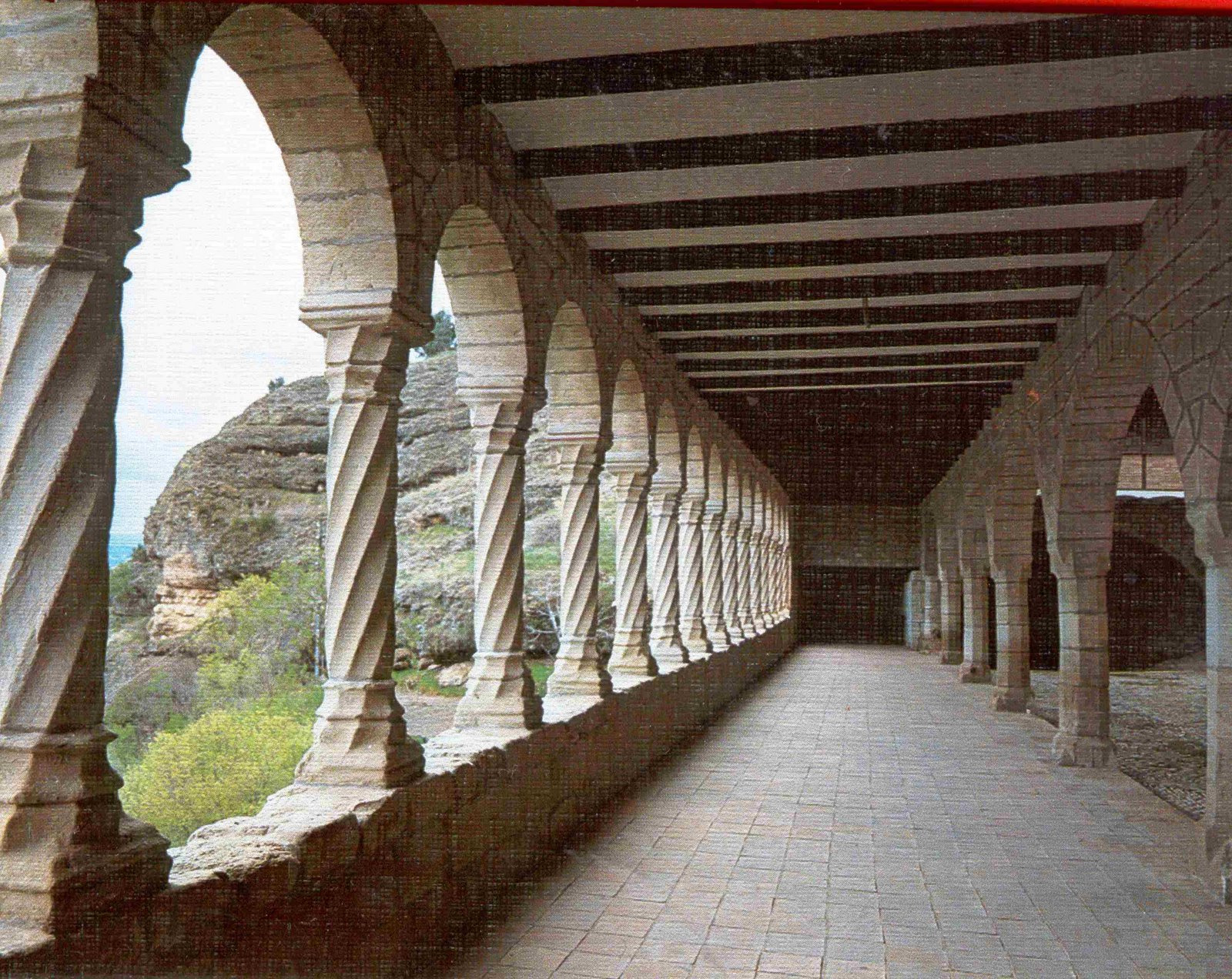}
                \put(25,83){Input image}
            \end{overpic}
            \hfill
            \begin{overpic}[width=0.166\linewidth]{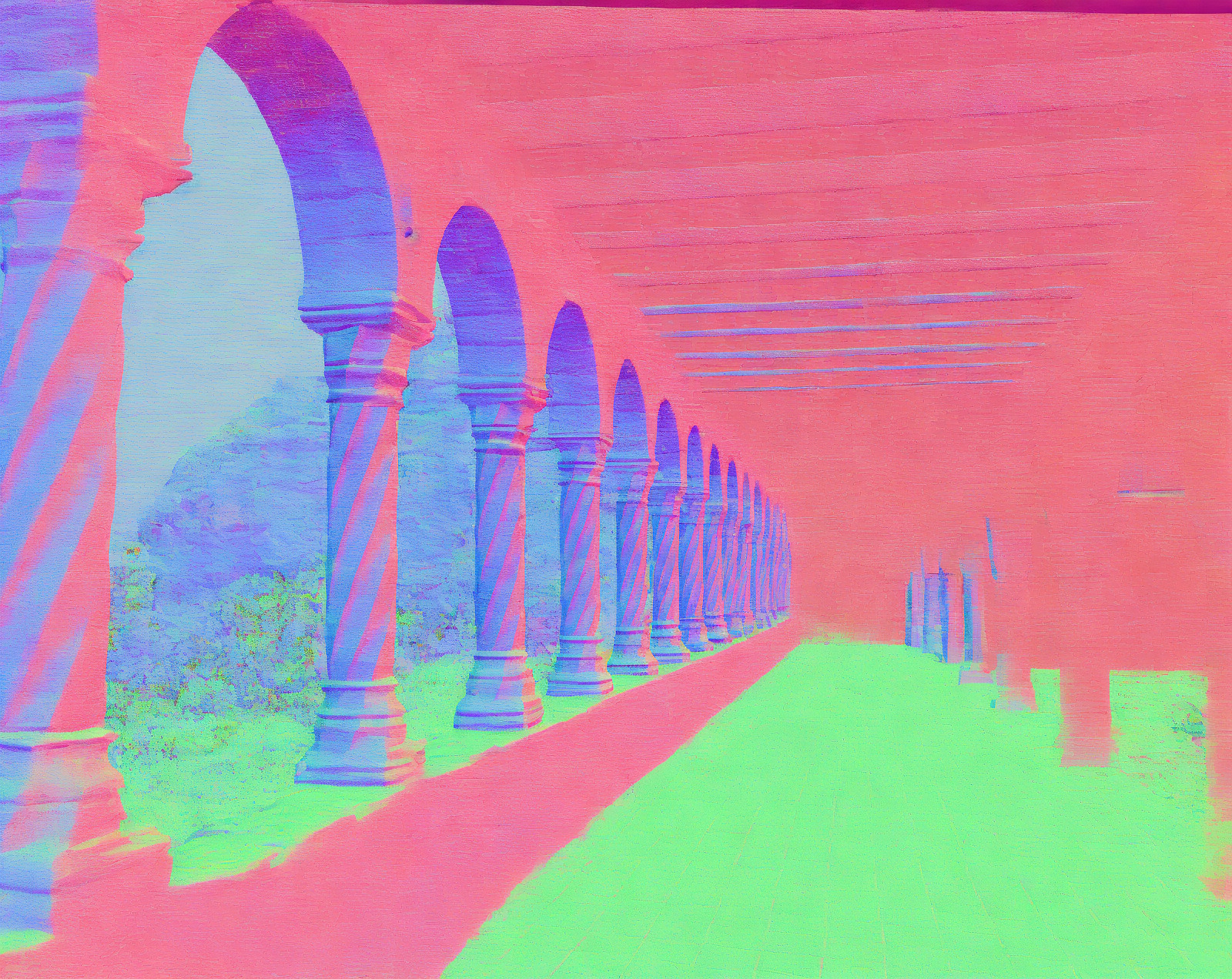}
                \put(0,83){\textbf{Our \rgbtoxx normal} }
            \end{overpic}
            \hfill
            \begin{overpic}[width=0.166\linewidth]{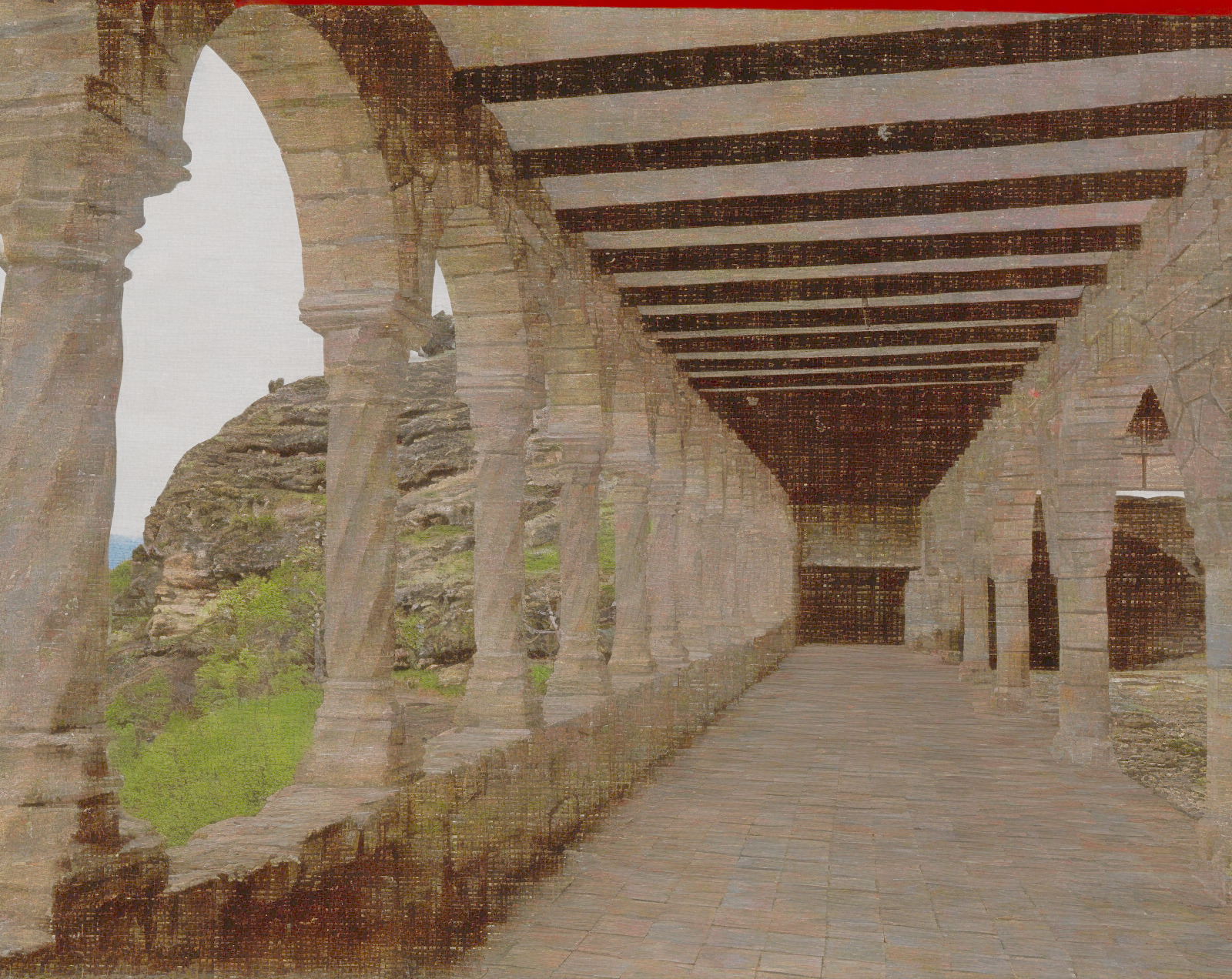}
                \put(2,83){\textbf{Our \rgbtoxx albedo}}
            \end{overpic}
            \hfill
            \begin{overpic}[width=0.166\linewidth]{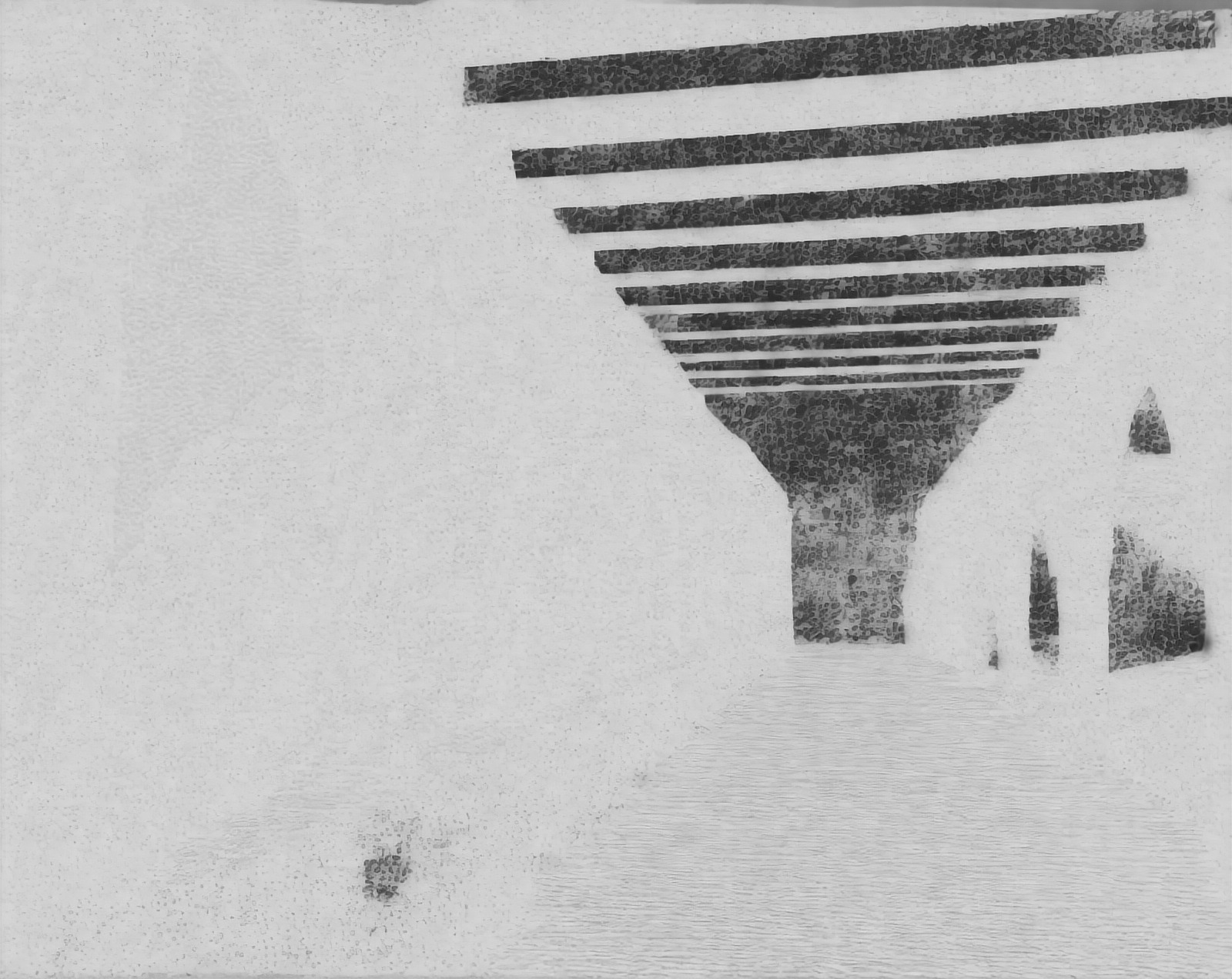}
                \put(2,83){\textbf{Our \rgbtoxx rough.}}
            \end{overpic}
            \hfill
            \begin{overpic}[width=0.166\linewidth]{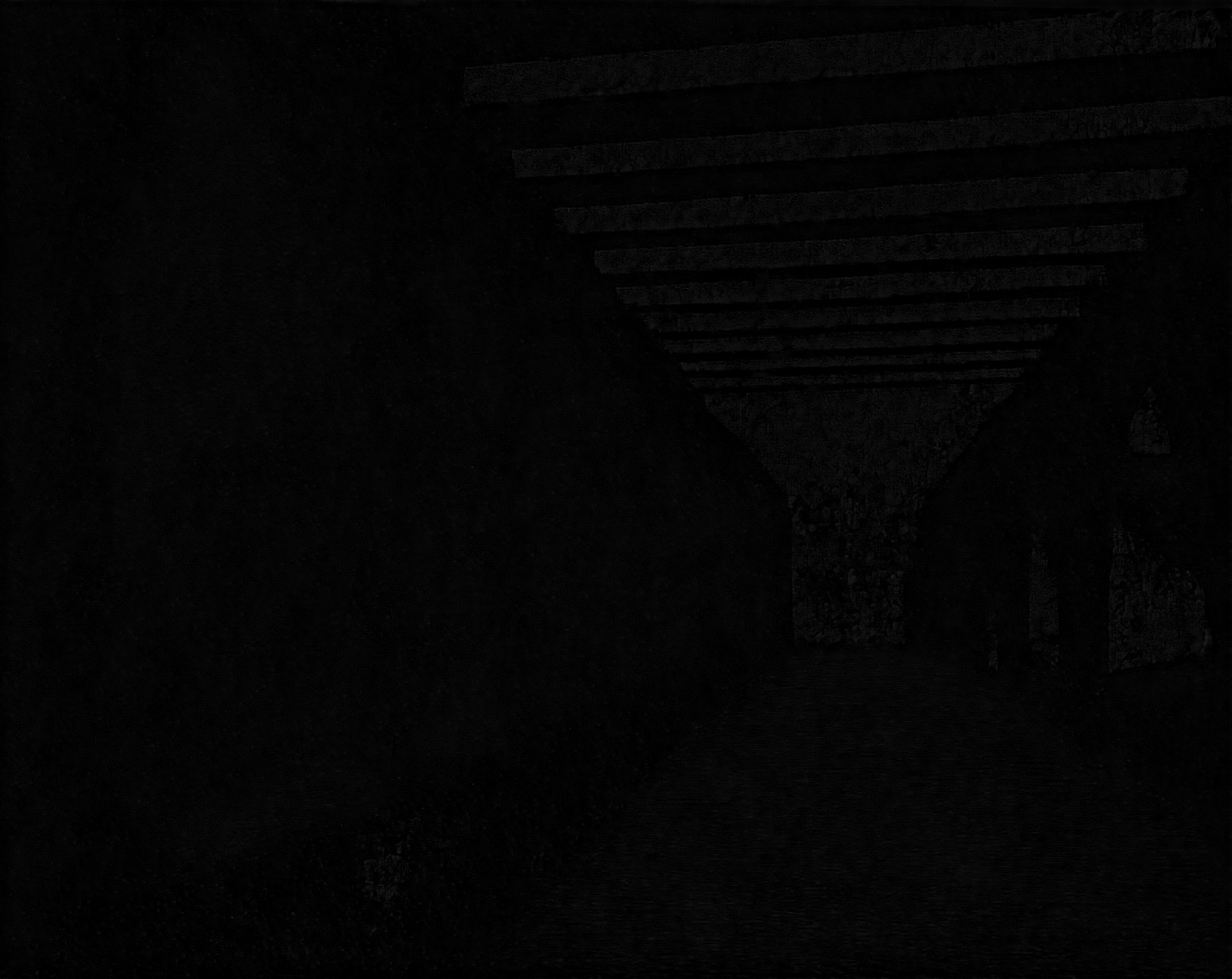}
                \put(2,83){\textbf{Our \rgbtoxx metal.}}
            \end{overpic}
            \hfill
            \begin{overpic}[width=0.166\linewidth]{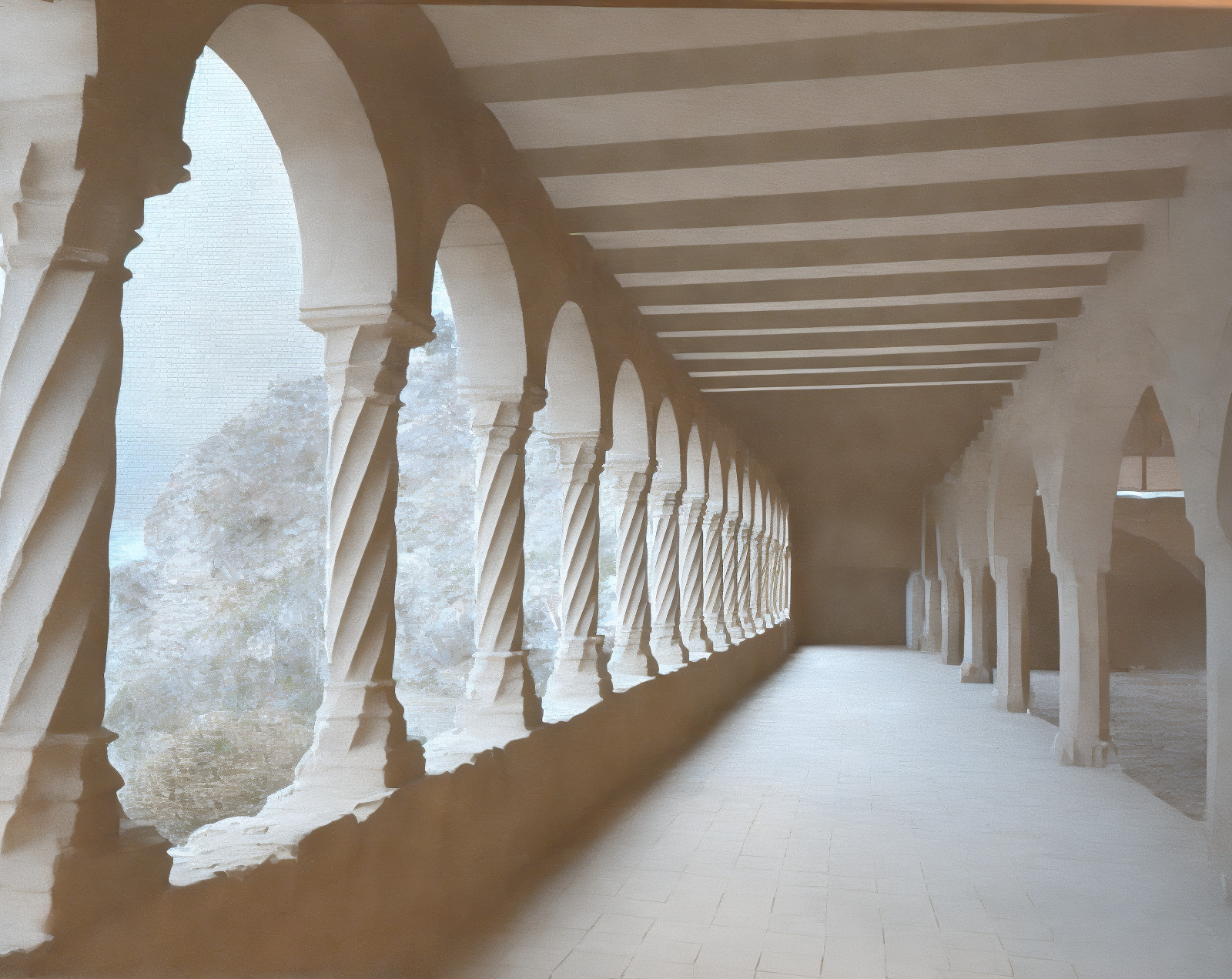}
                \put(4,83){\textbf{Our \rgbtoxx irra.}}
            \end{overpic}
        }
    \end{minipage}\par\smallskip
\end{minipage}

%% file: supp/ablation-rgb2x/ablation-rgb2x-5.tex
\begin{minipage}{0.9\linewidth}
    \begin{minipage}{\linewidth}
    \end{minipage}\par\smallskip
    \centering
    \begin{overpic}[width=0.162\linewidth]{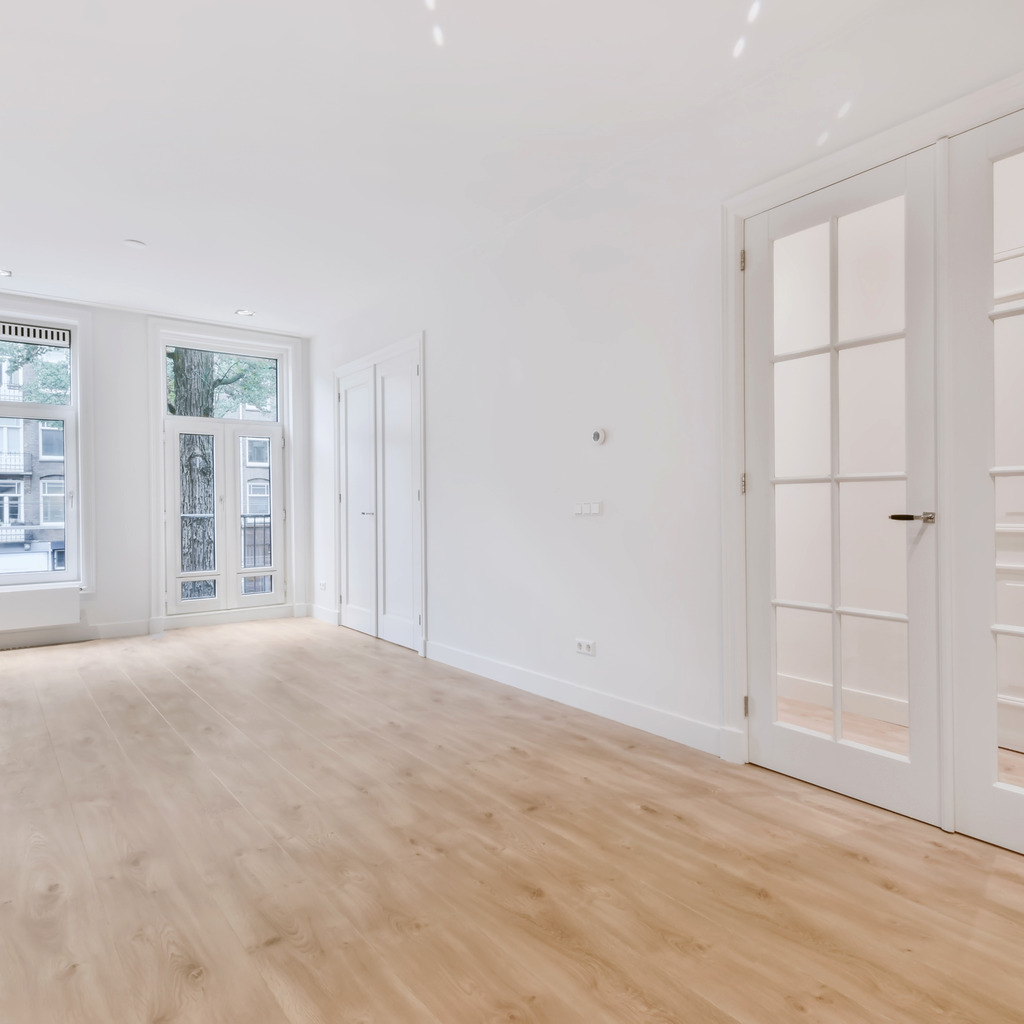}
        \put(25,104){Input image}
    \end{overpic}
    \hfill
    \begin{overpic}[width=0.162\linewidth]{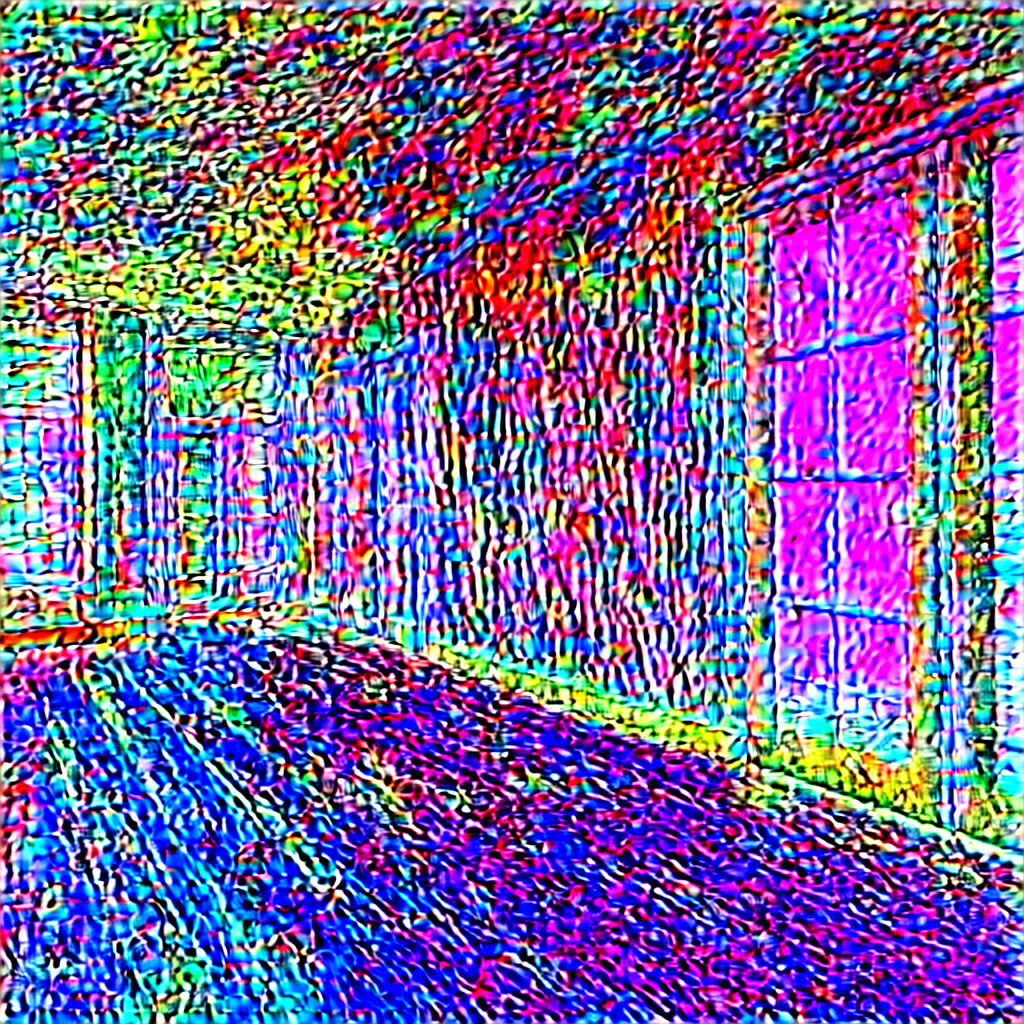}
        \put(1,115){\footnotesize Model outputs 5 channels}
        \put(30,104){normal}
    \end{overpic}
    \hfill
    \begin{overpic}[width=0.162\linewidth]{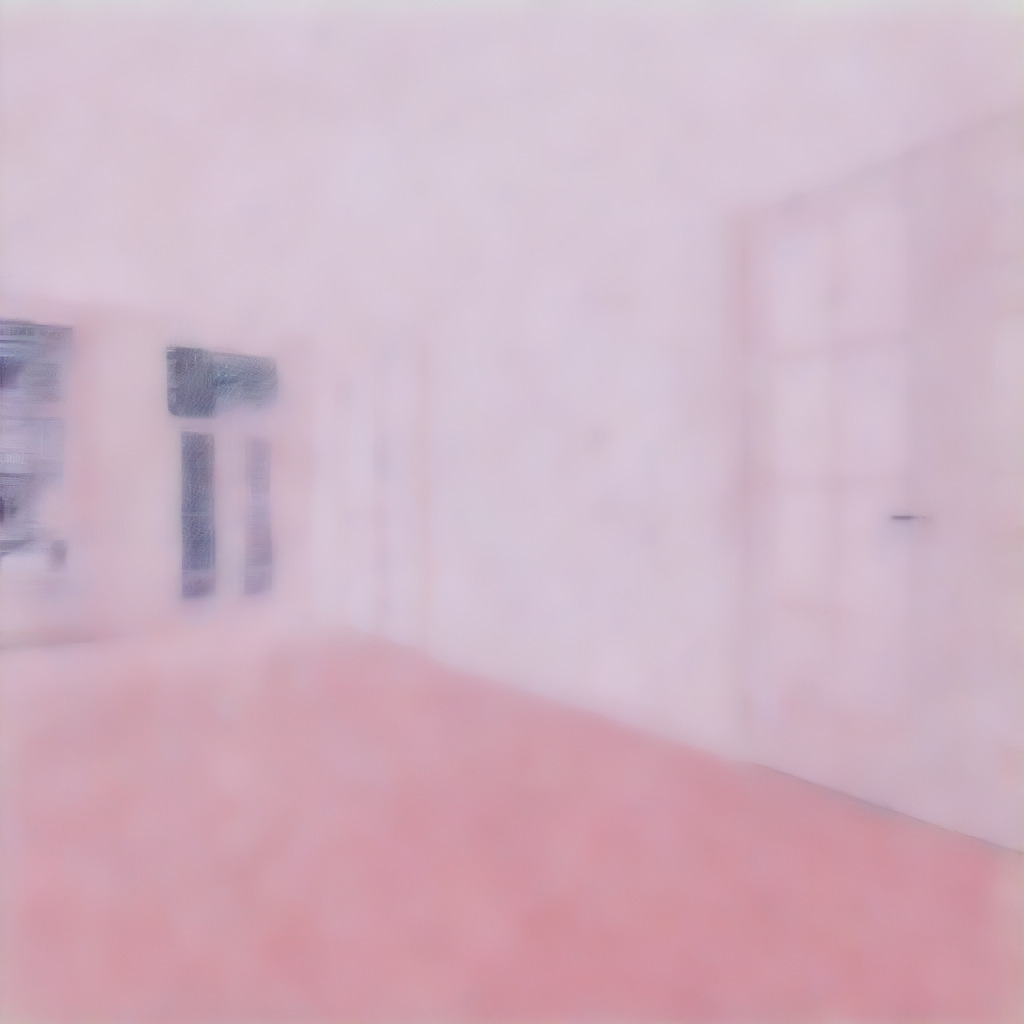}
        \put(1,115){\footnotesize Model outputs 5 channels}
        \put(33,104){albedo}
    \end{overpic}
    \hfill
    \begin{overpic}[width=0.162\linewidth]{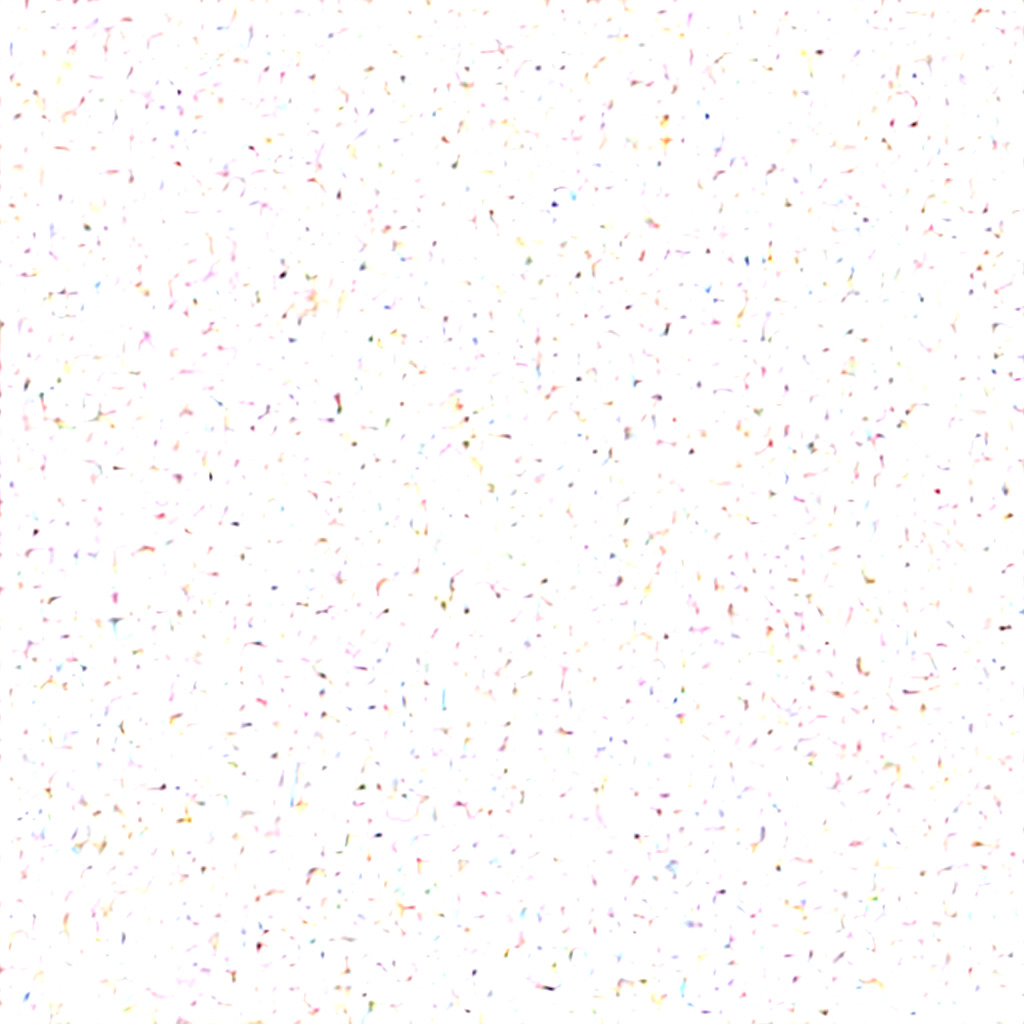}
        \put(1,115){\footnotesize Model outputs 5 channels}
        \put(27,104){roughness}
    \end{overpic}
    \hfill
    \begin{overpic}[width=0.162\linewidth]{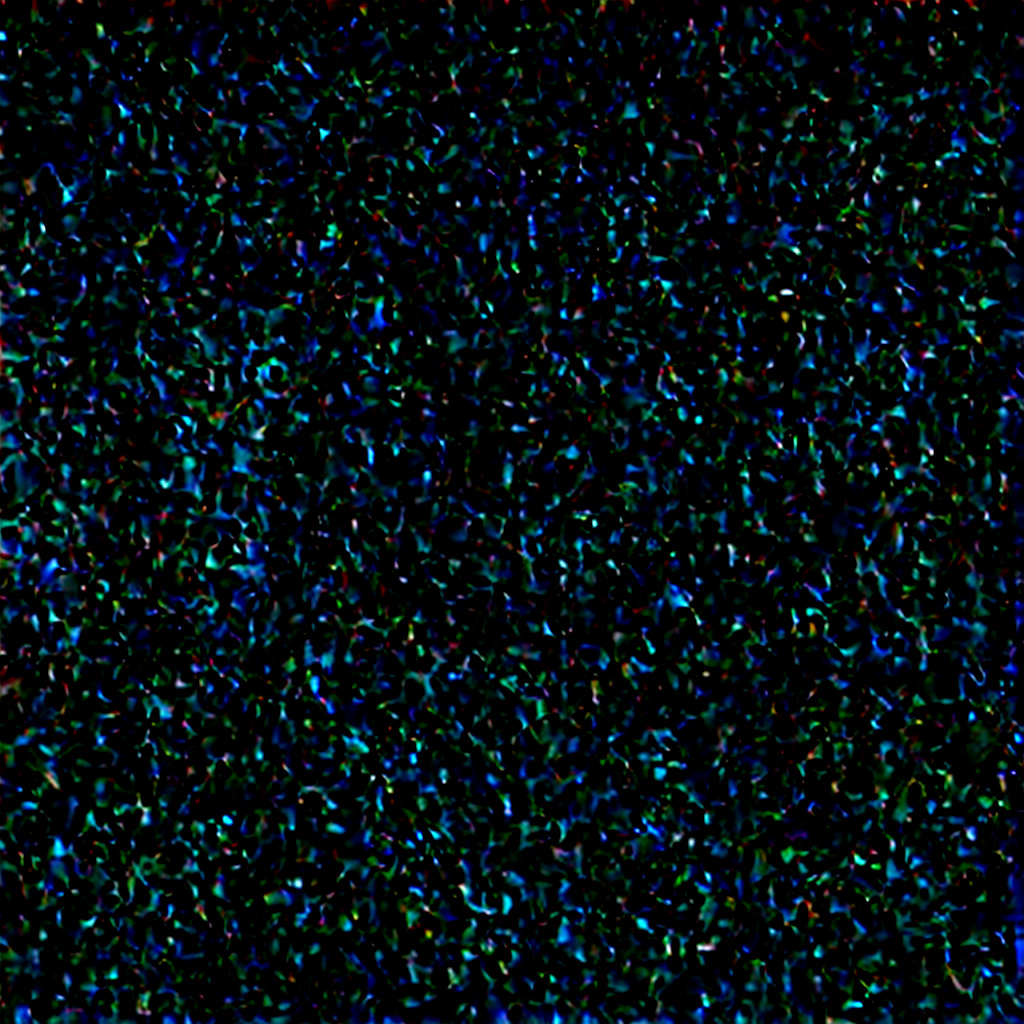}
        \put(1,115){\footnotesize Model outputs 5 channels}
        \put(27,104){metallicity}
    \end{overpic}
    \hfill
    \begin{overpic}[width=0.162\linewidth]{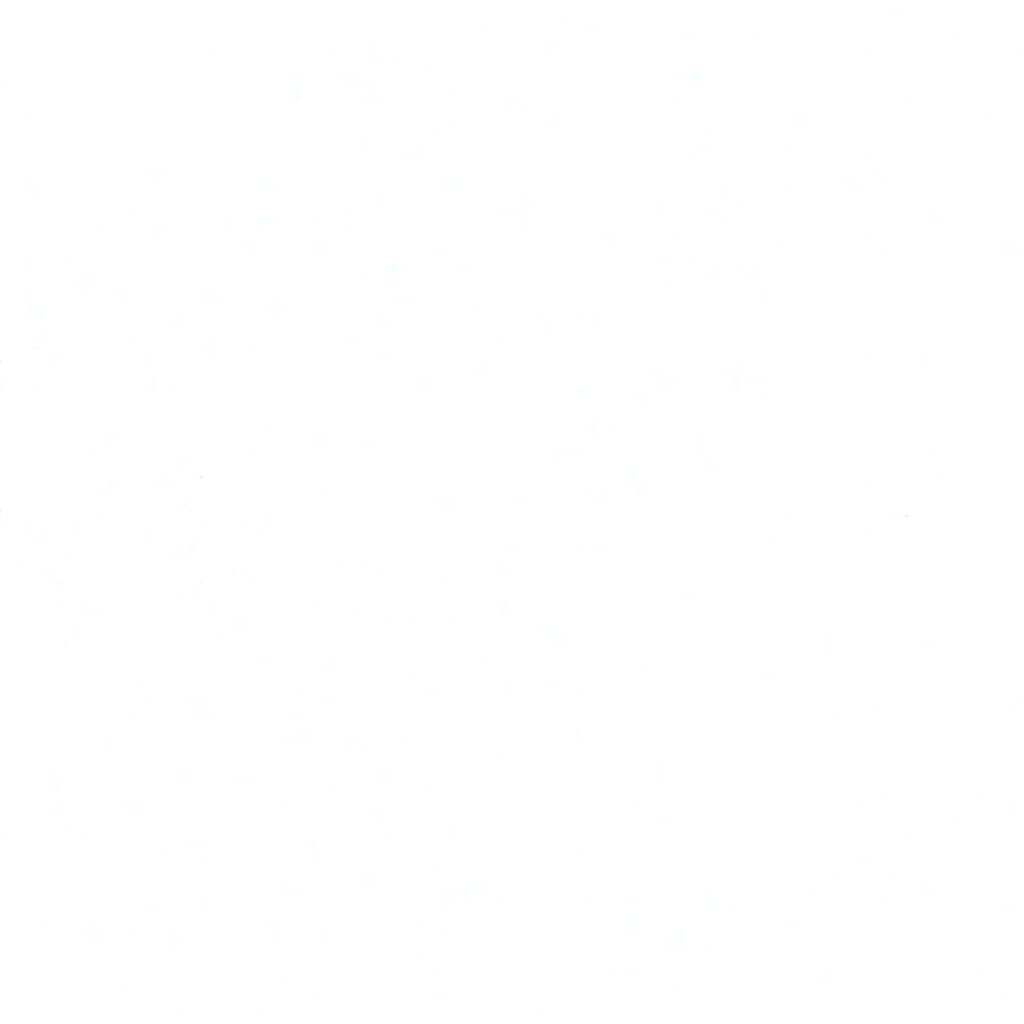}
        \put(1,115){\footnotesize Model outputs 5 channels}
        \put(27,104){irradiance}
    \end{overpic}
\end{minipage}\par\bigskip\bigskip
\begin{minipage}{0.9\linewidth}
    \centering
    \begin{overpic}[width=0.162\linewidth]{supp/ablation-rgb2x/scene-1-photo}
        \put(25,104){Input image}
    \end{overpic}
    \hfill
    \begin{overpic}[width=0.162\linewidth]{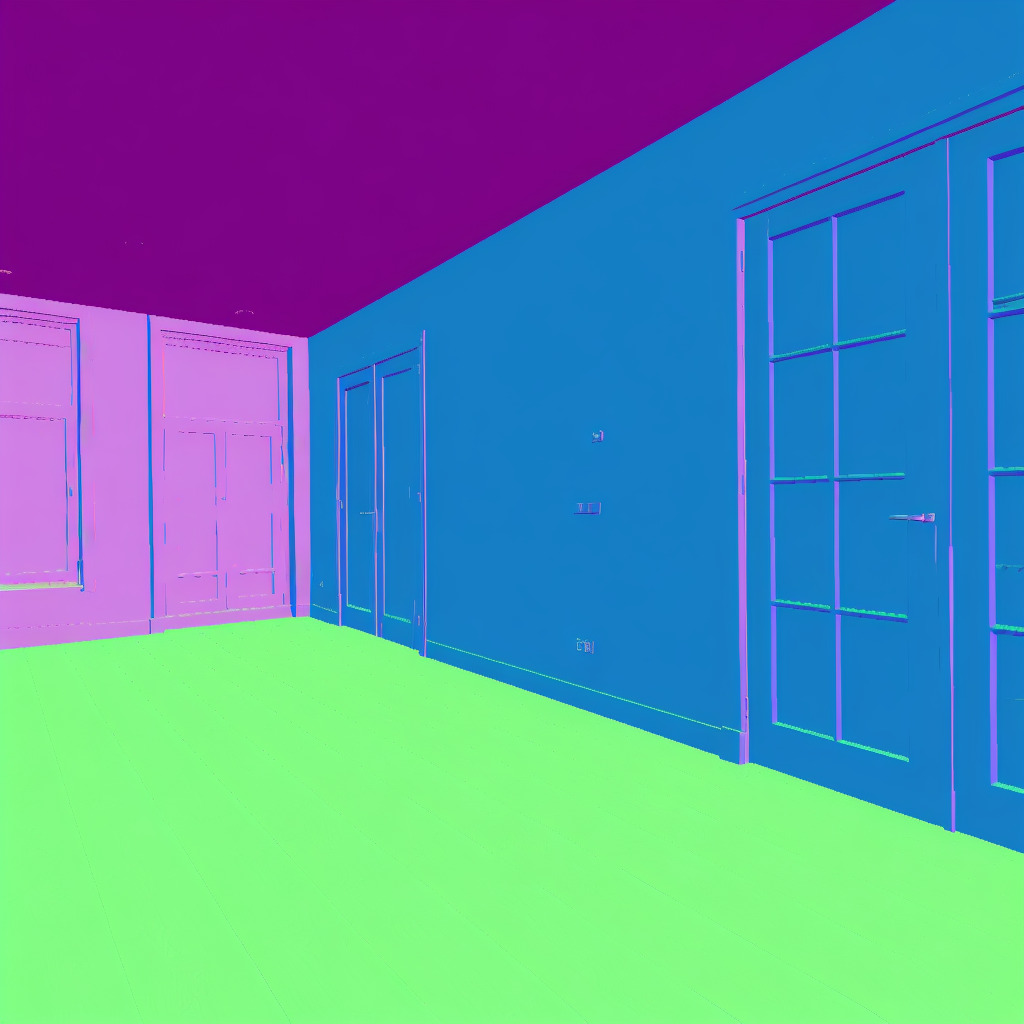}
        \put(0,104){\small \textbf{Our \rgbtoxx normal} }
    \end{overpic}
    \hfill
    \begin{overpic}[width=0.162\linewidth]{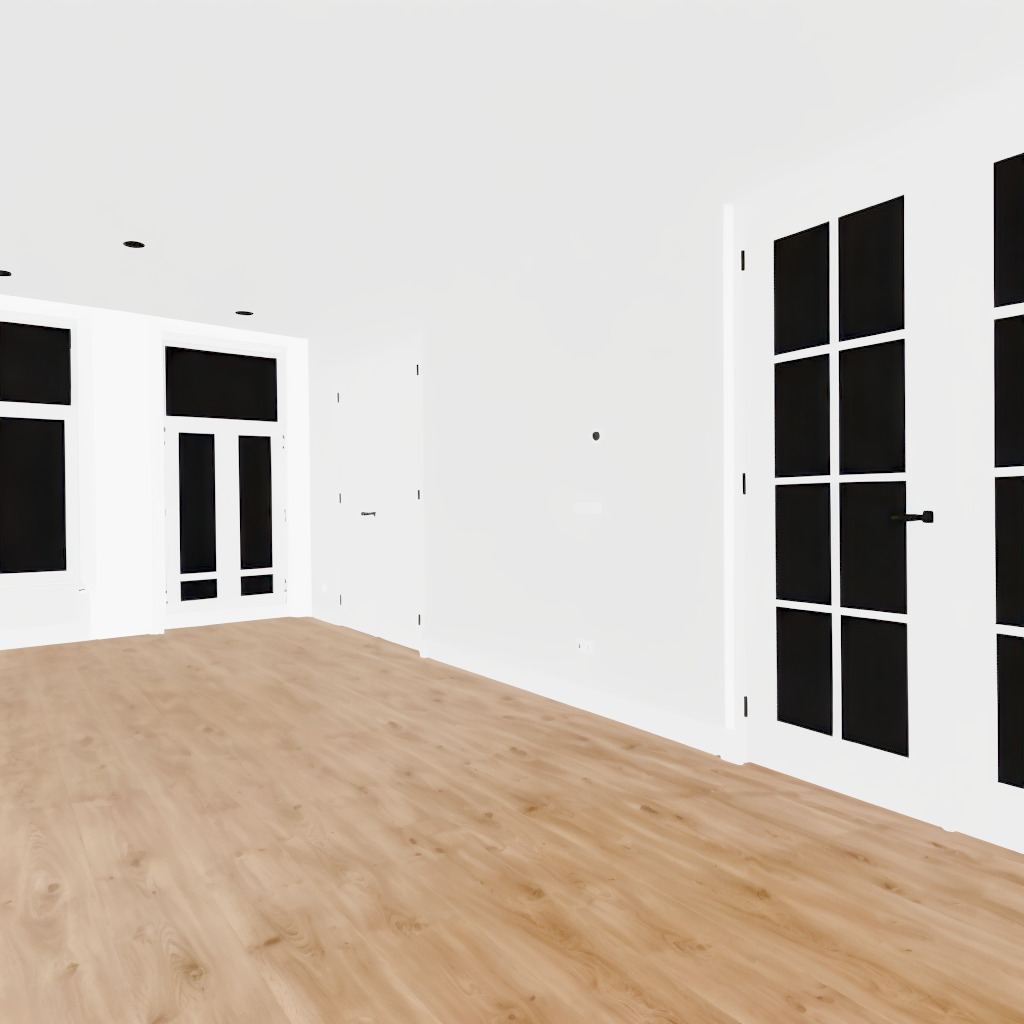}
        \put(2,104){\small \textbf{Our \rgbtoxx albedo}}
    \end{overpic}
    \hfill
    \begin{overpic}[width=0.162\linewidth]{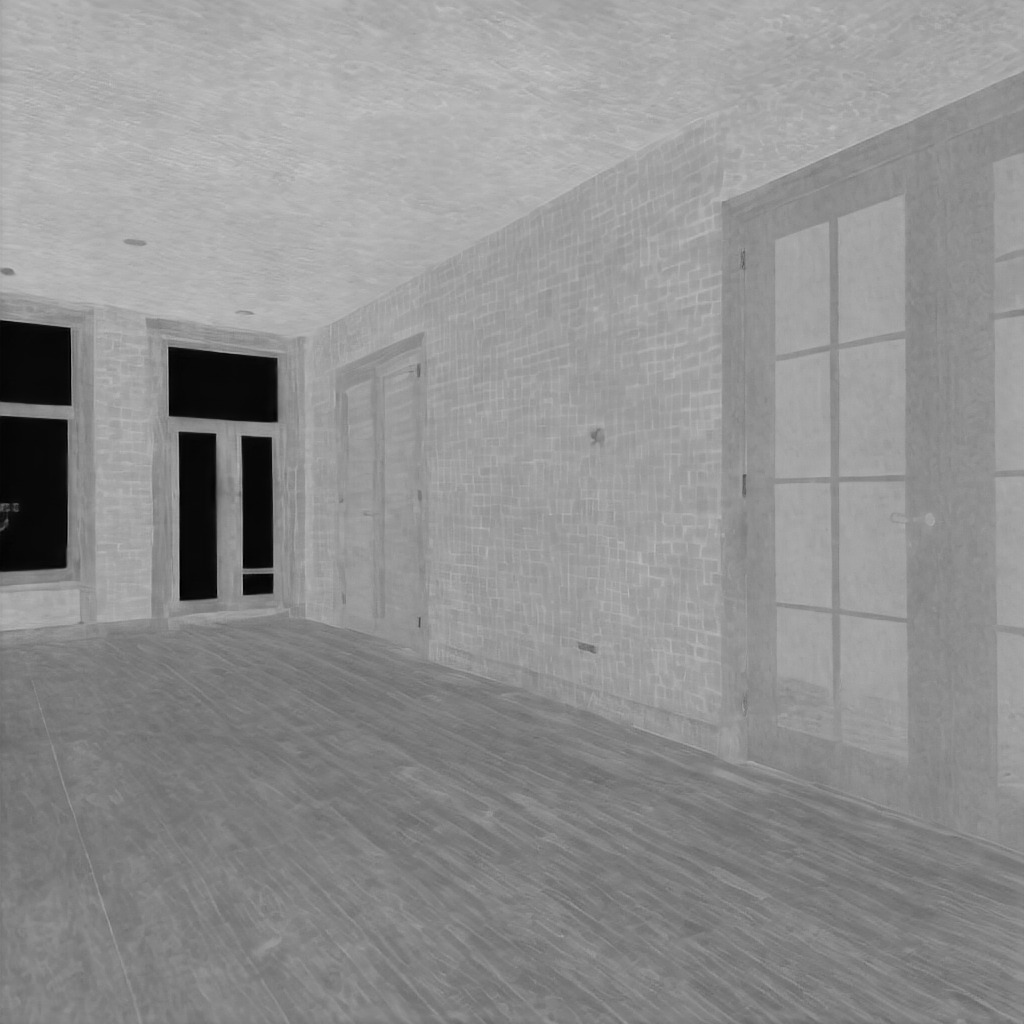}
        \put(2,104){\small \textbf{Our \rgbtoxx rough.}}
    \end{overpic}
    \hfill
    \begin{overpic}[width=0.162\linewidth]{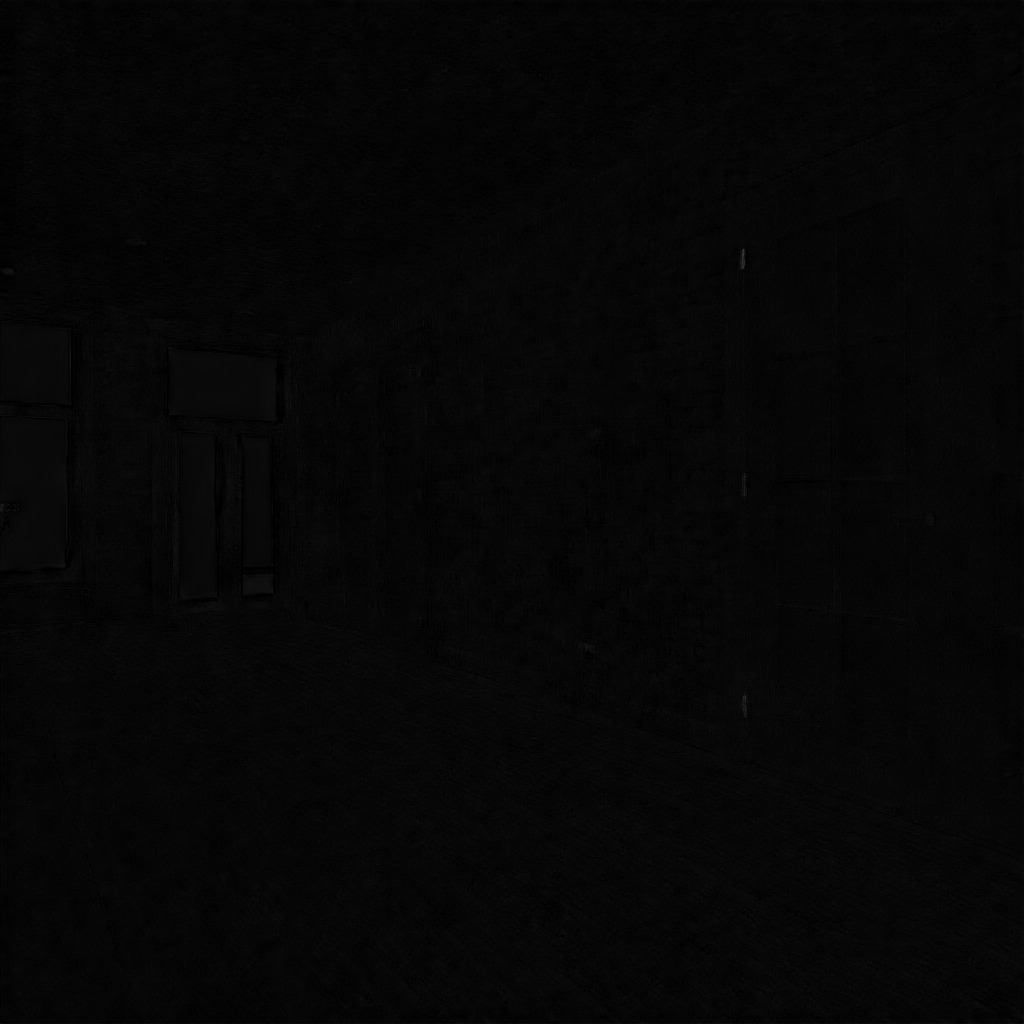}
        \put(2,104){\small \textbf{Our \rgbtoxx metal.}}
    \end{overpic}
    \hfill
    \begin{overpic}[width=0.162\linewidth]{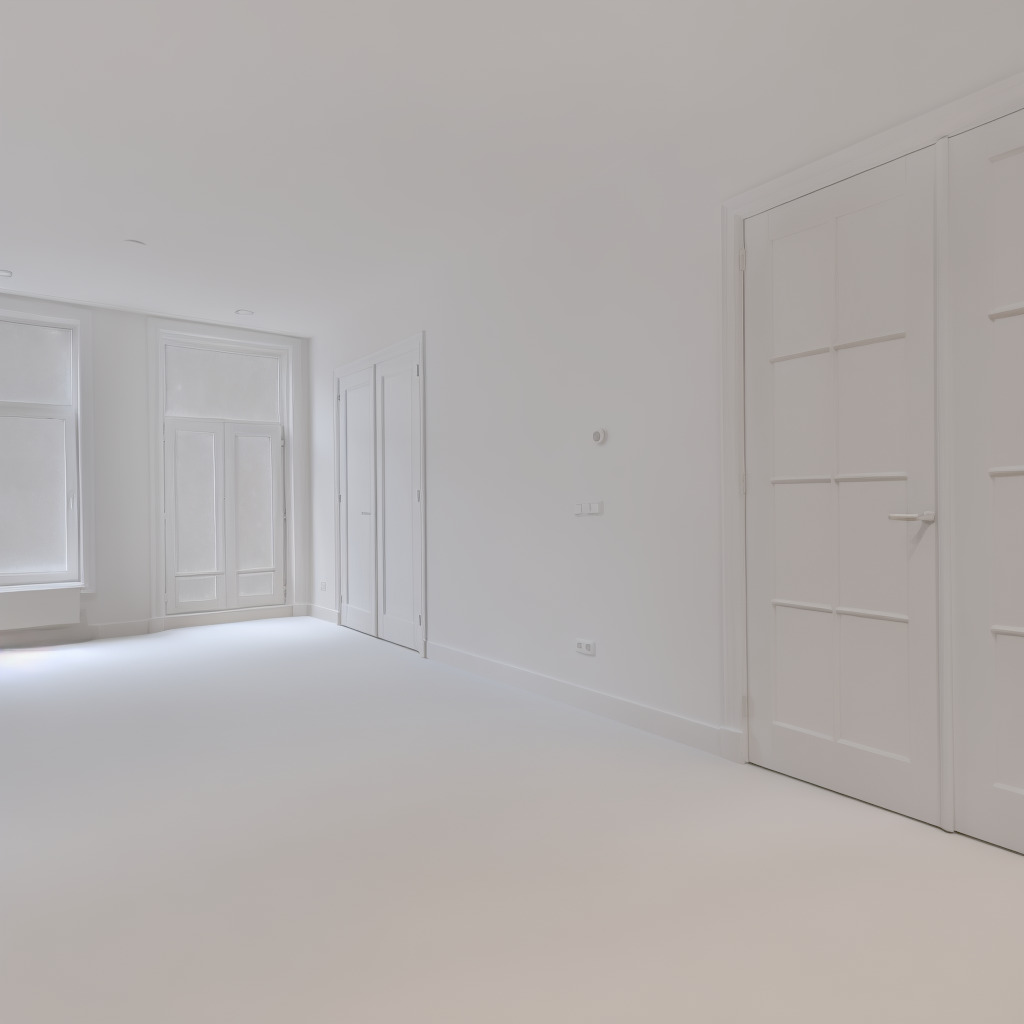}
        \put(4,104){\small \textbf{Our \rgbtoxx irra.}}
    \end{overpic}
\end{minipage}\par\smallskip

%% file: supp/ablation-rgb2x/ablation-rgb2x-4.tex
\begin{minipage}{0.9\linewidth}
    \begin{minipage}{\linewidth}
    \end{minipage}\par\bigskip
    \begin{overpic}[width=0.195\linewidth]{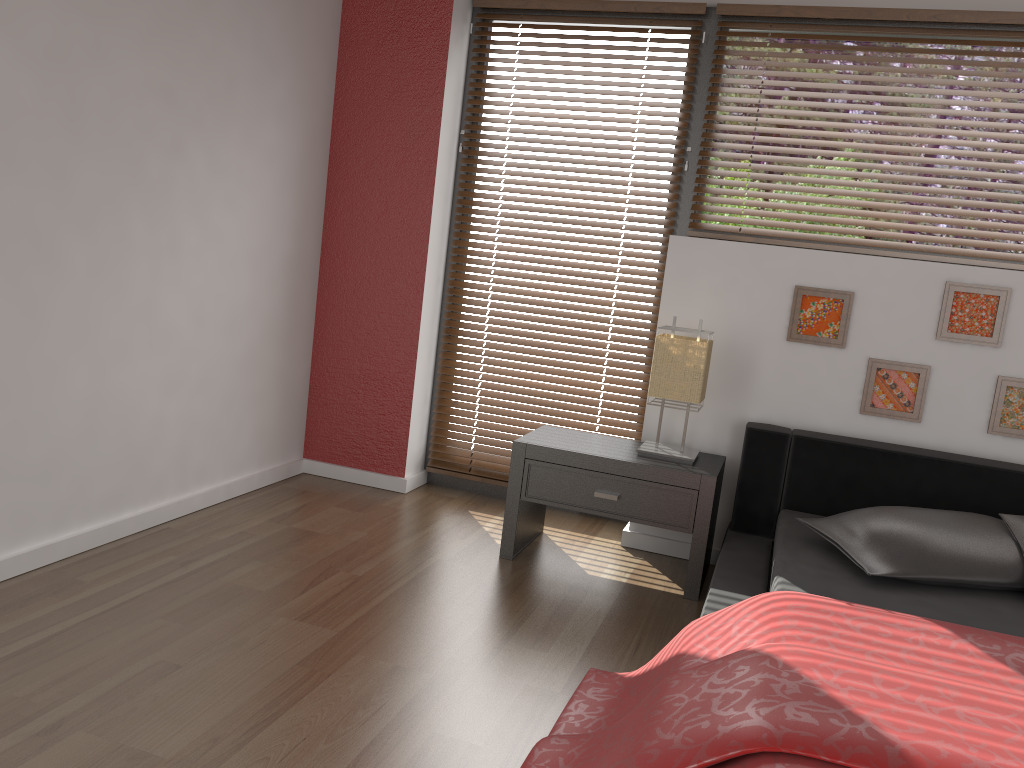}
        \put(26,78){Input image}
    \end{overpic}
    \hfill
    \begin{overpic}[width=0.195\linewidth]{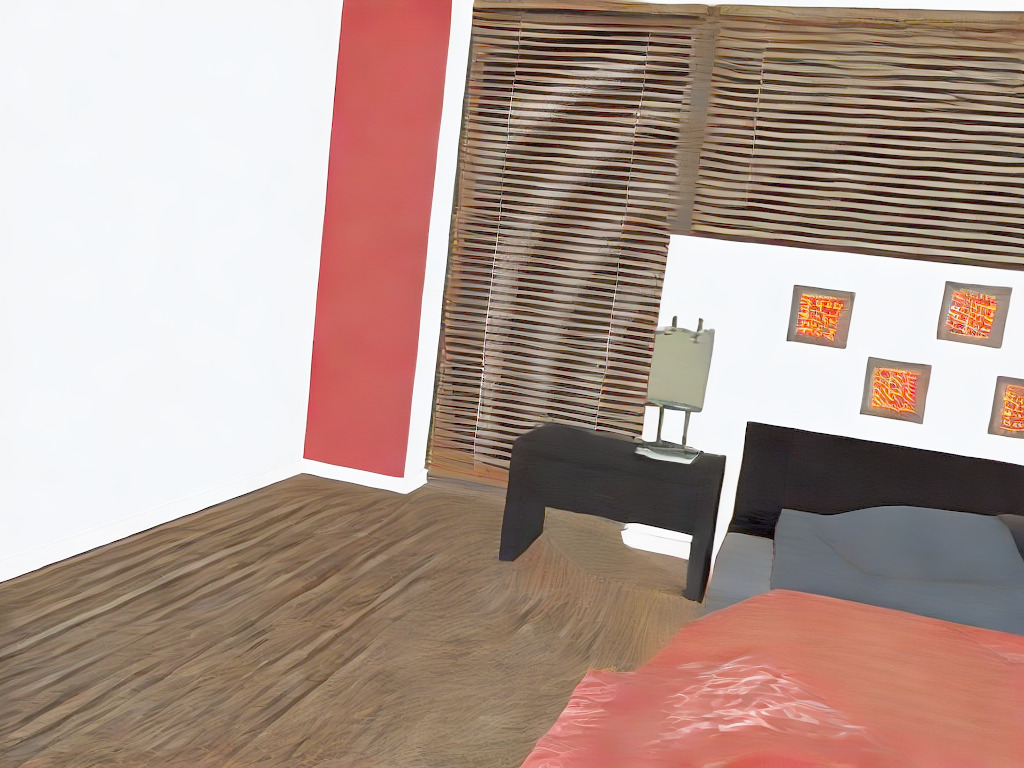}
        \put(9,89){\footnotesize Model outputs 4 channels}
        \put(36,78){albedo}
    \end{overpic}
    \hfill
    \begin{overpic}[width=0.195\linewidth]{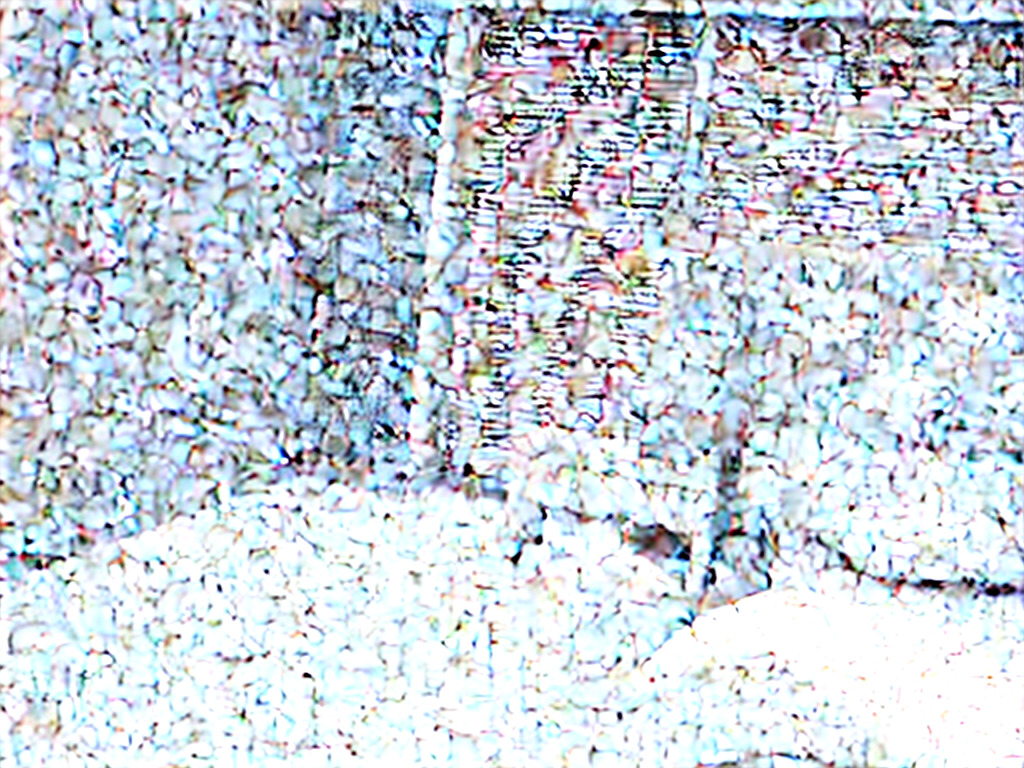}
        \put(9,89){\footnotesize Model outputs 4 channels}
        \put(30,78){roughness}
    \end{overpic}
    \hfill
    \begin{overpic}[width=0.195\linewidth]{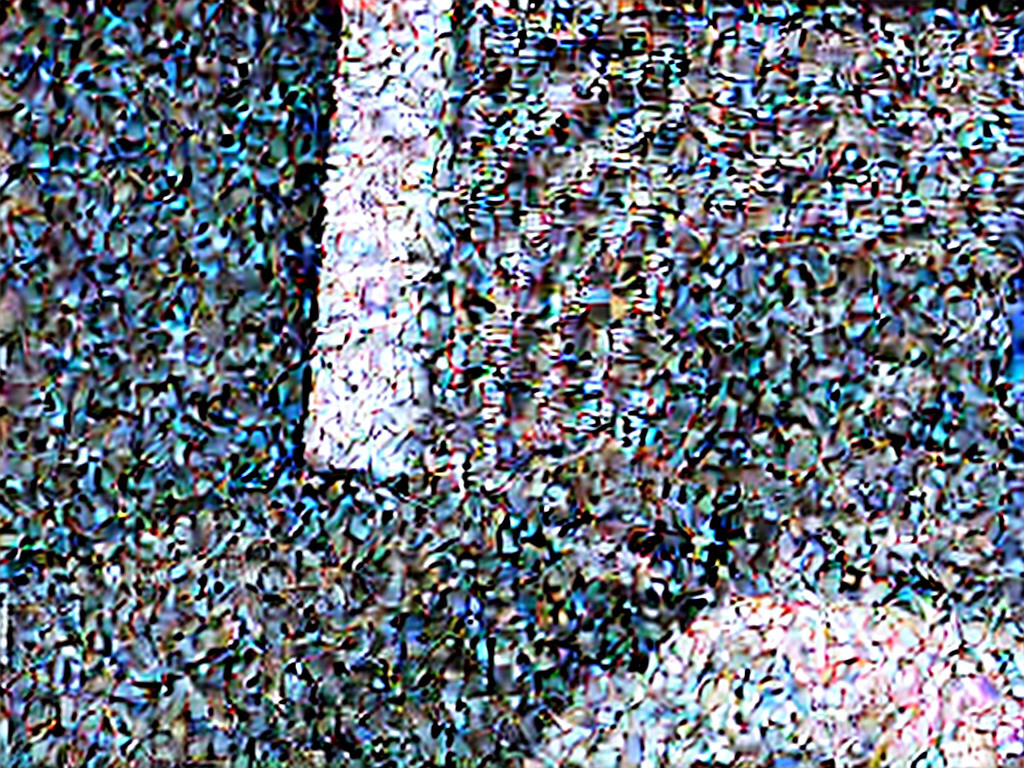}
        \put(9,89){\footnotesize Model outputs 4 channels}
        \put(30,78){metallicity}
    \end{overpic}
    \hfill
    \begin{overpic}[width=0.195\linewidth]{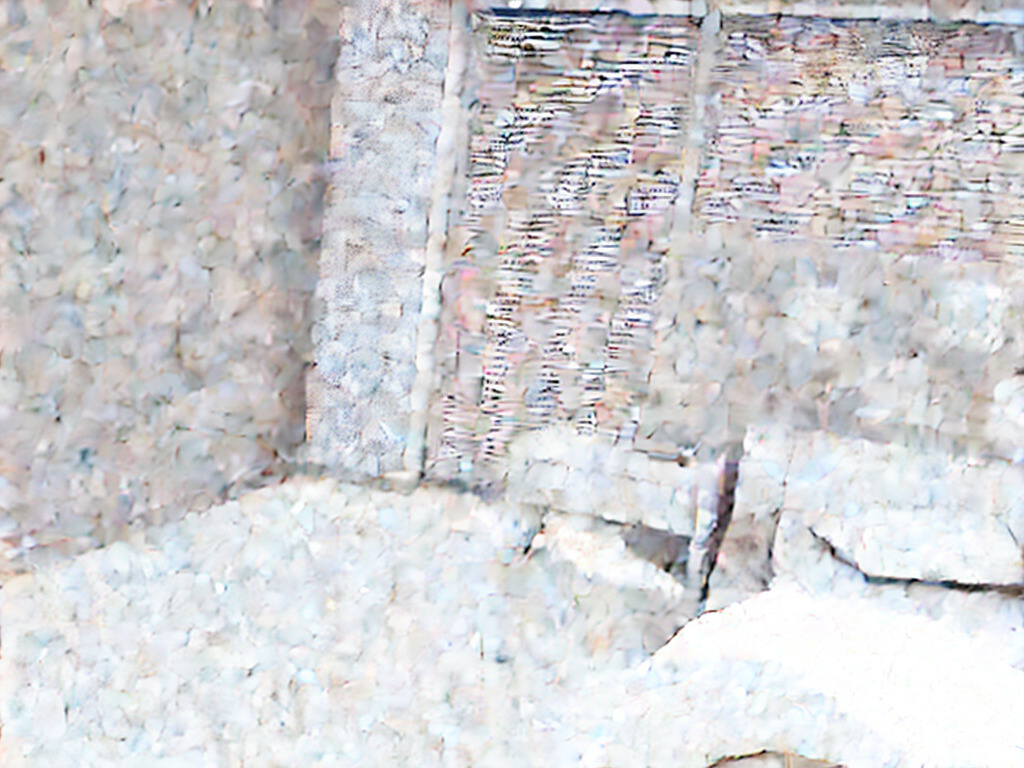}
        \put(9,89){\footnotesize Model outputs 4 channels}
        \put(30,78){irradiance}
    \end{overpic}
\end{minipage}\par\bigskip\bigskip
\begin{minipage}{0.9\linewidth}
    \begin{overpic}[width=0.195\linewidth]{supp/ablation-rgb2x/scene-3-photo}
        \put(26,78){Input image}
    \end{overpic}
    \hfill
    \begin{overpic}[width=0.195\linewidth]{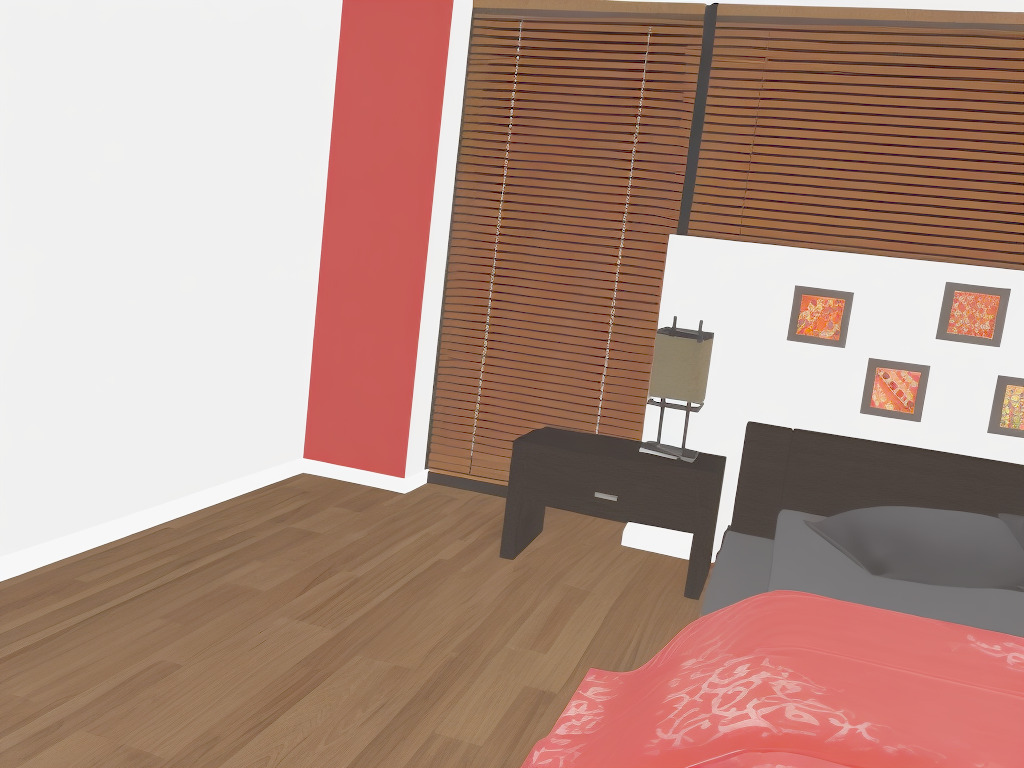}
        \put(6,78){\textbf{Our \rgbtoxx albedo}}
    \end{overpic}
    \hfill
    \begin{overpic}[width=0.195\linewidth]{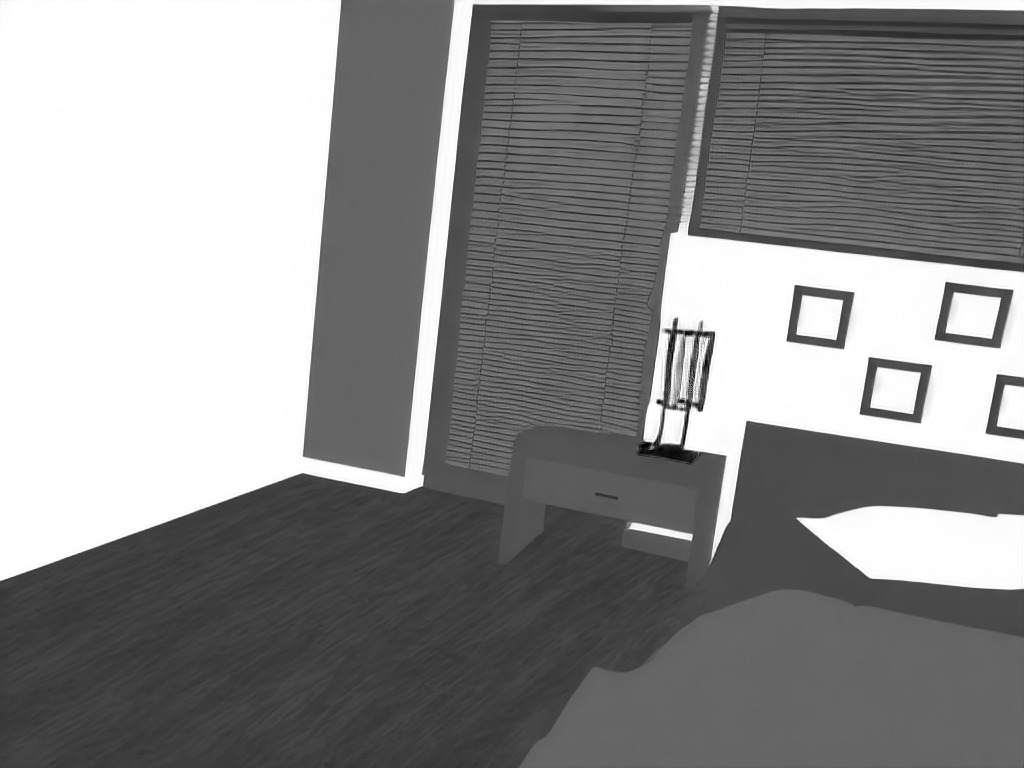}
        \put(7,78){\textbf{Our \rgbtoxx rough.}}
    \end{overpic}
    \hfill
    \begin{overpic}[width=0.195\linewidth]{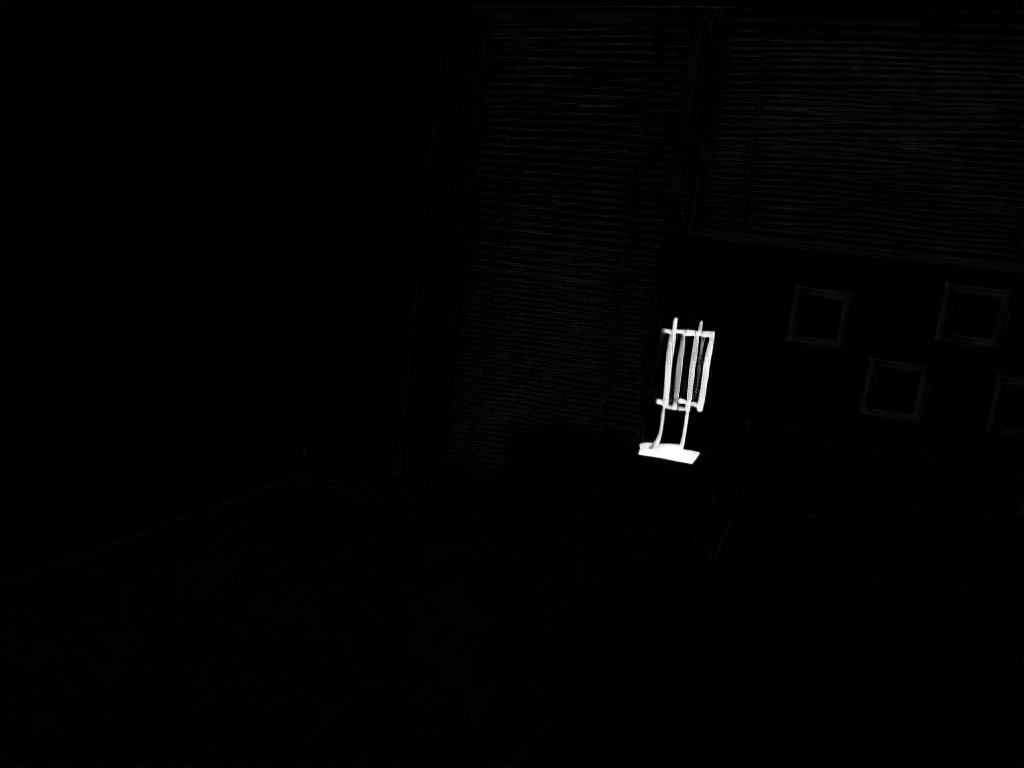}
        \put(7,78){\textbf{Our \rgbtoxx metal.}}
    \end{overpic}
    \hfill
    \begin{overpic}[width=0.195\linewidth]{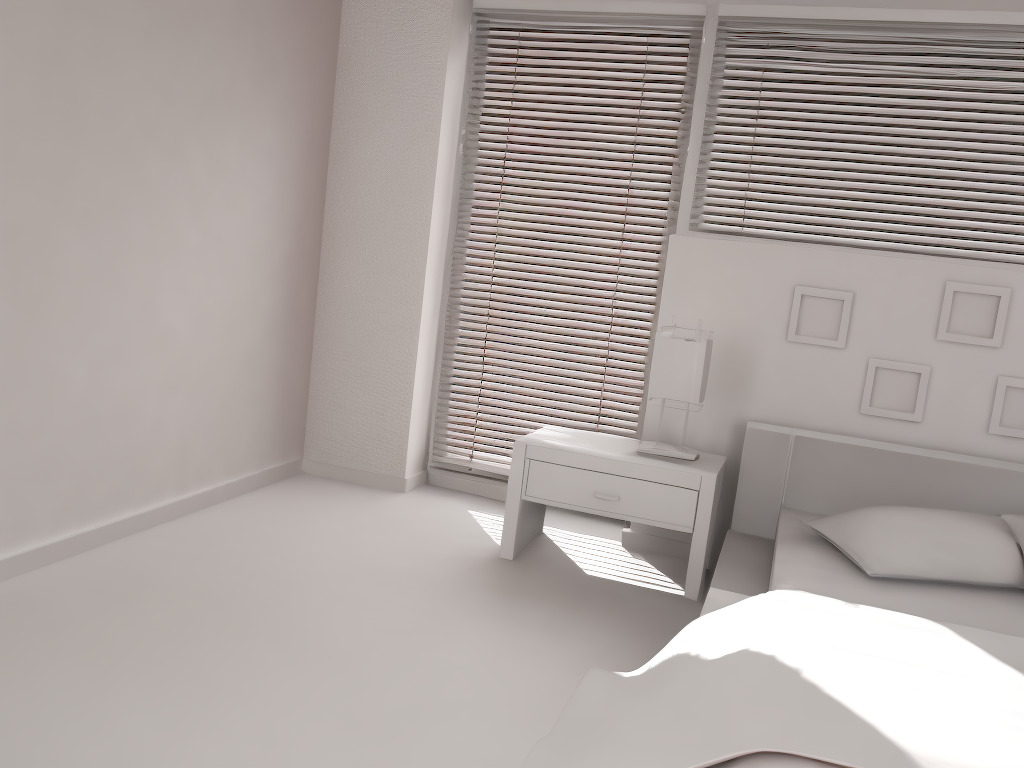}
        \put(9,78){\textbf{Our \rgbtoxx irra.}}
    \end{overpic}
\end{minipage}\par

%% file: supp/ablation-rgb2x/ablation-rgb2x-3.tex
\begin{minipage}{0.8\linewidth}
    \begin{minipage}{\linewidth}
    \end{minipage}\par\bigskip
    \hspace*{\fill}
    \begin{overpic}[width=0.162\linewidth]{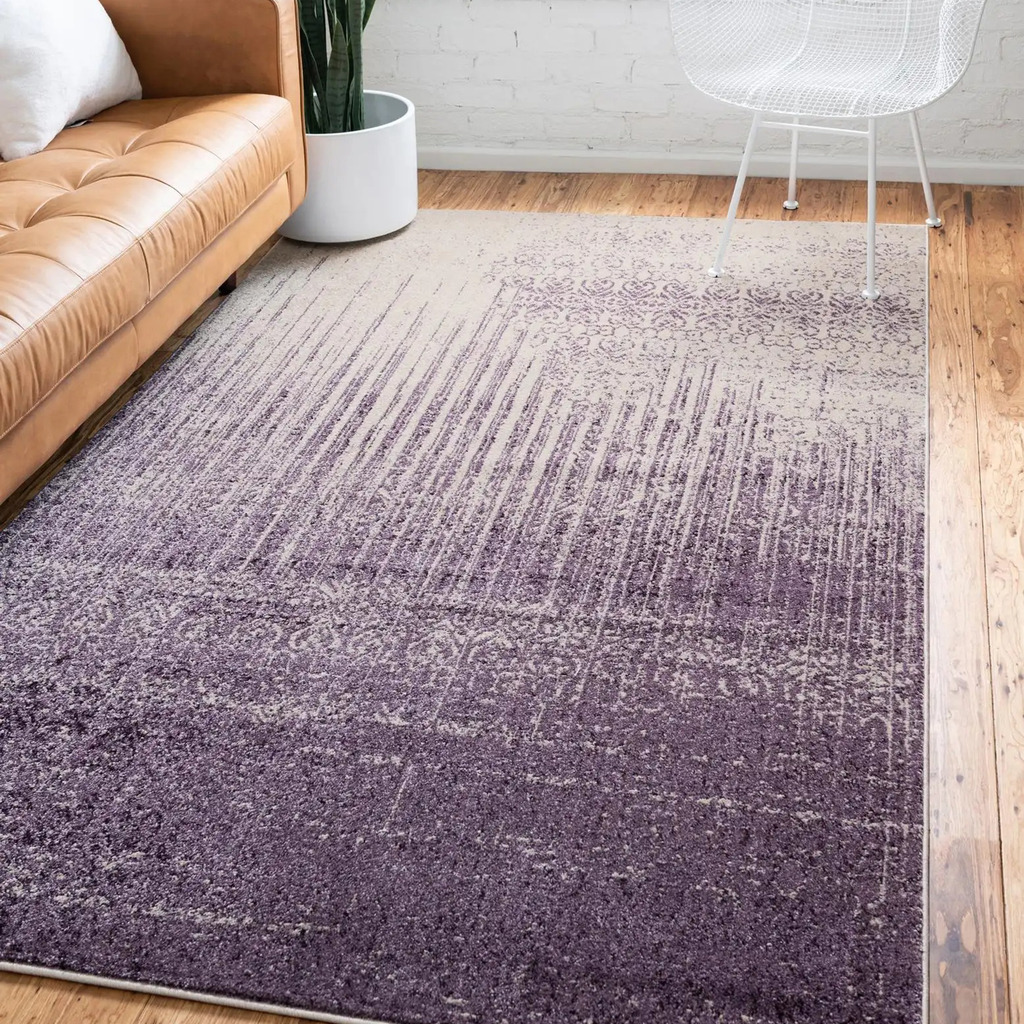}
        \put(20,104){Input image}
    \end{overpic}
    \begin{overpic}[width=0.162\linewidth]{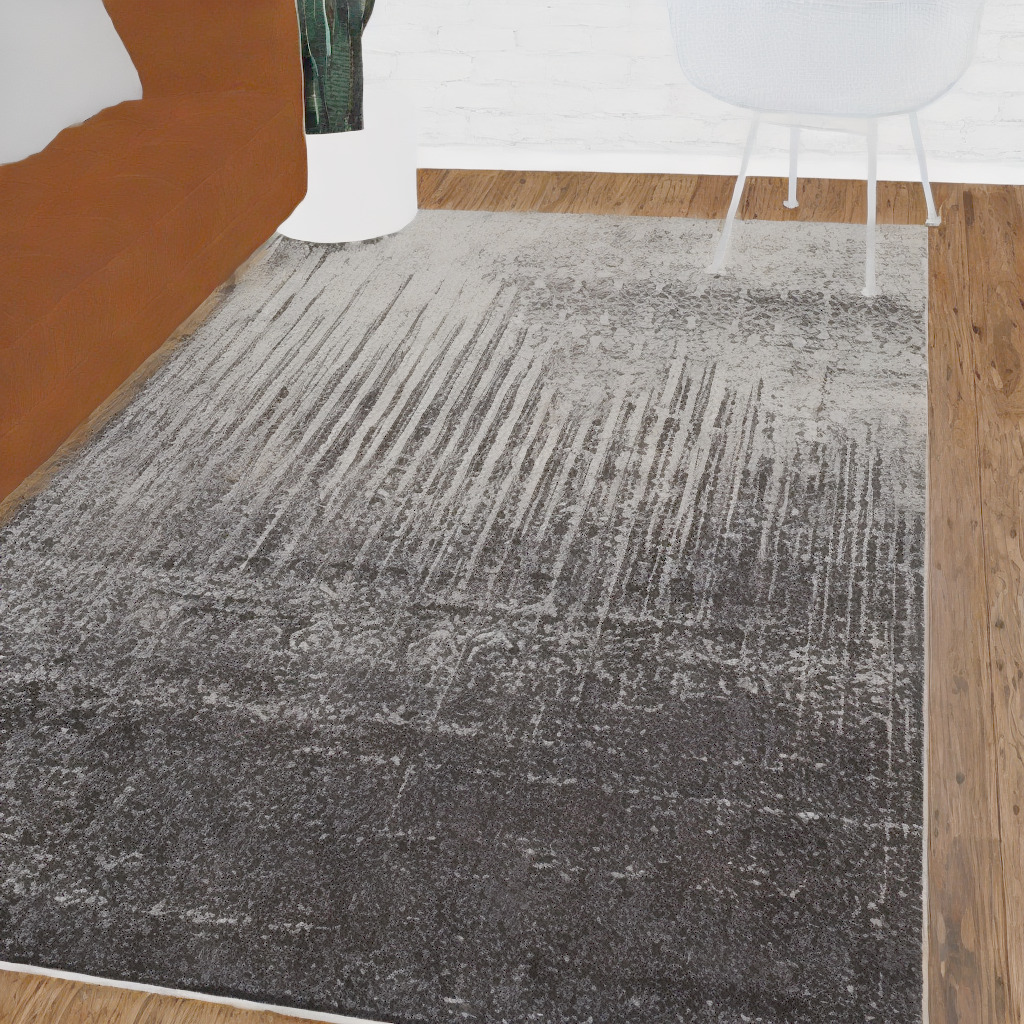}
        \put(2,116){\scriptsize Model outputs 3 channels}
        \put(33,104){albedo}
    \end{overpic}
    \begin{overpic}[width=0.162\linewidth]{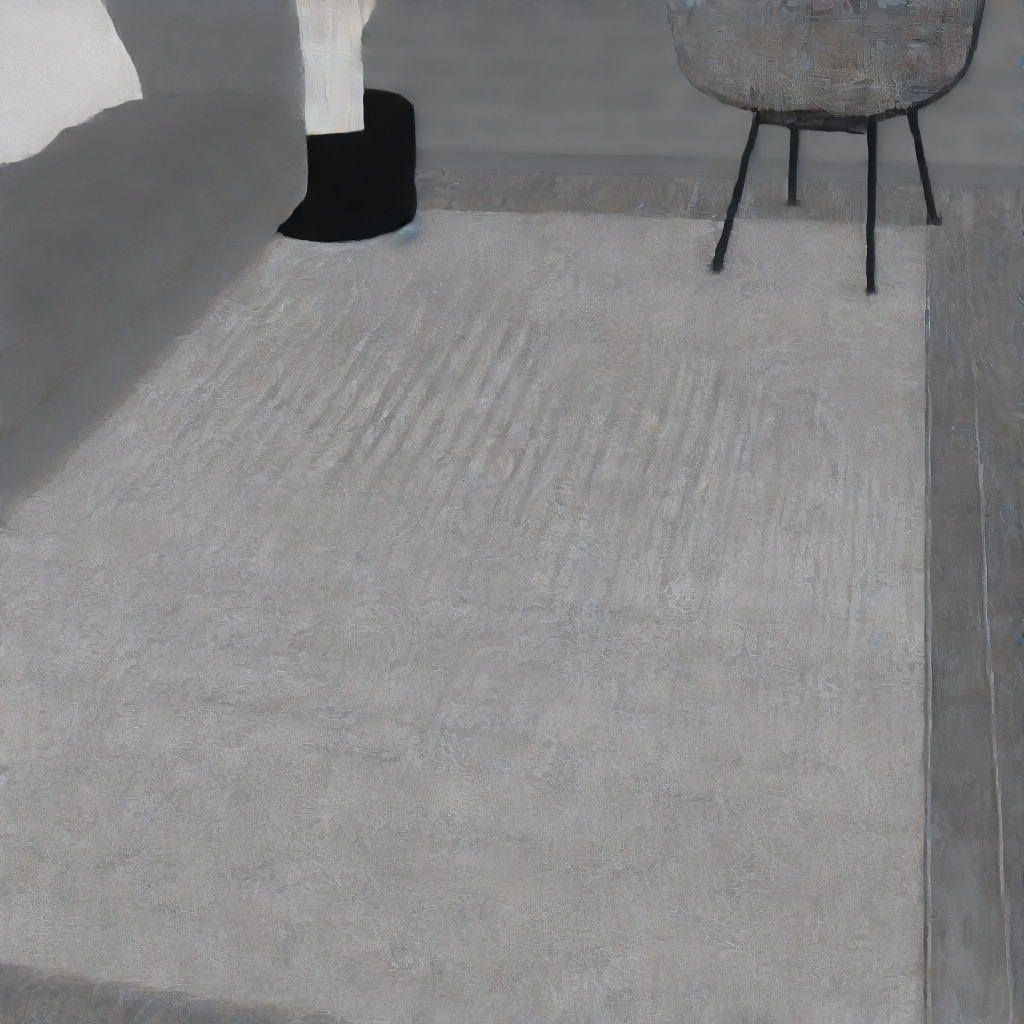}
        \put(2,116){\scriptsize Model outputs 3 channels}
        \put(22,104){roughness}
    \end{overpic}
    \begin{overpic}[width=0.162\linewidth]{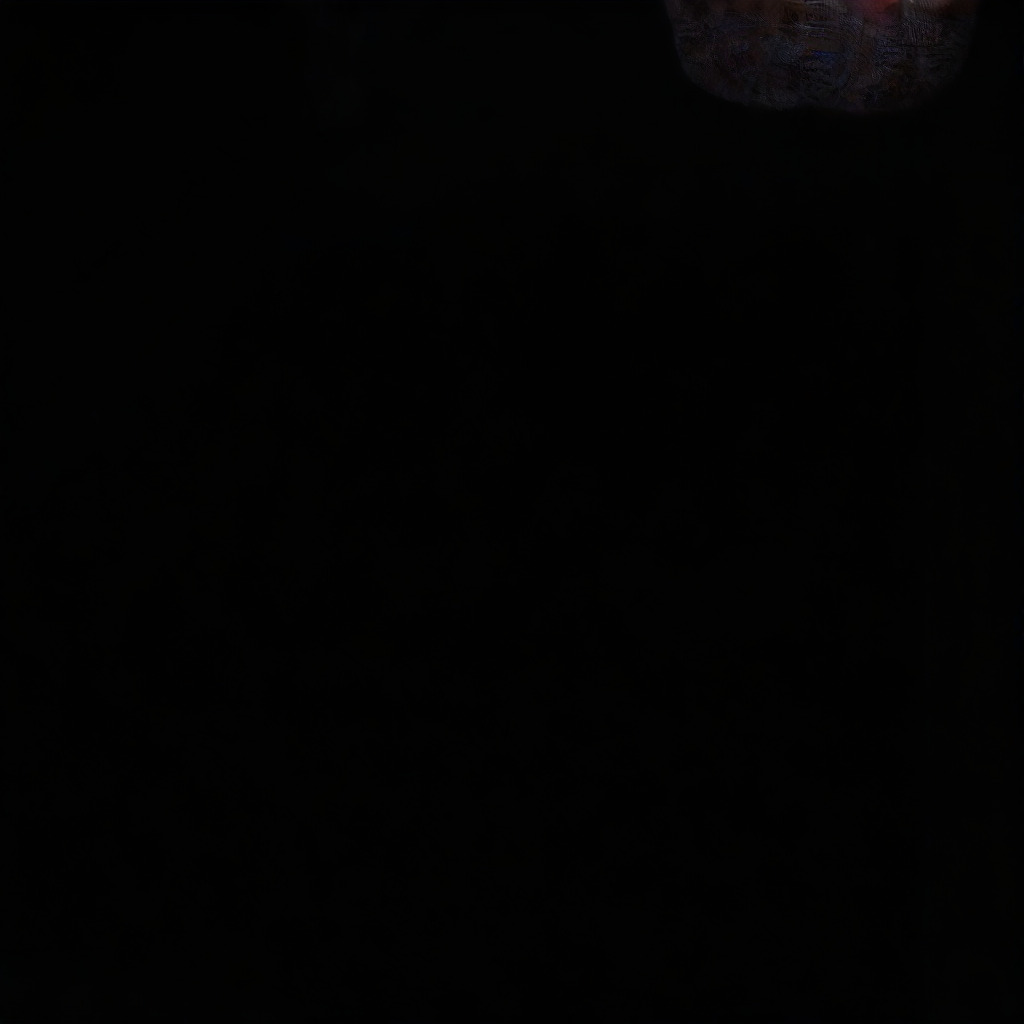}
        \put(2,116){\scriptsize Model outputs 3 channels}
        \put(22,104){metallicity}
    \end{overpic}
    \hspace*{\fill}
\end{minipage}\par\bigskip\bigskip
\begin{minipage}{0.8\linewidth}
    \begin{minipage}{\linewidth}
    \end{minipage}\par\medskip
    \hspace*{\fill}
    \begin{overpic}[width=0.162\linewidth]{supp/ablation-rgb2x/scene-5-photo}
        \put(20,104){Input image}
    \end{overpic}
    \begin{overpic}[width=0.162\linewidth]{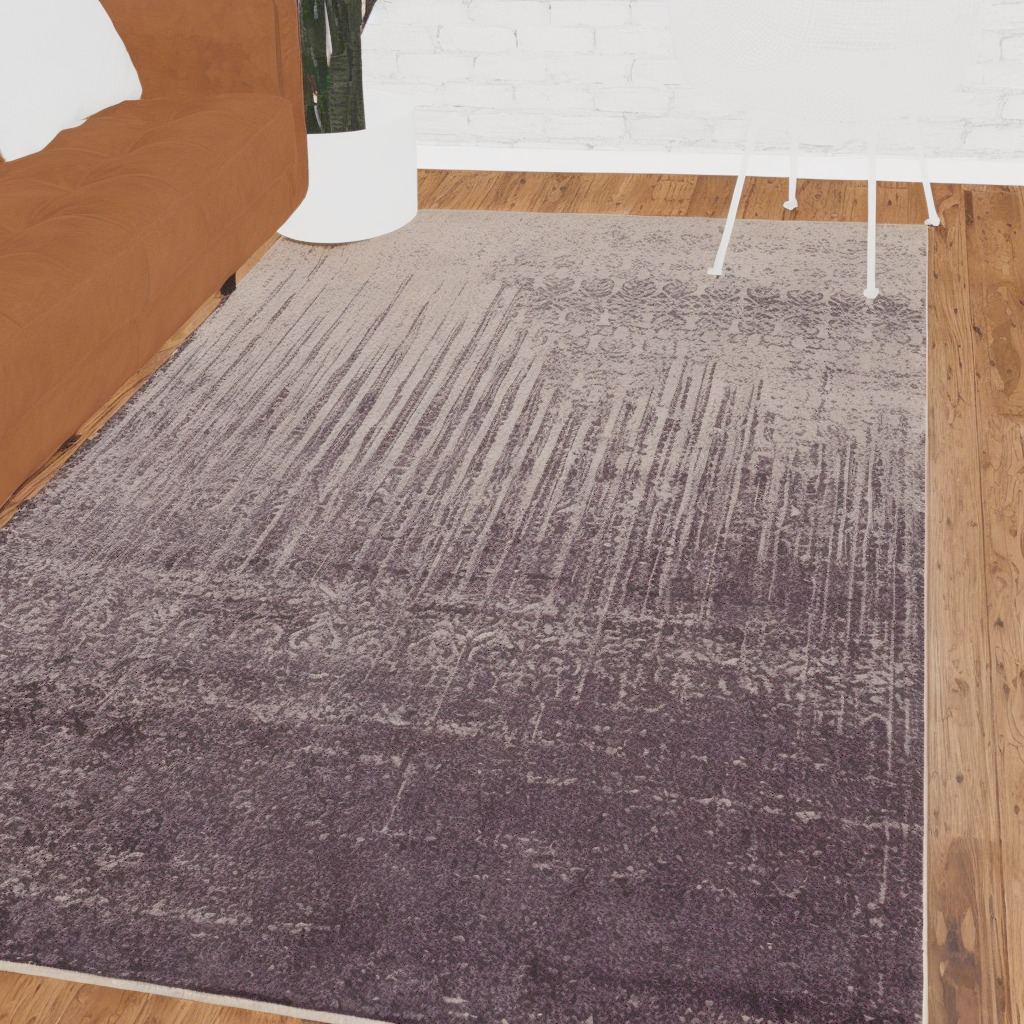}
        \put(2,104){\footnotesize \textbf{Our \rgbtoxx albedo}}
    \end{overpic}
    \begin{overpic}[width=0.162\linewidth]{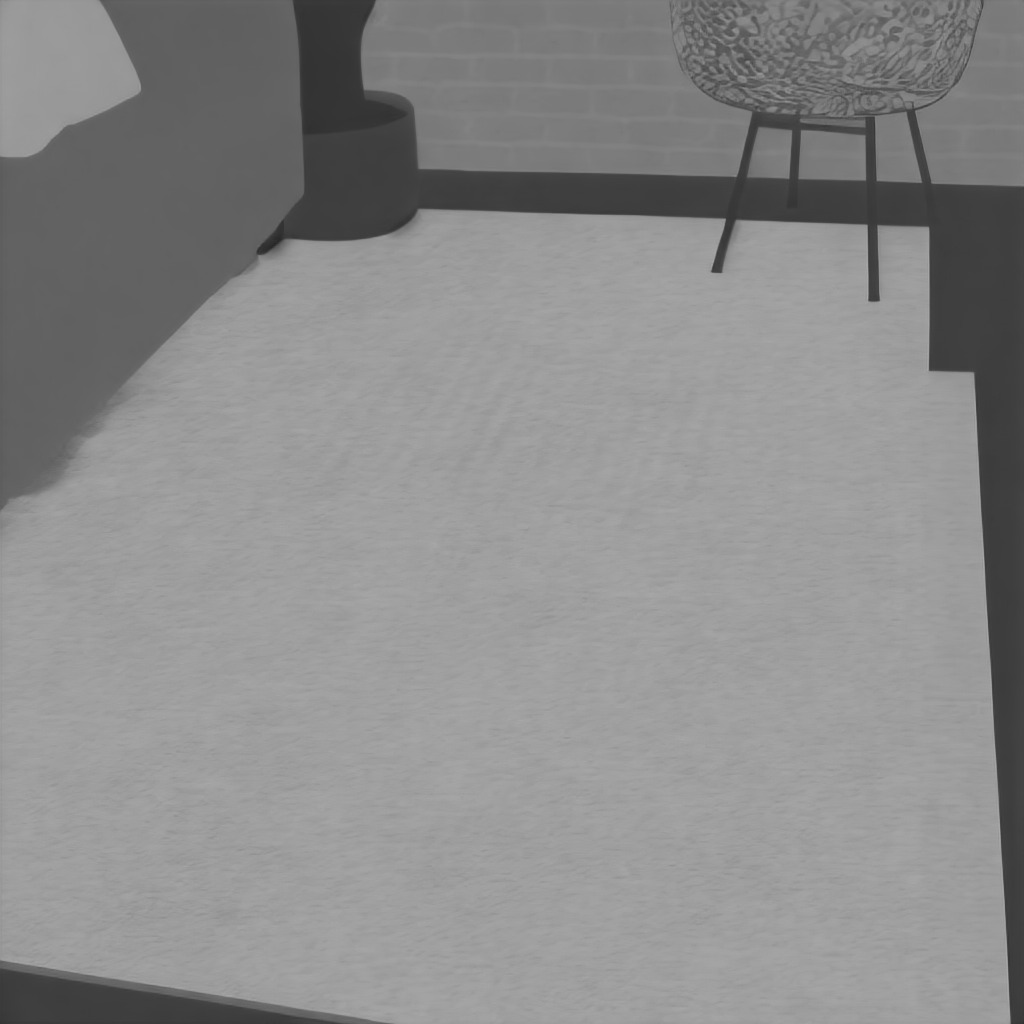}
        \put(2,104){\footnotesize \textbf{Our \rgbtoxx rough.}}
    \end{overpic}
    \begin{overpic}[width=0.162\linewidth]{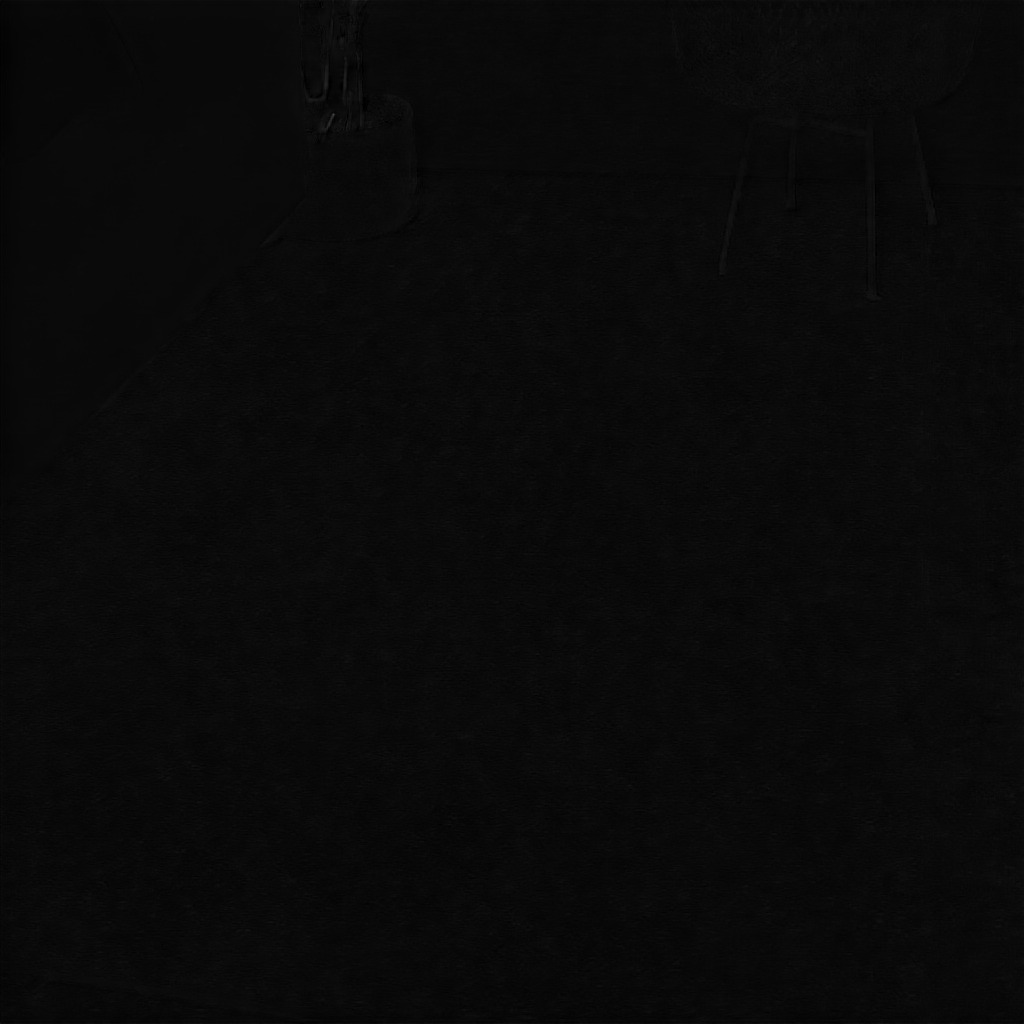}
        \put(2,104){\footnotesize \textbf{Our \rgbtoxx metal.}}
    \end{overpic}
    \hspace*{\fill}
\end{minipage}\par